\documentclass[conference]{IEEEtran}

\IEEEoverridecommandlockouts
\usepackage{amsmath,amssymb,amsfonts}
\usepackage{graphicx} 
\usepackage{algorithm}
\usepackage{algpseudocode}

\usepackage{amsmath} 

\usepackage{textcomp}
\usepackage{xcolor}
\usepackage{tcolorbox} 

\usepackage{nameref}

\usepackage{fancyhdr}
\usepackage[style=numeric]{biblatex}

\usepackage[T1]{fontenc}

\usepackage[justification=justified,singlelinecheck=false]{caption}

\usepackage{xcolor}
\usepackage{url}
\usepackage{hyperref}
\hypersetup{
    colorlinks=true,
    linkcolor=blue,
    filecolor=magenta,     
    urlcolor=cyan,
    bookmarks=true
}

\def\BibTeX{{\rm B\kern-.05em{\sc i\kern-.025em b}\kern-.08em
    T\kern-.1667em\lower.7ex\hbox{E}\kern-.125emX}}

\definecolor{mypurple}{HTML}{8172B2}
\definecolor{myred}{HTML}{C44E52}
\definecolor{myblue}{HTML}{4C72B0}
\definecolor{mygreen}{HTML}{55A868}
\definecolor{mycyan}{HTML}{64B5CD}

\begin{document}

\title{Navigating the Landscape of Large Language Models: A Comprehensive Review and Analysis of Paradigms and Fine-Tuning Strategies\\
}

\author{\IEEEauthorblockN{1\textsuperscript{st} Benjue Weng}}

\maketitle






\begin{abstract}
With the surge of ChatGPT,the use of large models has significantly increased,rapidly rising to prominence across the industry and sweeping across the internet. This article is a comprehensive review of fine-tuning methods for large models. This paper investigates the latest technological advancements and the application of advanced methods in aspects such as task-adaptive fine-tuning,domain-adaptive fine-tuning,few-shot learning,knowledge distillation,multi-task learning,parameter-efficient fine-tuning,and dynamic fine-tuning.
\end{abstract}


\section{Introduction}
The advent of Transformer models marks a significant milestone in the field of natural language processing (NLP). Initially conceived to address the limitations of recurrent neural networks (RNNs~\cite{RNN}) and convolutional neural networks (CNNs~\cite{CNN}) in handling long-range dependencies,the Transformer architecture,introduced by Vaswani et al. in 2017~\cite{Vaswani2017AttentionIA},has revolutionized the way we approach language understanding and generation tasks.

\textbf{Background of Transformer Architecture:} The Transformer model emerged from the need to process sequential data more effectively than traditional models. Its unique architecture,devoid of recurrence and convolutions,leverages attention mechanisms to draw global dependencies between input and output,offering remarkable improvements in processing efficiency and model performance.

\textbf{Encoder~\cite{Devlin2019BERTPO},Decoder~\cite{Radford2018ImprovingLU}~\cite{Radford2019LanguageMAgpt2}~\cite{gpt3},and Encoder-Decoder~\cite{T5} Architectures: } The Transformer architecture is primarily characterized by its encoder and decoder components. The encoder processes the input sequence,creating a rich contextual representation of each word. In contrast,the decoder generates an output sequence,often in a language translation task,using the encoded information. 

\textbf{The distinction lies in their roles: } encoders are context-aware processors of input,while decoders generate predictions based on encoded inputs. The encoder-decoder architecture,often used in sequence-to-sequence tasks,combines these two components,facilitating complex tasks like machine translation,where the encoder processes the source language and the decoder generates the target language.

\textbf{Emergence of Fine-Tuning in Large Models: } The concept of fine-tuning large language models stems from the challenge of adapting these models,pre-trained on vast,diverse datasets,to specific tasks or domains. Fine-tuning adjusts the model's weights,tailored to particular tasks,enhancing its ability to generalize from broad linguistic patterns to specific application requirements. This approach has become increasingly vital as models grow in size and complexity,necessitating more nuanced adaptation techniques to harness their full potential.

This paper is structured to provide a comprehensive overview of the methodologies and advancements in fine-tuning large language models. The subsequent sections are organized as follows:

\textbf{Literature Review: } Examines the evolution of language models leading to the Transformer architecture,highlighting key developments and foundational concepts.

\textbf{Theoretical Foundations:} Delves into the theoretical underpinnings of the Transformer model,including the mechanics of attention mechanisms,encoders,and decoders.

\textbf{Fine-Tuning Strategies:} Discusses various fine-tuning approaches,such as task-specific,domain-specific adaptations,and advanced techniques like few-shot learning and dynamic fine-tuning.

\textbf{Challenges and Future Directions:} Identifies current challenges in fine-tuning methodologies and explores potential future research directions in this rapidly evolving field.

This paper introduces the paradigm of large language models based on the Transformer architecture and provides a detailed overview of commonly used large model fine-tuning methods. It concludes with a comparative experiment focusing on the model size and the LoRA fine-tuning paradigm across six text classification datasets. The code for the experiments has been made available on GitHub\footnote{Github Code:\label{section:llms-peft-cook}\url{https://github.com/wengbenjue/llms-peft-cook}}.

\section{RELATED WORK}
The Transformer architecture,introduced in ``Attention Is All You Need'' by Vaswani et al. (2017)~\cite{Vaswani2017AttentionIA},revolutionized NLP by replacing recurrent layers with self-attention mechanisms.

\subsection{Evolution of Transformer-Based Models}

\textbf{BERT}: ``BERT: Pre-training of Deep Bidirectional Transformers for Language Understanding'' by Devlin et al. (2018)~\cite{Devlin2019BERTPO} introduced bidirectional training for deeper context comprehension. Key features of BERT include:

\textbf{Bidirectional Training}: BERT uniquely allows for considering both left and right context in the text,unlike previous unidirectional models.

\textbf{Pre-training and Fine-tuning}: It is first pre-trained on a large corpus and then fine-tuned for specific tasks.

\textbf{Transformer Architecture}: BERT is based on the Transformer model,particularly its encoder component.

\textbf{Wide Applications}: Since its introduction,BERT has achieved remarkable performance across various NLP tasks like sentiment analysis,question-answering systems and text summarization.

\textbf{Open Source Availability:} Google has made the pre-trained BERT models publicly available, facilitating research and application.

These characteristics have made BERT a milestone in the field of natural language processing, laying the groundwork for subsequent models like RoBERTa~\cite{zhuang-etal-2021-robustly-roberta} and ALBERT~\cite{Lan2019ALBERTAL}\footnote{BERT opensource code:\label{section:bert}\url{https://github.com/google-research/bert}}.

\textbf{GPT Series:} Radford et al.'s (2018) ``Improving Language Understanding by Generative Pre-Training''~\cite{Radford2018ImprovingLU} initiated the GPT series, advancing generative tasks in NLP. GPT-2~\cite{Radford2019LanguageMAgpt2}: An expansion over GPT, it featured more parameters and a larger dataset, enhancing the coherence and diversity of text generation. GPT-3~\cite{gpt3}: Further scaled up, as detailed by Brown et al. (2020), it introduced enhanced generative capabilities and better task generalization, showcasing zero-shot and few-shot learning abilities on various tasks. These models are characterized by their use of the Transformer's decoder architecture and learn rich language patterns through extensive unsupervised learning. The success of the GPT series eventually led to applications like ChatGPT, which is specifically optimized for conversational systems to produce more natural and fluent dialogues\footnote{ChatGPT: \url{https://openai.com/chatgpt}}.

\subsection{In-Context Learning and Prompt Engineering}
GPT-3~\cite{gpt3} demonstrated in-context learning,enabling language models to adapt to tasks through relevant prompts. In-context learning and prompt engineering are two key features of large language models (like GPT-3),playing a significant role in natural language processing (NLP).

\textbf{In-Context Learning:} A technique allowing models to adapt to specific tasks without explicit fine-tuning.
The model analyzes provided examples (the context) to understand and perform tasks.

For instance,presenting GPT-3 with a series of questions and answers enables it to``learn'' how to respond to similar subsequent questions.

(Pan,Jane and Gao et al.,2023)~\cite{pan-etal-2023-context} primarily investigates how Large Language Models (LLMs) learn tasks through In-Context Learning (ICL) and examined the mechanisms behind this learning approach. The researchers tested 16 classification datasets through experiments using three cutting-edge LLM families: GPT-3~\cite{gpt3},LLaMA~\cite{Touvron2023LLaMA},and OPT~\cite{Liu2021OPTOP}. Additionally,the paper discusses the difference between Task Recognition (TR) and Task Learning (TL) in ICL,and how LLMs enhance TL with increasing model scales. Finally,the paper explores the implications of these findings for future ICL research and how they can be applied to improve the performance of LLMs in various classification tasks. (Qingxiu Dong et al.,2022)~\cite{ASurveyIncontextLearning} initially presents a formal definition of In-Context Learning (ICL) and clarifies its relation to related research. It then introduces advancements and advanced techniques in ICL,including training strategies,demonstration design strategies,and relevant analyses. Training strategies encompass gradient-based,sample-based,and label-based methods; demonstration design strategies include template-based,interaction-based,and natural language-based methods. Furthermore,the paper covers the application areas of ICL,such as text classification,named entity recognition,and machine translation. Finally,it discusses the challenges in ICL,including data scarcity,domain adaptability,model interpretability,and offers potential directions for further research,like enhancing ICL's efficiency and accuracy and addressing data scarcity issues.

\textbf{Prompt Engineering:} The process of designing and optimizing prompts (input texts) to guide the model to generate specific outputs.
This method relies on carefully crafted prompts to elicit specific behaviors or responses from the model.
For example,precisely formulated questions or statements can guide the model to perform better on specific tasks or generate more relevant outputs.
These techniques are particularly important in large pre-trained models,offering a flexible,efficient way to leverage the models' capabilities without the need for laborious fine-tuning. In-context learning and prompt engineering allow users to better control and direct the model's output,thereby achieving better results in various NLP tasks.(Jules White et al.,2023)~\cite{White2023APP}
discusses the application of LLMs to automate software development tasks. It provides a framework for documenting patterns for structuring prompts to solve a range of problems,which can then be adapted to different domains.(Lisa P et al.,2022.)~\cite{Argyle2022AnIA} takes an information-theoretic approach to prompt engineering. It focuses on how pre-trained language models,which acquire substantial linguistic and factual knowledge from massive corpora,can be aligned to specific tasks through prompt engineering.(Qinyuan Ye et al.,2023.)~\cite{ye2023prompt} highlights the complexities involved in prompt engineering,emphasizing the need for intricate reasoning to analyze a model's errors,determine what might be missing or misleading in the current prompt,and communicate tasks clearly.(Andrew Gao et al.,2023.) discusses the influence of large language models,like OpenAI's ChatGPT and Google's Bard,in various aspects of life and work. It explores the nuances of prompt engineering in the context of these advanced models.

\subsection{Fine-Tuning in Large Language Models}
\begin{figure}[htbp]
\centering{\includegraphics[width=0.8\linewidth]{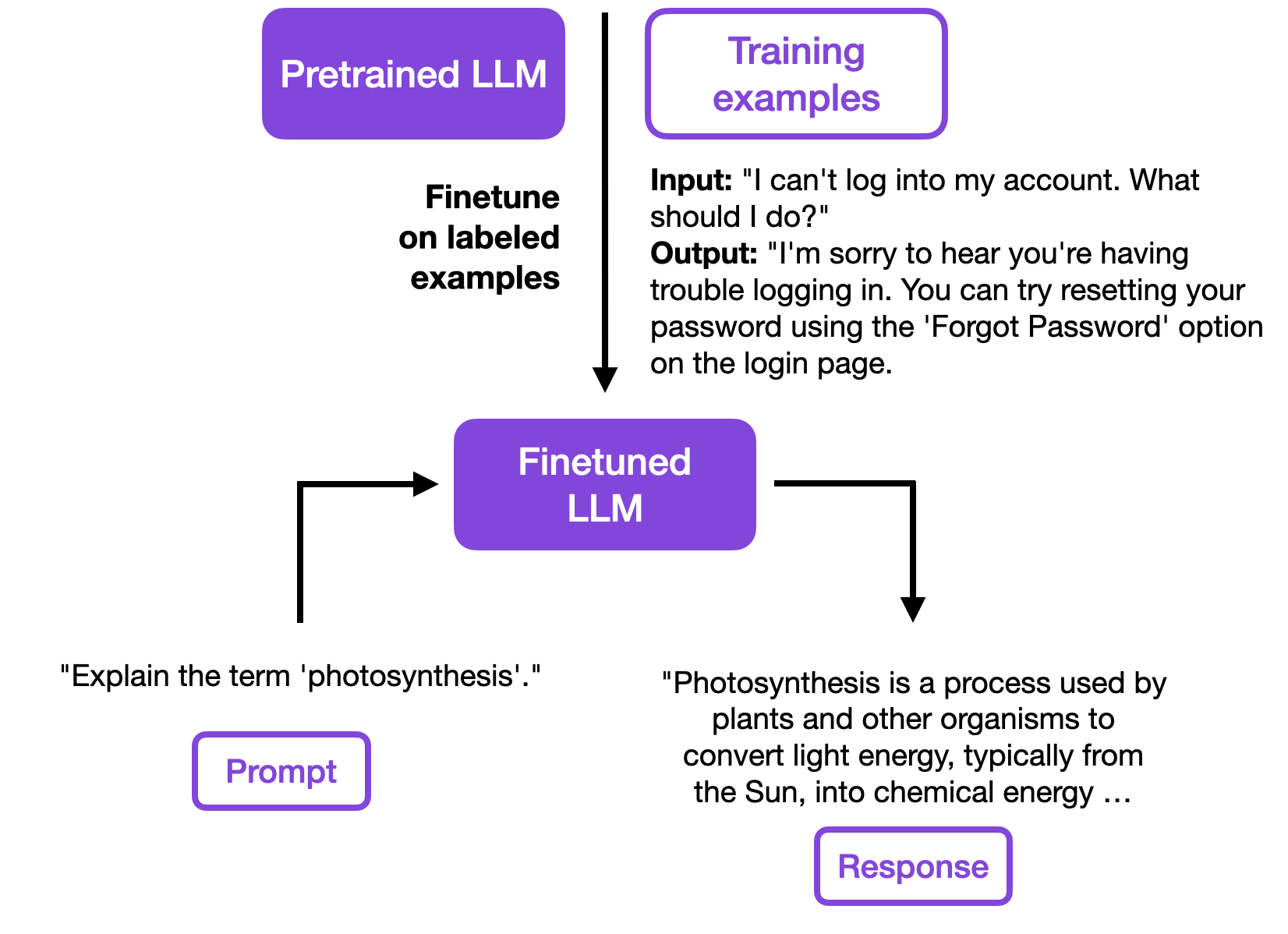}}
\caption{Finetuning a pretrained LLM to follow instructions~\cite{FintuingLLM}.}
\label{fig-fintuing-pretrained-llm}
\end{figure}
The fine-tuning concept,crucial for adapting large pre-trained models to specific tasks(Fig.~\ref{fig-fintuing-pretrained-llm}),was advanced by Howard and Ruder in~\cite{howard2018universal}.
The Evolution of Fine-Tuning Strategies:

\textbf{Task-Specific Adaptation:} is a method of fine-tuning large language models,aimed at specially adjusting a general model to suit specific applications or tasks. The key aspects of this approach include:
\begin{itemize}
\item \textbf{Selecting Task-Specific Datasets}: Data is collected from specific tasks such as sentiment analysis,question-answering systems,etc.
\item \textbf{Fine-Tuning Training}: These data are used to further train the pre-trained general model,enabling it to better understand and perform specific tasks.
\item \textbf{Evaluation and Optimization}: The model's performance on specific tasks is evaluated and adjusted as needed. \textit{ULMFiT}~\cite{howard-ruder-2018-universal} proposes a universal method for fine-tuning,with a special emphasis on how to use pre-trained language models for task-specific fine-tuning. \textit{BERT}~\cite{Devlin2019BERTPO} demonstrates how it can be adapted to various NLP tasks through fine-tuning.
\end{itemize}

\textbf{Domain-Specific Fine-Tuning: } This refers to adjusting large pre-trained language models to meet the needs of specific industries or professional fields.The steps include:
\begin{itemize}
\item \textbf{Collecting Domain-Specific Data:} This may include industry reports,professional papers,technical documents,etc.
\item \textbf{Fine-Tuning Pre-Trained Models:} Using this data to adjust the model,enhancing its understanding and handling of domain-specific content.
\item \textbf{Evaluation and Iteration:} Testing the model's performance in a specific domain and optimizing it based on feedback.BioBERT~\cite{BioBERT2019} demonstrates how to perform domain-specific fine-tuning on the BERT model,making it more suitable for biomedical text mining.In ClinicalBERT~\cite{ClinicalBERT2019},the authors fine-tune the BERT model to adapt it for the analysis of clinical medical notes.
\end{itemize}

\textbf{Few-Shot Learning:} Few-Shot Learning refers to the method of training machine learning models with only a small amount of labeled data. This is particularly important in the field of large pre-trained models,as it allows the model to quickly adapt to new tasks without the need for extensive data.
Key steps and features of few-shot learning include:
\begin{itemize}
\item \textbf{High-Quality Samples:} Select a small but high-quality,representative set of samples for training.
\item \textbf{Utilizing Pre-Trained Models:} Make use of models that have been pre-trained on large datasets to leverage their rich prior knowledge.
\item \textbf{Meta-Learning or Transfer Learning Techniques:} Employ special training techniques,such as meta-learning or transfer learning,to enhance the model's generalization capability.
\end{itemize}

Model-Agnostic Meta-Learning~\cite{Finn2017ModelAgnosticMF} introduced the MAML (Model-Agnostic Meta-Learning) framework,a milestone in the field of few-shot learning.Few-Shot Text Classification~\cite{Geng2019FewShotTC} focused on few-shot learning for text classification,proposing an effective network-based method.Few-shot learning is particularly important in fields with scarce data,as it can significantly reduce the need for training data,accelerating the adaptation speed and flexibility of models.

\begin{figure}[htbp]
\centering
\includegraphics[width=0.8\linewidth]{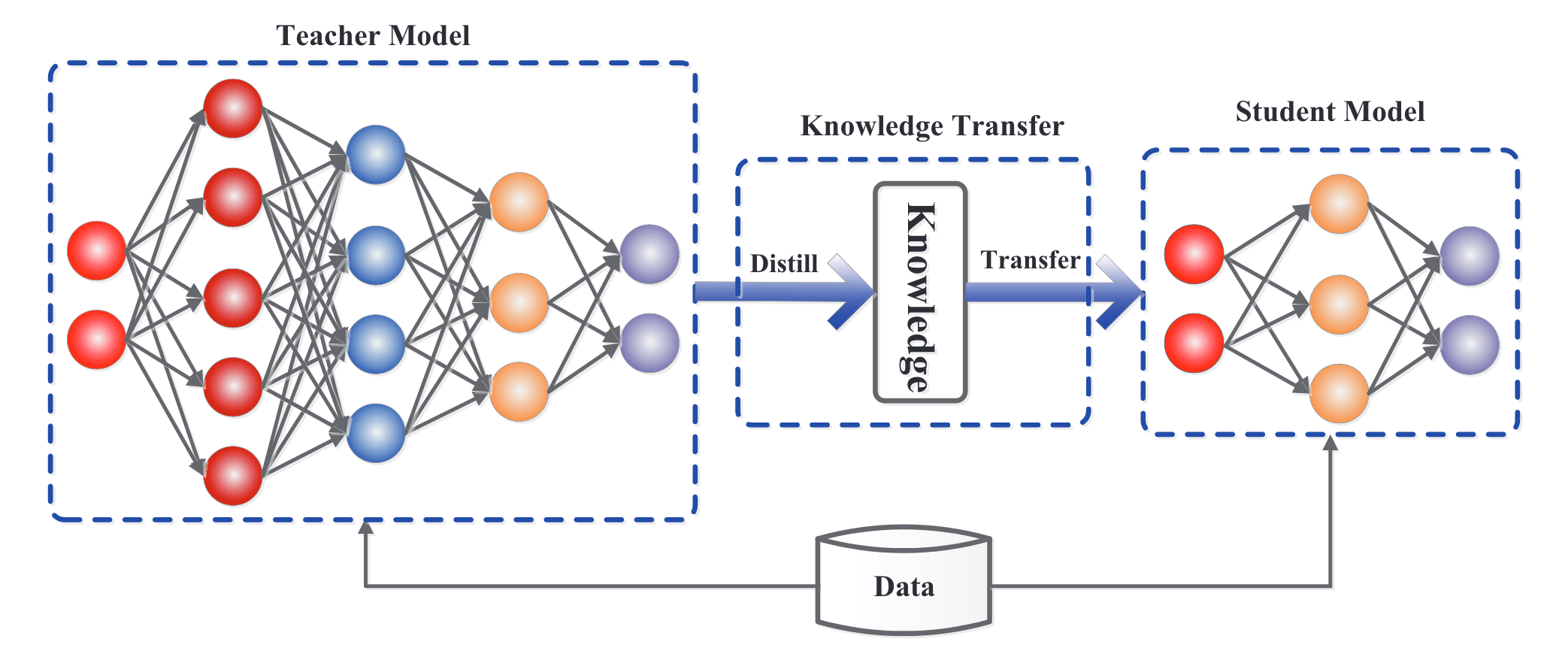}
\caption{The generic teacher-student framework for knowledge distillation.~\cite{Gou2020KnowledgeDA}}
\label{fig-teacher-student}
\end{figure}

\textbf{Knowledge Distillation: } Knowledge Distillation(Fig.~\ref{fig-teacher-student}) is a model compression technique used to transfer knowledge from large models (teacher models) to smaller models (student models). This method enables smaller models to mimic the behavior of large models while maintaining a smaller size and higher operational efficiency.

Key steps in knowledge distillation include:
\begin{itemize}
\item \textbf{Training Large Models:} train a large,high-performance model.
\item \textbf{Transfer Learning:} Use the output of the large model to guide the training of the small model.
\item \textbf{Optimization and Evaluation:} Adjust the small model to maximally mimic the behavior of the large model and evaluate its performance.
\end{itemize}

''Distilling the Knowledge in a Neural Network'' by Hinton et al . (2015).~\cite{Hinton2015DistillingTK} This pioneering paper in the field of knowledge distillation was authored by Geoffrey Hinton and others.''TinyBERT: Distilling BERT for Natural Language Understanding'' by Jiao et al. (2019)~\cite{Jiao2019TinyBERTDB}. This paper focuses on distilling the BERT model into a smaller model for natural language understanding tasks.

\textbf{Multi-Task Learning: } Multi-Task Learning (MTL) is a method for training machine learning models to learn simultaneously across multiple related tasks. The core advantage of this approach lies in shared representation learning,allowing models to apply knowledge learned in one task to others.Key features include:
\begin{itemize}
\item \textbf{Shared Architecture:} Models typically have shared layers to learn common features across all tasks.
\item \textbf{Task-Specific Layers:} Additionally,each task might have its dedicated layers for learning task-specific features.
\item \textbf{Joint Optimization:} During training,the loss function across all tasks is optimized jointly.
\end{itemize}

''An Overview of Multi-Task Learning in Deep Neural Networks'' by Sebastian Ruder (2017).~\cite{Ruder2017MultiTaskLearning}  provides a detailed introduction to the principles and methods of multi-task learning.''Multi-Task Deep Neural Networks for Natural Language Understanding'' by Liu et al. (2019).~\cite{MultiTaskDeepNeuralNetworks}  focuses on using deep neural networks for multi-task learning in the field of Natural Language Processing (NLP).Multi-task learning enhances the generalization capabilities of models,reduces the need for large amounts of labeled data,and improves data utilization efficiency.

\textbf{Parameter-Efficient Fine-Tuning:} Parameter-Efficient Fine-Tuning refers to a method that makes the fine-tuning process of models more efficient,particularly suitable for large pre-trained models. The goal of these methods is to adapt to new tasks by altering fewer parameters,thus saving computational resources and storage space.

\begin{itemize}
\item \textbf{Adapter:} Adapter layers are inserted between layers of the pre-trained model,with only these small adapter layers being fine-tuned instead of the entire model. ``Parameter-Efficient Transfer Learning for NLP'' by Houlsby et al. (2019).~\cite{houlsby2019parameterefficient} This paper primarily discusses a transfer learning method based on adapter modules,which enables parameter sharing and scalability without sacrificing performance. The authors demonstrate the effectiveness of this approach by transferring the BERT Transformer model to 26 text classification tasks,achieving near state-of-the-art performance with only a small number of additional parameters for each task.
\item \textbf{Prefix-Tuning:} Involves adding a small,learnable prefix to the input sequence to guide the model in generating specific outputs. ``Prefix-Tuning: Optimizing Continuous Prompts for Generation'' by Li and Liang (2021).~\cite{li2021prefixtuning} This paper primarily discusses a natural language generation method called Prefix-Tuning,which optimizes downstream tasks by tuning a small,continuous task-specific vector (referred to as a prefix) instead of modifying all language model parameters. By learning only $0.1\%$ of the parameters,Prefix-Tuning achieves comparable performance in full-data settings,outperforms Fine-Tuning in low-data scenarios,and demonstrates better extrapolation capabilities for examples of topics not seen during training.
\item \textbf{LoRA (Low-Rank Adaptation):} Adjusts pre-trained models by making low-rank modifications to the model's weight matrices,allowing effective fine-tuning of large models without adding too many parameters.``LoRA: Low-Rank Adaptation of Large Language Models'' by Hu et al. (2021).~\cite{Hu2021LoRA} primarily discusses a technique called LoRA,which stands for Low-Rank Adaptation for large language models. This method significantly reduces the number of trainable parameters for downstream tasks while still maintaining high model quality.
\end{itemize}

These techniques limit the number of parameters that need to be changed during the fine-tuning process,making the fine-tuning of large models more efficient and feasible.

\textbf{Dynamic Fine-Tuning: } Dynamic Fine-Tuning is a method that involves continuous optimization of a model during training using real-time data. This approach is particularly suitable for datasets and applications that change over time.Features and steps include:
\begin{itemize}
\item \textbf{Real-Time Data Integration:} Dynamic fine-tuning continuously integrates new,real-time data samples into the training process.
\item \textbf{Ongoing Optimization:} Model parameters are constantly adjusted and optimized based on new data.
\item \textbf{Strong Adaptability:} This method allows the model to better adapt to changes in data,enhancing its responsiveness to new information.
''Fine-tuning Language Models from Human Preferences'' by Ziegler et al. (2019).~\cite{Ziegler2019FineTuningLM}  demonstrates how to dynamically fine-tune language models using real-time user feedback.
''Continual Learning for Natural Language Generation in Task-Oriented Dialogue Systems'' by Mi et al. (2020).~\cite{Mi2020ContinualLF}  explores the use of dynamic fine-tuning in task-oriented dialogue systems.

Dynamic fine-tuning is particularly important for applications that need to reflect the latest data and trends promptly,such as news generation,financial market analysis,or social media content monitoring.
\end{itemize}

\section{Transformer Architecture}

\begin{figure}[htbp]
\centering
\includegraphics[width=0.8\linewidth]{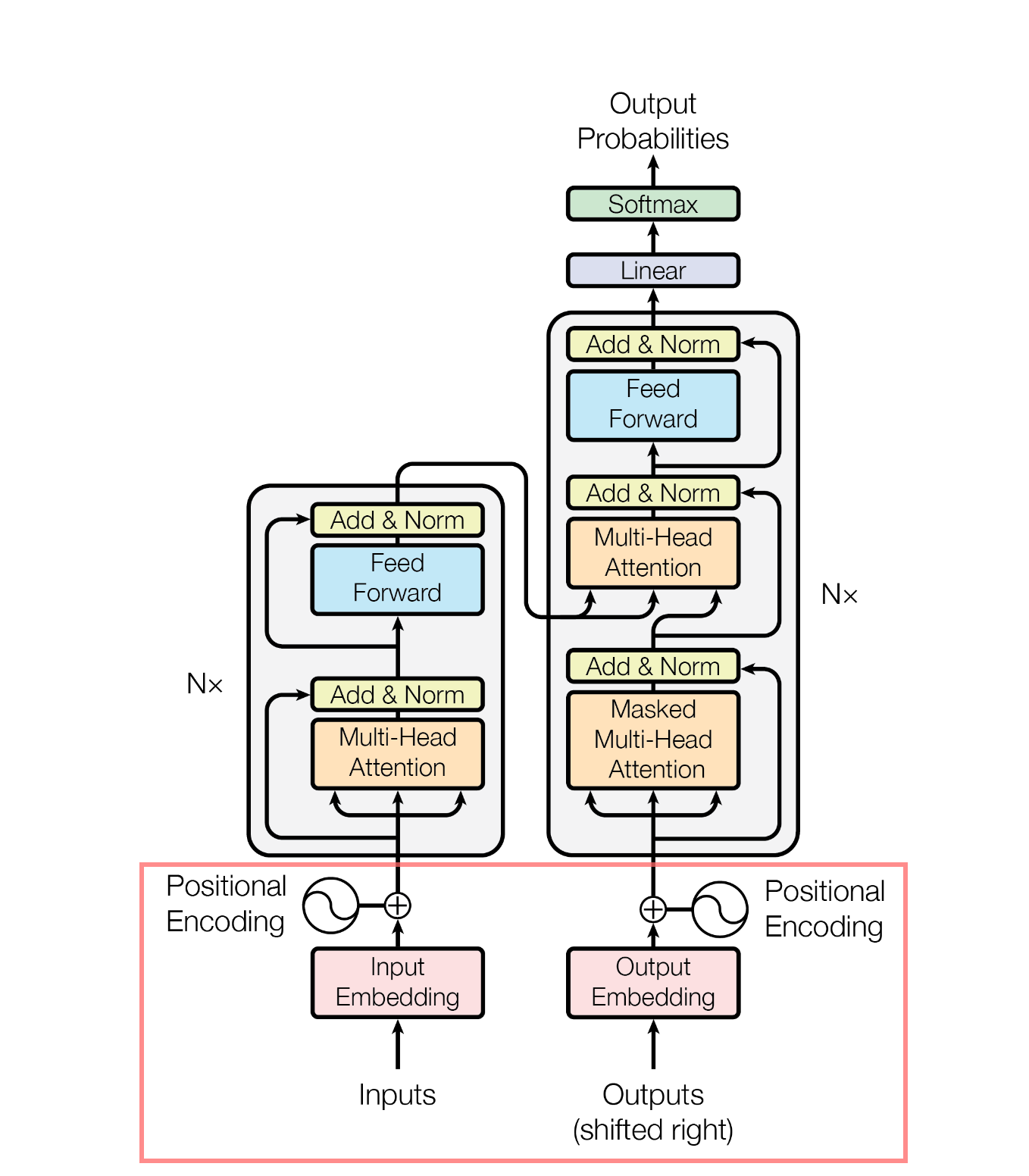}
\caption{The Transformer-model architecture.~\cite{Vaswani2017AttentionIA}}
\label{fig-transormer-architecture}
\end{figure}

The Transformer architecture(Fig.~\ref{fig-transormer-architecture}),introduced in the seminal paper “Attention Is All You Need” by Vaswani et al. in 2017~\cite{Vaswani2017AttentionIA},revolutionized the approach to natural language processing tasks. It forms the backbone of many modern NLP models,including BERT~\cite{Devlin2019BERTPO},GPT~\cite{Radford2018ImprovingLU},and T5~\cite{T5}.

\textbf{Key Components of the Transformer Architecture: }

\subsection{Attention Mechanism } 
\begin{figure}[htbp]
\centering{\includegraphics[width=0.8\linewidth]{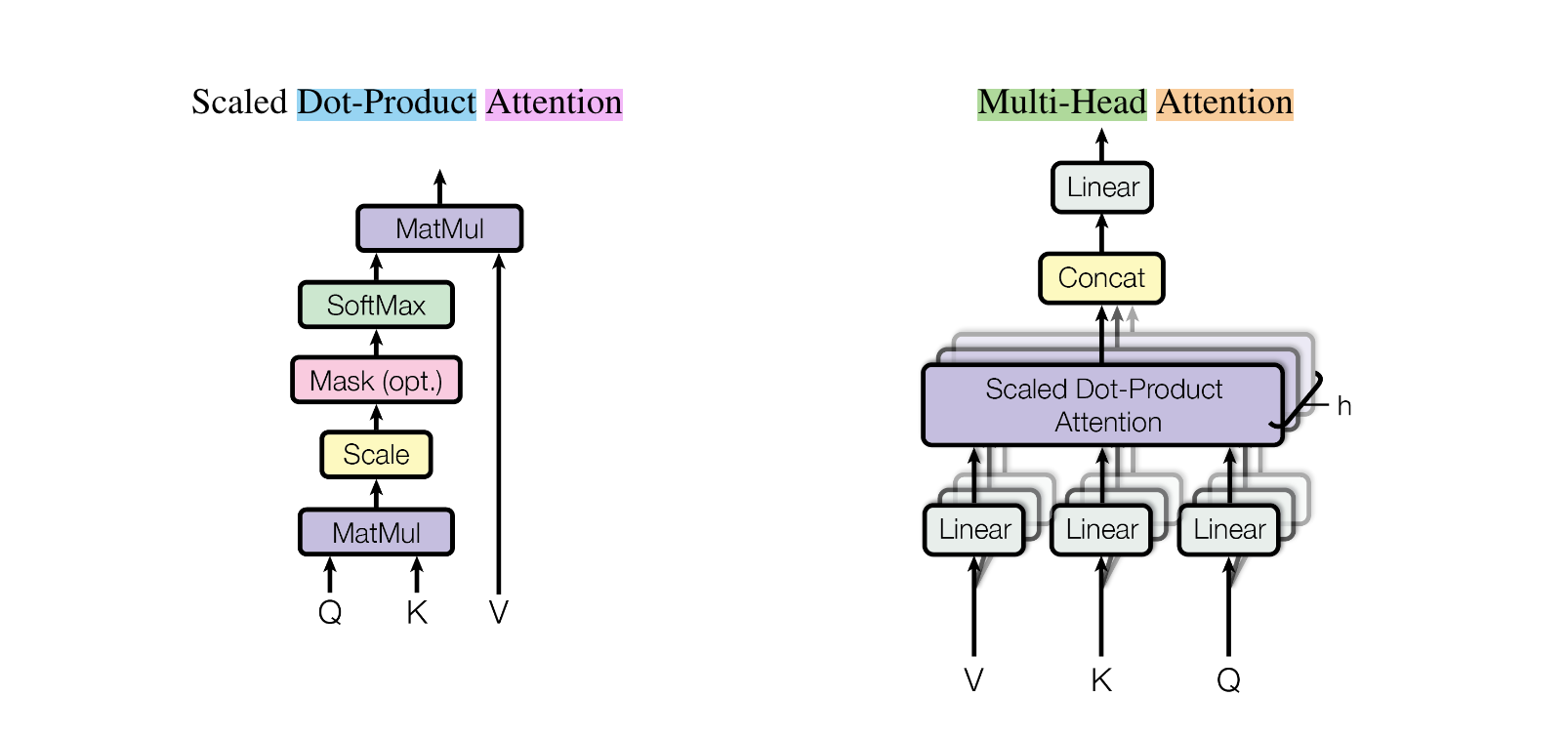}}
\caption{(left) Scaled Dot-Product Attention. (right) Multi-Head Attention consists of several attention layers running in parallel.~\cite{Vaswani2017AttentionIA}}
\label{fig-Attention}
\end{figure}
The central idea of the Transformer is the attention mechanism,specifically``self-attention''.
Self-attention (Fig.~\ref{fig-Attention}) enables the model to assign significance to different words in a sentence,considering the whole sequence,which aids in grasping context and relationships within the text.
\begin{equation}
\operatorname{Attention}(Q,K,V)=\operatorname{softmax}\left(\frac{Q K^T}{\sqrt{d_k}}\right) V \label{eq-selfattention}
\end{equation}

Equation~\eqref{eq-selfattention},$Q$,$K$,and $V$ respectively represent the Query,Key,and Value matrices,which are derived from the input sequence through linear transformations. $d_k$ denotes the dimensionality of the key vectors. The core of the self-attention mechanism lies in computing the relationships between these vectors.

\textbf{Attention Score Calculation: } Initially,the dot product of the queries and keys is computed,followed by scaling with $\sqrt{d_k}$. This step is designed to measure the similarity of each query with all keys.

\begin{figure}[htbp]
\centering{\includegraphics[width=0.8\linewidth]{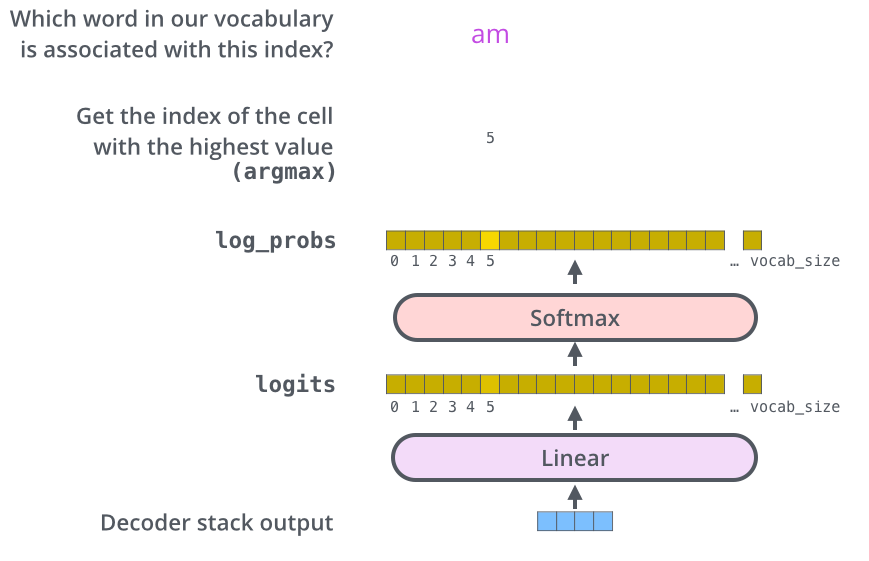}}
\caption{transformer decoder output softmax.~\cite{tranformer_understand}}
\label{fig-transformer-softmax}
\end{figure}

\textbf{Applying the Softmax Function: } Next,the softmax function (Fig.~\ref{fig-transformer-softmax}) is applied to the scores of each query. This transforms the scores into a probability distribution,indicating the importance of each value in the output.

\textbf{Weighted Summation: } Finally,these probabilities are used to perform a weighted summation of the values,resulting in the final attention output.

\begin{figure}[htbp]
\centering{\includegraphics[width=0.8\linewidth]{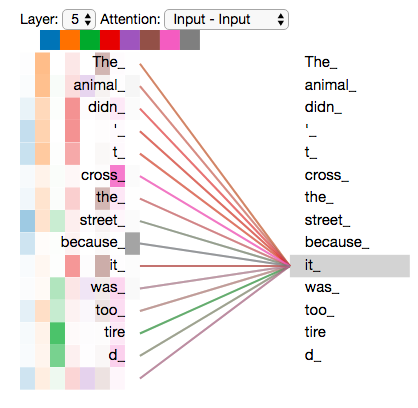}}
\caption{transformer self-attention visualization.~\cite{tranformer_understand}}
\label{fig-attention-it}
\end{figure}

The advantage of the self-attention mechanism is its ability to capture the relationship between any two elements in the sequence,regardless of their position(Fig.~\ref{fig-attention-it}). This is particularly crucial for understanding context and long-range dependencies in text. Moreover,since the computations for all positions can be processed in parallel,self-attention is also computationally more efficient than traditional recurrent neural networks.

\subsection{Encoder and Decoder Blocks }

\begin{figure}[htbp]
\centering{\includegraphics[width=0.8\linewidth]{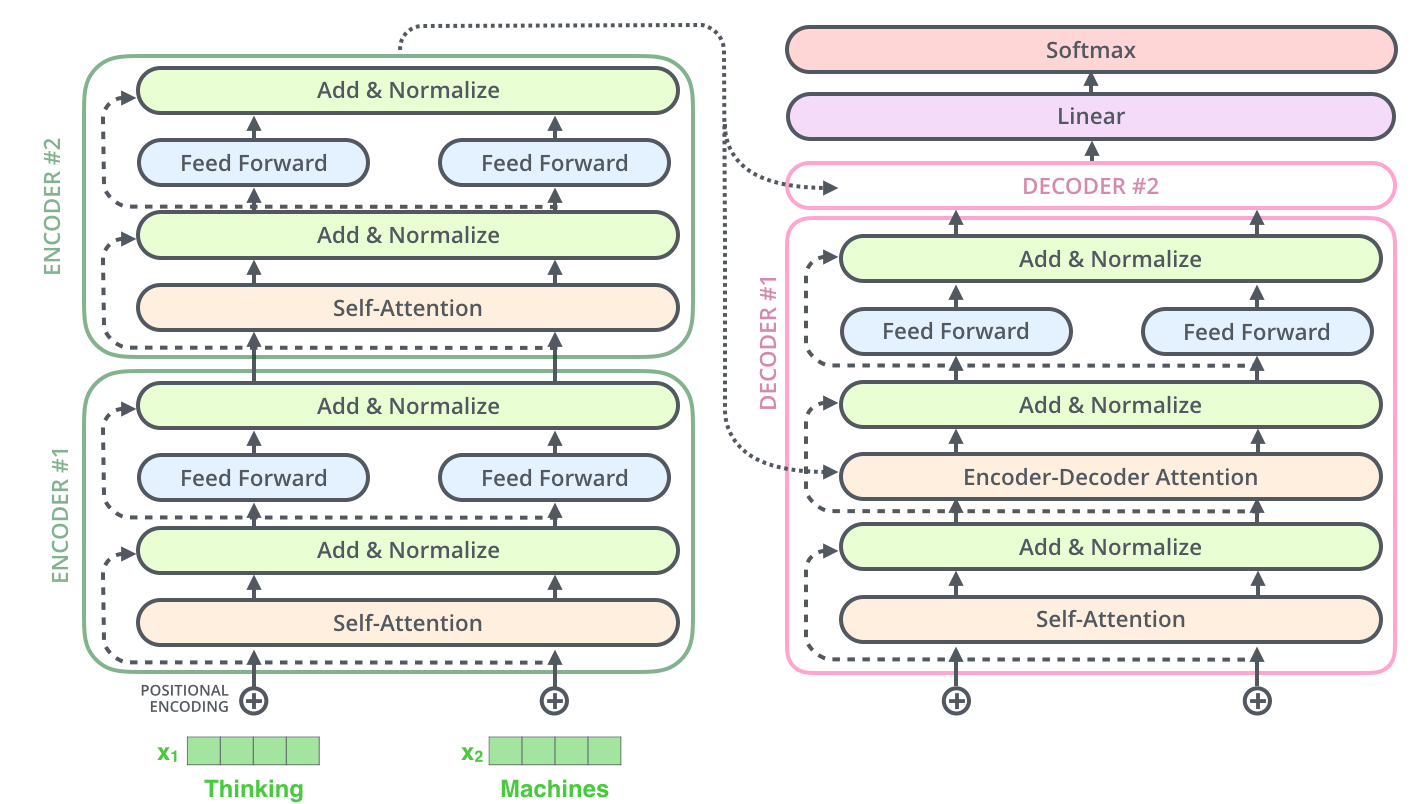}}
\caption{transformer encoding.~\cite{tranformer_understand}}
\label{fig-transformer_encoding}
\end{figure}

\begin{figure}[htbp]
\centering{\includegraphics[width=0.8\linewidth]{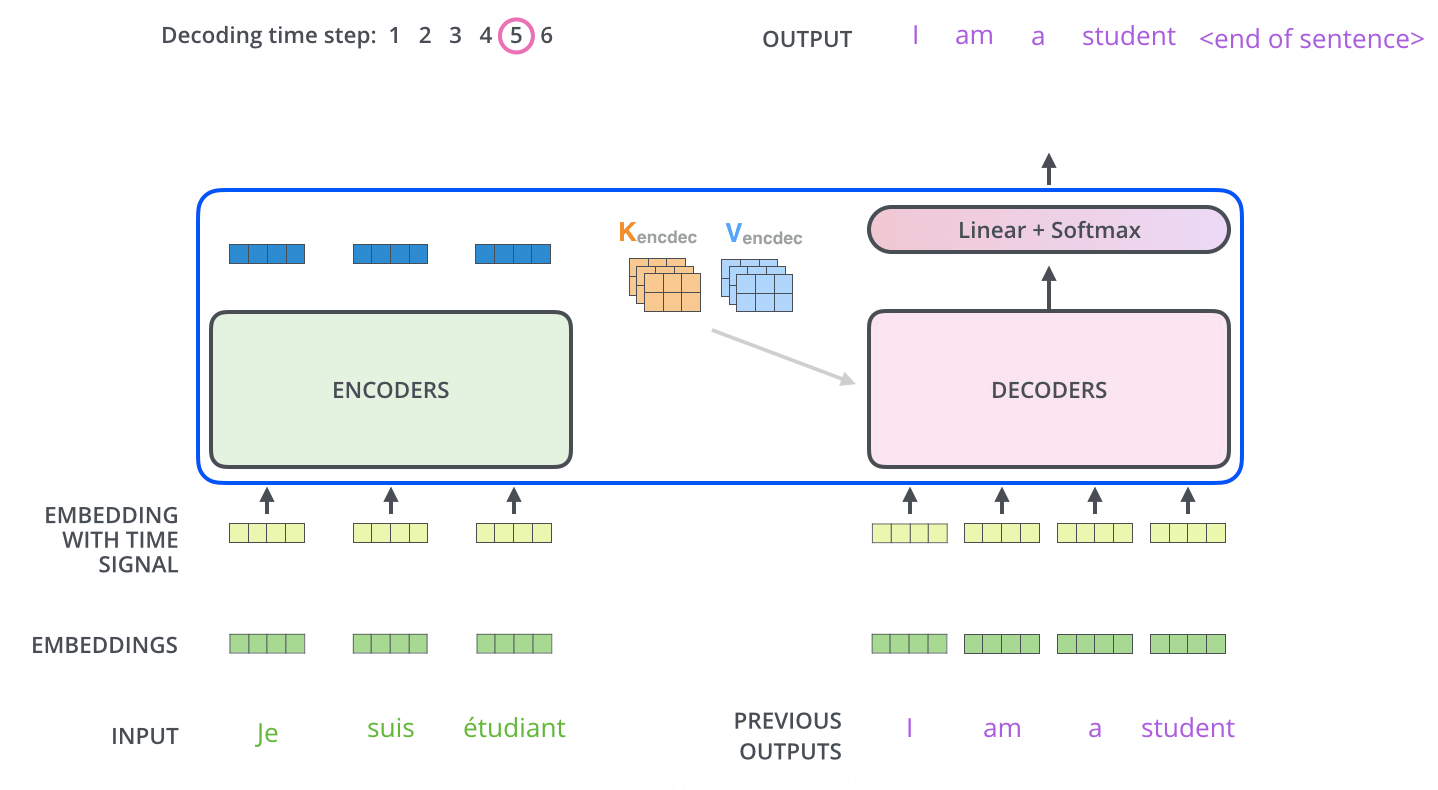}}
\caption{transformer decoding.~\cite{tranformer_understand}}
\label{fig-transformer_decoding}
\end{figure}

The Transformer consists of an encoder and a decoder,each with multiple identical layers.
The encoder processes the input text,and the decoder generates the output.
Each encoder layer has a self-attention layer and a feed-forward neural network (Fig.~\ref{fig-transformer_encoding}). The decoder layer adds a third sub-layer for attention over the encoder's output (Fig.~\ref{fig-transformer_decoding}).

\subsection{Positional Encoding }

\begin{figure}[htbp]
\centering{\includegraphics[width=0.8\linewidth]{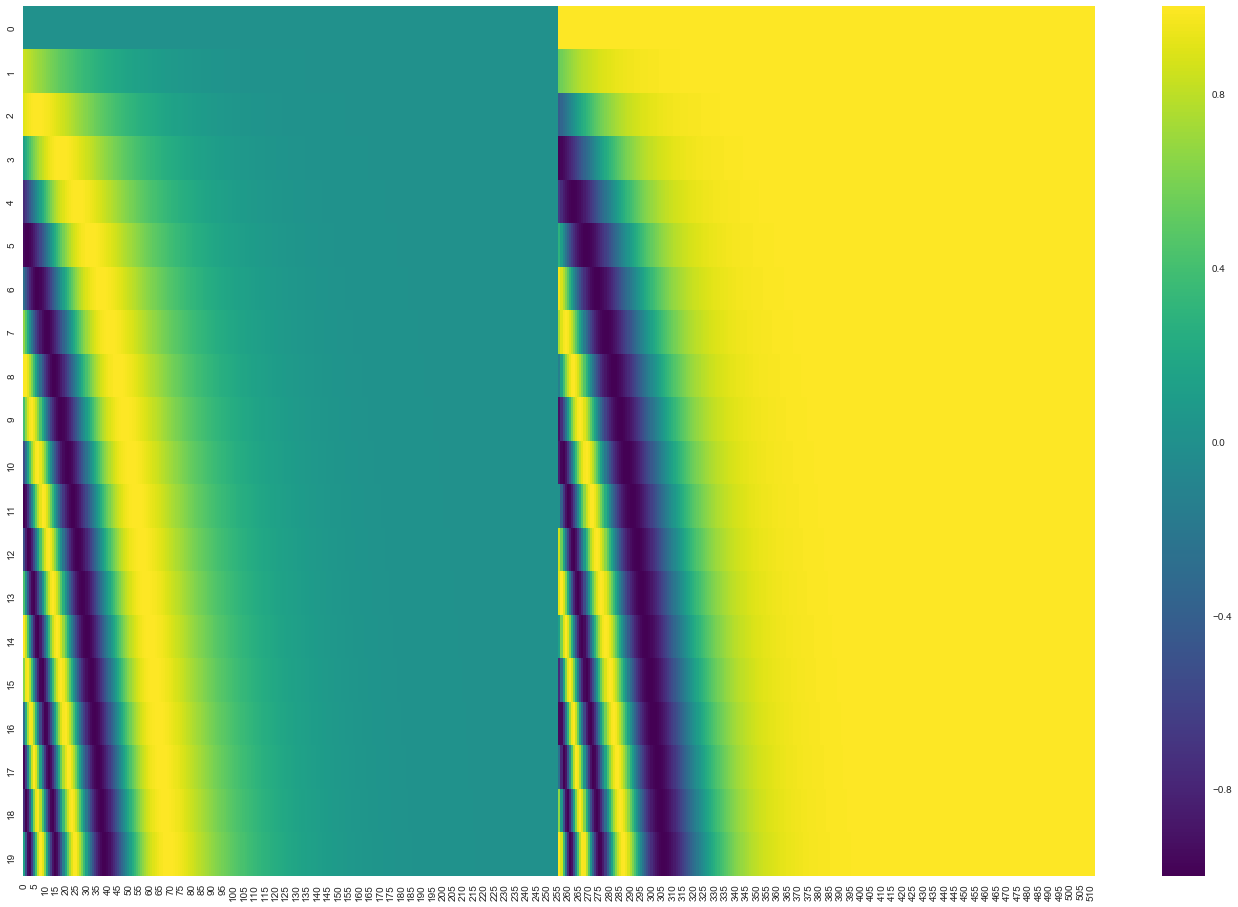}}
\caption{transformer positional encoding.~\cite{tranformer_understand}}
\label{fig-transformer-positional-encoding}
\end{figure}

Transformers lack inherent understanding of word order or position due to the absence of recurrent or convolutional elements.
Positional encodings are added to input embeddings to give positional information of words in the sequence (Fig.~\ref{fig-transformer-positional-encoding}).
\begin{equation}
\begin{aligned}
P E_{(p o s,2 i)} & =\sin \left(\text { pos } / 10000^{2 i / d_{\mathrm{model}}}\right) \\
P E_{(p o s,2 i+1)} & =\cos \left(\text { pos } / 10000^{2 i / d_{\mathrm{model}}}\right)
\end{aligned} \label{eq-position-encoding}
\end{equation}

\cite{su2023roformer} introduced a new encoding method,Rotary Position Encoding.In traditional Transformers,positional information is incorporated by adding position encodings to the word embeddings in a sequence,enabling the model to understand the position of words within a sentence. However,this method doesn't directly reflect the relative positions between words.
\begin{equation}
f_{\{q,k\}}\left(\boldsymbol{x}_m,m\right) =\boldsymbol{R}_{\Theta,m}^d \boldsymbol{W}_{\{q,k\}} \boldsymbol{x}_m 
\end{equation}
\textbf{RoFormer} improves upon this with rotary position encoding. Imagine each word attached to a rotating disk,where different positions on the disk represent different positional information. By rotating the disk,we can simulate the relative positional relationships between words. This method more naturally represents the relative distances and order between words,enhancing the model's ability to handle word order and contextual information.
\begin{equation}
\begin{small}
\boldsymbol{R}_{\Theta,m}^d =\left(
\begin{array}{ccccc}
\cos m \theta_1 & -\sin m \theta_1 & \cdots & 0 & 0 \\
\sin m \theta_1 & \cos m \theta_1 & \cdots & 0 & 0 \\
\vdots & \vdots & \ddots & \vdots & \vdots \\
0 & 0  &\cdots & \cos m \theta_{d / 2} & -\sin m \theta_{d / 2} \\
0 & 0  &\cdots & \sin m \theta_{d / 2} & \cos m \theta_{d / 2}
\end{array}\right) 
\end{small} \label{eq-rorormer}
\end{equation}\cite{tranformer_understand}
The core formula of RoFormer \eqref{eq-rorormer} involves calculating this rotary effect. Simplified,each word's representation is multiplied by a specific rotation matrix that varies based on the word's position. Thus,each word's representation includes its positional information.

\subsection{Multi-Head Attention }

The self-attention mechanism in Transformers employs``multi-head'' attention (Fig.~\ref{fig-Attention}),allowing the model to attend to information from different representation subspaces at different positions.
This helps the model to interpret text from various angles.
\begin{equation}
\begin{aligned}
\operatorname{MultiHead}(Q,K,V) & =\operatorname{Concat}\left(\operatorname{head}_1,\ldots,\operatorname{head}_{\mathrm{h}}\right) W^O \\
\text { where head } & =\operatorname{Attention}\left(Q W_i^Q,K W_i^K,V W_i^V\right)
\end{aligned} \label{eq-multi-head-attention}
\end{equation}

\subsection{Layer Normalization and Residual Connections }

\begin{figure}[htbp]
\centering{\includegraphics[width=0.8\linewidth]{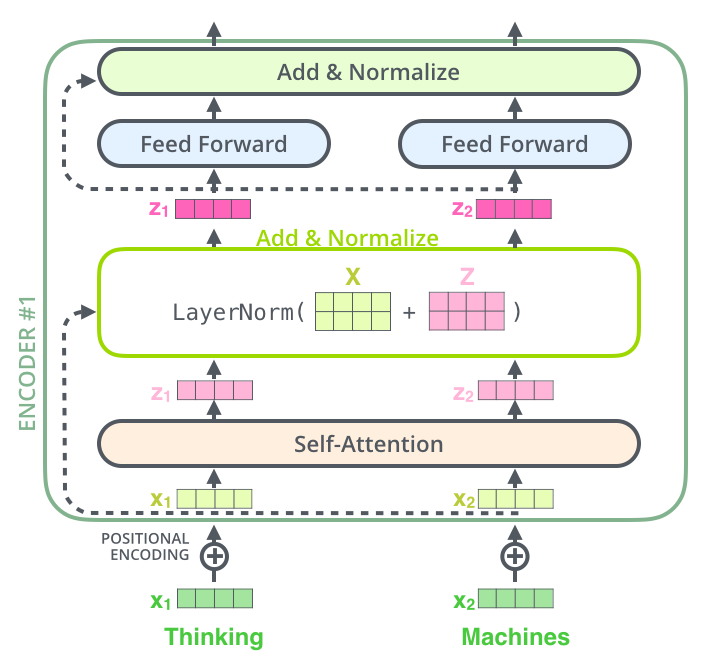}}
\caption{transformer layer normalization.~\cite{tranformer_understand}}
\label{fig-transformer-layer-normalization}
\end{figure}

Each sub-layer in the encoder and decoder is equipped with a residual connection,followed by layer normalization (Fig.~\ref{fig-transformer-layer-normalization}).
This aids in preventing the vanishing gradient problem and stabilizes training.

The Transformer architecture has greatly influenced machine learning,leading to advancements in creating more efficient and accurate models for processing language and other data types.

\section{LLMs Paradigm}

\subsection{LLMs family}

\begin{figure}[htbp]
\centering{\includegraphics[width=0.8\linewidth]{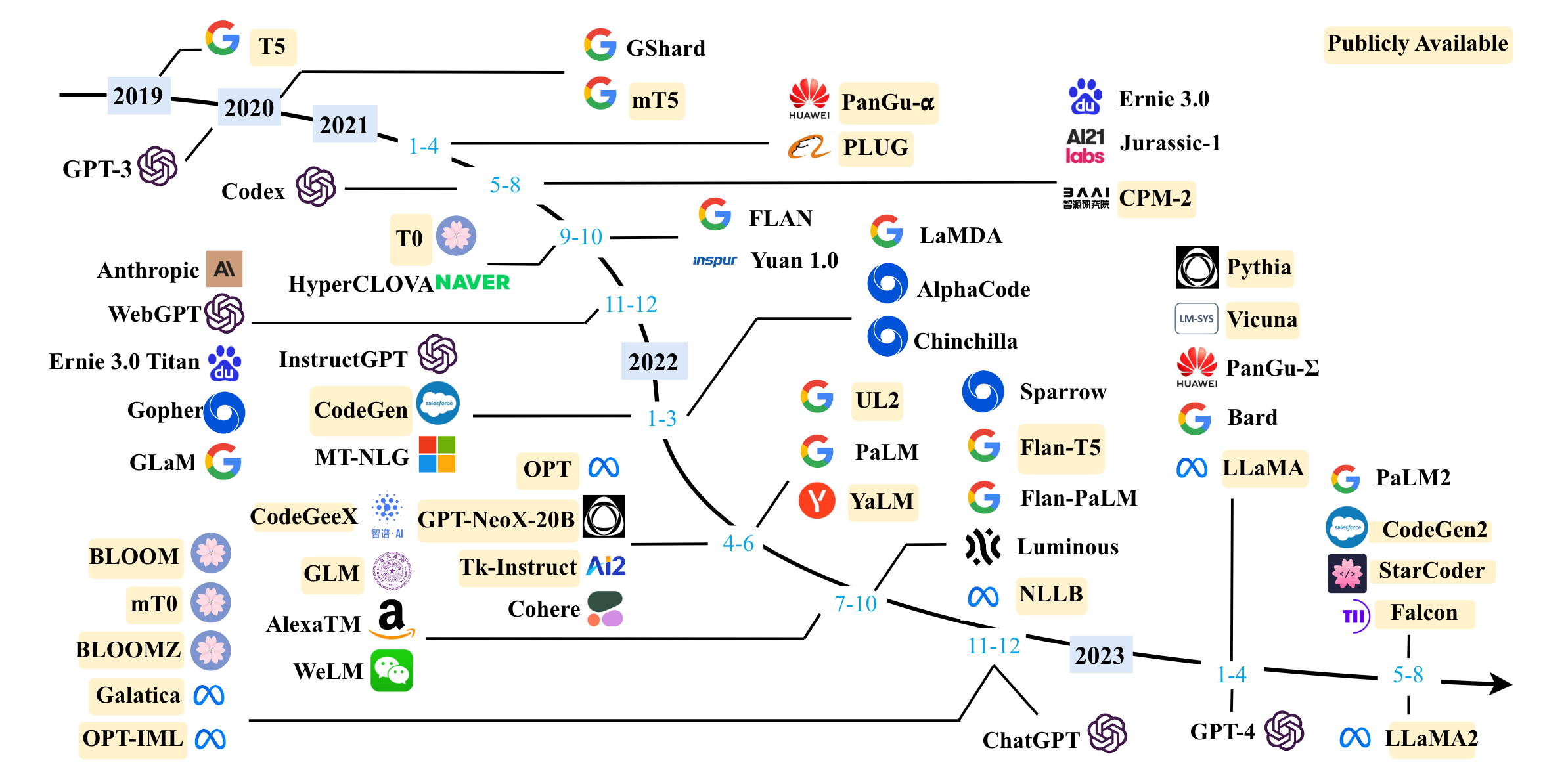}}
\caption{A timeline of existing large language models (having a size larger than 10B) in recent years.~\cite{zhaoSurveyLargeLanguage2023}}
\label{fig-llms-models}
\end{figure}

Since the introduction of the Transformer architecture in 2017~\cite{Vaswani2017AttentionIA},the development of large language models (LLMs) in the field of natural language processing (NLP) has been rapid and revolutionary~\cite{zhaoSurveyLargeLanguage2023}. These models have achieved significant success in text understanding and generation and have demonstrated immense potential in image and multimodal tasks. Below are some notable LLMs and their features:

\textbf{BERT~\cite{Devlin2019BERTPO}:} Launched by Google in 2018,it uses bidirectional Transformers to better understand context.

\textbf{GPT Series (Generative Pre-trained Transformer): } A series of models developed by OpenAI,including GPT~\cite{Radford2018ImprovingLU},GPT-2~\cite{Radford2019LanguageMAgpt2},GPT-3~\cite{gpt3},and GPT-4~\cite{OpenAI2023GPT4TR}.Each new version has increased in scale and complexity,focusing on generating coherent and informative text.

\begin{figure}[htbp]
\centering{\includegraphics[width=0.8\linewidth]{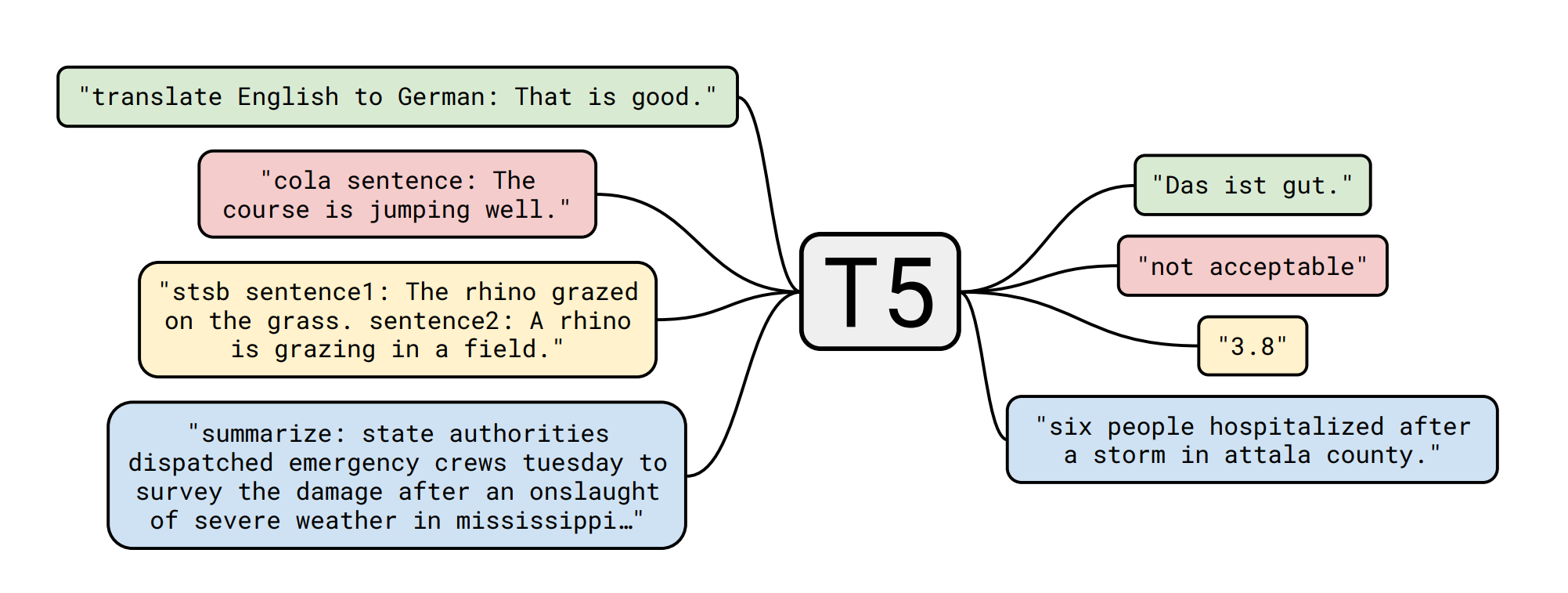}}
\caption{A diagram of our text-to-text framework.~\cite{T5}}
\label{fig-t5}
\end{figure}
\textbf{T5:} T5(Text-to-Text Transfer Transformer)~\cite{T5}(Fig.~\ref{fig-t5}) is a large-scale pretrained language model developed by the Google Brain team. T5 adopts the Transformer architecture and stands out for its unique approach of treating all natural language processing tasks as text-to-text problems.

The fundamental idea behind T5~\cite{T5} is to frame all natural language processing tasks as text-to-text tasks. Both input and output are represented as text strings,with the task type specified through textual prompts. For instance,in text classification,the input might be``classification:'' followed by a text segment,and the output corresponds to the relevant label for classification. This consistent input-output format makes T5 highly versatile,allowing it to adapt to a wide range of text processing tasks,including text classification,text generation,question answering,summarization,and more.

During its pretraining phase,T5~\cite{T5} leverages massive amounts of textual data from sources like the internet,books,news,and more to acquire extensive language knowledge. In downstream tasks,T5~\cite{T5} can be fine-tuned with task-specific prompts to adapt to particular applications. This uniform input-output format and large-scale pretraining make T5 a versatile tool for natural language processing tasks.

In comparison to some traditional models,T5~\cite{T5} has exhibited remarkable performance across multiple natural language processing tasks,drawing significant attention from both research and industry. Its versatility,high performance,and ease of use make it suitable for a wide range of applications,including automatic summarization,text translation,chatbots,and more. In summary,T5 represents a significant technological advancement in the field of natural language processing,providing a powerful tool for text data processing.

\textbf{BART:} BART (Bidirectional and Auto-Regressive Transformers)~\cite{lewis-etal-2020-bart} is a large-scale pretrained language model developed by Facebook AI. Similar to T5 (Text-to-Text Transfer Transformer),BART is based on the Transformer architecture,but it possesses unique features and applications.

Firstly,BART~\cite{lewis-etal-2020-bart} is equipped with bidirectional encoding and auto-regressive decoding capabilities,allowing it to handle bidirectional textual information simultaneously,making it suitable for a wide range of natural language processing tasks. BART adopts a masked language modeling approach during pretraining,enabling the model to consider both left and right context in input text,facilitating a better understanding and generation of text.

BART~\cite{lewis-etal-2020-bart} finds applications across various domains,excelling in tasks such as text generation,text summarization,text translation,and text compression. For instance,it can be used to generate high-quality article summaries,perform automatic text translation,or compress lengthy text for specific purposes. Its auto-regressive decoding strategy makes it particularly effective in text generation tasks,especially when context comprehension and long-text generation are required.

In comparison to T5~\cite{T5},one key distinction of BART lies in its bidirectional encoding capability. While T5~\cite{T5} predominantly employs auto-regressive decoding,BART~\cite{lewis-etal-2020-bart} combines bidirectional encoding during pretraining,potentially making it more proficient at handling context information and generating coherent text in certain tasks. Additionally,BART boasts a broader range of applications,including text summarization and text compression tasks,while T5 primarily focuses on text-to-text transformation tasks.

\textbf{ChatGPT~\ref{section:chatgpt},InstructGPT~\cite{Ouyang2022InstructGPT}:} A variant of GPT-3~\cite{gpt3},optimized specifically for producing fluid dialogue.The introduction of InstructGPT~\cite{Ouyang2022InstructGPT},with its alignment fine-tuning using reinforcement learning,paved the way for the development of ChatGPT~\ref{section:chatgpt},further enhancing the model's ability to understand and generate natural language conversations. InstructGPT~\cite{Ouyang2022InstructGPT},by learning from human feedback,optimized the way the model responds,ensuring that its output is not only smooth and informative but also closely aligned with the user's instructions and expectations.

The uniqueness of ChatGPT~\ref{section:chatgpt} lies in its ability to generate coherent and relevant dialogues,understand complex queries,and provide detailed and accurate answers. Its optimization extends beyond improving the quality of language generation to enhancing the understanding of user inputs,making conversations more natural and human-like.

Moreover,ChatGPT\ref{section:chatgpt} combines various technologies,including but not limited to knowledge retrieval,sentiment recognition,and contextual understanding,enabling it to be applied in multiple domains,from simple Q\&A to complex problem-solving. This versatility and adaptability make ChatGPT~\ref{section:chatgpt} a powerful tool,suitable not only for customer service and personal assistant domains but also in industries like education,healthcare,and entertainment.

As technology continues to advance,we expect ChatGPT ~\ref{section:chatgpt} and similar large language models to evolve further,incorporating more advanced features and greater efficiency,while also improving in maintaining the naturalness and fluidity of dialogues. The advancements in these models will further propel the application of artificial intelligence in our daily lives,making communication between machines and humans more seamless and efficient
\footnote{chatgpt:\label{section:chatgpt}\url{https://chat.openai.com/}}.

\textbf{LLaMA,LLaMA2:} LLaMA~\cite{Touvron2023LLaMA} focused on language understanding and generation,emphasizing performance and scalability.The LLAMA model,emphasizing language understanding and generation,set the stage for its successor,LLAMA2~\cite{Touvron2023Llama2}. Building on the foundation laid by the original LLAMA~\cite{Touvron2023LLaMA},LLAMA2~\cite{Touvron2023Llama2} extends its capabilities with enhanced performance and scalability,making it an even more powerful tool in the realm of NLP.

LLAMA2~\cite{Touvron2023Llama2} inherits the strengths of its predecessor,particularly in its ability to understand and generate natural language. However,it takes these abilities to new heights with improved algorithms and a more extensive training dataset. This results in a model that not only understands the nuances of human language more effectively but can also generate more coherent,contextually relevant,and diverse responses.

One of the key advancements in LLAMA2~\cite{Touvron2023Llama2} is its enhanced scalability. The model is designed to handle a larger variety of tasks and datasets,making it adaptable to a broader range of applications. This scalability is achieved without sacrificing performance,ensuring that LLAMA2~\cite{Touvron2023Llama2} remains efficient even as it tackles more complex and diverse linguistic challenges.

In addition to scalability,LLAMA2~\cite{Touvron2023Llama2} focuses on performance optimization. This involves fine-tuning the model to reduce latency and increase processing speed,making it suitable for real-time applications. The model also features improved accuracy in language understanding and generation,which is critical for applications requiring high levels of reliability and precision.

LLAMA2's~\cite{Touvron2023Llama2} advancements in language understanding and generation,combined with its scalability and performance optimization,position it as a leading model in NLP. Its ability to adapt to various tasks and its efficiency in processing make it a valuable asset for researchers and practitioners alike,paving the way for more innovative applications in artificial intelligence and machine learning.

\textbf{Codex:} Codex~\cite{Chen2021-CodeX} Developed by OpenAI,Codex~\cite{Chen2021-CodeX} is an advanced language model developed by OpenAI,specifically designed for understanding and generating programming code. Based on the GPT-3~\cite{gpt3} model,Codex~\cite{Chen2021-CodeX} can automatically generate code snippets,comprehend natural language instructions,and translate them into effective programming code. It supports multiple programming languages and is suitable for rapid programming,code correction,and as a tool for programming education. Codex~\cite{Chen2021-CodeX} demonstrates the potential of artificial intelligence in the fields of software development and programming education.

\textbf{WebGPT:} WebGPT~\cite{Nakano2021WebGPTBQ} is a language model developed by OpenAI that combines the functionalities of traditional large language models with the ability to perform web searches. This model can execute internet searches to gather information,thereby answering questions or completing specific tasks. The key feature of WebGPT~\cite{Nakano2021WebGPTBQ} is its capability to utilize up-to-date information from the internet to provide more accurate and comprehensive responses. This integration of language modeling with web search capabilities makes WebGPT~\cite{Nakano2021WebGPTBQ} potentially powerful in a variety of applications,such as question-answering systems,market research,and data analysis.

\textbf{GLaM:}  GLaM~\cite{Du2021GLaMES} Developed by Google,GLaM is a large multi-task language model.It utilizes a dense Transformer architecture capable of handling a variety of different language processing tasks.The design of GLaM aims to enhance the model's generalization capabilities and processing efficiency,enabling it to better understand and generate natural language.

\textbf{BLOOM:}  BLOOM~\cite{Scao2022BLOOMA1} is an open-source multilingual large language model developed by the BigScience research team.
It focuses on multilingual capabilities,aiming to provide broader language coverage,including languages typically underrepresented in large language models.
BLOOM's~\cite{Scao2022BLOOMA1} open-science approach allows the research community better access to and utilization of the model.

\textbf{GLM:}  GLM~\cite{Du2021GLMGL}~\cite{Zeng2022GLM130BAO} refers to a general language model designed for various natural language processing tasks.It typically denotes large models trained for specific tasks,excelling in understanding and generating natural language.The precise features and applications of GLM may vary depending on the specific implementation and research objectives.

\textbf{GPT-NeoX-20B~\cite{Black2022GPTNeoX20BAO} :} GPT-NeoX-20B~\cite{Black2022GPTNeoX20BAO} is a large language model developed by the EleutherAI team,serving as an open-source alternative to GPT-3. The model,with approximately 20 billion parameters,is a deep learning model based on the Transformer architecture,focusing on text generation and language understanding.

Key features of GPT-NeoX-20B include:
\begin{itemize}
    \item \textbf{Large Scale Parameters:} With 20 billion parameters,it exhibits high precision and flexibility in handling complex language tasks.
    \item \textbf{Open-source Accessibility:} Unlike GPT-3,GPT-NeoX-20B,as an open-source model,provides broader access and usage possibilities for researchers and developers.
    \item \textbf{Text Generation and Understanding:} The model excels in natural language generation and understanding tasks,capable of creating coherent and realistic text.
    \item \textbf{Suitable for Various Applications:} GPT-NeoX-20B effectively supports tasks ranging from content creation to natural language understanding.
\end{itemize}

\textbf{OPT:} OPT~\cite{Zhang2022OPTOP} (Open Pre-trained Transformer) is a large-scale open-source language model developed by Meta AI. This model is designed to provide functionalities similar to GPT-3~\cite{gpt3},but in an open-source format,making it accessible to a broader community. Available in various sizes,from millions to billions of parameters,OPT~\cite{Zhang2022OPTOP} is suitable for a wide range of natural language processing tasks,including text understanding and generation. As an open-source project,OPT promotes innovation and application of technology.

\textbf{mT5:} mT5~\cite{Xue2020mT5AM} is a large-scale multilingual pre-trained Transformer model developed by Google AI. It is the multilingual version of the T5 model (Text-to-Text Transfer Transformer),designed to handle natural language processing tasks across multiple languages. mT5~\cite{Xue2020mT5AM} has been pre-trained on a vast amount of multilingual datasets,enabling it to understand and generate text in over 100 different languages.

\textbf{GShard:} GShard~\cite{Lepikhin2020GShardSG} is a large-scale model training framework developed by Google,designed to facilitate the efficient training of massive neural network models through distributed training techniques. The key feature of GShard~\cite{Lepikhin2020GShardSG} is its ability to process huge models,such as multilingual translation models and large language models,distributively across hundreds or thousands of processors.

\begin{figure}[htbp]
\centering{\includegraphics[width=0.8\linewidth]{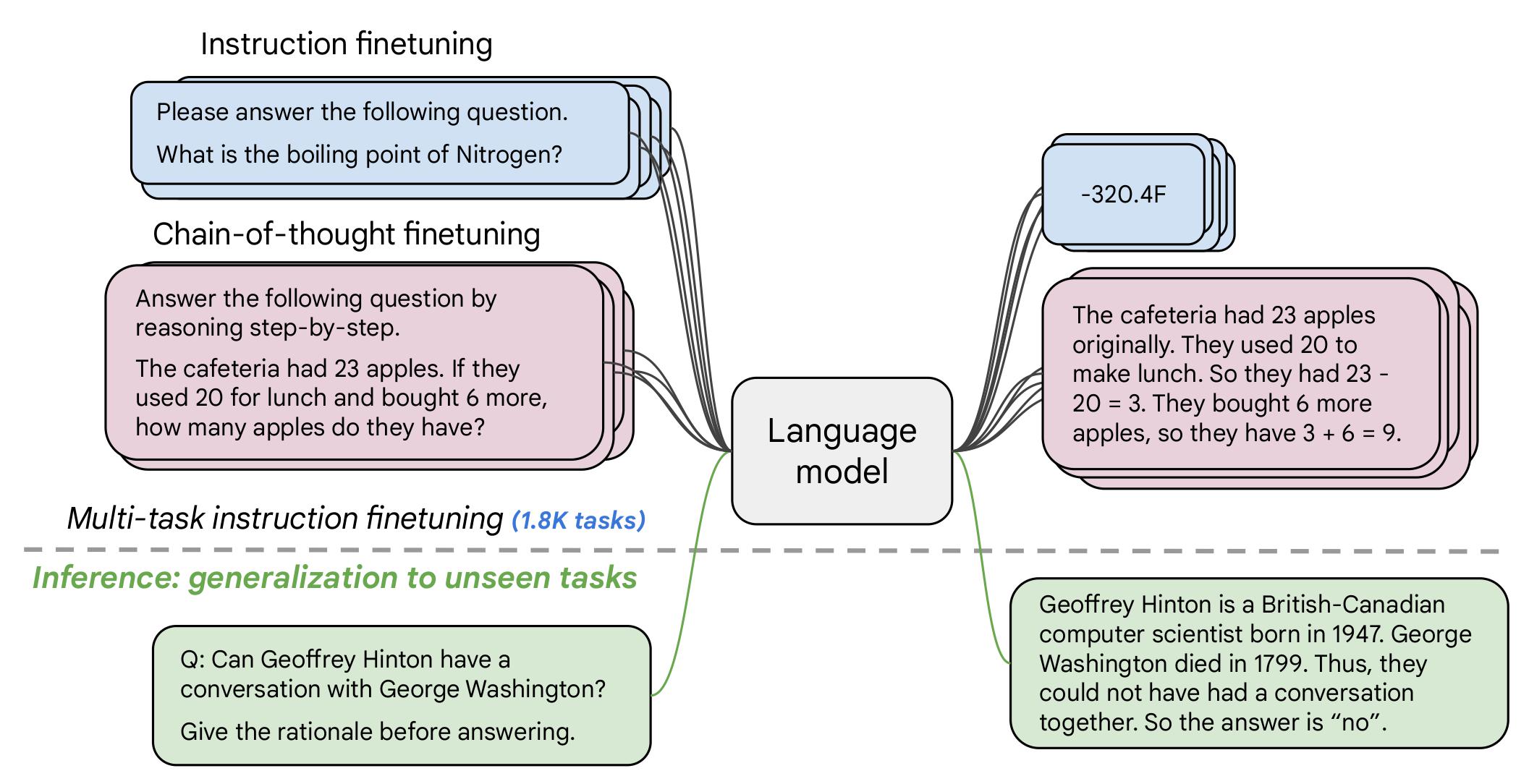}}
\caption{A diagram of flan2.~\cite{Chung2022ScalingILFLAN}}
\label{fig-flan2}
\end{figure}
\textbf{FLAN} FLAN~\cite{Chung2022ScalingILFLAN}(Fig.~\ref{fig-flan2}) models are based on the principle of fine-tuning language models on a collection of datasets phrased as instructions. This approach has shown to improve model performance and generalization to unseen tasks.
The focus of FLAN~\cite{Chung2022ScalingILFLAN} is on scaling the number of tasks,scaling the model size,and fine-tuning on chain-of-thought data.
The models include Flan-PaLM~\cite{Chung2022ScalingILFLAN} and Flan-T5~\cite{Chung2022ScalingILFLAN},with Flan-PaLM~\cite{Chung2022ScalingILFLAN} 540B demonstrating state-of-the-art performance on several benchmarks,such as achieving 75.2\% on five-shot MMLU.
Flan-T5~\cite{Chung2022ScalingILFLAN}(Fig.~\ref{fig-flan2}) is specifically noted for its strong few-shot performance,even compared to much larger models like PaLM 62B.
Instruction fine-tuning,as used in FLAN~\cite{Chung2022ScalingILFLAN},is a general method for improving the performance and usability of pre-trained language models.

\textbf{Yuan:} Yuan~\cite{wu2021yuan} is a large-scale multimodal pre-trained model developed by Baidu. This model aims to combine the processing capabilities of text,images,and audio to achieve more comprehensive and efficient natural language understanding and generation. The Yuan model utilizes the latest deep learning technologies and algorithms,with a particular emphasis on innovation in multimodal data processing.

Key features of Yuan include: 

\textbf{1. Multimodal Integration:} Yuan is capable of processing text,images and audio simultaneously,excelling in multimodal tasks such as image description,speech-to-text conversion,and more.

\textbf{2. Large-scale Pre-training:} The model has been pre-trained on a vast array of multimodal datasets,including text,images,and audio,enabling it to learn rich patterns and representations.

\textbf{3. Cross-modal Understanding and Generation:} Yuan can not only understand data from a single modality but also interact and integrate information across different modalities,providing more comprehensive responses and generation capabilities.

\textbf{4. Applicability to Various Tasks:} The model can be used for a wide range of NLP and multimodal tasks,such as automatic text generation,speech recognition,image recognition,and generation.

As an advanced multimodal model,Yuan~\cite{wu2021yuan} demonstrates great potential in handling complex interactive tasks,especially in scenarios that involve the combination of text,images,and audio. With ongoing technological advancements,Yuan is likely to play a significant role in fields such as multimodal interaction,automated content creation,and intelligent assistants.

\textbf{LaMDA:} LaMDA~\cite{Thoppilan2022LaMDALM} (Language Model for Dialogue Applications) is a large language model developed by Google,specifically designed for dialogue applications. This model aims to generate more natural and meaningful conversations,regardless of the breadth or complexity of the topics.

\textbf{PaLM,PaLM2:} The PaLM~\cite{Chowdhery2022PaLMSL}~\cite{Anil2023PaLM2T} model,developed by Google,is a large-scale language model. This model was trained with a 540-billion parameter,densely activated setup,demonstrating exceptional performance across a variety of natural language tasks using few-shot learning.

\textbf{Sparrow:} Sparrow\cite{Glaese2022Sparrow} is a conversational system developed by DeepMind,based on a large language model. It can engage in natural,helpful,and safe conversations with users. It uses reinforcement learning and human feedback to improve its behavior,avoiding inaccurate,inappropriate,or harmful responses.It can leverage Google search on the internet to integrate the latest and relevant information into its responses,enhancing its reliability and usefulness.

It adheres to a set of conversational guidelines to ensure its responses align with ethical,legal,and societal standards,preventing misleading,harmful,or offensive content.

\textbf{NLLB:} NLLB\cite{Koishekenov2022MemoryefficientNLLB},developed by Meta (formerly Facebook) AI,is a large-scale machine translation model. Its full name is``No Language Left Behind'' reflecting its mission to ensure that no language is forgotten. 
It can translate over 200 different languages,including many African and other minority languages that are not supported by most current translation tools,covering over 90\% of the global population.
It is a single neural network model with a massive parameter size of 545 billion,enabling automatic translation between all language pairs,resulting in 20,400,402 translation directions.
It employs a hybrid neural network architecture that combines dense and sparse neural networks,utilizing sparse attention mechanisms to enhance the model's efficiency and scalability.
It was trained on 18 billion parallel sentence pairs covering 202 languages and 2,440 language directions. The post-training model exhibits an average quality improvement of 44\% over existing translation tools on multiple metrics. In some African and Indian languages,the improvement is even higher,surpassing the latest translation systems by up to 70\%.

Its code and models have been open-sourced on GitHub\footnote{Code For NLLB:\label{section:NLLB-code}\url{https://github.com/facebookresearch/fairseq/tree/nllb/}},making them available for free use by anyone. It also collaborates with the Wikimedia Foundation to improve the translation system of Wikipedia.

\textbf{CPM-2:} CPM-2~\cite{Zhang2021CPM2LC} is a large-scale and efficient pretrained language model developed by the team led by Zhiyuan Liu at Tsinghua University. 

It employs knowledge inheritance techniques,leveraging the knowledge from existing pretrained models to expedite the training process of new models,thereby conserving computational resources and time.CPM-2~\cite{Zhang2021CPM2LC} explores Prompt fine-tuning methods,reducing task-specific parameters by designing suitable prompts to enhance the model's generalization ability and efficiency.It implements the InfMoE \footnote{InfMoE:\label{section:InfMoE}\url{https://github.com/TsinghuaAI/InfMoE}} (Inference Mixture of Experts) tool,enabling the inference of models with hundreds of billions of parameters on a single GPU,lowering the barrier to using such large models.CPM-2~\cite{Zhang2021CPM2LC} includes models with 110 billion parameters for Chinese,110 billion parameters for Chinese-English,and a massive 1.98 trillion parameters for Chinese-English MoE (Mixture of Experts),offering comprehensive language understanding and generation capabilities for both Chinese and English.

Across seven major machine language tasks,including recognition,reading,classification,reasoning,cross-lingual understanding,generation,and summarization,CPM-2 outperforms existing open-source pretrained models significantly in terms of overall performance.CPM-2 represents a notable advancement in the field of natural language processing,particularly in terms of efficiency,knowledge transfer,and superior performance across a range of language tasks.

\textbf{Ernie 3.0:} Ernie 3.0~\cite{Sun2021ERNIE3L} is a large pretrained language model developed by Baidu AI Open Platform. It is built upon a knowledge-enhanced multi-paradigm unified pretraining framework,combining both autoregressive and autoencoding networks,enabling it to handle various natural language understanding and generation tasks. 
It leverages a variety of pretraining tasks,including entity prediction,causal relationship judgment between sentences,article sentence structure reconstruction,and knowledge-enhanced pretraining tasks using knowledge graph data to enhance the model's semantic understanding capabilities.
It adopts the Transformer-XL~\cite{dai-etal-2019-transformer} architecture,enabling it to model long text sequences and process input sequences that exceed 512 tokens.

Ernie 3.0~\cite{Sun2021ERNIE3L} achieves state-of-the-art results across a wide range of tasks,including 45 natural language understanding datasets across 14 types,9 natural language generation tasks,18 zero-shot learning tasks,and the SuperGLUE benchmark. This demonstrates its powerful generalization capabilities and robustness.
Ernie 3.0~\cite{Sun2021ERNIE3L} represents a significant advancement in pretrained language models,offering versatility and superior performance across a spectrum of natural language processing tasks.

\textbf{Vicuna:} Vicuna is an open-source large language model developed collaboratively by researchers from institutions such as UC Berkeley,CMU (Carnegie Mellon University),Stanford,and others. It comes in two versions,one with 7 billion parameters and another with 13 billion parameters. Vicuna is fine-tuned on user-generated conversational data collected from the ShareGPT website\footnote{ShareGPT:\label{section:ShareGPT}\url{https://sharegpt.com/}} and aims to create a chatbot capable of engaging in natural,fluent,and enjoyable multi-turn conversations with humans.

In terms of performance,Vicuna achieves over 90\% on certain evaluation metrics,surpassing both ChatGPT and Bard in more than 90\% of cases. It outperforms other open-source models like Alpaca\footnote{Alpaca:\label{section:Alpaca}\url{https://github.com/tatsu-lab/stanford_alpaca}} and Koala\footnote{Koala:\label{section:Koala}\url{https://bair.berkeley.edu/blog/2023/04/03/koala/}} in a majority of scenarios. Additionally,the training cost for Vicuna is relatively low,with the 7B model costing approximately \$140 and the 13B model costing around \$300.

The code and models for Vicuna are open-sourced on GitHub  \footnote{Code For Vicuna:\label{section:Vicuna-code}\url{https://github.com/tatsu-lab/stanford_alpaca}},\footnote{Code For Chinese Vicuna:\label{section:Chinese-Vicuna-code}\url{https://github.com/Facico/Chinese-Vicuna}},making them available for free use by anyone.

\textbf{CodeGen2:} CodeGen2~\cite{Nijkamp2023CodeGen2LF} is a large language model developed by Salesforce for program synthesis,which involves generating computer programs based on natural language descriptions provided by users. It is the second generation of the CodeGen model and comes in four versions with parameters ranging from 1 billion to 16 billion. 
It can perform infilling,meaning it can fill in missing parts of code,enabling more flexible program synthesis.
It supports a wider range of programming languages,including Python,Java,C,C++,Go,and JavaScript,allowing for code conversion and migration between different languages.

It is built upon the evaluation framework of GPT-4,utilizing GPT-4 as an evaluator to assess the performance of chatbots by designing diverse and challenging questions.

Its code and models are open-sourced on GitHub \footnote{Code For CodeGen2:\label{section:CodeGen2-code}\url{https://github.com/salesforce/CodeGen2}},making them freely available for anyone to use.
CodeGen2 represents a significant advancement in the field of program synthesis,offering increased flexibility,language support,and a robust evaluation framework.

\textbf{Falcon:} Falcon~\cite{Penedo2023TheRDFalcon} is a large language model developed and open-sourced by the Technology Innovation Institute (TII)\footnote{TII:\label{section:tii}\url{https://www.tii.ae/}} in Abu Dhabi. It comes in multiple versions,including models with 1.3 billion,7.5 billion,40 billion,and 180 billion parameters. 
It was trained on a high-quality dataset consisting of nearly 5 trillion tokens,extracted from the public web,and subjected to filtering and deduplication processes.

The Falcon model consists of two primary models: Falcon-40B\footnote{Huggingface falcon-40b:\label{section:falcon-40b}\url{https://huggingface.co/tiiuae/falcon-40b}} and its smaller counterpart,Falcon-7B\footnote{Huggingface falcon-7b:\label{section:falcon-7b}\url{https://huggingface.co/tiiuae/falcon-7b}}. The Falcon-40B,with its 40 billion parameters,currently holds the highest position on the Open LLM Leaderboard\footnote{huggingface open llm leaderboard:\label{section:open_llm_leaderboard}\url{https://huggingface.co/spaces/HuggingFaceH4/open_llm_leaderboard}} rankings. In contrast,the Falcon-7B model,with 7 billion parameters,leads its respective weight class and performance category.This is the Falcon-Chat demo\footnote{Falcon-Chat Demo:\label{section:falcon-chat-demo}\url{https://huggingface.co/spaces/HuggingFaceH4/falcon-chat}} with falcon-40b.

\begin{figure}[htbp]
\centering{\includegraphics[width=0.8\linewidth]{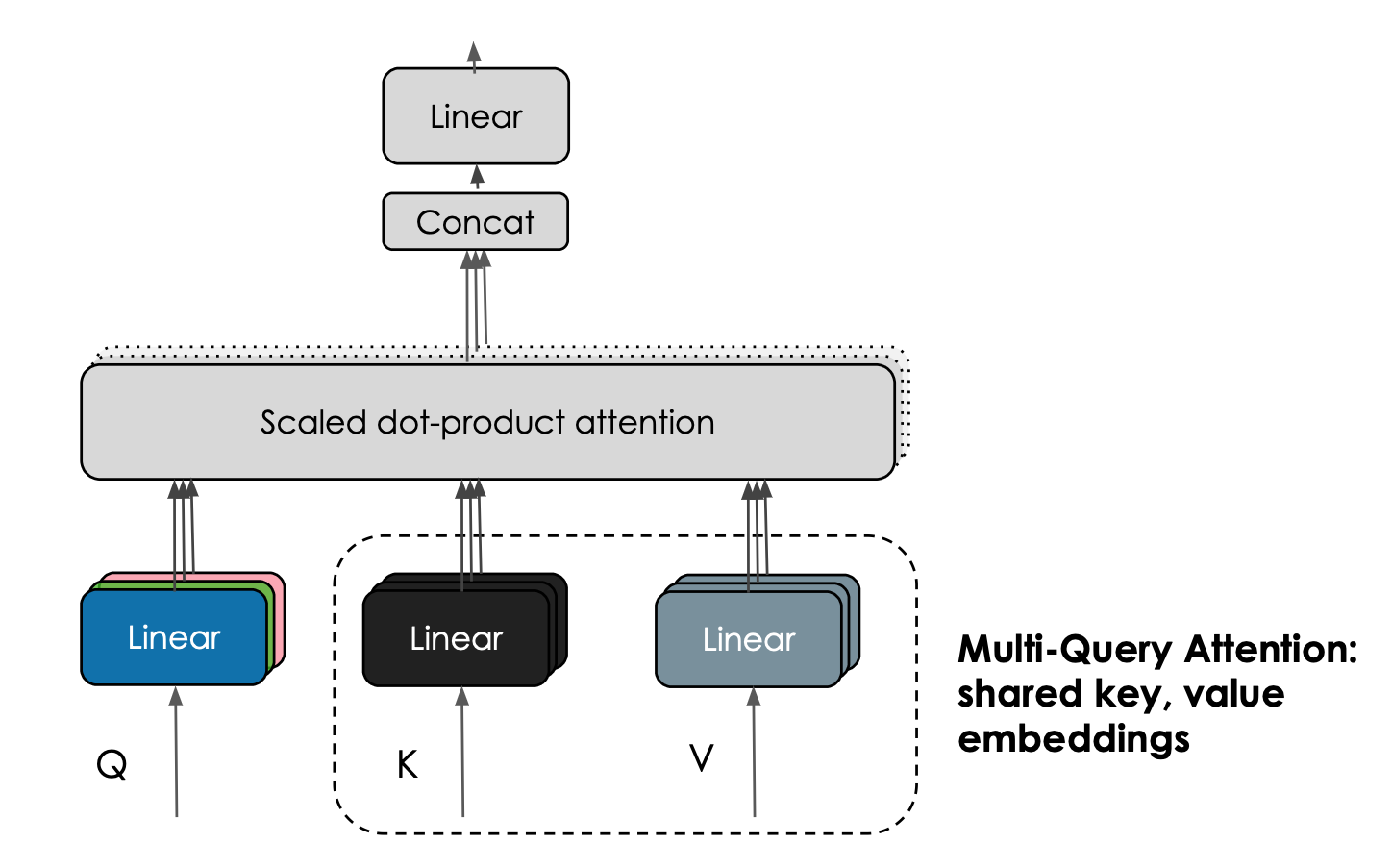}}
\caption{multi query attention.~\cite{FalcoEcosystem}}
\label{fig-multi-query-attention}
\end{figure}

Falcon models utilize a multiquery attention(Fig.~\ref{fig-multi-query-attention}) mechanism~\cite{Shazeer2019FastMultiQueryAttention},which differs from the traditional multihead attention scheme. In a typical multihead attention setup,each head has its own query,key,and value. However,in the case of multiquery attention employed by Falcon models,a single key and value are shared across all heads.
Falcon utilizes a custom distributed training library called Gigatron~\cite{Miao2022GalvatronET},which employs 3D parallelism strategies,ZeRO,Triton,and other technologies to enhance training efficiency and scalability.
It is a autoregressive decoder model based on the Transformer architecture,making it suitable for various natural language processing tasks,such as text generation,summarization,question-answering,dialogue,and machine translation.

Falcon is fully open-source and commercially usable,with models available for free download and usage on the Hugging Face platform. Additionally,it provides a dialogue model called Falcon-180B-Chat\footnote{Falcon-180B-Chat:\label{section:Falcon-180B-Chat}\url{https://huggingface.co/tiiuae/falcon-180B-chat}},capable of engaging in natural,enjoyable,and safe conversations with users.

\begin{figure}[htbp]
\centering{\includegraphics[width=0.8\linewidth]{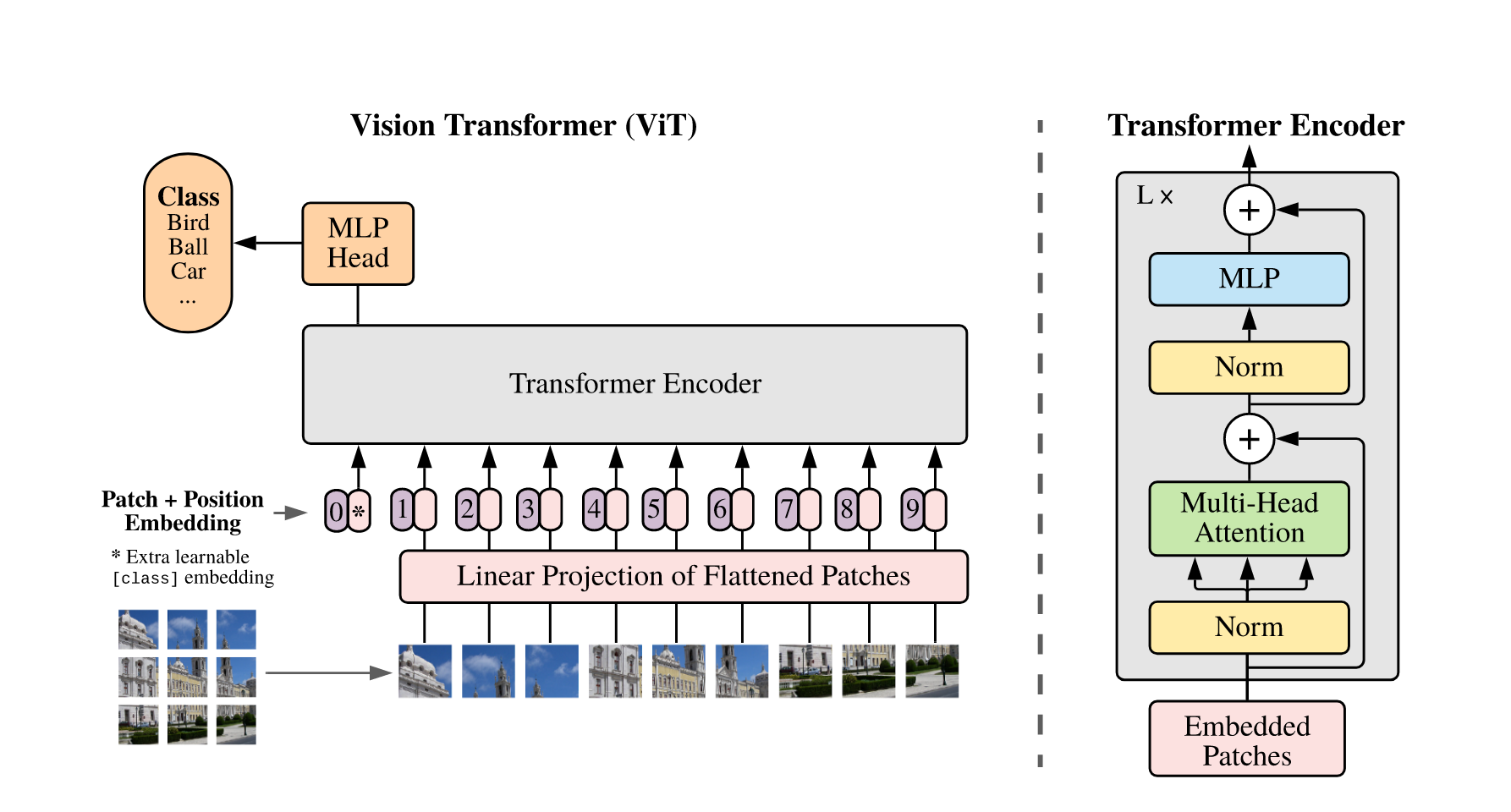}}
\caption{VIT Model overview.~\cite{Dosovitskiy2020VIT}}
\label{fig-vit-model}
\end{figure}

\textbf{VIT:} ViT~\cite{Dosovitskiy2020VIT}(Fig.~\ref{fig-vit-model}),short for Vision Transformer,is a model proposed by the Google team for applying the Transformer architecture to image classification. 
It divides the input image into multiple small patches,then projects each patch into a vector,which is then fed into the Transformer encoder. Finally,the output of a special token is used as the representation of the image for classification or other tasks.

ViT~\cite{Dosovitskiy2020VIT} does not use any convolutional layers or traditional computer vision operations. Instead,it relies entirely on the Transformer's self-attention mechanism to capture both global and local information within the images.
It exhibits strong scalability,meaning that larger models tend to perform better. ViT~\cite{Dosovitskiy2020VIT} can be pretrained on large-scale datasets and then fine-tuned or zero-shot transferred to various visual tasks,achieving or surpassing the current state-of-the-art performance.

ViT~\cite{Dosovitskiy2020VIT} represents a novel approach to computer vision by leveraging the power of the Transformer architecture,which was originally developed for natural language processing,and adapting it to the domain of image classification and understanding.

\begin{figure}[htbp]
\centering{\includegraphics[width=0.8\linewidth]{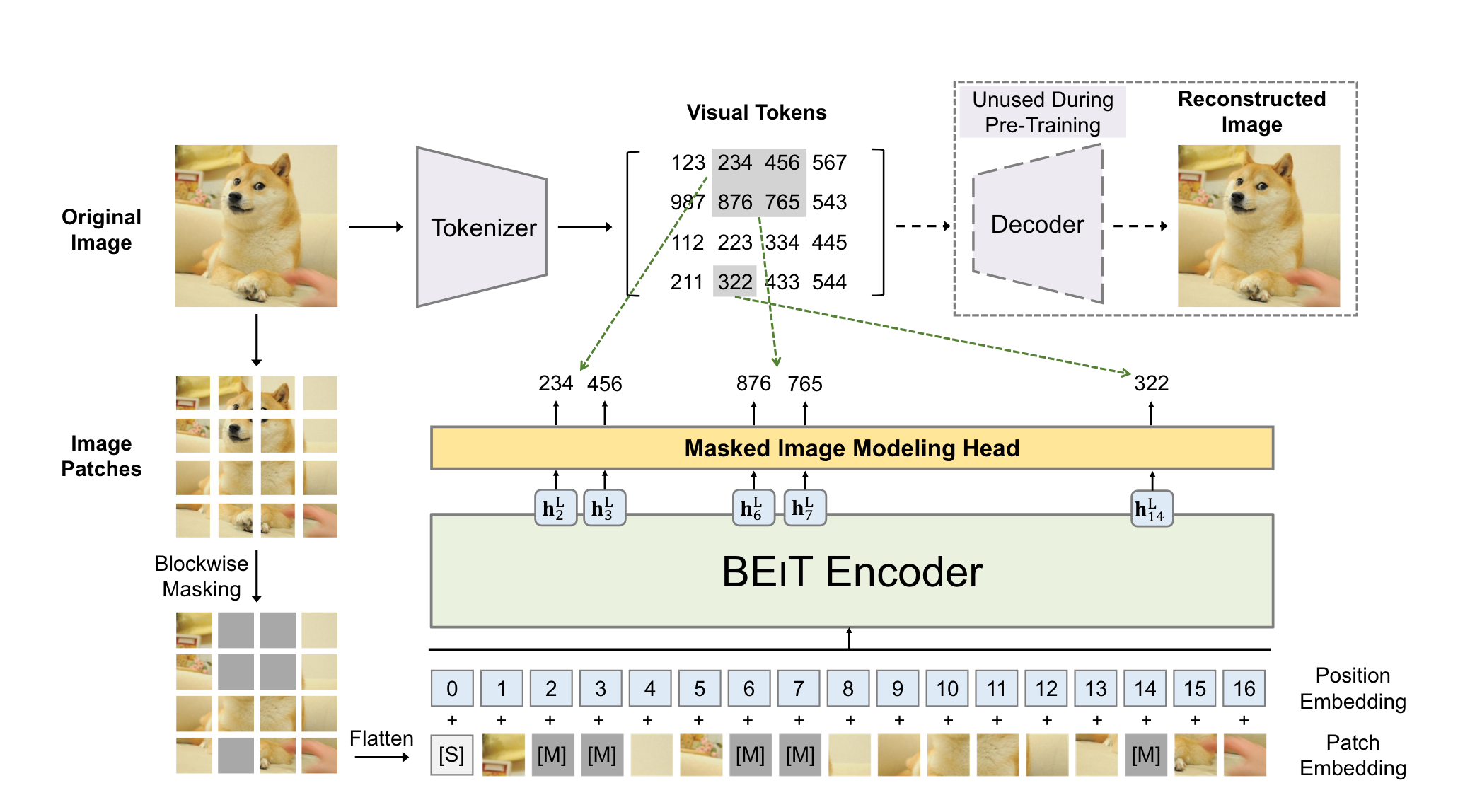}}
\caption{Overview of BEITpre-training.~\cite{Bao2021BEiTBP}}
\label{fig-beit-model}
\end{figure}

\textbf{BEiT:} BEiT~\cite{Bao2021BEiTBP}(Fig.~\ref{fig-beit-model})~\cite{Peng2022BEiTVM} is a self-supervised visual representation model that employs a masked image modeling approach similar to BERT~\cite{Devlin2019BERTPO} for pretraining visual transformers. It has achieved competitive results in image classification and semantic segmentation tasks,surpassing previous pretrained methods.

The core idea behind BEiT~\cite{Bao2021BEiTBP} is to divide an image into small blocks,randomly mask some of these blocks,and then task the model with predicting the content of the masked blocks based on the remaining ones. This way,the model can learn both global and local features of the image as well as the relationships between different blocks. BEIT utilizes a visual transformer~\cite{Dosovitskiy2020VIT} as its backbone network,which is a neural network based on self-attention mechanisms,capable of effectively processing serialized representations of images. The pretraining objective of BEIT is to minimize the reconstruction error of the masked blocks,similar to BERT's masked language modeling in the field of natural language processing.

One of the strengths of BEiT~\cite{Bao2021BEiTBP} is its ability to perform pretraining directly on images without requiring additional data or labels. It also allows for easy fine-tuning or linear probes on downstream tasks by adding task-specific layers on top of the pretrained encoder. Experimental results demonstrate that BEiT~\cite{Bao2021BEiTBP} performs well in both image classification and semantic segmentation tasks,even outperforming models pretrained on large-scale supervised data in some cases.

\begin{figure}[htbp]
\centering{\includegraphics[width=0.8\linewidth]{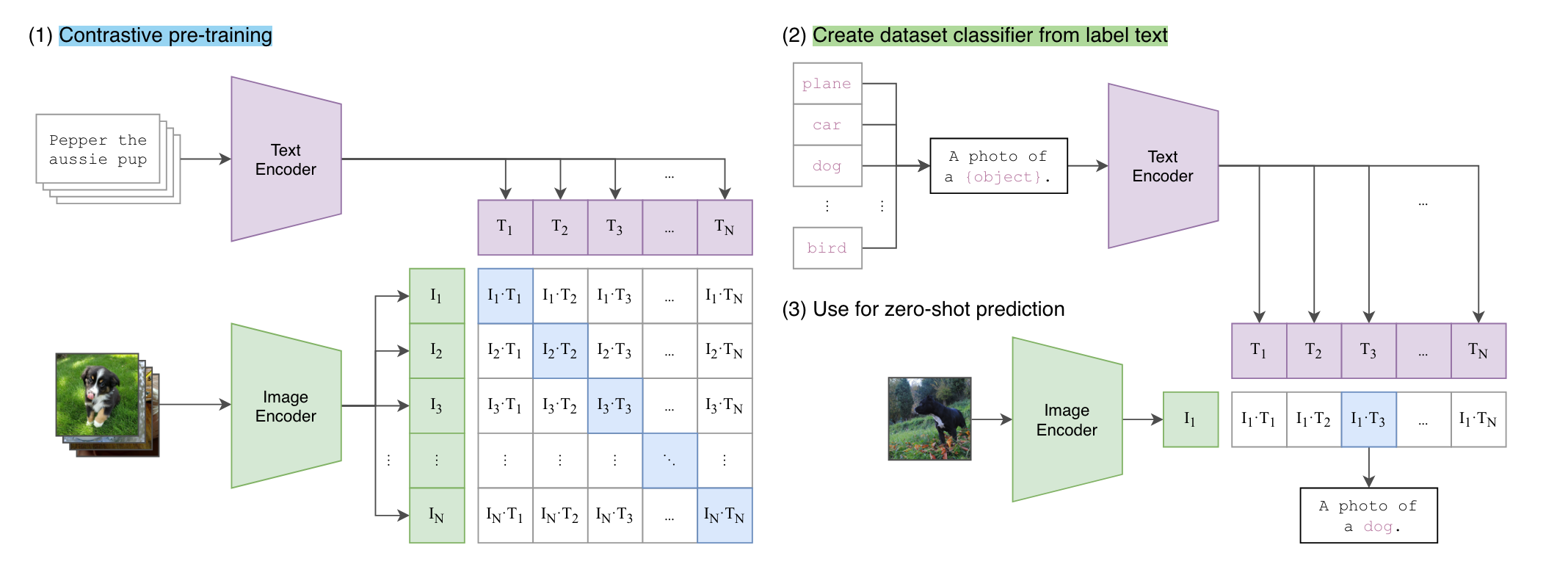}}
\caption{Summary of CLIP approach.~\cite{Radford2021LearningTVCLIP}}
\label{fig-clip-model}
\end{figure}

\begin{figure}[htbp]
\centering{\includegraphics[width=0.8\linewidth]{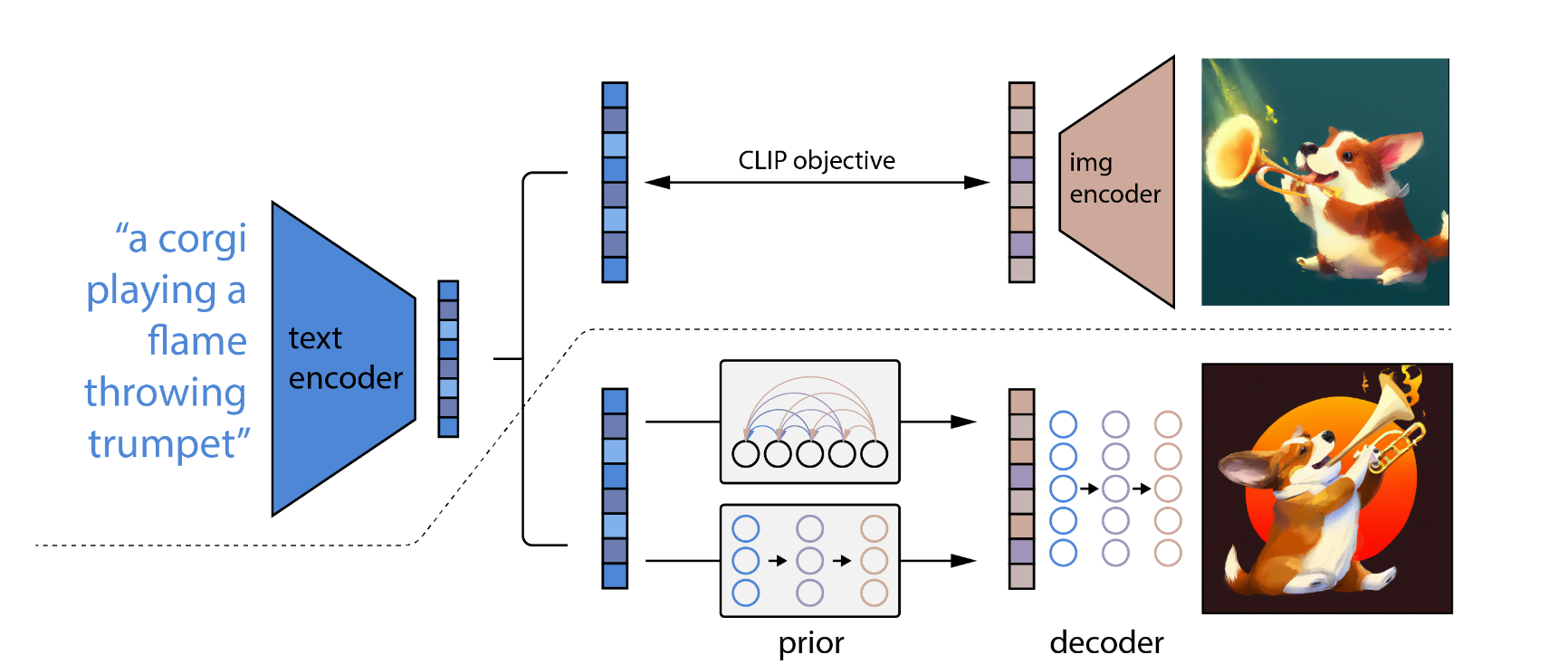}}
\caption{A high-level overview of unCLIP.~\cite{Ramesh2022HierarchicalTIDALLE2}}
\label{fig-unclip-model}
\end{figure}

\textbf{DALLE2:} DALLE2~\cite{Ramesh2022HierarchicalTIDALLE2} is a large language model developed by OpenAI for creating images and art forms based on natural language text descriptions. It is the second-generation model of DALLE and is capable of generating more realistic and accurate images with a fourfold increase in resolution. 
It utilizes CLIP~\cite{Radford2021LearningTVCLIP}(Fig.~\ref{fig-clip-model}),priors,and unCLIP(Fig.~\ref{fig-unclip-model}) models to generate images,with the quality of the generated images depending on the specificity of the textual prompts provided.

DALLE2~\cite{Ramesh2022HierarchicalTIDALLE2} can perform outpainting,which involves adding new content to the edges of an image,allowing for more natural image extension.
It can perform inpainting,filling in missing parts of an image,offering greater flexibility for image editing.
DALLE2~\cite{Ramesh2022HierarchicalTIDALLE2} is capable of generating images in various styles and themes,including photography,painting,cartoons,abstract art,and more,catering to diverse creative needs.

DALLE2~\cite{Ramesh2022HierarchicalTIDALLE2} represents a significant advancement in the field of AI-generated art and image synthesis,offering improved image quality and a broader range of creative possibilities compared to its predecessor.

\begin{figure}[htbp]
\centering{\includegraphics[width=0.8\linewidth]{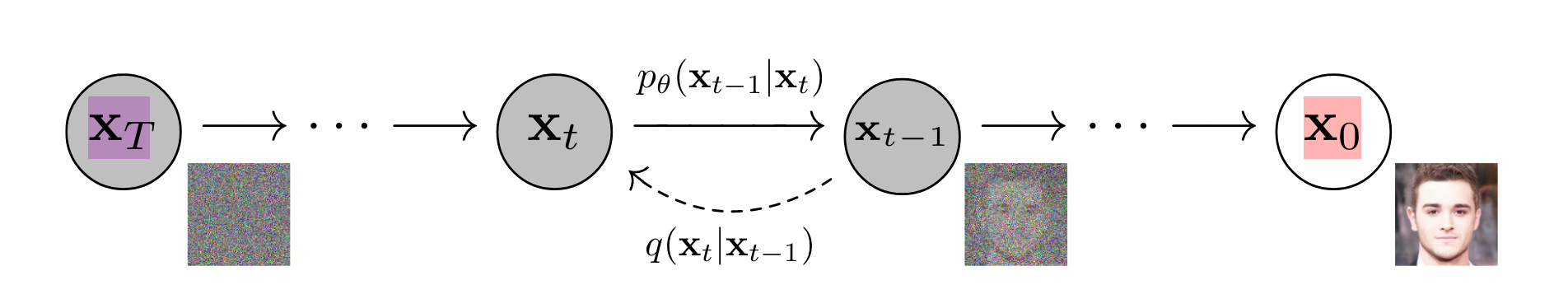}}
\caption{The directed graphical model of diffusion model.~\cite{Ho2020DenoisingDP}}
\label{fig-ddpm-model}
\end{figure}

\textbf{Imagen} Imagen~\cite{Saharia2022PhotorealisticTDImagen} is an advanced text-to-image diffusion model~\cite{Ho2020DenoisingDP}(Fig.~\ref{fig-ddpm-model}) developed by Google,capable of generating high-quality,high-resolution images based on text prompts. The core of Imagen~\cite{Saharia2022PhotorealisticTDImagen} lies in its combination of a powerful language model with a conditional diffusion model~\cite{Zhang2023AddingCC} to generate images,allowing it to understand complex text descriptions and produce detailed images that match them.

The workflow of Imagen~\cite{Saharia2022PhotorealisticTDImagen} is as follows:
\begin{enumerate}
    \item First,the text prompt is input into a frozen text encoder,resulting in text embeddings (which contain all the information from the text).
    \item These text embeddings are then fed into a generative model to create an image based on this information.
The process starts with generating a low-resolution image,which is then passed through two super-resolution networks in sequence. The inputs for these networks are the previously generated low-quality image and text embeddings.
    \item The final output is a high-quality image.
\end{enumerate}

Google found that general large-scale language models pre-trained on pure text corpora,such as T5~\cite{T5},are surprisingly effective in encoding text for image synthesis. The size of such models and the richness of their training data enable Imagen~\cite{Saharia2022PhotorealisticTDImagen} to excel in generating images that match text descriptions.

\textbf{Midjourney} Midjourney\footnote{Midjourney:\label{section:Midjourney}\url{https://www.midjourney.com/}} is an AI painting tool founded by David Holz and launched in March 2022. This tool can quickly generate images based on text prompts entered by users,usually in less than a minute. After the launch of its beta version,Midjourney quickly became a topic of discussion and has been integrated into the Discord community. It supports various artistic styles,such as Andy Warhol,Leonardo da Vinci,Salvador Dalí,and Pablo Picasso,and can recognize specific camera or photographic terms. Compared to other AI drawing tools like Google's Imagen~\cite{Saharia2022PhotorealisticTDImagen} and OpenAI's DALL·E~\cite{Ramesh2022HierarchicalTIDALLE2},Midjourney was the first to rapidly produce AI-generated images and open the platform for public application.

\begin{figure}[htbp]
\centering{\includegraphics[width=0.8\linewidth]{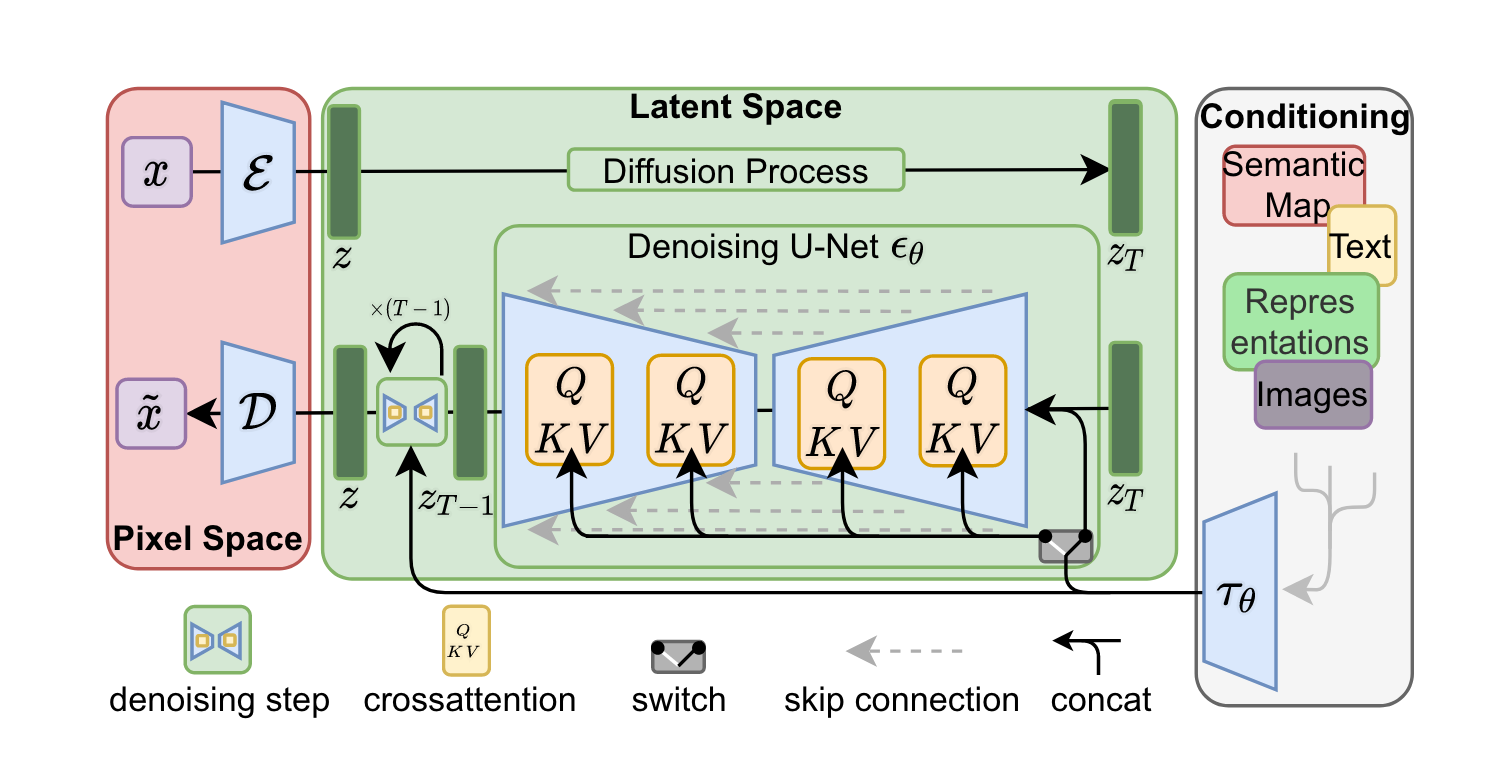}}
\caption{Overview of LDMs.~\cite{Rombach2021HighResolutionISStableDiffusion}}
\label{fig-ldm-model}
\end{figure}

\textbf{StableDiffusion:} Stable Diffusion\cite{Rombach2021HighResolutionISStableDiffusion} is an open-source deep learning model for text-to-image generation,developed through collaboration between the LMU Munich Computer Vision Group,Stability AI,and Runway. 
It is primarily used for generating detailed images based on textual descriptions,although it can also be applied to other tasks such as inpainting,outpainting,and image transformation guided by prompt words.
It relies on algorithms from the latent diffusion model(Fig.~\ref{fig-ldm-model}) and utilizes variational autoencoders and transformer networks to achieve high-resolution image synthesis.

It can be pretrained on various datasets to enhance its generality and practicality and can also be fine-tuned for specific domains or styles to improve its specialization and personalization.
Stable Diffusion represents a significant advancement in the field of text-to-image generation,offering capabilities for high-quality image synthesis and adaptability to different applications and domains.

\textbf{Kandinsky}~\cite{Razzhigaev2023KandinskyAI} The large-scale model Kandinsky,developed by the Russian AI research team AI Forever,is the largest open-source text-to-image model. It features an exceptionally large text encoder with 8.6 billion parameters,bringing the total model size to 11.9 billion parameters. Kandinsky-3\footnote{Kandinsky-3:\label{section:Kandinsky-3}\url{https://huggingface.co/kandinsky-community/kandinsky-3}} is characterized by its ability to generate high-quality images directly from textual prompts,support for longer text inputs,and the use of attention pooling to derive a global feature from text features,further enhancing the consistency and quality of image generation.

\begin{figure}[htbp]
\centering{\includegraphics[width=0.8\linewidth]{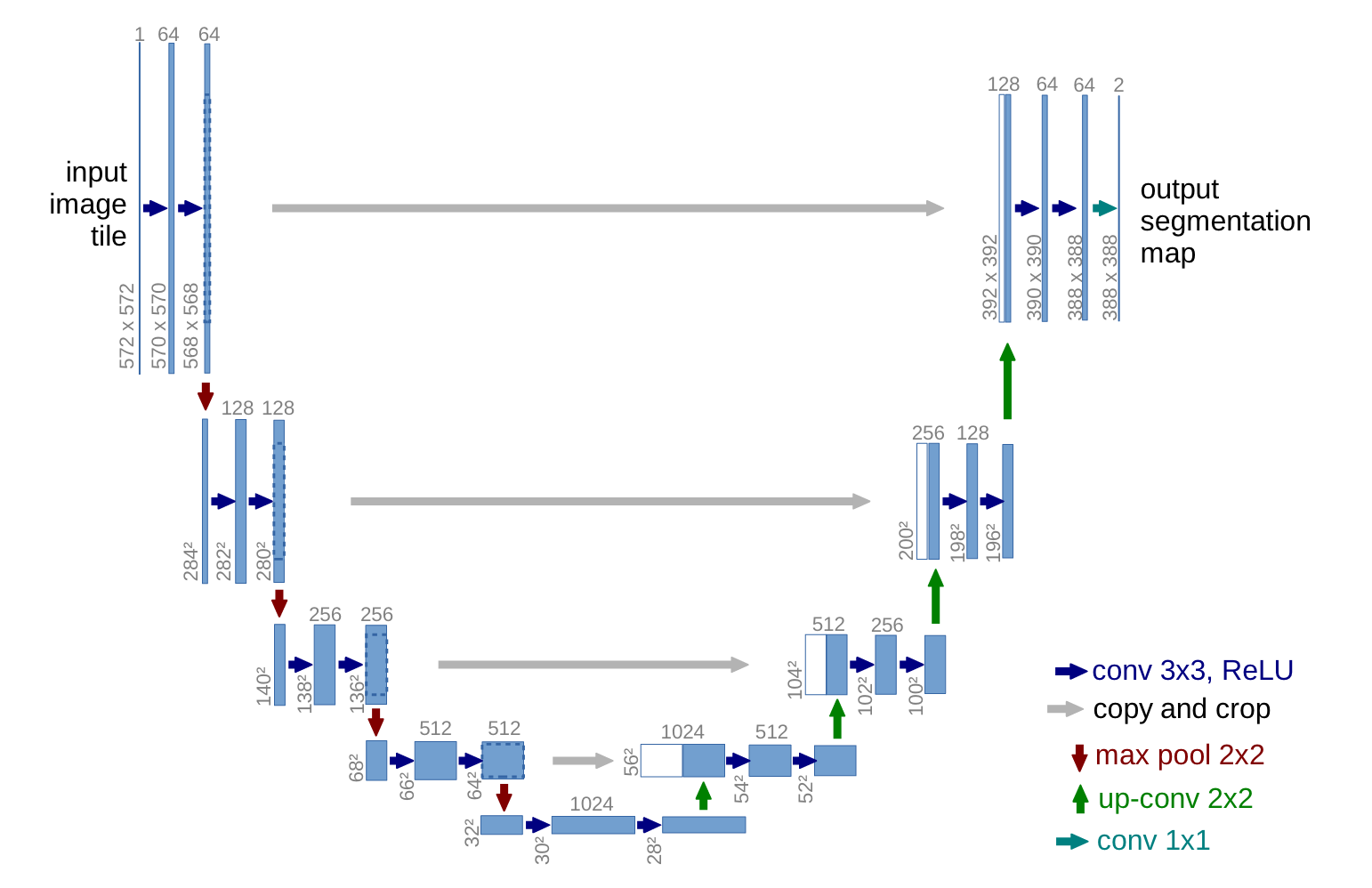}}
\caption{U-net architecture (example for 32x32 pixels in the lowest resolution).~\cite{Ronneberger2015UNetCN}}
\label{fig-unet-model}
\end{figure}

The structure of the Kandinsky-3 model includes a text encoder based on Flan-UL2\footnote{Flan-UL2:\label{section:Flan-UL2}\url{https://huggingface.co/google/flan-ul2}} and a Latent Diffusion model\cite{Rombach2021HighResolutionISStableDiffusion}(Fig.~\ref{fig-ldm-model}). Flan-UL2 is a large-scale language model by Google,with a total of 20 billion parameters,and Kandinsky-3 employs its encoder part. This model's UNet~\cite{Ronneberger2015UNetCN}(Fig.~\ref{fig-unet-model}) architecture has 3 billion parameters,slightly larger than Stable Diffusion XL (2.6 billion)\footnote{Stable Diffusion XL:\label{section:Stable Diffusion XL}\url{https://stablediffusionweb.com/StableDiffusionXL}}. Kandinsky-3's training strategy involves multi-stage training,ultimately enabling it to directly generate images of 1024x1024 resolution.

\textbf{LCM-LoRA} LCM-LoRA~\cite{Luo2023LCMLoRAAU} is an innovative image synthesis technique that combines Latent Consistent Model (LCM) and Low-Rank Adaptation (LoRA) distillation techniques applied to stable diffusion models~\cite{Rombach2021HighResolutionISStableDiffusion} like SD-V1.5,SSD-1B,and SDXL. This technology significantly reduces memory requirements while enhancing the quality of image generation. As a universal acceleration module,LCM-LoRA can be directly applied to various fine-tuned stable diffusion models and LoRA~\cite{Hu2021LoRA} models without the need for additional training,making it a versatile accelerator supporting various image generation tasks.

The core mechanism of LCM-LoRA~\cite{Luo2023LCMLoRAAU} lies in refinement techniques,utilizing the latent space of pre-trained autoencoders to refine the guided diffusion model into an LCM model. This process focuses on maintaining the trajectory consistency of generated samples,thus producing high-quality images while reducing the number of sampling steps. LCM-LoRA~\cite{Luo2023LCMLoRAAU} can serve as a plug-and-play neural PF-ODE solver,predicting solutions in latent space without the need for iterative solving through numerical ODE solvers,efficiently synthesizing high-resolution images in just 1-4 steps of inference.

The advantages of LCM-LoRA~\cite{Luo2023LCMLoRAAU} include efficiency,fast generation of high-quality images,flexibility,and versatility,with profound implications for the field of image generation. It has applications in various domains,from artistic creation to design and media,leveraging its ability to rapidly generate high-quality images from text.

You can review the source code\footnote{LCM-LoRA Github:\label{section:lcm-lora-code}\url{https://github.com/luosiallen/latent-consistency-model}} and demo\footnote{LCM-LoRA Demo:\label{section:lcm-lora-demo}\url{https://huggingface.co/spaces/SimianLuo/Latent_Consistency_Model}}  of LCM-LoRA.

\textbf{Cogview,Cogview2} Cogview~\cite{Ding2021CogViewMT} and Cogview2~\cite{Ding2022CogView2FA} are both text-to-image generation models capable of generating images corresponding to arbitrary text input in both Chinese and English. They both use a visual transformer~\cite{Dosovitskiy2020VIT} as their backbone network and visual tokens to represent images. 

\begin{figure}[htbp]
\centering{\includegraphics[width=0.8\linewidth]{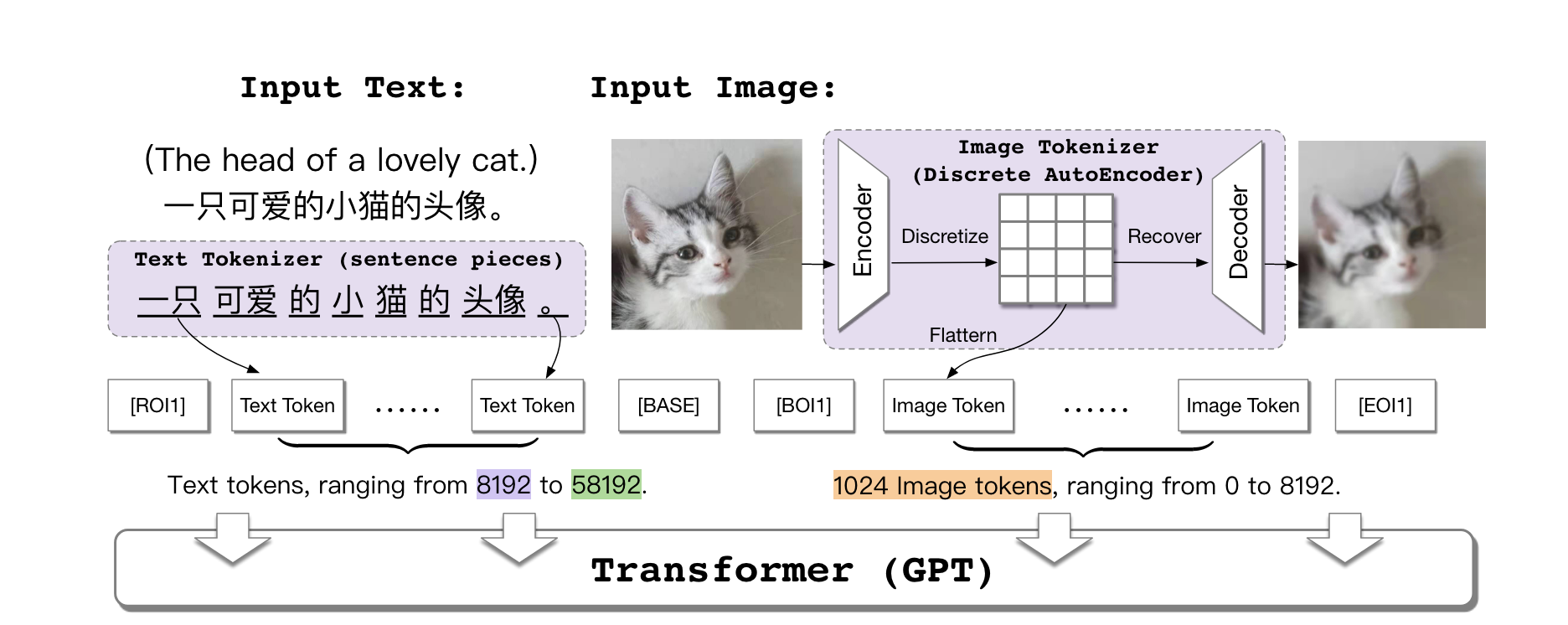}}
\caption{The framework of CogView. [ROI1], [BASE1], etc., are seperator tokens.~\cite{Ding2021CogViewMT}}
\label{fig-cogview-model}
\end{figure}

Cogview~\cite{Ding2021CogViewMT}(Fig.~\ref{fig-cogview-model}) is an autoregressive generation-based model that employs Visual Token Knowledge Distillation (VQKD) to learn the encoding of visual tokens from a large-scale image dataset. It then uses a unidirectional language model (UniLM~\cite{Dong2019UniLM},\cite{Bao2020UniLMv2PL}) to encode text and interact with the visual transformer to generate images from text descriptions.

\begin{figure}[htbp]
\centering{\includegraphics[width=0.8\linewidth]{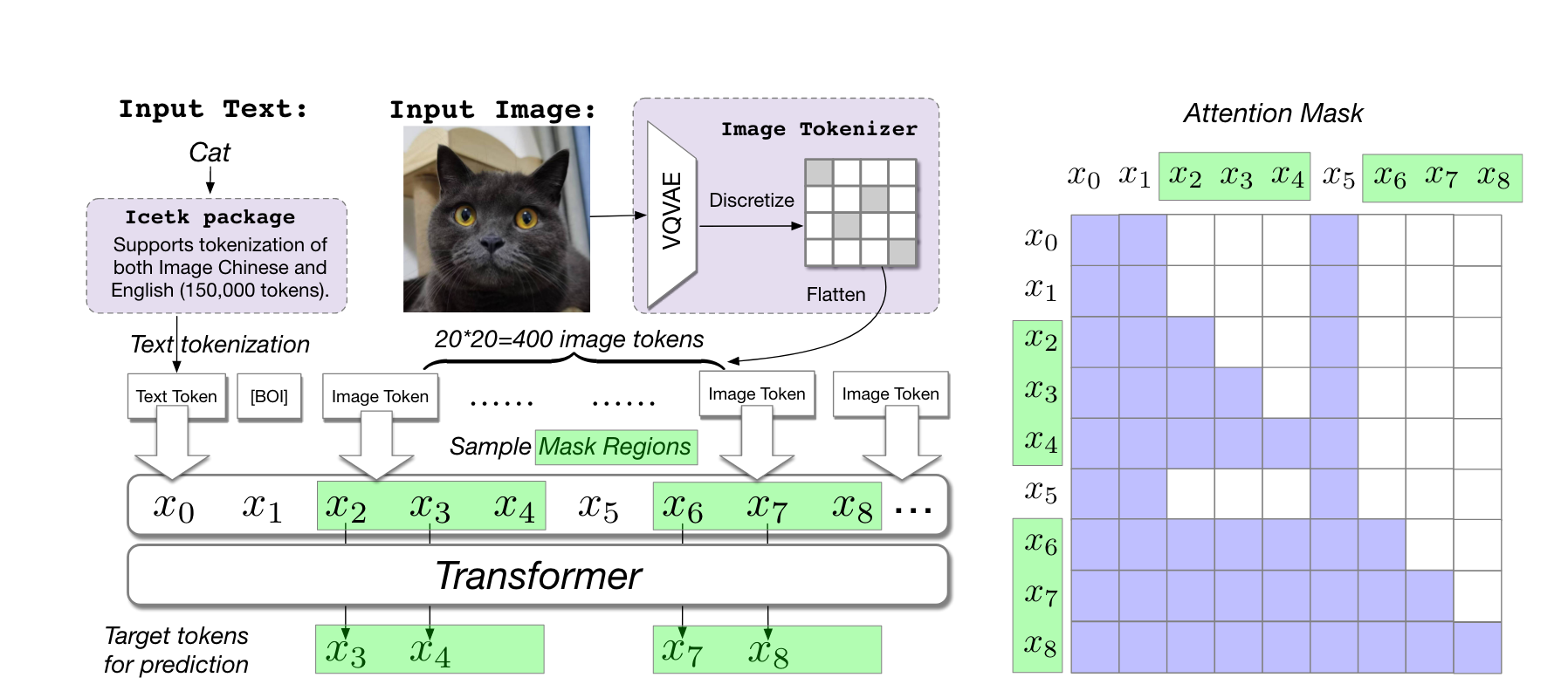}}
\caption{CogLM. (Left) The sequence consists of both text and image tokens. [BOI] (Begin-OfImage) is the separator token. Mask regions are sampled according to different strategies. Only the second-to-last tokens in the mask regions are predicted to compute the loss. (Right) The mask will only change the attention map, not the input sequence, where rows and columns of all the masked tokens together form a low-triangle attention mask matrix.~\cite{Ding2022CogView2FA}}
\label{fig-cogview2-model}
\end{figure}

Cogview2~\cite{Ding2022CogView2FA}(Fig.~\ref{fig-cogview2-model}),on the other hand,is based on hierarchical transformers and local parallel autoregressive generation. It uses Masked Image Modeling (MIM) as a pretraining objective,similar to masked language modeling in BERT for natural language processing. Cogview2 employs a bidirectional language model (CogLM) to encode text and interact with the visual transformer for text-to-image generation.

Cogview2~\cite{Ding2022CogView2FA} has larger model parameters,ranging from 6 billion to 9 billion parameters,compared to Cogview's~\cite{Ding2021CogViewMT} 4 billion,enabling it to generate higher-resolution and higher-quality images.

Cogview2~\cite{Ding2022CogView2FA} uses Local Parallel Autoregressive Generation (LoPAR) as a generation strategy,allowing it to divide images into multiple regions and generate visual tokens in parallel within each region. This accelerates the generation speed while maintaining global consistency.
Cogview2~\cite{Ding2022CogView2FA} employs CogLM as the text encoder,which is a bidirectional language model based on BERT. It can effectively handle both Chinese and English text and interacts with the visual transformer for text-to-image generation.

In summary,Cogview2~\cite{Ding2022CogView2FA} offers improvements in model size,generation strategy,and text encoding capabilities,resulting in the generation of high-quality images from textual descriptions.

\textbf{Pika} Pika\footnote{Pika:\label{section:pika}\url{https://pika.art/}} is a text-to-video AI tool,allowing users to generate video content by inputting text prompts. What sets Pika apart is its capability to dynamically transform elements in the scene based on the input prompts without causing overall visual distortion. Additionally,Pika can discern elements within the scene and generate content that logically fits within the image,avoiding any distortions in the visuals.

\textbf{Table-GPT:} Table-GPT~\cite{Li2023TableGPTTG} is a language model specifically designed for understanding and generating tabular data. This model combines traditional text processing capabilities with an understanding of table data structures,enabling it to excel in handling text that includes tables. Core functionalities of Table-GPT~\cite{Li2023TableGPTTG} include extracting information from tables,generating textual content within tables,and understanding and answering questions related to table data.

It is commonly used in scenarios that require extensive data organization and analysis,such as data analytics,financial reporting,and academic research. With Table-GPT~\cite{Li2023TableGPTTG},users can more easily extract valuable information from complex datasets or automate certain data processing tasks. The advent of this model marks a significant advancement in natural language processing technology,especially in handling structured data.

\begin{figure}[htbp]
\centering{\includegraphics[width=0.8\linewidth]{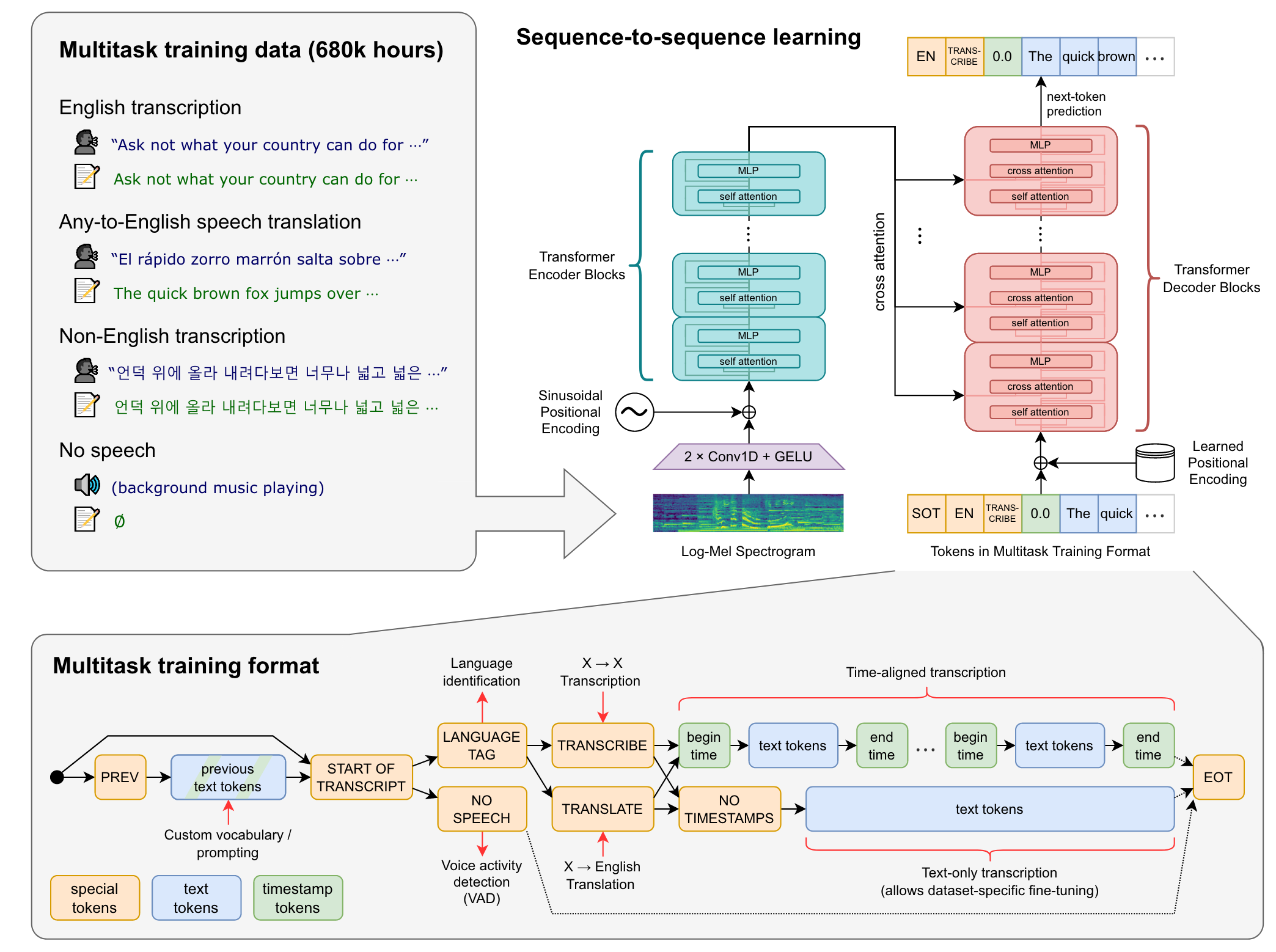}}
\caption{Overview of Whisper. A sequence-to-sequence Transformer model is trained on many different speech processing tasks, including multilingual speech recognition, speech translation, spoken language identification, and voice activity detection. All of these tasks are jointly represented as a sequence of tokens to be predicted by the decoder, allowing for a single model to replace many different stages of a traditional speech processing pipeline. The multitask training format uses a set of special tokens that serve as task specifiers or classification targets.~\cite{Radford2022RobustSRWhisper}}
\label{fig-whisper-model}
\end{figure}

\textbf{Whisper} Whisper~\cite{Radford2022RobustSRWhisper}(Fig.~\ref{fig-whisper-model}) is a large-scale pretrained language model released by OpenAI in 2022. It incorporates functionalities such as multilingual Automatic Speech Recognition (ASR),speech translation,and language identification. It has been trained on a substantial amount of weakly labeled data,eliminating the need for fine-tuning for specific tasks and allowing for direct multitask learning. Whisper's performance in languages like English and Spanish is approaching human-level capabilities. Additionally,it demonstrates strong robustness and generalization across tasks and languages.

\textbf{VideoLDM} VideoLDM~\cite{Blattmann2023AlignYLVideoLDM} is a novel video generation model developed collaboratively by NVIDIA and the research team at Cornell University. It is a type of Latent Diffusion Model (LDM),specifically designed for high-resolution video synthesis,capable of automatically generating video content based on textual descriptions.

The working principle of VideoLDM~\cite{Blattmann2023AlignYLVideoLDM} involves initially training a diffusion model in a compressed,low-dimensional latent space to achieve high-quality image synthesis while avoiding excessive computational demands. The model then transforms into a video generator by introducing a temporal dimension to the latent space diffusion model and fine-tuning it on encoded image sequences (i.e.,videos). Additionally,VideoLDM~\cite{Blattmann2023AlignYLVideoLDM} aligns the upsampler of the diffusion model temporally,enabling it to generate time-consistent video super-resolution models.

VideoLDM~\cite{Blattmann2023AlignYLVideoLDM} has a wide range of applications,including the simulation of wilderness driving data and text-to-video modeling for creative content. It achieves state-of-the-art performance in synthesizing realistic driving videos and can generate videos with resolutions up to 1280 x 2048 pixels. This opens exciting new directions for future content creation.

\textbf{DINOv2} DINOv2~\cite{Oquab2023DINOv2LR} is a self-supervised visual Transformer model by Meta AI. It is trained using unsupervised learning methods,enabling it to learn general visual features from a large corpus of unlabeled images. A key characteristic of DINOv2 is its ability to train on any collection of images without the need for any associated metadata,meaning it can learn from all images it receives,not just those with a specific set of labels or captions.

The main advantages of DINOv2~\cite{Oquab2023DINOv2LR} include:
\begin{enumerate}
\item \textbf{Self-Supervised Training:} The model learns image features autonomously without the need for labeled data.
\item \textbf{No Need for Fine-Tuning:} The trained model can be directly transferred to downstream tasks such as classification,segmentation,and image retrieval.
\item R\textbf{eady-to-Use:} As a foundational large model for vision,DINOv2 offers high-performance features that can be directly used as input for simple linear classifiers.
\item \textbf{Strong Generalization Capabilities:} It performs exceptionally well in downstream tasks like depth estimation,even surpassing supervised learning methods.
\end{enumerate}
DINOv2~\cite{Oquab2023DINOv2LR} has pioneered a new direction in the field of computer vision,demonstrating that high-quality visual features can be learned through unsupervised learning. This provides insights for future large-scale learning of visual features.

\textbf{Segment Anything} Segment Anything~\cite{Kirillov2023SegmentA}\footnote{Segment Anything Github:\label{section:Segment-Anything}\url{https://github.com/facebookresearch/segment-anything}} is a new image segmentation project released by Meta AI\footnote{Meta AI:\label{section:Meta AI}\url{https://ai.meta.com/}},comprising a new task,model,and dataset. The aim of the project is to create a foundational model capable of returning effective segmentation masks for any given segmentation prompt. The model is designed to be promptable,meaning it can zero-shot transfer to new image distributions and tasks.

Highlights of the project include:
\begin{enumerate}
\item \textbf{Largest Segmentation Dataset:} Over a billion masks on 11 million licensed and privacy-respectful images.
\item \textbf{Strong Zero-Shot Generalization:} The model can handle new image distributions and tasks without additional training.
\item \textbf{Flexible Prompting:} Users can guide the model for image segmentation through text prompts.
\item \textbf{Real-Time Mask Calculation:} Allows for interactive use with the ability to rapidly generate segmentation masks.
\item \textbf{Open Source Model and Dataset:} The SAM (Segment Anything Model) and the corresponding SA-1B dataset\footnote{SA-1B dataset:\label{section:SA-1B-dataset}\url{https://segment-anything.com/dataset/index.html}} are publicly available to facilitate research in foundational models of computer vision.
\end{enumerate}

The emergence of this project unifies downstream applications of segmentation tasks and demonstrates the potential of large models in the field of computer vision. It could bring substantial transformations to computer vision research,enabling many tasks to be handled in a unified manner.

\textbf{LINGO-1} LINGO-1~\cite{LINGO-1} is an open-loop driving commentator model that combines visual,language,and action aspects. It was developed by the UK autonomous driving startup company Wayve with the aim of enhancing the understanding,explanation,and training effectiveness of basic driving models. LINGO-1~\cite{LINGO-1} achieves this by collecting data from expert drivers that includes images,language,and actions,creating a scalable and diverse dataset. This data is collected from drivers across various locations in the UK while they drive and provide commentary.

At the core of LINGO-1~\cite{LINGO-1} is the Visual-Language-Action Model (VLAM),which combines the reasoning capabilities of large language models with images and videos,performing various visual tasks such as image classification,text-to-image retrieval,and visual question-answering. Additionally,LINGO-1 can perform tasks based on simple prompts,such as reasoning about scene understanding questions and identifying the primary reasons behind driving decisions.

A key feature of LINGO-1~\cite{LINGO-1} is its ability to generate continuous commentary,explaining the reasons behind driving actions,helping to understand the model's reasoning process. For instance,it can describe actions taken when overtaking a parked car or how it slows down and stops when approaching a zebra crossing.

Despite showing significant potential in the field of autonomous driving,LINGO-1~\cite{LINGO-1} also has some limitations,such as limited generalization ability,issues related to hallucinations,and long-term memory retention. These limitations will need to be overcome through further research and development. Overall,LINGO-1~\cite{LINGO-1} offers new possibilities for the safety and interpretability of autonomous driving,driving research into end-to-end autonomous driving.

\begin{figure}[htbp]
\centering{\includegraphics[width=0.8\linewidth]{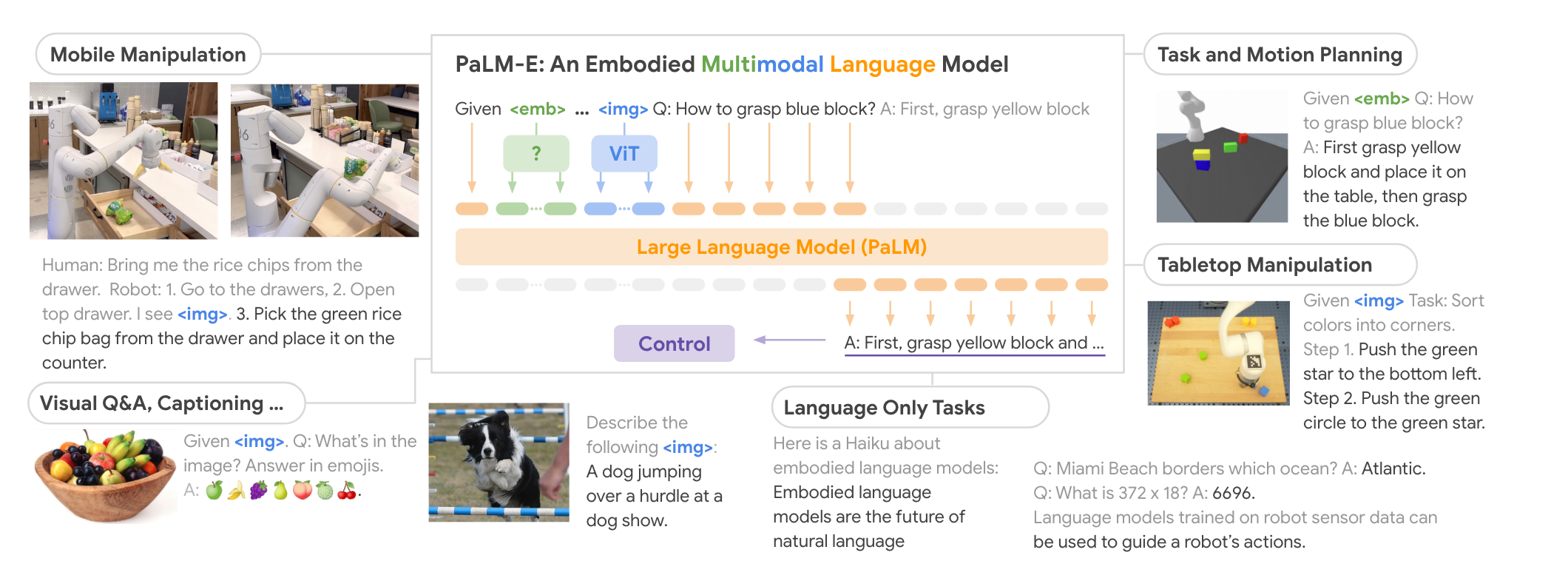}}
\caption{PaLM-E is a single general-purpose multimodal language model for embodied reasoning tasks, visual-language tasks, and language tasks. PaLM-E transfers knowledge from visual-language domains into embodied reasoning – from robot planning in environments with complex dynamics and physical constraints, to answering questions about the observable world. PaLM-E operates on multimodal sentences, i.e. sequences of tokens where inputs from arbitrary modalities (e.g. images, neural 3D representations, or states, in green and blue) are inserted alongside text tokens (in orange) as input to an LLM, trained end-to-end.~\cite{Driess2023PaLMEAE}}
\label{fig-PaLM-E-model}
\end{figure}

\textbf{PaLM-E} PaLM-E~\cite{Driess2023PaLMEAE}(Fig.~\ref{fig-PaLM-E-model}) is a large multimodal embodied visual language model (VLM) released by Google,boasting a staggering 562 billion parameters. It has the ability to integrate visual and language information into robot control,allowing it to perform various tasks without the need for specialized training. What sets PaLM-E~\cite{Driess2023PaLMEAE} apart is its capability to comprehend images,generate language,and combine both to process complex robot commands. For instance,it can instruct robots to complete a wide range of intricate tasks and also generate language descriptions for images.

One notable feature of PaLM-E~\cite{Driess2023PaLMEAE} is its strong transferability,meaning that when compared to single-task robot models trained in different domains,PaLM-E~\cite{Driess2023PaLMEAE} trained in various domains exhibits significantly improved performance. Additionally,PaLM-E demonstrates outstanding abilities in tasks such as multi-modal chain-of-thought reasoning and multi-image reasoning,achieving a new State of the Art (SOTA) on the OK-VQA~\cite{Marino2019OKVQAAV} benchmark.

PaLM-E~\cite{Driess2023PaLMEAE} also showcases unprecedented flexibility and adaptability,marking a significant leap in the field of human-machine interaction. Researchers have demonstrated that training across different combinations of tasks involving multiple robots and general visual-language tasks can lead to several approaches for transitioning from visual language to embodied decision-making,allowing robots to effectively leverage data when planning tasks.

\textbf{RT-2} RT-2\footnote{RT-2:\label{section:RT-2}\url{https://robotics-transformer2.github.io/}}~\cite{rt22023arxiv},developed by Google's DeepMind,is a novel robot model,which stands for Robotics Transformer 2. It is a Visual-Language-Action (VLA) model capable of learning from both web and robot data and translating this knowledge into general instructions for robot control. The RT-2 model is trained on large-scale web datasets,enabling it to recognize visual or language patterns and operate across different languages.

The core features of RT-2~\cite{rt22023arxiv} include its generalization and emergence capabilities. It can perform tasks not explicitly included in the training data and improves the performance of baseline models by three times in new skill evaluations. This demonstrates RT-2's outstanding ability to understand new commands and perform basic reasoning,such as reasoning about object categories or advanced descriptions.

Furthermore,RT-2 incorporates chain-of-thought reasoning into the model,allowing it to engage in multi-stage semantic reasoning,such as deciding which object can serve as an improvised hammer (a stone) or which beverage is best suited for a tired person (an energy drink). This chain-of-thought reasoning capability enables RT-2 to execute more complex commands that require intermediate steps to fulfill user instructions.

The RT-2~\cite{rt22023arxiv} model also performs well in real-world applications and can generalize to new objects and environments. For example,even in scenarios where no objects other than blue cubes were present in the training dataset,RT-2~\cite{rt22023arxiv} still performs well in real robot Language Table tasks.

\textbf{RoboCat} RoboCat~\cite{bousmalis2023robocat} is an AI agent developed by Google's DeepMind team,designed for general-purpose robotics. It possesses self-improvement capabilities,being able to complete tasks after only 100 training iterations and refine its performance from self-generated data. RoboCat~\cite{bousmalis2023robocat} can operate various robotic arms to perform a range of tasks,including ring stacking,building with blocks,and fruit picking,which challenge its precision and comprehension in manipulating robotic arms.

The emergence of RoboCat~\cite{bousmalis2023robocat} represents a significant advancement in the field of AI,as it reduces the need for human supervision during training and marks an important step towards creating versatile robots. It is built upon DeepMind's multimodal model Gato and leverages a large training dataset comprising image sequences and actions of various robotic arms,addressing hundreds of tasks. As it learns from an increasing number of tasks,RoboCat becomes better at learning and solving additional new tasks.

The research and development of RoboCat~\cite{bousmalis2023robocat} not only drive progress in robotics technology but also open up possibilities for the future application of robots in various domains,including industry,healthcare,and the service sector. Its success rate has already doubled in its initial stages,showcasing its ability for rapid learning and self-improvement.

\textbf{CICERO} CICERO~\cite{ghosal-etal-2022-cicero} is a recent AI model launched by Meta AI,and it excels in human negotiation games,demonstrating effective communication and collaboration with human players. CICERO's core is powered by a dialogue engine and a strategic reasoning engine,enabling it to understand language,motivations,devise conversational strategies,and adjust its wording appropriately. In experiments,CICERO participated incognito in a diplomatic game,engaging with 82 human players in 40 games,without being identified as an AI.

CICERO's~\cite{ghosal-etal-2022-cicero} dialogue model is trained based on the BART~\cite{lewis-etal-2020-bart} model,combining features of GPT~\cite{Radford2019LanguageMAgpt2} and BERT~\cite{Devlin2019BERTPO},making it suitable for text generation scenarios and capable of bi-directionally understanding contextual information. The strategic reasoning engine utilizes planning algorithms to compute optimal choices based on the current situation and is trained through reinforcement learning to make the model's strategies more reasonable and human-like. CICERO~\cite{ghosal-etal-2022-cicero} outperformed human players in the game,with an average score exceeding twice that of human players,ranking in the top 10\%.

\textbf{RFdiffusion} RFdiffusion\cite{Watson2022} is an open-source protein design framework that combines structure prediction networks and diffusion generative models to achieve state-of-the-art performance in various protein design challenges. This framework enables structural generation with or without conditional information,such as specific motifs or targets.

The core features of RFdiffusion~\cite{Watson2022} include:
\begin{itemize}
\item \textbf{Motif Scaffolding:} Creating a stable scaffold to support the target motif.
\item \textbf{Unconditional Protein Generation:} Generating protein structures without relying on specific conditions.
\item \textbf{Symmetric Unconditional Generation:} Generating protein structures with cyclic,dihedral,and tetrahedral symmetry.
\item \textbf{Symmetric Motif Scaffolding:} Designing cyclic oligomers that satisfy symmetric structures.
\item \textbf{Binder Design:} Designing proteins that can bind to other molecules.
\item \textbf{Diverse Design:} Sampling around a design to generate structurally diverse solutions.
\end{itemize}

The models trained with RFdiffusion~\cite{Watson2022} can design proteins with complex shapes, such as $\alpha$-helices and $\beta$-strands, which exhibit features similar to real proteins. It can also accurately predict scaffolds for catalytic sites in a range of enzymes, containing multiple side chains and backbone functional groups.

RFdiffusion~\cite{Watson2022} has broad application prospects,not only in drug design and biotechnology but also as a tool for vaccine platforms,drug delivery,and catalyst design.

\textbf{Med-PaLM 2} Med-PaLM 2~\cite{Singhal2023MedPaLM2} is a medical question-answering large language model jointly released by Google Research and DeepMind,based on PaLM 2~\cite{Singhal2023MedPaLM2}. It aims to provide high-quality medical question-answering services,achieving expert-level proficiency in medical knowledge retrieval,reasoning,and answering medical questions.

Key features of Med-PaLM 2~\cite{Singhal2023MedPaLM2} include:
\begin{itemize}
\item \textbf{High Accuracy:} Med-PaLM 2~\cite{Singhal2023MedPaLM2} scores 86.5\% on the MedQA dataset,over 19\% improvement from its predecessor Med-PaLM,setting a new state-of-the-art benchmark.
\item \textbf{Expert-Level Performance:} In medical question-answering,Med-PaLM 2~\cite{Singhal2023MedPaLM2} performs close to or surpasses the latest technological levels in multiple medical question-answering datasets.
\item \textbf{Human Evaluation:} Through detailed human evaluations,doctors tend to prefer Med-PaLM 2's answers across multiple axes related to clinical utility.
\item \textbf{Safety and Limitations:} Med-PaLM 2~\cite{Singhal2023MedPaLM2} shows significant improvements on every evaluation axis for the newly introduced adversarial question dataset,highlighting the importance of comprehensive evaluation.
\end{itemize}

The development of Med-PaLM 2~\cite{Singhal2023MedPaLM2} reflects the rapid advancement towards physician-level performance in the medical question-answering field. However,further research is necessary to validate the effectiveness of these models in real-world settings.

\textbf{TimeGPT} TimeGPT~\cite{Garza2023TimeGPT1} is the industry's first foundational model for time series prediction,developed by the American company Nixtla. It is trained on over 100 billion data points from multiple domains including finance,meteorology,energy,and web data,and allows users to fine-tune the model with their own data. TimeGPT~\cite{Garza2023TimeGPT1} is designed to support various forecasting and anomaly detection tasks,claiming to be the first foundational model that consistently surpasses other options with minimal complexity.

Key features of TimeGPT~\cite{Garza2023TimeGPT1} include:
\begin{itemize}
\item \textbf{Large-Scale Training Data:} TimeGPT~\cite{Garza2023TimeGPT1} is trained on the largest publicly available time series collection to date,encompassing data from finance,economics,demographics,healthcare,weather,IoT sensors,energy,web traffic,sales,transportation,and banking.
\item \textbf{Transformer Architecture:} TimeGPT~\cite{Garza2023TimeGPT1} utilizes a Transformer-based model architecture,processing input historical data through an encoder-decoder and incorporating positional encoding to capture temporal information.
\item \textbf{Probabilistic Forecasting:} TimeGPT~\cite{Garza2023TimeGPT1} is capable of probabilistic forecasting,requiring no assumptions about data distribution and providing prediction intervals at specified levels of precision.
\end{itemize}
TimeGPT~\cite{Garza2023TimeGPT1} has broad application prospects,not only in time series forecasting but also in anomaly detection and other related tasks.

\textbf{Mamba~} Mamba~\cite{Gu2023MambaLS} is a large-scale sequence modeling model developed by the Sonta team at MIT CSAIL\footnote{MIT CSAIL:\label{section:MIT_CSAIL}\url{https://www.csail.mit.edu/}}. It is an upgraded version of the S4 (Structured State Spaces) model\footnote{S4 Code:\label{section:S4_code}\url{https://github.com/state-spaces/s4}}~\cite{Gu2021EfficientlyMLS4}. The core idea of the Mamba~\cite{Gu2023MambaLS} model is to make the parameters in the S4 model data-dependent,thereby enhancing the model's performance and efficiency. This improvement enables Mamba~\cite{Gu2023MambaLS} to perform better in processing long sequence data,such as text or time series.

Key features of the Mamba~\cite{Gu2023MambaLS} model include:
\begin{itemize}
\item \textbf{Data-Dependent Parameters:} The parameters of the Mamba model are no longer static but dynamically change according to the input data,enabling the model to better adapt to different data characteristics.
\item \textbf{Efficient Training Algorithm:} Mamba~\cite{Gu2023MambaLS} introduces an IO-aware parallel scanning algorithm for efficient model training,reducing overall read and write operations and improving wall-time efficiency.
\item \textbf{Simplified Model Structure:} Mamba~\cite{Gu2023MambaLS} proposes a new minimalist-style model block (Mamba block),combining token mixing and channel mixing into one,thus simplifying the model structure.
\end{itemize}
Mamba~\cite{Gu2023MambaLS} excels in multiple benchmark tests,particularly in processing long sequence data,where it can rival or even surpass Transformer models. This makes Mamba~\cite{Gu2023MambaLS}widely applicable in natural language processing and other sequence modeling tasks.

\textbf{AlphaCode} AlphaCode~\cite{Li2022CompetitionlevelCGAlphaCode} is an artificial intelligence system developed by the DeepMind\footnote{DeepMind:\label{section:DeepMind}\url{https://deepmind.google/}} team,capable of automatically generating competition-level programming code. Tested on the globally renowned algorithmic platform Codeforces\footnote{Codeforces:\label{section:Codeforces}\url{https://codeforces.com/}},AlphaCode~\cite{Li2022CompetitionlevelCGAlphaCode} performed impressively in 10 challenges solved alongside 5000 users,ultimately achieving a ranking within the top 54\% of human programmers. This indicates that AlphaCode's~\cite{Li2022CompetitionlevelCGAlphaCode} coding capabilities are comparable to nearly half of the programmers who have participated in tests on Codeforces~\ref{section:Codeforces}.

The working principle of AlphaCode~\cite{Li2022CompetitionlevelCGAlphaCode} is to generate code using a large Transformer language model. The model's training is divided into two stages: initial pre-training using a subset of GitHub code,followed by fine-tuning with more targeted competitive programming data. For each new and unseen problem,AlphaCode~\cite{Li2022CompetitionlevelCGAlphaCode} generates a large number of program samples,filters them based on execution results,and then clusters the remaining samples to produce a small number of candidate codes for final submission and evaluation.

Despite AlphaCode's~\cite{Li2022CompetitionlevelCGAlphaCode} commendable performance in programming competitions,the DeepMind team also notes that AlphaCode~\cite{Li2022CompetitionlevelCGAlphaCode} is currently only applicable to competitive programming contests. This suggests that while AlphaCode~\cite{Li2022CompetitionlevelCGAlphaCode}has made breakthroughs in the field of automatic coding,it does not imply that it can completely replace programmers,especially in the context of complex business scenarios and unclear problem descriptions.

AlphaCode2~\cite{AlphaCode2_Tech_Report},an enhanced version of the programming tool powered by the Gemini model,has demonstrated exceptional performance in programming competitions,particularly in solving complex problems such as dynamic programming. In the context of competitive programming on the Codeforces~\ref{section:Codeforces} platform,AlphaCode2~\cite{AlphaCode2_Tech_Report},which utilizes programming languages such as Python,Java,C++,and Go,has shown a notable improvement over its predecessor,outperforming approximately 85\% of competitors on average.

This advanced system employs a series of strategic models to generate code samples. It utilizes clustering algorithms to select the optimal code solution. Looking forward,AlphaCode2~\cite{AlphaCode2_Tech_Report} holds promise as a collaborative tool that could significantly aid the entire software development process. Such a tool represents a substantial step forward in the field of automated programming,offering potential benefits in terms of efficiency and accuracy in code generation.

\textbf{Gemini} Google's Gemini~\cite{gemini_1_report}\cite{GeminiIntro}\footnote{Gemini Technologies:\label{section:Gemini_technologies}\url{https://deepmind.google/technologies/gemini}} model represents a significant advancement in the field of Large Language Models (LLMs). It integrates the advanced text generation capabilities of GPT-4~\cite{OpenAI2023GPT4TR} with innovative training techniques from AlphaGo~\cite{AlphaGo},including reinforcement learning and tree search algorithms. Gemini is expected to significantly surpass the computational capabilities of GPT-4~\cite{OpenAI2023GPT4TR},with its total pre-training floating-point operations per second (FLOPS) anticipated to be more than five times that of GPT-4~\cite{OpenAI2023GPT4TR}.

Key characteristics of the Gemini model include:
\begin{itemize}
\item \textbf{Multimodal Capabilities:} Designed inherently as a multimodal model,Gemini~\cite{gemini_1_report} is not only capable of understanding and generating text and code but can also process and generate images,indicating its potential in handling cross-modal data.
\item \textbf{Efficient Integration:} The model demonstrates efficiency in tools and application programming interface (API) integration,allowing it to effectively process and understand various types of data.
\item \textbf{Reinforcement Learning and Tree Search Techniques:} Drawing from the technology used in AlphaGo,particularly in planning and memory applications,it is expected to enhance the overall performance of Gemini~\cite{gemini_1_report}.
\item \textbf{Training with YouTube Data:} Gemini~\cite{gemini_1_report} may utilize YouTube video data for training,potentially enhancing its capabilities in video recognition and processing.
\end{itemize}
Overall,the development of the Gemini model is Google's competitive response to OpenAI's ChatGPT and other AI conversational systems. Its multimodal design enables it to handle and understand different forms of data and shows efficiency in tool and API integration,allowing it not only to process text but also to proficiently handle various types of inputs including images and videos.

\begin{figure}[htbp]
\centering{\includegraphics[width=0.8\linewidth]{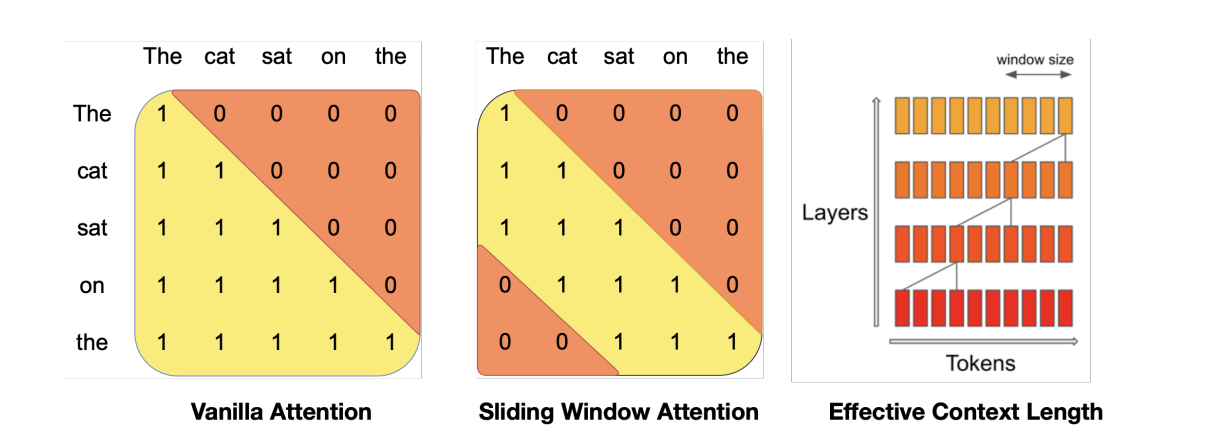}}
\caption{Sliding Window Attention.~\cite{Jiang2023Mistral7}}
\label{fig:slidingwindowattention}
\end{figure}

\textbf{Mistral}
Mistral~\cite{Jiang2023Mistral7} AI is a large language model project developed by a team composed of former DeepMind~\ref{section:DeepMind} and Meta employees. The team aims to create high-performance models capable of handling complex tasks and possessing multilingual abilities. Mistral AI’s models utilize innovative architectural choices,such as Sliding Window Attention(Fig.~\ref{fig:slidingwindowattention}) and Grouped Query Attention (GQA)\cite{Ainslie2023GQATG},enabling the models to process longer text sequences while reducing the demands on inference speed and memory usage.
\begin{figure}[htbp]
\centering{\includegraphics[width=0.8\linewidth]{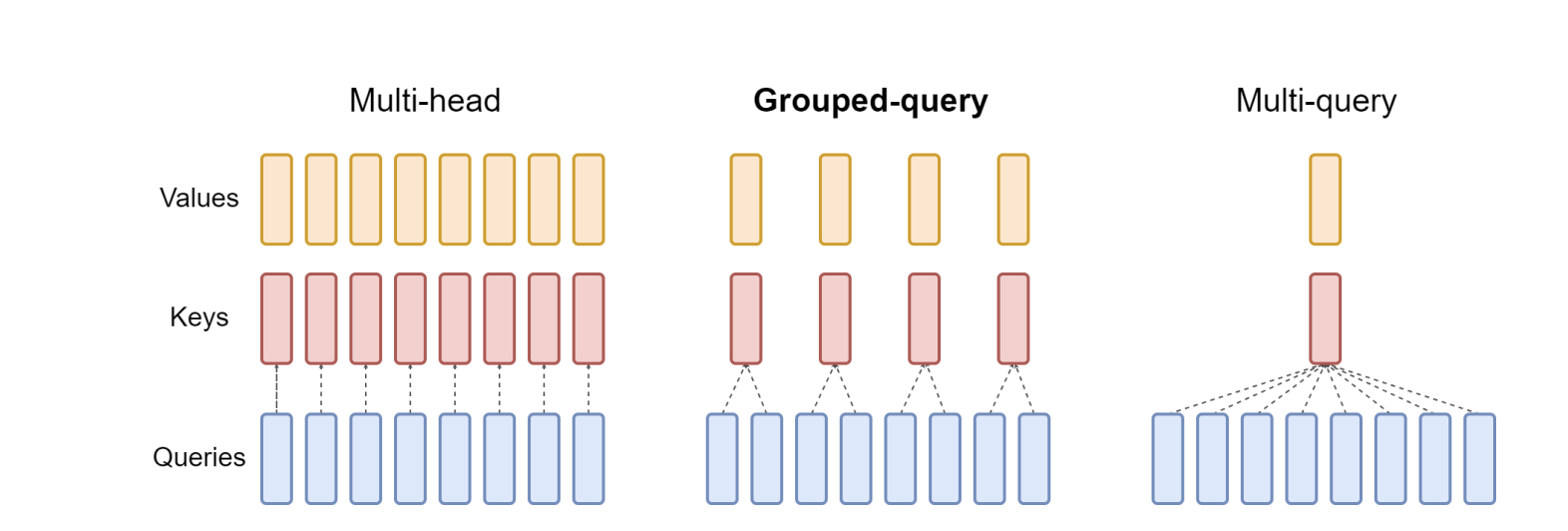}}
\caption{ Overview of grouped-query method. Multi-head attention has H query,key,and value heads. Multi-query
attention shares single key and value heads across all query heads. Grouped-query attention instead shares single
key and value heads for each group of query heads,interpolating between multi-head and multi-query attention.~\cite{Ainslie2023GQATG}}
\label{fig:grouped-query-attention}
\end{figure}
Mistral~\cite{Jiang2023Mistral7} AI’s models are open-source \footnote{Mistral Github:\label{section:Mistral}\url{https://github.com/mistralai/mistral-src}},licensed under the very permissive Apache 2.0 license,allowing free use,modification,and distribution. This makes Mistral AI’s models highly attractive to researchers and developers,as they can freely explore and improve the models without the restrictions of commercial licensing.

The first model from Mistral~\cite{Jiang2023Mistral7} AI is Mistral-7B-v0.1,a pre-trained generative text model with 7 billion parameters. This model outperformed the Llama 2 13B model in all tested benchmarks and showed excellent performance in code,math,and reasoning evaluations. Additionally,Mistral~\cite{Jiang2023Mistral7} AI has also released the Mistral-7B-Instruct-v0.1 model \footnote{Mistral-7B-Instruct-v0.1 Huggingface:\label{section:Mistral-7B-Instruct-v0.1}\url{https://huggingface.co/mistralai/Mistral-7B-Instruct-v0.1}},an instructionally fine-tuned version of the Mistral-7B-v0.1 model \footnote{Mistral-7B-v0.1 Huggingface:\label{section:Mistral-7B-v0.1}\url{https://huggingface.co/mistralai/Mistral-7B-v0.1}} ,fine-tuned using various publicly available dialogue datasets.

The Mistralai/Mistral-8x7B-v0.1\footnote{Mixtral-8x7B-v0.1 Huggingface:\label{section:Mixtral-8x7B-v0.1}\url{https://huggingface.co/mistralai/Mixtral-8x7B-v0.1}} is a version of the large language model developed by the Mistral AI team. This model version consists of 8 sub-models,each with 7 billion parameters,totaling 56 billion parameters. This model structure allows for higher performance and flexibility when processing large amounts of data and complex tasks. Each sub-model can be considered an “expert” focused on handling specific types of data or tasks.

The Mistral-8x7B-v0.1~\ref{section:Mixtral-8x7B-v0.1} model incorporates innovative techniques such as Sliding Window Attention and Grouped Query Attention (GQA)\cite{Ainslie2023GQATG},which enable the model to handle longer text sequences while reducing the demands on inference speed and memory usage. Additionally,this version of the model may also integrate the MOE (Mixture of Experts)~\cite{MOE} architecture,further enhancing the model’s efficiency and flexibility.

The Mistral-8x7B-v0.1~\ref{section:Mixtral-8x7B-v0.1} model is open-source and its reference implementation can be found on GitHub. There are also platforms that offer API services for running this model,such as Replicate,and the Mistral-7B-Instruct-v0.1 model on the Hugging Face platform,which is an instructionally fine-tuned version of the Mistral-7B-v0.1 model~\ref{section:Mixtral-8x7B-v0.1}.

\textbf{ChatGLM3}
ChatGLM3~\footnote{ChatGLM3 Github:\label{section:chatglm3-github}\url{https://github.com/THUDM/ChatGLM3}}~\cite{Zeng2022GLM130BAO} is a new generation of pre-trained dialogue models jointly released by Zhipu AI and the Knowledge Engineering Group (KEG) at Tsinghua University. ChatGLM3-6B~\footnote{ChatGLM3-6B Huggingface:\label{section:ChatGLM3-6B}\url{https://huggingface.co/THUDM/chatglm3-6b}},the open-source model in the ChatGLM3 series,retains many excellent features of the previous two generations,such as smooth dialogue and low deployment threshold,while introducing the following features:

More powerful base model: The base model,ChatGLM3-6B-Base~\footnote{ChatGLM3-6B-base Huggingface:\label{section:ChatGLM3-6B-base}\url{https://huggingface.co/THUDM/chatglm3-6b-base}},uses more diverse training data,more adequate training steps,and more reasonable training strategies. Evaluations on datasets from different perspectives,such as semantics,mathematics,reasoning,code,and knowledge,show that ChatGLM3-6B-Base~\ref{section:ChatGLM3-6B-base} has the strongest performance among pre-trained models under 10 billion parameters.
More complete functional support: ChatGLM3-6B~\ref{section:ChatGLM3-6B-base} adopts a newly designed prompt format,which not only supports normal multi-turn dialogue but also natively supports complex scenarios such as tool invocation (Function Call),code execution (Code Interpreter),and Agent tasks.
More comprehensive open-source sequence: In addition to the dialogue model ChatGLM3-6B\ref{section:ChatGLM3-6B-base},the base model ChatGLM-6B-Base and the long-text dialogue model ChatGLM3-6B-32K~\footnote{ChatGLM3-6B-32K Huggingface:\label{section:ChatGLM3-6B-32K}\url{https://huggingface.co/THUDM/chatglm3-6b-32k}} have also been open-sourced. All weights are fully open for academic research and are also allowed for free commercial use after registration through a questionnaire.

The ChatGLM3~\cite{Zeng2022GLM130BAO} series models excel in dialogue generation,semantic understanding,and knowledge integration,and are suitable for various application scenarios,including but not limited to chatbots,customer service assistants,educational tutoring,etc.

\subsection{In-Context Learning (ICL)} In-Context Learning represents a paradigm in NLP where language models are trained to understand and execute tasks through a set of examples or instructions. This approach hinges on the model's ability to learn from task-relevant,analogous samples. The methodology involves curating examples in a specific format and then concatenating these with the current input through a prompt,which collectively serves as the input for the language model.

The advantages of ICL are manifold:
\begin{itemize}
\item \textbf{Interpretability:} The use of natural language for crafting examples offers a highly interpretable means of interacting with language models,thereby fostering clearer communication.
\item \textbf{Analogy-based Learning:} This method mirrors the human decision-making process of learning through analogy,where learning is extrapolated from one instance to broader applications.
\item \textbf{Reduced Computational Cost:} In comparison to supervised learning,ICL negates the need for model retraining for new tasks,significantly lowering the computational costs associated with adapting to new tasks.
\end{itemize}
In essence,ICL is a transformative approach that leverages the inherent capabilities of language models for efficient and intuitive task adaptation,making it a notable advance in the field of AI and NLP.

The paper ''Rethinking the Role of Demonstrations: What Makes In-Context Learning Work?''~\cite{Min2022RethinkingTRICL} find that real labels in demonstrations are not essential. Randomly replacing labels in demonstrations has minimal impact on performance for a range of classification and multiple-choice tasks,even across 12 different models,including GPT-3. The other aspects of demonstrations,such as examples of label space,distribution of input text,and overall sequence format,are the crucial drivers of task performance. This discovery offers a new perspective on how In-Context Learning (ICL) functions and suggests that large language models learn limited knowledge solely through inference. This research opens new directions for future exploration,especially in understanding the content that large language models can learn through reasoning.

The paper ''An Explanation of In-context Learning as Implicit Bayesian Inference''~\cite{Xie2021AnEOICL} examines how large language models like GPT-3 perform tasks through In-Context Learning (ICL) without explicit task-specific pre-training. Its main contributions include a theoretical framework for understanding ICL when pre-trained documents have long-term coherence,necessitating LMs to infer implicit document-level concepts for coherent token generation. The study shows ICL occurs even with distribution mismatches between prompts and pre-training data. It introduces a small synthetic dataset (GINC),where both Transformer and LSTM models demonstrate ICL. Findings indicate that larger models improve contextual accuracy,and example sequence affects performance. This paper offers a new perspective on ICL,suggesting it occurs through implicit Bayesian inference,providing a theoretical foundation for understanding how large LMs use context cues to``anchor'' concepts learned previously for ICL tasks.

\subsection{Chain-of-Thought Reasoning (CoT)} CoT~\cite{Wei2022ChainOTCoT} is an advanced reasoning method that solves complex problems by breaking them down into a series of logical steps. The key to this method is that it allows models to process multi-step reasoning problems through a series of intermediate steps,allocating more computational resources for problems that require deep reasoning.This process can be represented by the following illustrative mathematical formula:
\begin{equation}
\begin{split}
\text{CoT}(Q) &= \text{Step}_1(\text{Init}) \rightarrow \text{Step}_2(\text{Step}_1) \rightarrow \cdots \\
&\rightarrow \text{Step}_n(\text{Step}_{n-1}) \rightarrow \text{Answer}
\end{split}
\label{eq-cot}
\end{equation}
In Equation (~\ref{eq-cot}):

- \( Q \) is the problem we are trying to solve.

- \( \text{Init} \) represents the initial state or known information of the problem-solving process.

- \( \text{Step}_i \) denotes the reasoning process at the \( i \)th step,which is based on the result of the previous step \( \text{Step}_{i-1} \).

- \( \text{Answer} \) is the final answer,derived from the result of the last step of reasoning \( \text{Step}_n \).

Key features of CoT~\cite{Wei2022ChainOTCoT} include:
\begin{itemize}
\item \textbf{Step-by-step reasoning:} The CoT~\cite{Wei2022ChainOTCoT} method generates a reasoning chain through step-by-step reasoning in natural language before presenting the final answer.
\item \textbf{Explainability:} CoT~\cite{Wei2022ChainOTCoT} provides a window that allows us to see how the model solves problems step by step,increasing the transparency and explainability of the reasoning process.
\item \textbf{Broad applicability:} CoT~\cite{Wei2022ChainOTCoT} can be applied to mathematical problems,common sense reasoning,symbolic operations,and more,even suitable for any problem that needs to be solved through language.
\end{itemize}

\begin{figure}[htbp]
\centering{\includegraphics[width=0.8\linewidth]{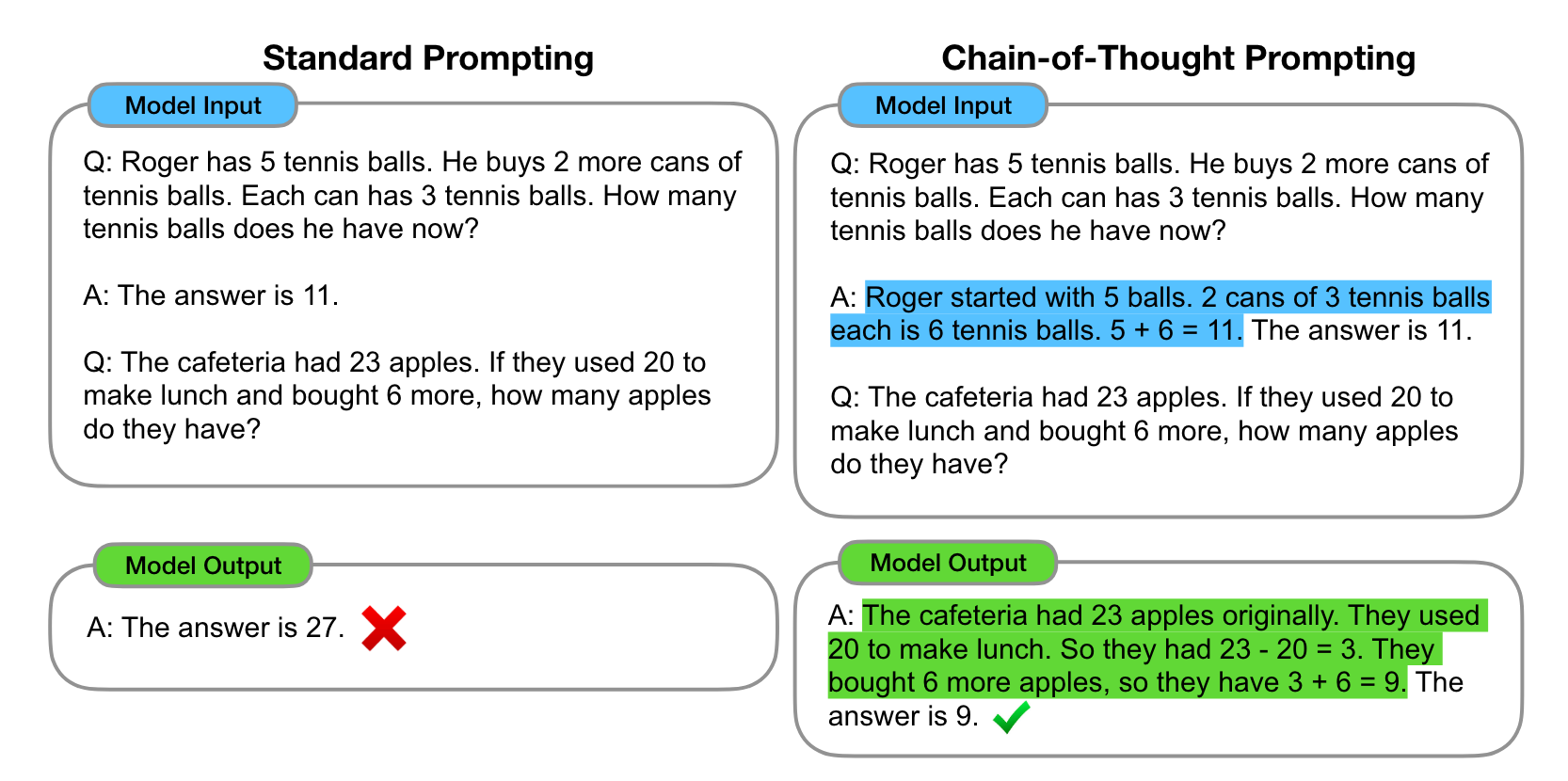}}
\caption{Chain-of-thought prompting enables large language models to tackle complex arithmetic, commonsense, and symbolic reasoning tasks. Chain-of-thought reasoning processes are highlighted.~\cite{Wei2022ChainOTCoT}}
\label{fig:o-cot}
\end{figure}

\textbf{Implementation of CoT:} The implementation of CoT~\cite{Wei2022ChainOTCoT} relies on the capabilities of large language models,which can guide reasoning abilities even without directly provided samples.
The inception of CoT~\cite{Wei2022ChainOTCoT} is attributed to two important papers,with “Chain-of-Thought Prompting Elicits Reasoning in Large Language Models”~\cite{Wei2022ChainOTCoT}(Fig.~\ref{fig:o-cot}) being the seminal work authored by Jason Wei and others. The second paper,“Large Language Models are Zero-Shot Reasoners,”~\cite{Kojima2022LargeLM} further simplifies the CoT process by adding \textbf{“Let’s think step by step”} to the prompt,guiding the model to generate reasoning chains automatically.

\textbf{Advantages of CoT:}
CoT~\cite{Wei2022ChainOTCoT} not only enhances the logical reasoning capabilities of models but also provides better result explainability,making the reasoning process more credible.
CoT~\cite{Wei2022ChainOTCoT} significantly improves the reasoning abilities of large language models through step-by-step thinking.

\begin{figure}[htbp]
\centering{\includegraphics[width=0.8\linewidth]{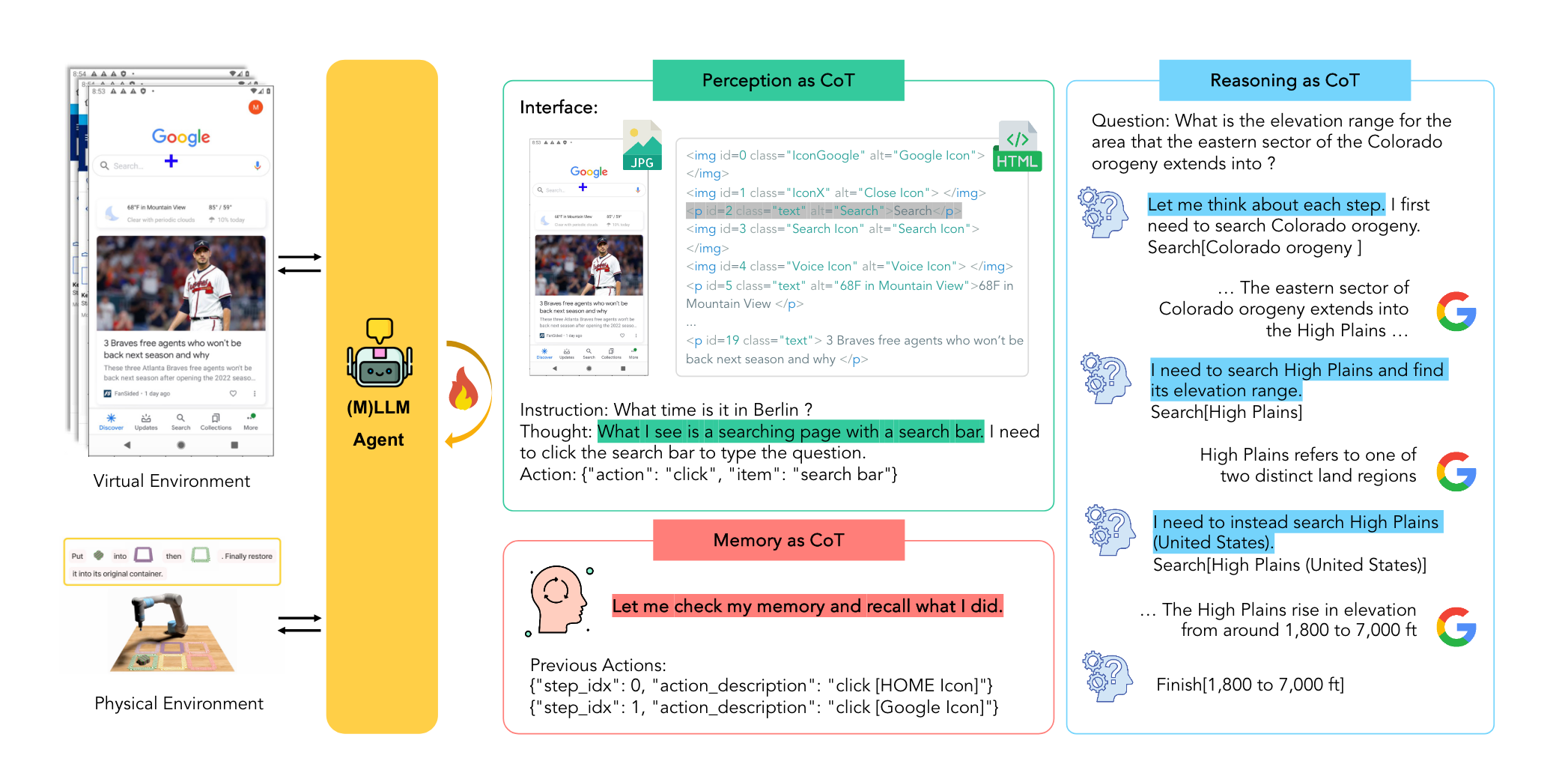}}
\caption{An overview of language agent framework empowered with the chain-of-thought (CoT) mechanism in perception, memory, and reasoning.~\cite{Zhang2023IgnitingLI}}
\label{fig:cot-of-llms}
\end{figure}

The paper “Igniting Language Intelligence: The Hitchhiker’s Guide From Chain-of-Thought Reasoning to Language Agents\cite{Zhang2023IgnitingLI}” delves into the significant advancements of large language models (LLMs) in the field of language intelligence,especially their exceptional performance in complex reasoning tasks. The paper emphasizes the crucial role of Chain-of-Thought (CoT) reasoning techniques in enhancing the efficiency of LLMs(Fig.~\ref{fig:cot-of-llms}) in handling complex reasoning tasks. CoT reasoning not only improves reasoning performance but also enhances explainability,controllability,and flexibility.

\subsection{LLMs Agent}

\begin{figure}[htbp]
\centering{\includegraphics[width=0.8\linewidth]{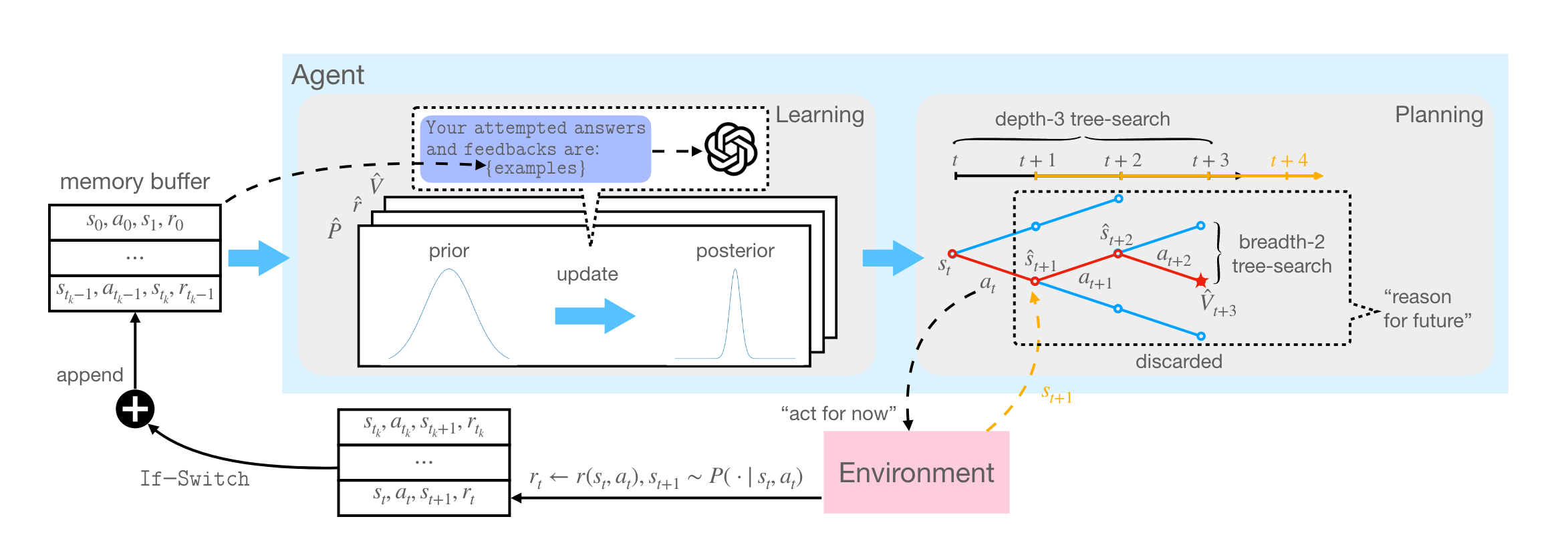}}
\caption{Illustration of the RAFA (``reason for future, act for now'') framework.~\cite{Liu2023ReasonFFXAgent}}
\label{fig:xagent}
\end{figure}

\textbf{X-Agent} X-Agent~\cite{Liu2023ReasonFFXAgent}(Fig.~\ref{fig:xagent}) is an Artificial Intelligence Agent,a smart entity capable of perceiving its environment,making decisions,and executing actions. Powered by large model technology,X-Agent~\cite{Liu2023ReasonFFXAgent} enables high automation in executing and handling professional or complex work tasks through natural language interaction,significantly freeing up human effort.

An AI Agent,also known as an``intelligent agent'' aims to assist users in completing a variety of tasks through natural language interaction. These tasks include knowledge queries,content generation,business processing,and management decision-making. Combining the powerful computational abilities of large models with specialized domain knowledge,it can provide precise services in specific industry scenarios.

The development of X-Agent~\cite{Liu2023ReasonFFXAgent} is rooted in the surge of large language models,which has rapidly advanced AI Agent-related research. It represents a primary exploration path toward General Artificial Intelligence (AGI). The vast training datasets of large models contain extensive human behavior data,laying a solid foundation for simulating human-like interactions. As these models continue to grow in size,they have developed various abilities similar to human thought processes,such as context learning,reasoning,and cognitive chaining.

\textbf{AutoGPT} Auto-GPT~\cite{Yang2023AutoGPTFO} is an open-source application\footnote{AutoGPT:\label{section:AutoGPT}\url{https://github.com/Significant-Gravitas/AutoGPT}} that leverages OpenAI's large language model,GPT-4~\cite{OpenAI2023GPT4TR},to autonomously execute multi-step tasks which would typically require user prompts when directly using GPT-4. Auto-GPT allows the GPT model to operate independently,without the need for manual prompting for each action. It is capable of generating human-like text,answering questions,translating languages,summarizing text,and providing recommendations. Auto-GPT is publicly available on GitHub\textsuperscript{~\ref{section:AutoGPT}},but requires some programming experience to use,as it runs on Python and needs OpenAI and Pinecone API keys.

Key features of Auto-GPT~\cite{Yang2023AutoGPTFO} include:
\begin{enumerate}
\item \textbf{Autonomy:} The ability to operate without continuous user prompts.
\item \textbf{Goal-Oriented Functionality:} Capable of rewriting its own code,autonomously searching the internet,and performing tasks such as saving files to a computer.
\item \textbf{Versatility:} Able to generate text,answer questions,translate languages,summarize text,and provide recommendations.
\item \textbf{User-Friendly Interface:} While it requires some technical knowledge,its level of independence allows it to operate without in-depth programming expertise.
\item \textbf{Long-Term and Short-Term Memory:} Capable of storing and recalling information as needed,particularly useful in generating longer texts.
\end{enumerate}
Auto-GPT~\cite{Yang2023AutoGPTFO} remains an experimental project,requiring programming skills for practical deployment in any substantial capacity. Its key features include autonomy,goal-oriented functionality,versatility,and a user-friendly interface. AutoGPT~\cite{Yang2023AutoGPTFO} also provides internet access for data collection and research,and has improved long-term and short-term memory capabilities.

\begin{figure}[htbp]
\centering{\includegraphics[width=0.8\linewidth]{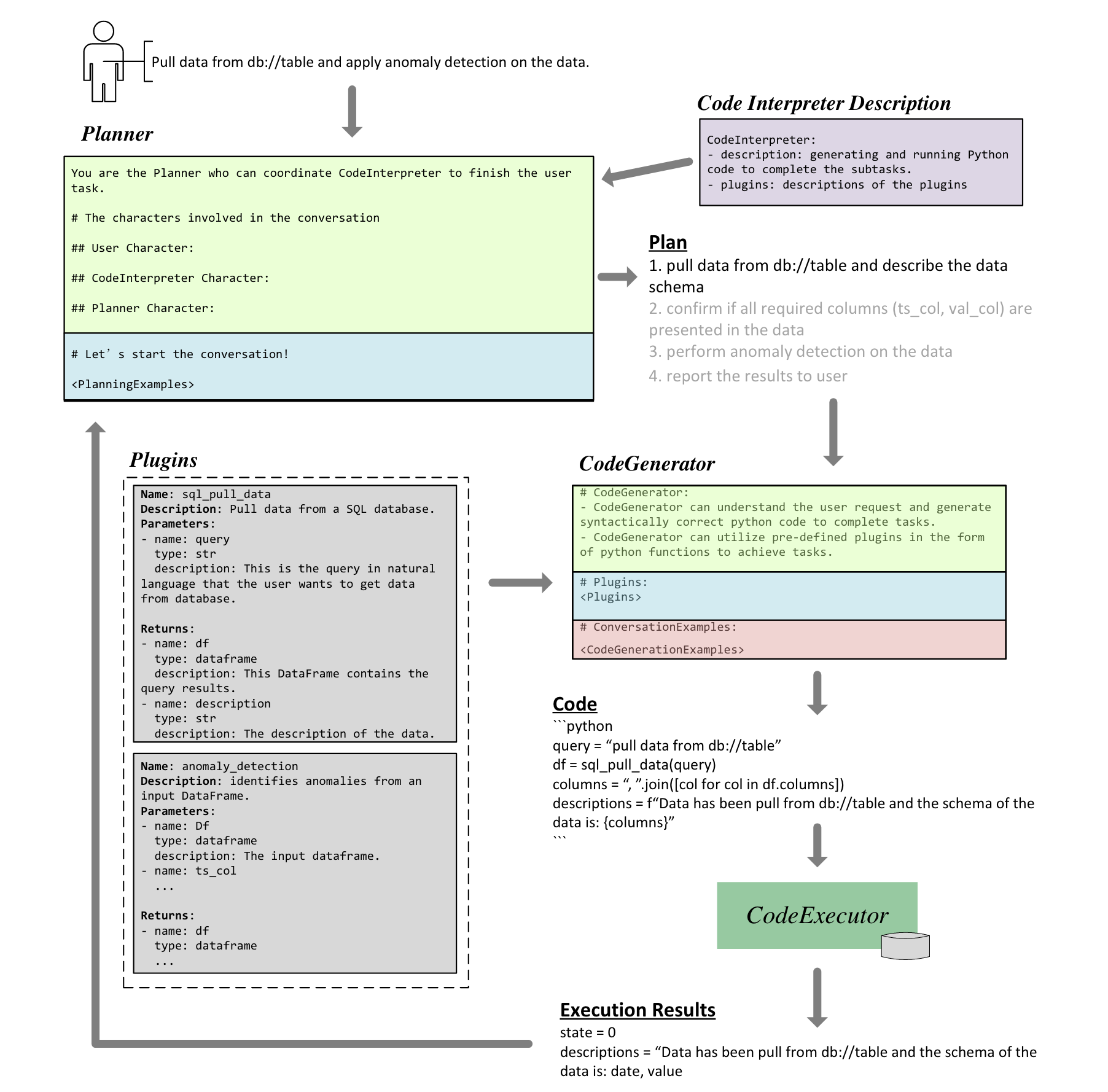}}
\caption{Workflow of TaskWeaver.~\cite{Qiao2023TaskWeaverAC}}
\label{fig-WorkflowOfTaskWeaver}
\end{figure}

\textbf{TaskWeaver} TaskWeaver~\cite{Qiao2023TaskWeaverAC} is a code-first AI agent framework developed by Microsoft,designed specifically for seamless planning and execution of data analysis tasks. It interprets user requests through code snippets and efficiently coordinates a series of plugins,which exist as functions,to perform data analysis tasks. Here’s a detailed explanation of TaskWeaver’s~\cite{Qiao2023TaskWeaverAC} architecture(Fig.~\ref{fig-TaskWeaver_Overview},~\ref{fig-WorkflowOfTaskWeaver}
):
\begin{figure}[htbp]
\centering{\includegraphics[width=0.8\linewidth]{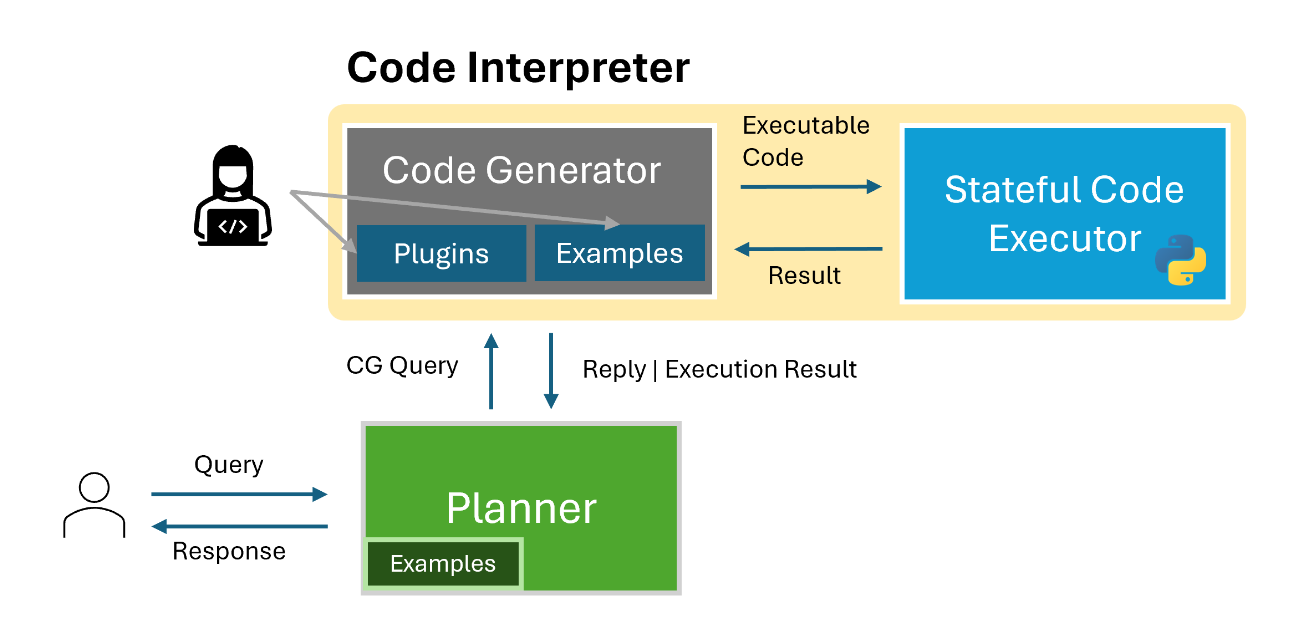}}
\caption{The overview of TaskWeaver.~\cite{Qiao2023TaskWeaverAC}}
\label{fig-TaskWeaver_Overview}
\end{figure}
\begin{itemize}
\item \textbf{User Request Interpretation:}
TaskWeaver~\cite{Qiao2023TaskWeaverAC} interprets user requests through code snippets. Users can describe the data analysis tasks they want to perform by writing code snippets. These snippets are then interpreted by TaskWeaver and transformed into executable tasks.
\item \textbf{Plugin System:} One of the core features of TaskWeaver~\cite{Qiao2023TaskWeaverAC} is its plugin system. Users can encapsulate their algorithms into plugins (in the form of Python functions) and orchestrate them to complete complex tasks. These plugins can be data processing,machine learning models,visualization tools,etc.
\item \textbf{Rich Data Structure Support:} Unlike other frameworks,TaskWeaver~\cite{Qiao2023TaskWeaverAC} allows the use of rich data structures in Python,such as DataFrames,rather than being limited to text strings. This provides greater flexibility and efficiency in handling complex data analysis tasks.
\item \textbf{Stateful Dialogue:} TaskWeaver~\cite{Qiao2023TaskWeaverAC} supports stateful dialogue,meaning it can remember the context of the conversation and use this information to improve the user experience. This is particularly useful for tasks that require multi-step input or depend on previous interactions.
\item \textbf{Code Verification:} TaskWeaver~\cite{Qiao2023TaskWeaverAC} has a code verification feature designed to detect potential issues in the generated code and provide suggestions for fixes,ensuring the safety and reliability of the code.
\item \textbf{Ease of Use and Debugging:} TaskWeaver~\cite{Qiao2023TaskWeaverAC} provides a set of example plugins and tutorials to help users get started,allowing them to easily create their own plugins based on the examples. Additionally,TaskWeaver~\cite{Qiao2023TaskWeaverAC} provides detailed logs to help users understand what happens during the invocation of LLM,code generation,and execution processes,making the debugging process simpler.
\item \textbf{Open Source Code:}
The code for TaskWeaver~\cite{Qiao2023TaskWeaverAC} is open source\footnote{TaskWeaver Github:\label{section:TaskWeaver}\url{https://github.com/microsoft/TaskWeaver}},meaning users can access,modify,and contribute to the code. This provides a platform for the community to collaboratively improve and expand the functionality of the framework.
\end{itemize}
Overall,TaskWeaver’s~\cite{Qiao2023TaskWeaverAC} architecture offers a powerful and flexible framework for creating intelligent dialogue agents capable of handling complex tasks and adapting to domain-specific scenarios. 

\textbf{AutoGen} AutoGen~\cite{Wu2023AutoGenEN} is a framework developed by Microsoft that allows developers to create applications through conversational agents that can leverage the capabilities of large language models (LLMs). These agents can converse with each other to complete complex tasks,while also being able to interact with human users.
\begin{figure}[htbp]
\centering{\includegraphics[width=0.8\linewidth]{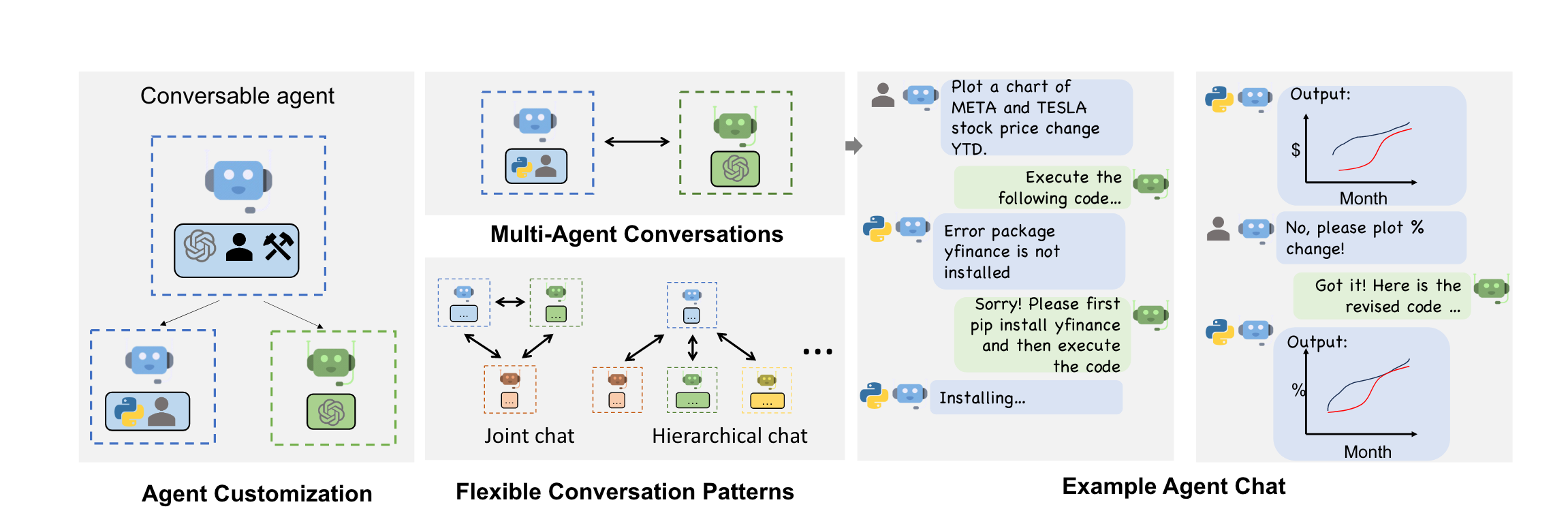}}
\caption{AutoGen enables diverse LLM-based applications using multi-agent conversations. (Left) AutoGen agents are conversable,customizable,and can be based on LLMs,tools,humans,or even a combination of them. (Top-middle) Agents can converse to solve tasks. (Right) They can form a chat,potentially with humans in the loop. (Bottom-middle) The framework supports flexible conversation patterns.~\cite{Wu2023AutoGenEN}}
\label{fig:AutoGen}
\end{figure}
In AutoGen~\cite{Wu2023AutoGenEN},agents are customizable and can be adjusted according to the specific needs of an application. They are conversational,meaning they can communicate through natural language. Additionally,the AutoGen~\cite{Wu2023AutoGenEN} framework supports multi-agent dialogue,which means multiple agents can participate in a task simultaneously,each taking on different roles and responsibilities.

AutoGen~\cite{Wu2023AutoGenEN} is designed to simplify and optimize workflows using large language models. It automates and orchestrates the dialogue between agents,making it easier for developers to build and maintain complex LLM applications. Furthermore,AutoGen~\cite{Wu2023AutoGenEN} provides enhanced LLM inference APIs,which help improve performance and reduce costs.

Overall,AutoGen~\cite{Wu2023AutoGenEN} is a powerful tool that helps developers fully utilize the potential of LLMs to create smarter and more interactive applications.

\textbf{Stanford Virtual Town} Generative Agents: Interactive Simulacra of Human Behavior~\cite{Park2023GenerativeAI}\footnote{Stanford Virtual Town:\label{section:generative_agents}\url{https://github.com/joonspk-research/generative_agents}} introduces a computational agent based on language models and natural language, which can simulate credible human behavior. These agents can be used for interactive applications, such as immersive environments, interpersonal communication practice spaces, prototype design tools, etc. The paper describes an architecture that extends a large language model into a complete agent, using natural language to store the agent’s experiences, synthesizing them into high-level reflections, and dynamically retrieving them to plan behavior. The paper also implements an interactive sandbox environment inspired by The Sims, allowing users to interact with 25 agents in a small town using natural language.

\begin{figure}[htbp]
\centering{\includegraphics[width=0.8\linewidth]{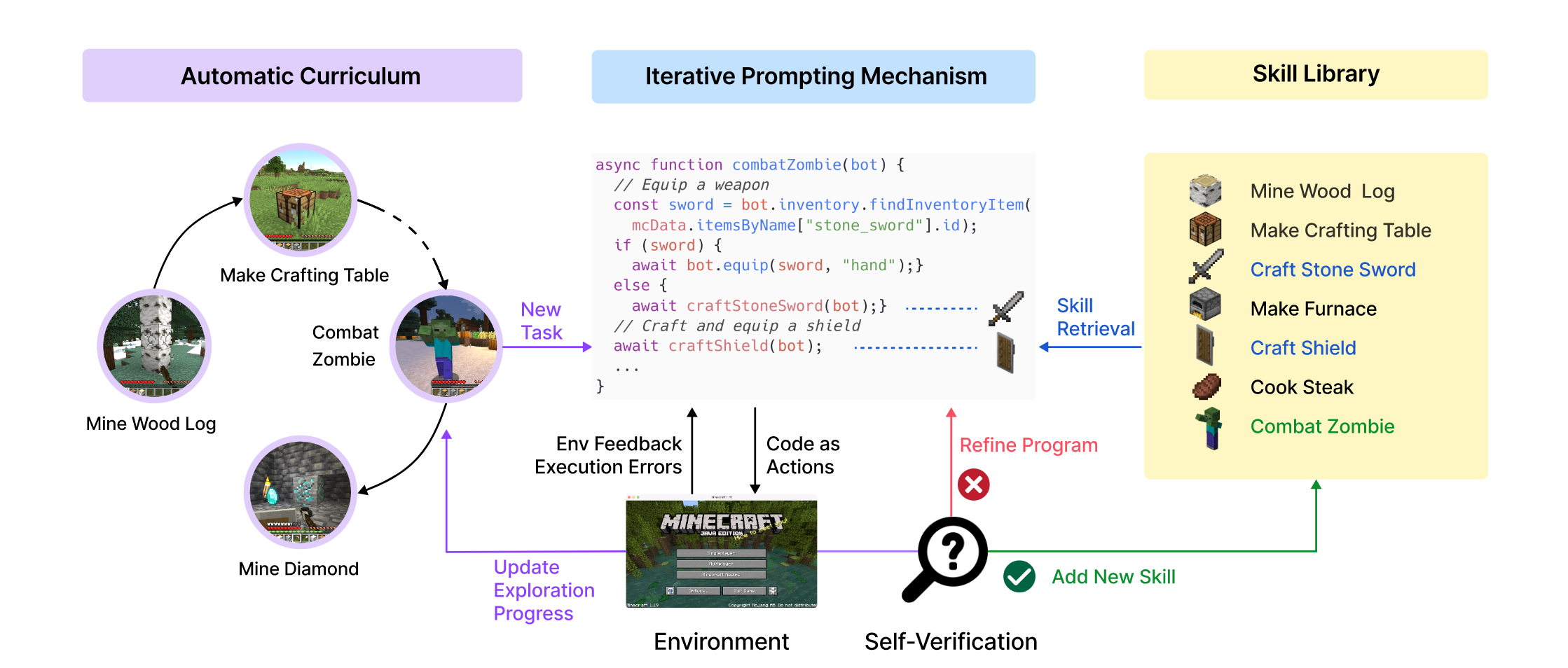}}
\caption{VOYAGER consists of three key components: an automatic curriculum for open-ended exploration, a skill library for increasingly complex behaviors, and an iterative prompting mechanism that uses code as action space.~\cite{Wu2023AutoGenEN}}
\label{fig:voyager}
\end{figure}

\textbf{NVIDIA AI Agent Voyager} NVIDIA AI Agent Voyager~\cite{Wang2023VoyagerAO}(Fig.~\ref{fig:voyager})\footnote{Voyager:\label{section:Voyager}\url{https://github.com/MineDojo/Voyager}} is an embodied lifelong learning agent based on large language models (LLMs), which can continuously explore, learn and develop new skills in the open world game Minecraft. Voyager~\cite{Wang2023VoyagerAO} leverages the world knowledge and programming ability of GPT-4, and generates executable code through natural language, to complete various tasks proposed by automatic curriculum. Voyager can also store the learned code in a skill library, for reuse or combination in similar situations. The advantage of Voyager is that it can achieve efficient exploration, interpretable skills and zero-shot generalization without human intervention.

\subsection{ReAct}
ReAct~\cite{Yao2022ReActSR} is an application paradigm for large language models (LLMs) that combines reasoning and acting processes to enhance the model’s accuracy and interpretability when solving complex problems. The core idea of the ReAct~\cite{Yao2022ReActSR} paradigm is to have the LLM not just generate answers,but to find answers through an explicit reasoning process and corresponding actions.

In ReAct~\cite{Yao2022ReActSR},the model first reasons,using its semantic understanding capabilities to analyze the problem. Then,the model takes action,which may include capabilities beyond the model itself,such as searching,computing,or user-defined actions. This process can be iteratively used in a single interaction to solve more complex problems.

For example,to solve a problem about an Apple remote control,the ReAct~\cite{Yao2022ReActSR} method would have the model first search for the earliest software that the Apple remote could control,then based on the information obtained (such as the “Front Row” software),proceed with further searches or actions,ultimately arriving at the correct answer (such as “keyboard function keys”).

The advantage of ReAct~\cite{Yao2022ReActSR} is that it generates task-solving trajectories similar to human reasoning and action,which not only improves the accuracy of the answers but also enhances interpretability,allowing us to understand how the model arrived at the answer. This is particularly useful in situations that require debugging and human intervention.

Furthermore,ReAct~\cite{Yao2022ReActSR} can also improve accuracy through fine-tuning,similar to the human process of “internalizing” knowledge. By inputting correct reasoning and action trajectories into the LLM for fine-tuning,accuracy can be significantly improved. This method shows significant advantages compared to other traditional methods.

\subsection{Code Interpreter} 
Large Model Code Interpreter typically refers to a tool that uses large language models (LLMs) to interpret and execute code. These tools can understand instructions in programming languages and translate them into actions that machines can execute. The core feature of Large Model Code Interpreters is that they often combine deep learning algorithms,especially the Transformer architecture,to understand and generate code.

The advantage of these interpreters is that they can handle complex programming tasks such as code generation,code interpretation,code optimization,and error detection. They provide a deep understanding of code,helping developers to comprehend and maintain codebases more quickly.

For example,Denigma\footnote{Denigma:\label{section:Denigma}\url{https://denigma.app/}} is an AI tool that can explain code in human-understandable English.

Open Interpreter\footnote{Open Interpreter:\label{section:open-interpreter}\url{https://github.com/KillianLucas/open-interpreter}} is an open source tool that allows language models to run code on your computer. It is a localized implementation of OpenAI’s Code Interpreter, which can directly interpret and generate code from natural language descriptions. You can interact with Open Interpreter through a ChatGPT-like interface in the terminal, executing code in various languages such as Python, Javascript, Shell, etc. Open Interpreter can also access the internet, use any package or library, and is not limited by time or file size.

The application range of Large Model Code Interpreters is broad,not limited to code interpretation but also including code generation,natural language processing,machine translation,and more. These tools help developers save time,accelerate the development process,and improve code quality by understanding and executing code.

\subsection{LangChain}
LangChain\footnote{LangChain:\label{section:LangChain}\url{https://github.com/langchain-ai/langchain}}~\footnote{Chat LangChain:\label{section:chat-langchain}\url{https://chat.langchain.com/}} is an open-source framework designed to simplify the process of creating applications with large language models (LLMs). It provides a standard interface for ``chaining'' different components together to create more advanced use cases around LLMs. The core idea of LangChain is that we can create more complex applications,such as chatbots,generative question answering (GQA),document analysis and summarization,etc.,by combining multiple components.

Key concepts of LangChain include:

\textbf{Components:} Modular building blocks that can be used to build powerful applications,including LLM wrappers,prompt templates,and indices.

\textbf{Chains:} Chains allow us to combine multiple components to solve specific tasks,making the implementation of complex applications more modular,easier to debug,and maintain.

\textbf{Agents:} Agents allow LLMs to interact with their environment,such as performing specific actions using external APIs.

LangChain has a wide range of applications and can be used to build various LLM-based applications. For example,it can be used to build chatbots that interact naturally with users,analyze code to find potential errors or security vulnerabilities,answer questions using various sources,and even perform machine translation,etc.“Automating Customer Service using LangChain: Building custom open-source GPT Chatbot for organizations”~\cite{Pandya2023AutomatingCS} - This paper discusses how to use LangChain to automate customer service and build customized open-source GPT chatbots for organizations.“PromptChainer: Chaining Large Language Model Prompts through Visual Programming”~\cite{Wu2022PromptChainerCL}  - explores a method of chaining large language model prompts through visual programming.

\section{Task-Specific Adaptation Fine-Tuning}
Task-specific fine-tuning is a process of adjusting large pre-trained models,such as BERT~\cite{Devlin2019BERTPO},to optimize their performance for specific natural language processing (NLP) tasks. BERT,introduced by Google in 2018,is based on the Transformer architecture,particularly its encoder component.

\begin{figure}[htbp]
\centering{\includegraphics[width=0.8\linewidth]{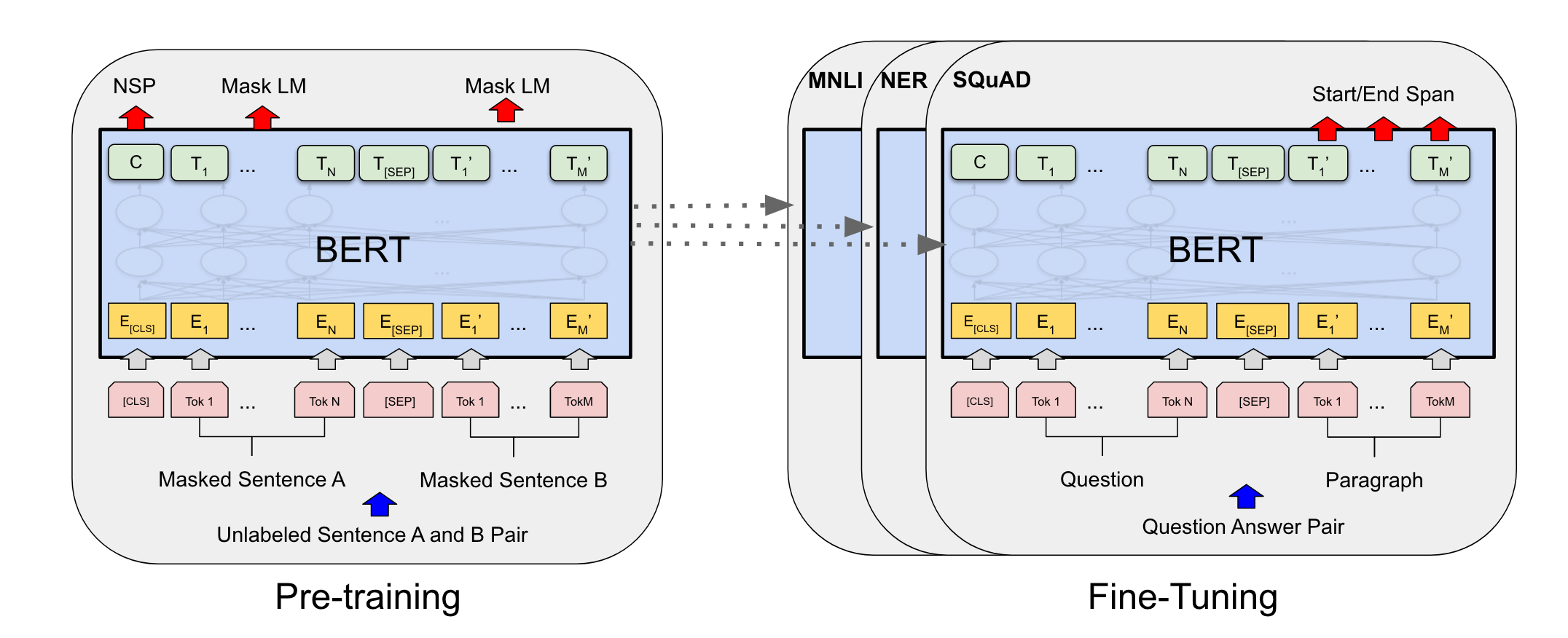}}
\caption{Overall pre-training and fine-tuning procedures for BERT.~\cite{Devlin2019BERTPO}}
\label{fig-bert}
\end{figure}

Task-specific fine-tuning mainly consists of two stages (Fig.~\ref{fig-bert}).
\begin{itemize}
    \item \textbf{Pre-trained BERT Model:} Initially,the BERT model is pre-trained on a vast corpus to learn rich language representations. It is trained to predict masked words in sentences (Masked Language Model,MLM) and to anticipate the next sentence (Next Sentence Prediction,NSP).
    \item \textbf{Fine-Tuning for Specific Tasks:} After pre-training,the BERT model can be further adjusted through a fine-tuning (Fine-Tuning) process to adapt to specific downstream tasks like sentiment analysis,question-answering,text classification,etc.
\end{itemize}

\textbf{Steps in the Fine-Tuning Process: }

\begin{itemize}
    \item \textbf{Task-Specific Data:} First,training data specific to the task at hand needs to be collected and prepared. For example,for a sentiment analysis task,this might include sentences labeled with positive or negative sentiments.
    \item \textbf{Modifying the Model Structure:} Depending on the task,some modifications to the output layer of the BERT model might be necessary. For instance,for classification tasks,a fully connected layer is typically added after the last layer of BERT.
    \item \textbf{Fine-Tuning Training:} The BERT model is trained using the data for the specific task. During this process,all parameters of BERT are updated based on the new task's data.
    \item \textbf{Evaluation and Optimization:} The model's performance should be evaluated on a validation set during fine-tuning. Based on the assessment,adjustments might be needed in training parameters like learning rate,batch size,etc.
\end{itemize}

The advantages of task-specific fine-tuning are as follows: 

\begin{itemize}
    \item \textbf{Efficiency:} As BERT has already been pre-trained on a large amount of data,the fine-tuning process is generally much quicker than training from scratch.
    \item \textbf{Flexibility:} The same pre-trained model can be adapted through fine-tuning for a variety of different NLP tasks.
    \item \textbf{Improved Performance:} Compared to models without pre-training,pre-trained BERT demonstrates superior performance on many NLP tasks.
\end{itemize}

Overall,task-specific fine-tuning is an effective approach to leverage large pre-trained models like BERT for efficient and high-quality performance across various NLP tasks.

\section{Few-Shot Learning And Meta Learning}

Few-Shot Learning is a method of training machine learning models with only a small amount of labeled data. It is particularly suited for scenarios where labeled data is scarce or expensive. In the context of large pre-trained models like BERT~\cite{Devlin2019BERTPO} or GPT~\cite{Radford2018ImprovingLU},few-shot learning typically involves the following key steps:
\begin{itemize}
    \item \textbf{Pre-training:} The model is first pre-trained on a large dataset to learn rich feature representations and language patterns.
    \item \textbf{Fine-Tuning:} Then,the model undergoes fine-tuning on a small amount of labeled data. This step adjusts the model parameters to better suit a specific task.
    \item \textbf{Meta-Learning or Transfer Learning:} Few-shot learning often utilizes Meta-Learning or Transfer Learning techniques,enabling the model to rapidly adapt to new tasks.
\end{itemize}

Model-Agnostic~\cite{Finn2017ModelAgnosticMF} introducing the concept of Model-Agnostic Meta-Learning (MAML)~\cite{Finn2017ModelAgnosticMF},also known as learn to learn, refers to enabling machines to ``learn how to learn'' empowering them with the capability of learning. MAML~\cite{Finn2017ModelAgnosticMF} aims to quickly adapt to new tasks even with minimal data.

\begin{algorithm}
\caption{Model-Agnostic Meta-Learning}
\label{alg:maml}
\begin{algorithmic}[1] 
\Require $p(\mathcal{T})$: distribution over tasks
\Require $\alpha,\beta$: step size hyperparameters
\State randomly initialize $\theta$
\While{not done}
    \State Sample batch of tasks $\mathcal{T}_i \sim p(\mathcal{T})$
    \ForAll{$\mathcal{T}_i$}
        \State Evaluate $\nabla_{\theta} \mathcal{L}_{\mathcal{T}_i} (f_{\theta})$ with respect to $K$ examples
        \State Compute adapted parameters with gradient descent: $\theta'_i = \theta - \alpha \nabla_{\theta} \mathcal{L}_{\mathcal{T}_i} (f_{\theta})$
    \EndFor
    \State Update $\theta \gets \theta - \beta \nabla_{\theta} \sum_{\mathcal{T}_i \sim p(\mathcal{T})} \mathcal{L}_{\mathcal{T}_i} (f_{\theta'_i})$
\EndWhile
\end{algorithmic}
\end{algorithm}
Principles of MAML(Algorithm:\ref{alg:maml}):
\begin{itemize}
\item \textbf{Model Agnosticism:} A key feature of MAML~\cite{Finn2017ModelAgnosticMF} is its agnosticism to specific model architectures,meaning it can be applied to any type of neural network,whether used for computer vision,natural language processing,or other tasks.
\item \textbf{Meta-Learning Task:} MAML’s goal is to train a model that can rapidly adapt to a new task after observing only a few samples. This is achieved by training the model across a variety of different tasks to find a good initialization of model parameters that requires only a few gradient update steps for good performance on a new task.
\item \textbf{Two-Level Optimization:} MAML involves two levels of optimization. The inner optimization adjusts the model parameters for a specific task. the outer optimization updates the initial parameters so that the model can better adapt to new tasks.
\end{itemize}

How to Fine-Tune with MAML:
\begin{itemize}
\item \textbf{Initial Training:} First,the model is trained across multiple tasks to find a good point of parameter initialization. Each task has its own dataset,and the model undergoes short-term training on each task (i.e.,inner optimization).
\item \textbf{Gradient Updates:} The model undergoes gradient updates on each task to optimize performance. These updated results are then used to adjust the initial parameters (i.e.,outer optimization).
\item \textbf{Adapting to New Tasks:} When faced with a new task,the model starts with the previously found initial parameters and quickly adapts to the new task with a few gradient updates.
\end{itemize}

The advantage of MAML is that it allows a model to leverage knowledge learned across a range of tasks,thereby enabling rapid adaptation to new,unseen tasks. This is particularly valuable in few-shot learning scenarios,as it reduces the need for collecting large amounts of labeled data.

\section{Knowledge Distillation And Transfer Learning}

In the domain of Knowledge Distillation (KD),the seminal work by Hinton et al. titled “Distilling the Knowledge in a Neural Network” (2015)~\cite{Hinton2015DistillingTK} laid the foundational framework for this concept. Furthermore,TinyBERT~\cite{Jiao2019TinyBERTDB} serves as an exemplar demonstration of fine-tuning BERT models for natural language understanding tasks using knowledge distillation.

\begin{figure}[htbp]
\centering{\includegraphics[width=0.8\linewidth]{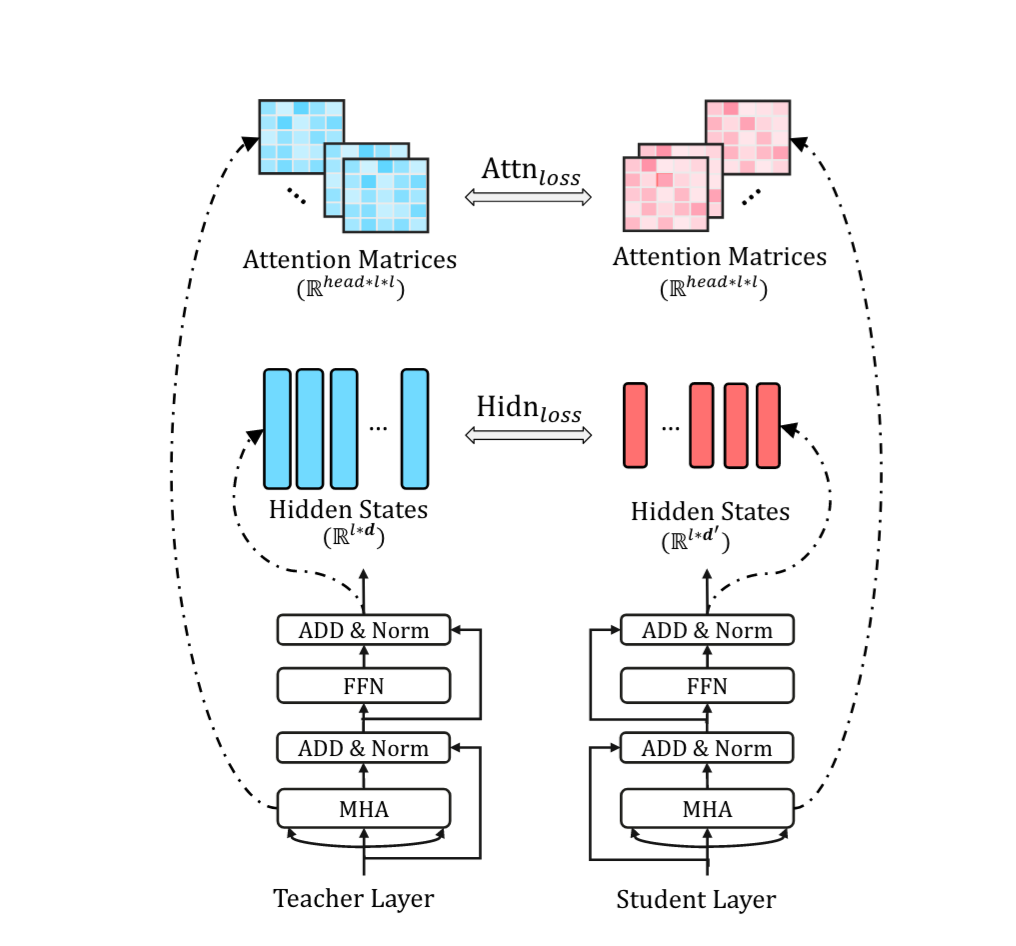}}
\caption{Transformer-layer distillation.~\cite{Jiao2019TinyBERTDB}}
\label{fig-tinybert}
\end{figure}

The principle of Knowledge Distillation (Fig.~\ref{fig-tinybert}) and its application in fine-tuning can be academically articulated as follows:

\begin{itemize}
    \item \textbf{Initially:} the teacher model — a complex model pre-trained on a large dataset — has acquired a profound knowledge of accurately categorizing input data. This knowledge is manifested not solely in the model’s final classification decisions but also in the probabilistic output distribution of the decision-making process. Upon the completion of the teacher model’s training,its output probability distribution,i.e.,the soft labels,are utilized as carriers of information.
    \item \textbf{Subsequently:} the student model — a simpler,less parameter-intensive model — is trained by emulating the soft outputs of the teacher model. During this process,a specific loss function,such as the Kullback-Leibler divergence~\cite{kl},is employed to quantify the disparity between the outputs of the student and teacher models. In training the student model,the distillation loss,which includes minimizing traditional classification loss (e.g.,cross-entropy),is incorporated,enabling the student model to learn the teacher model’s behaviors.
\end{itemize}

Knowledge distillation can be performed at the pre-training stage or the fine-tuning stage, or both. The fine-tuning stage of knowledge distillation is usually for specific downstream tasks, such as classification, sequence labeling, machine reading comprehension, etc. The basic steps of knowledge distillation at the fine-tuning stage are as follows:
\begin{itemize}
\item \textbf{Choose a suitable teacher model}, such as BERT-base or BERT-large, and fine-tune it on the target task, obtaining a fine-tuned teacher model.
\item \textbf{Choose a suitable student model}, such as DistillBERT~\cite{Sanh2019DistilBERTAD} or TinyBERT~\cite{Jiao2019TinyBERTDB}, and initialize it, using some layers of the teacher model or random initialization.
\item \textbf{Define a suitable distillation loss function}, such as the cross-entropy of the output probabilities of the teacher model and the student model, the mean squared error of the hidden vectors, the cosine similarity of the attention matrices, etc., or their combination. At the same time, you can also add the cross-entropy of the original labels as part of the loss function.
\item \textbf{Use the training data of the target task to train the student model}, optimize the distillation loss function, and make the student model fit the output and generalization ability of the teacher model as much as possible.
\item \textbf{Use the test data of the target task to evaluate the student model}, compare its performance and accuracy with the teacher model and the original model.
\end{itemize}

In the work of Jiao et al. (2019),the authors introduce TinyBERT~\cite{Jiao2019TinyBERTDB},a compressed version of the BERT~\cite{Devlin2019BERTPO} model,conceived to maintain BERT's performance while reducing the model size and inference time. Through knowledge distillation,TinyBERT learns to produce probability outputs similar to the original BERT model,achieving performance close to that of the original model despite a substantial reduction in the number of parameters. This achievement illustrates the significant potential of knowledge distillation for model compression and efficiency enhancement.

DistillBERT~\cite{Sanh2019DistilBERTAD} is a BERT-based model compression method that allows a small student model to learn the knowledge of a large teacher model, thereby improving performance and accuracy. DistillBERT~\cite{Sanh2019DistilBERTAD} reduces the model size by 40\% and increases the running speed by 60\%, while retaining more than 95\% of BERT’s language understanding ability. DistillBERT’s~\cite{Sanh2019DistilBERTAD} training process includes the pre-training stage and the fine-tuning stage, where the pre-training stage uses a triple loss function that combines language modeling, distillation, and cosine distance. DistillBERT~\cite{Sanh2019DistilBERTAD} is a small, fast, cheap, and lightweight Transformer model, suitable for resource-constrained scenarios.

In summary,knowledge distillation is an efficacious technique that enables smaller neural network models to learn from and inherit knowledge from larger neural network models. This technique allows for the reduction of model complexity and computational cost without sacrificing substantial performance. It holds significant implications for deploying deep learning models on resource-constrained devices and enhancing operational efficiency of models.

\section{Multi-Task Learning}

Multi-Task Learning (MTL) in machine learning is a paradigm that aims to enhance the generalization capability of a model by training it simultaneously on related tasks. In this strategy,the model shares representations,allowing it to capture and exploit commonalities and differences across multiple tasks.

When integrated with the Transformer model,MTL can be considered a method to utilize the Transformer's robust self-attention mechanism for capturing interrelationships between various tasks. The Transformer model,initially proposed by Vaswani et al. (2017)~\cite{Vaswani2017AttentionIA}  captures long-range dependencies through its self-attention mechanism and is not constrained by the sequential nature of the data.

\begin{figure}[htbp]
\centering{\includegraphics[width=0.9\linewidth]{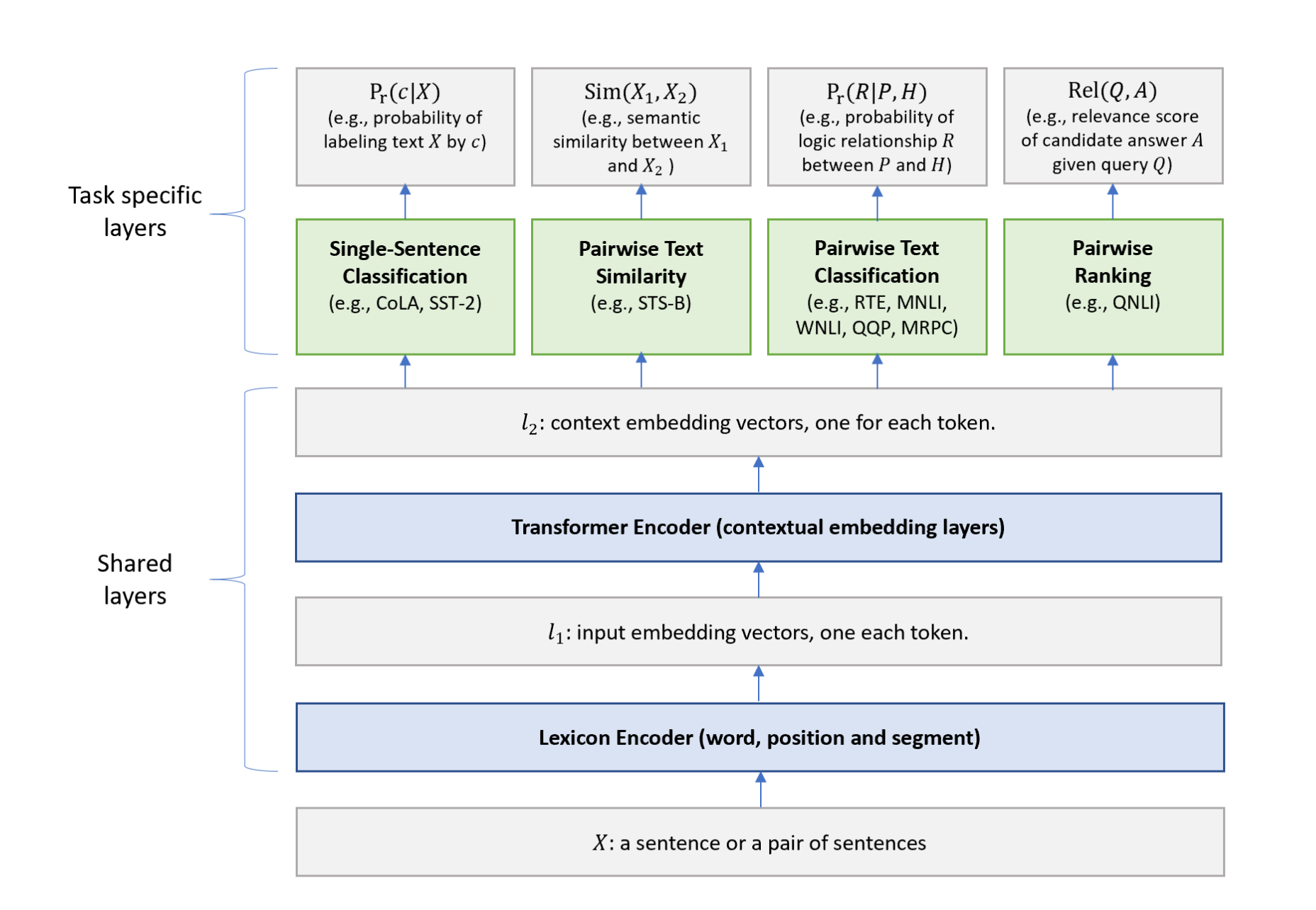}}
\caption{Architecture of the MT-DNN model for representation learning.~\cite{MultiTaskDeepNeuralNetworks}}
\label{fig-mt-dnn}
\end{figure} 

``Multi-Task Deep Neural Networks for Natural Language Understanding''\cite{MultiTaskDeepNeuralNetworks}(Fig.~\ref{fig-mt-dnn}) explores the use of deep neural networks for multi-task learning in the field of NLP.

In a multi-task learning setting,the Transformer model can be trained to perform various language processing tasks,such as sentiment analysis,named entity recognition,and question answering. The model's encoder serves as a shared foundation to capture universal language representations across tasks,while the decoder or task-specific output layers can be tailored for individual tasks.

The formulation of the MTL Transformer model can be represented as follows:

Let there be a set of tasks \(\mathcal{T} = \{T_1,T_2,...,T_n\}\). For each task \(T_i\),the model's parameters can be denoted as \(\theta_{shared}\) and \(\theta_{T_i}\). Here,\(\theta_{shared}\) are parameters shared across multiple tasks,and \(\theta_{T_i}\) are specific to task \(T_i\). The overall training objective can be formalized as:

\begin{equation}
\min _{\theta_{\text {shared }},\left\{\theta_{T_i}\right\}} \sum_{i=1}^n \mathcal{L}_{T_i}\left(f\left(x_i ; \theta_{\text {shared }},\theta_{T_i}\right)\right) \label{eq-mtl}
\end{equation}

In Equation \eqref{eq-mtl},\(f\) represents the model,\(x_i\) is the input for task \(T_i\),and \(\mathcal{L}_{T_i}\) is the loss function for task \(T_i\).

The advantage of MTL lies in its ability to improve data efficiency,as the same data samples can be used to train multiple tasks simultaneously. Moreover,MTL often enhances the model's generalizability,as it forces the model to learn more representative features useful across all tasks.

Multi-task learning and fine-tuning can be combined to improve the performance and generalization ability of the model. The fine-tuning process of multi-task learning is roughly as follows:
\begin{itemize}
\item \textbf{Choose a pre-trained model}, such as BERT or T5, as the base model. This model is usually pre-trained on a large-scale corpus with self-supervised learning, learning general language representations.
\item \textbf{Choose multiple related downstream tasks}, such as text classification, sequence labeling, question answering, etc. These tasks can share some common features or knowledge, and also have some different features or knowledge.
\item \textbf{Define a multi-task objective function}, such as the weighted sum of the loss functions of multiple tasks, or a unified generative loss function. At the same time, you can also consider the priority or difficulty of the tasks, and assign different weights or learning rates to different tasks.
\item \textbf{Fine-tune the base model on the training data of multiple tasks}, optimize the multi-task objective function, and make the model learn the features and knowledge of multiple tasks simultaneously. You can use hard sharing or soft sharing methods to control the degree of parameter sharing of the model, and also use dynamic sampling or multi-stage training strategies to balance the training efficiency of multiple tasks.
\end{itemize}
In summary,MTL offers significant advantages in enhancing model generalization and performance in data-scarce situations,especially when leveraging powerful architectures like the Transformer. The successful implementation of this approach can significantly enhance task performance in fields like NLP.

\section{Parameter-Efficient Fine-Tuning}

\begin{figure}[htbp]
\centering{\includegraphics[width=0.9\linewidth]{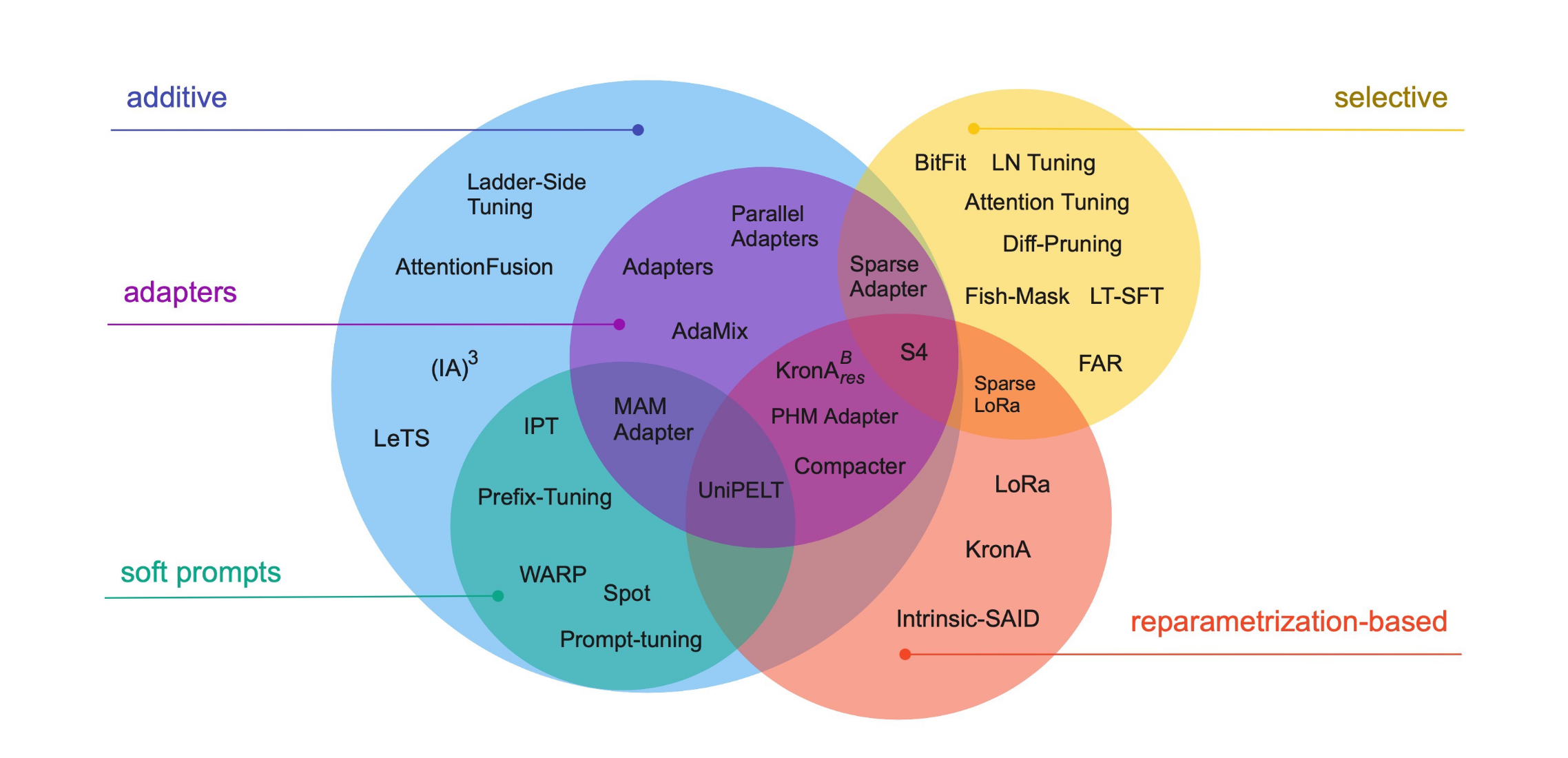}}
\caption{Parameter-efficient fine-tuning methods taxonomy.~\cite{Lialin2023ScalingDT}}
\label{fig-ppeft_all}
\end{figure} 
Parameter-Efficient Fine-Tuning (PEFT) is a technique used to efficiently adapt pretrained language models (PLMs) to various downstream applications without fine-tuning all model parameters. The PEFT method involves fine-tuning only a small number of additional parameters,significantly reducing computational and storage costs. PEFT approaches can achieve performance levels close to or surpassing full fine-tuning methods.PEFT strategies (Fig.~\ref{fig-ppeft_all}) are particularly effective in adapting these models to specific downstream tasks without the necessity of extensively retraining all parameters.

Hugging Face's PEFT \footnote{Hugging face's PEFT:\label{section:huggingface-peft}\url{https://huggingface.co/docs/peft/index}} are excellent toolkits for fine-tuning large language models. It can leverage a small number of additional parameters or text prompts to adapt pretrained language models to various downstream tasks,improving efficiency and performance. It also support multiple pretrained language models and downstream tasks,making them user-friendly and versatile. You can use them to implement the language applications you desire,such as text generation,text summarization,text classification,and more.

Here are several common classifications of Parameter-Efficient Fine-Tuning (PEFT) methods:

\textbf{Adapter Tuning:}

Small networks,known as adapters,are inserted between specific layers of the model.
Only the parameters of these adapter layers are trained,while the rest of the pre-trained model's parameters remain fixed.
This method allows the model to rapidly switch between different tasks while retaining the knowledge of the original model.

\textbf{Prompt Tuning: }

Trainable prefixes,referred to as prompts,are added to the input sequence.
These prompts,treated as additional inputs,participate in the inference process along with the model's original parameters.
By adjusting these prompts,the model's output can be guided to adapt to new tasks.

\textbf{Low-Rank Adaptation (LoRA): }

The weight matrices of the model are modified using low-rank updates.
This approach adjusts the original weights by adding low-rank matrices,allowing fine-tuning without significantly increasing the number of parameters.
LoRA maintains the model's performance while reducing the computational resources required.

\textbf{BitFit: }

Only the bias terms in the model's parameters are adjusted,without changing the weight matrices.
This method is highly parameter-efficient and suitable for resource-constrained scenarios.

\subsection{Prompt Tuning}
\begin{figure}[htbp]
\centering{\includegraphics[width=0.9\linewidth]{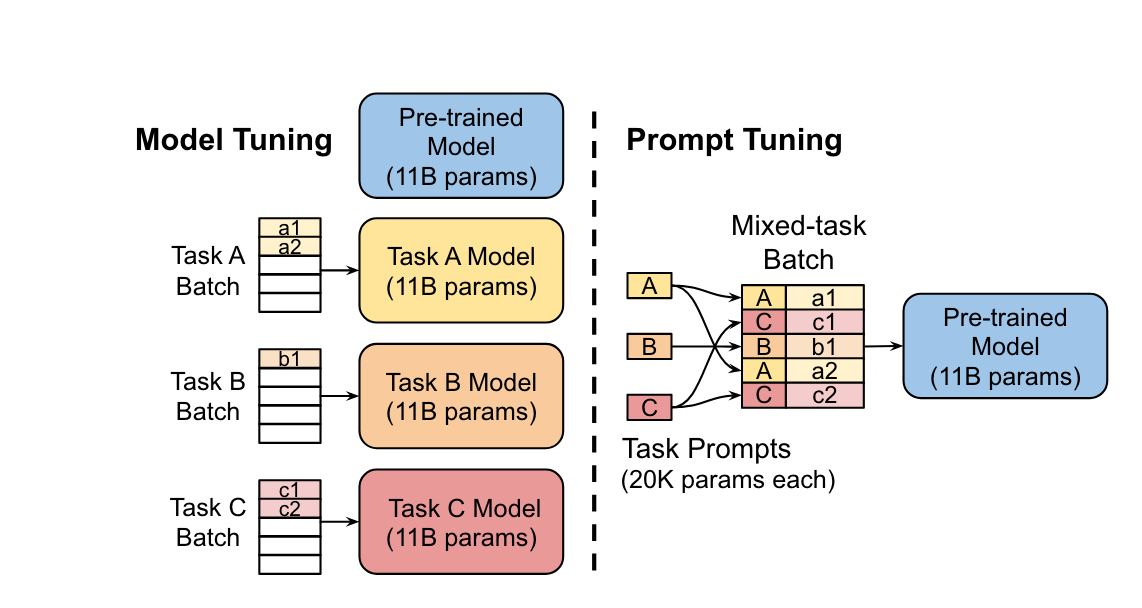}}
\caption{Model tuning requires making a taskspecific copy of the entire pre-trained model for each downstream task and inference must be performed in separate batches. Prompt tuning only requires storing a small task-specific prompt for each task,and enables mixed-task inference using the original pretrained model.~\cite{lester-etal-2021-Prompt-Tuning}}
\label{fig-prompt-tuning}
\end{figure} 

Prompt-Tuning~\cite{lester-etal-2021-Prompt-Tuning}(Fig.~\ref{fig-prompt-tuning}) is a method suitable for both Natural Language Understanding (NLU) and Natural Language Generation (NLG) tasks. It employs discrete textual tokens (such as \textbf{[CLS]},\textbf{[SEP]},etc.) as hard prompts,which are concatenated to the beginning and end of the input text to form a prompt. The embedding vectors of these hard prompts are fixed and do not participate in training,but they can be selected or initialized based on the target task. Prompt-Tuning~\cite{lester-etal-2021-Prompt-Tuning} also uses a linear layer as the output layer,whose parameters are trainable and can be optimized through backpropagation,enabling the model to learn the output mapping relevant to the target task.

In this paper~\cite{lester-etal-2021-Prompt-Tuning},the authors first present the fundamental principles and methods of prompt tuning. Specifically,they use a simple linear model to combine prompt information with the hidden representations of the pretrained model,thereby generating predictions for downstream tasks. Prompt information can take various forms,such as questions,answers,keywords,and more. To make prompt information more flexible and controllable,the authors introduce the concept of \textbf{soft prompts} which allows for weighting the prompt information to adjust the model's predictions.soft-prompts are represented as a parameter $P_e \in \mathbb{R}^{p \times e}$,where $p$ is the length of the prompt.

Next,the authors evaluate the performance of prompt tuning through experiments on multiple downstream tasks. The experimental results demonstrate that prompt tuning can significantly improve the model's performance across various tasks,including text classification,named entity recognition,natural language inference,and more. Furthermore,prompt tuning enhances the model's robustness and generalization,enabling it to adapt better to different data distributions and task settings.

Finally,the authors discuss some applications and future research directions of prompt tuning. They point out that prompt tuning can be applied to various natural language processing tasks,including text generation,dialogue systems,information retrieval,and more. Additionally,they raise interesting questions such as how to automate prompt generation and selection,as well as how to effectively combine and select from multiple prompts.

In summary, Prompt Tuning~\cite{lester-etal-2021-Prompt-Tuning} introduces a simple yet effective method for improving the performance of pretrained language models on downstream tasks,offering practical value and research significance.

\subsection{P-Tuning} 

\begin{figure}[htbp]
\centering{\includegraphics[width=0.9\linewidth]{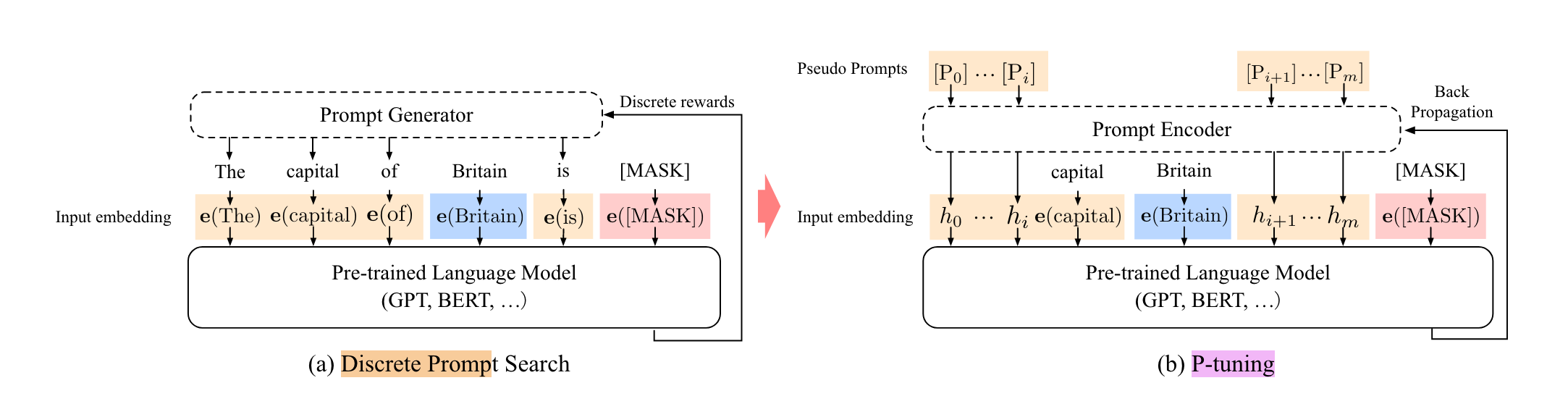}}
\caption{Discrete Prompt Search and P-Tuing.~\cite{Liu2021P-Tuning}}
\label{fig-p-tuning}
\end{figure} 

Compared to traditional model fine-tuning,the core idea of P-Tuning~\cite{Liu2021P-Tuning}(Fig.~\ref{fig-p-tuning}) is to introduce trainable prefixes (prompts) at the input stage,rather than adjusting all parameters of the model,significantly reducing the required computational resources and training time.

P-Tuning~\cite{Liu2021P-Tuning} is a method tailored for Natural Language Understanding (NLU) tasks,which utilizes special tokens (such as [P1],[P2],etc.) as soft prompts. These are concatenated to the beginning and end of the input text to form a template. The embedding vectors of these soft prompts are trainable and can be optimized through backpropagation,enabling the model to learn semantic information relevant to the target task. P-Tuning also employs an LSTM to encode the soft prompts,enhancing their expressive capabilities.

Specifically,the method employs a technique known as``hybrid soft prompting,'' where continuous prompt embeddings are combined with discrete prompts to improve the model's performance. This approach can be applied to various types of pre-trained language models (PLMs),including BERT~\cite{Devlin2019BERTPO},RoBERTa~\cite{zhuang-etal-2021-robustly-roberta},and GPT-2~\cite{Radford2019LanguageMAgpt2}. Moreover,it is also applicable to both frozen and fine-tuned language models,and it has shown significant performance improvements across various NLU tasks.
\begin{figure}[htbp]
\centering{\includegraphics[width=0.9\linewidth]{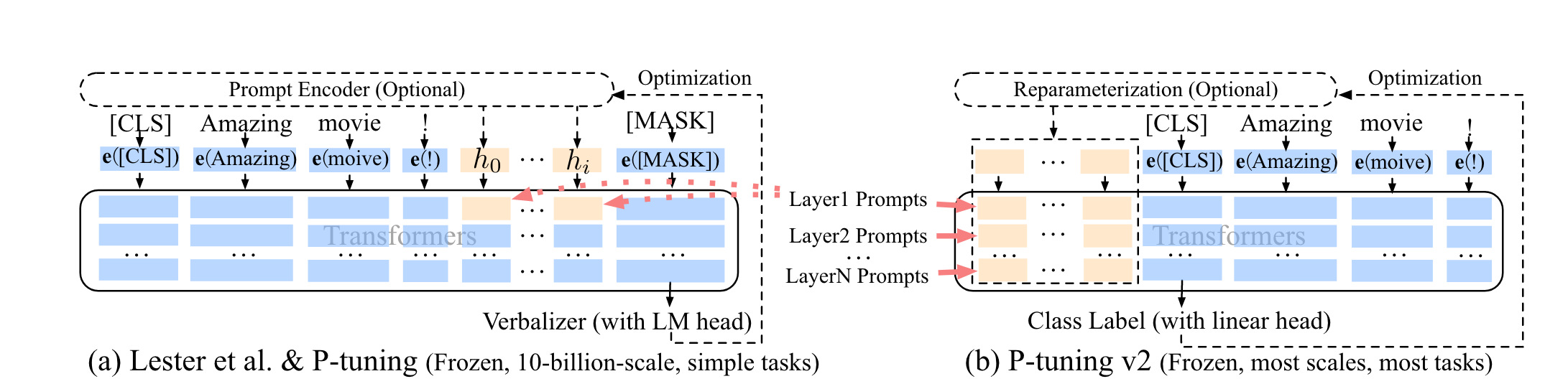}}
\caption{From P-tuning~\cite{Liu2021P-Tuning} to P-tuning v2~\cite{liu-etal-2022-p-tuning-v2}. Orange blocks (i.e.,h0,...,hi) refer to trainable prompt embeddings; blue blocks are embeddings stored or computed by frozen pre-trained language models.\cite{liu-etal-2022-p-tuning-v2}}
\label{fig-}
\end{figure} 

P-Tuning2~\cite{liu-etal-2022-p-tuning-v2} is an improvement over P-Tuning~\cite{Liu2021P-Tuning},utilizing technologies of hierarchical transformers and local parallel autoregressive generation to enhance Prompt-Tuning and P-Tuning,serving as a universal solution across scales and NLU tasks. P-Tuning2~\cite{liu-etal-2022-p-tuning-v2} introduces soft prompts in every layer,rather than just at the input layer,increasing the number of trainable parameters and enhancing the model's expressive power. P-Tuning2~\cite{liu-etal-2022-p-tuning-v2}  also reverts to the traditional classification label paradigm,instead of using label-word mappers,allowing it to adapt to more complex sequence labeling tasks.

Compared to fine-tuning,P-Tuning2~\cite{liu-etal-2022-p-tuning-v2}  only requires adjusting 0.1\% to 3\% of the parameters to achieve the same performance,while reducing storage and memory usage for each task.

\subsection{Prefix Tuning}
\begin{figure}[htbp]
\centering{\includegraphics[width=0.9\linewidth]{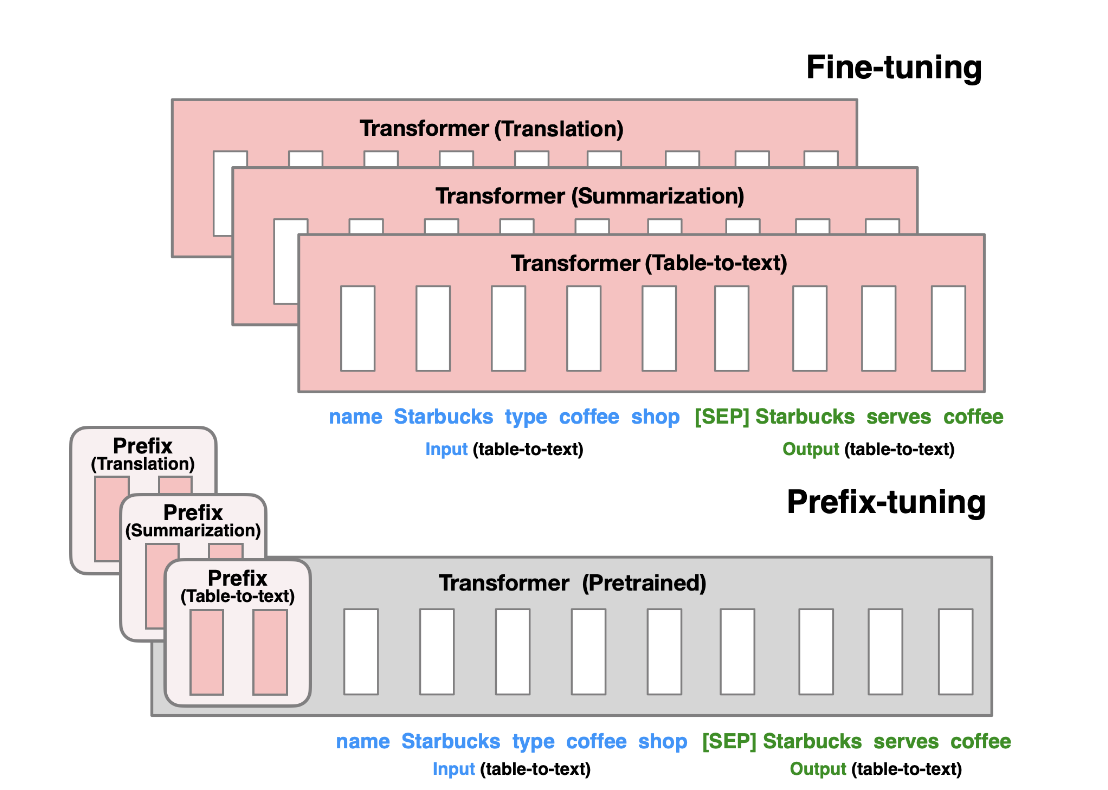}}
\caption{Fine-tuning (top) updates all Transformer parameters (the red Transformer box) and requires storing a full model copy for each task. Prefix-tuning (bottom),which freezes the Transformer parameters and only optimizes the prefix (the red prefix blocks).~\cite{li2021prefixtuning}}
\label{fig-prefix-tuning}
\end{figure} 

Prefix-Tuning~\cite{li2021prefixtuning} is a method tailored for Natural Language Generation (NLG) tasks,which utilizes a small multi-layer perceptron (MLP) as a prefix. The output of this prefix is concatenated with the input embeddings of the pre-trained model to form a new input. The parameters of this prefix are trainable and can be optimized through backpropagation,enabling the model to learn the generation strategies relevant to the target task. Prefix-Tuning~\cite{li2021prefixtuning}also adds prefixes to each transformer layer to enhance the model's efficiency and expressive power.

Prefix Tuning~\cite{li2021prefixtuning} optimizes a small continuous task-specific vector (referred to as the``prefix'') instead of modifying all language model parameters(Fig.~\ref{fig-prefix-tuning}). By learning only 0.1\% of the parameters,Prefix-Tuning achieves comparable performance on full data settings,outperforms Fine-Tuning in low-data settings,and exhibits better extrapolation capabilities for examples on topics unseen during training.

\begin{figure}[htbp]
\centering{\includegraphics[width=0.9\linewidth]{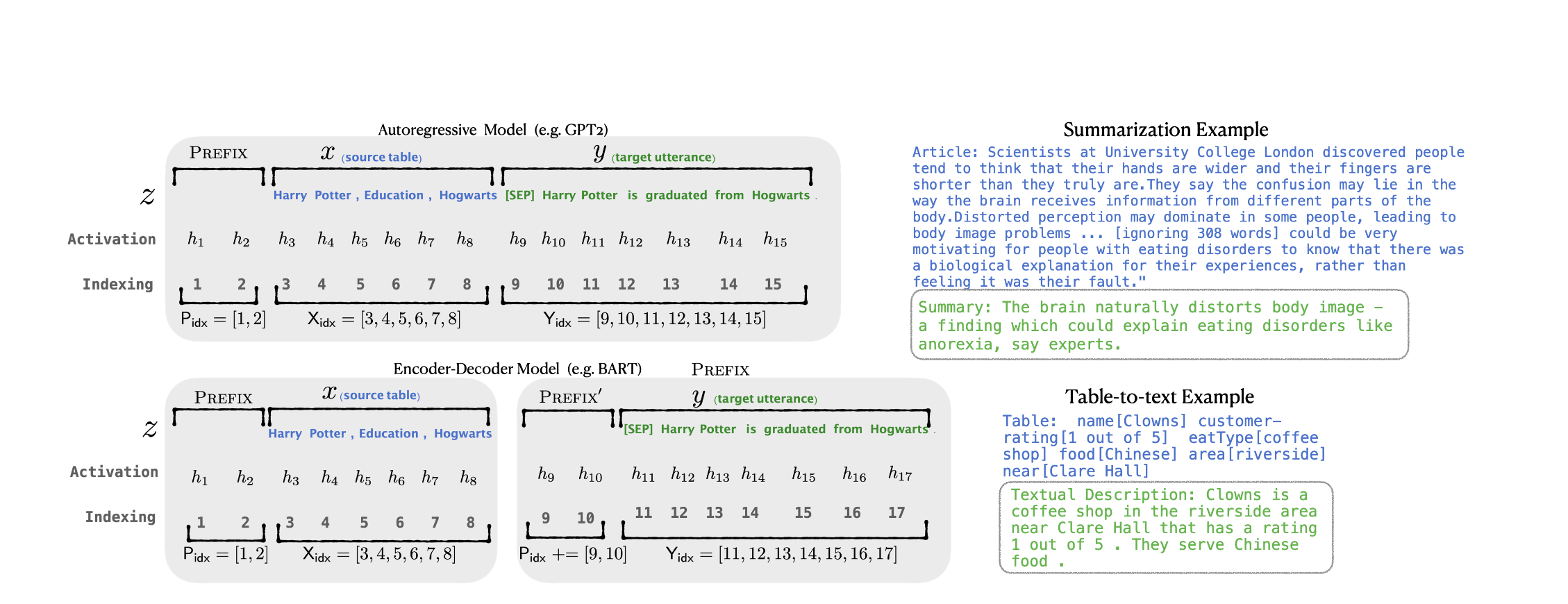}}
\caption{An annotated example of prefix-tuning using an autoregressive LM (top) and an encoder-decoder model (bottom)~\cite{li2021prefixtuning}.~\cite{li2021prefixtuning}}
\label{fig-prefix-tuning-example}
\end{figure} 
The paper also provides experimental results demonstrating the effectiveness of Prefix-Tuning across multiple natural language generation tasks,including GPT-2\cite{Radford2019LanguageMAgpt2} for table-to-text generation and BART~\cite{Lewis2019BARTDS} for abstractive summarization(Fig.~\ref{fig-prefix-tuning-example}). Furthermore,it compares Prefix-Tuning with other existing methods,showing superior performance in many cases.

Overall,Prompt-Tuning~\cite{lester-etal-2021-Prompt-Tuning},Prefix-Tuning~\cite{li2021prefixtuning},P-Tuning~\cite{Liu2021P-Tuning},and P-Tuning2~\cite{liu-etal-2022-p-tuning-v2} are all parameter-efficient fine-tuning methods. Compared to full model fine-tuning,they require training far fewer parameters to achieve good results. The main differences among them are:
\begin{itemize}
\item Both Prompt-Tuning~\cite{lester-etal-2021-Prompt-Tuning} and P-Tuning~\cite{Liu2021P-Tuning} use hard prompts,i.e.,fixed textual tokens,whereas Prefix-Tuning and P-Tuning2~\cite{liu-etal-2022-p-tuning-v2} use soft prompts,i.e.,trainable embedding vectors.
\item Prompt-Tuning~\cite{lester-etal-2021-Prompt-Tuning} and P-Tuning~\cite{Liu2021P-Tuning} add prompts at the input embedding layer,while Prefix-Tuning~\cite{li2021prefixtuning} and P-Tuning2~\cite{liu-etal-2022-p-tuning-v2} add prefixes or prompts at each transformer layer.
\item Prompt-Tuning~\cite{lester-etal-2021-Prompt-Tuning} is suitable for both NLU and NLG tasks,while Prefix-Tuning~\cite{li2021prefixtuning} is primarily for NLG tasks,and P-Tuning~\cite{Liu2021P-Tuning}  and P-Tuning2~\cite{liu-etal-2022-p-tuning-v2} are mainly for NLU tasks.
\end{itemize}

\subsection{Adapter Tuning}
\textbf{Adapter Tuning-PETL} PETL~\cite{Houlsby2019ParameterEfficientTL} is the employment of Adapter Modules for parameter-efficient transfer learning. An Adapter Module is a lightweight neural network module that,without altering the pre-trained model's architecture,introduces a minimal quantity of task-specific parameters to adapt to new downstream tasks. This methodology circumvents the need for comprehensive fine-tuning on each downstream task,thereby significantly diminishing the parameter count and computational overhead.

To elaborate,the method initially capitalizes on a pre-trained model (e.g.,BERT\cite{Devlin2019BERTPO}) honed on a large corpus,followed by the integration of Adapter Modules to transition the model to downstream tasks. The Adapter Modules comprise dual linear transformations alongside a nonlinear activation function,imbuing the model with the capacity to adapt to new tasks by appending a modest number of task-specific parameters,all without perturbing the pre-trained model's parameters. This approach enables the sharing of the pre-trained model's parameters across disparate downstream tasks,thereby substantially curtailing the parameter volume and computational expenditure.

This methodology has been rigorously evaluated across 26 textual classification tasks,achieving performance comparable to exhaustive fine-tuning while utilizing only a sparse array of task-specific parameters. Furthermore,the method exhibits scalability,with the capacity to incrementally train new downstream tasks without the obliteration of previously acquired knowledge.
\begin{figure}[htbp]
\centering{\includegraphics[width=0.9\linewidth]{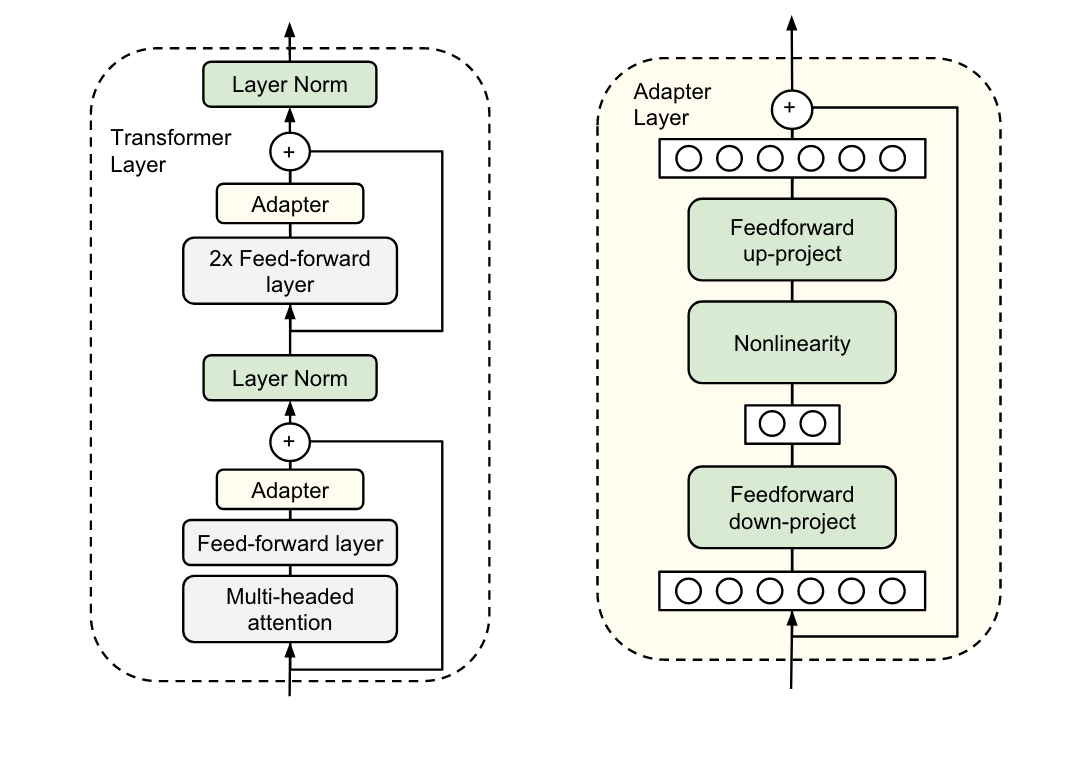}}
\caption{Architecture of the adapter module and its integration with the Transformer.~\cite{Houlsby2019ParameterEfficientTL}}
\label{fig-adapter-tuning}
\end{figure} 
(Fig.~\ref{fig-adapter-tuning}) shows  adapter architecture of PETL~\cite{Houlsby2019ParameterEfficientTL},Each layer of the Transformer consists of two primary sub-layers: an attention layer and a feedforward layer. These layers are immediately followed by a projection that remaps the feature size back to the size of the layer's input. A skip connection is implemented across each of the sub-layers. The output from each sub-layer is then fed into layer normalization. After each of these sub-layers,we insert two sequential adapters. The adapter is always applied right to the output of the sub-layer,after the features are projected back to the input size,but before the skip connection is reintroduced. Subsequently,the adapter's output is passed directly into the subsequent layer normalization,code\footnote{Code For Parameter-Efficient Transfer Learning for NLP:\label{section:petl-code}\url{https://github.com/google-research/adapter-bert}}.

\subsection{LoRA,QLoRA}
\begin{figure}[htbp]
\centering{\includegraphics[width=0.9\linewidth]{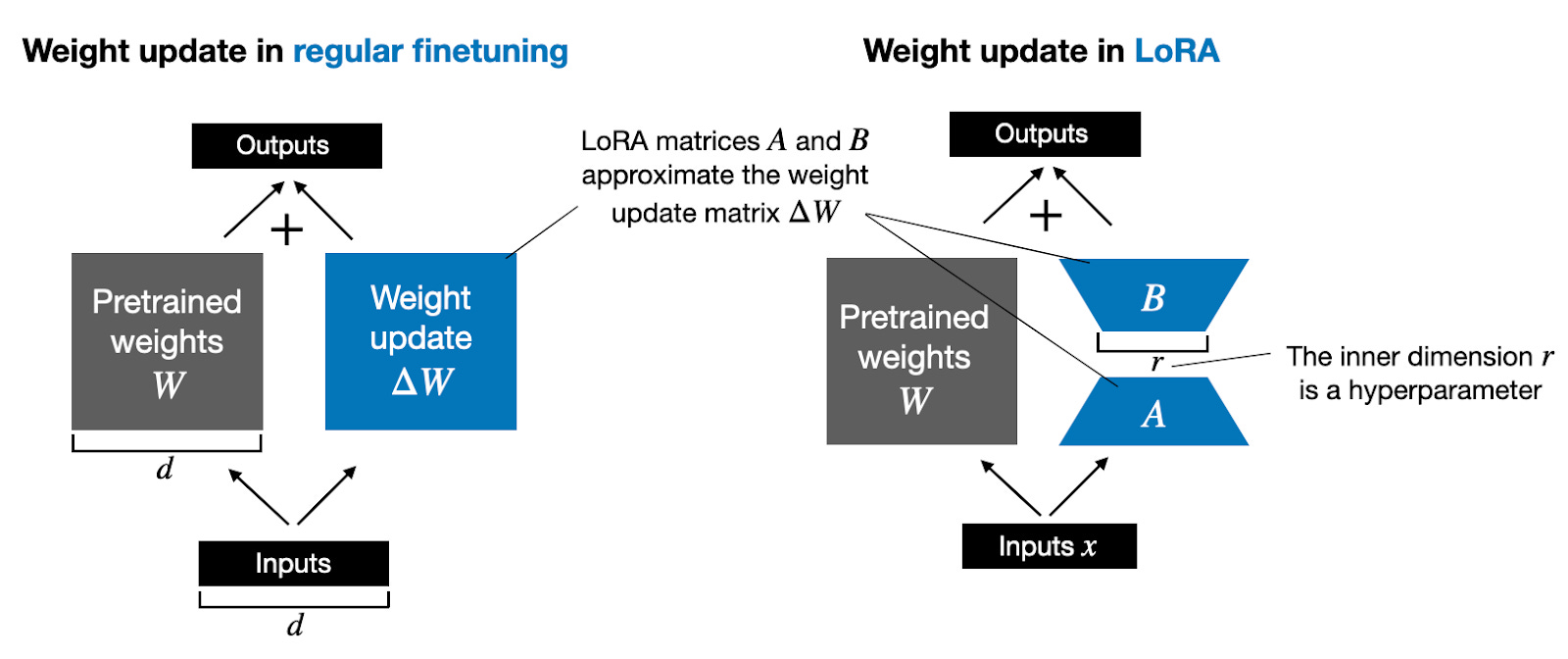}}
\caption{LoRA,only train A and B.~\cite{LoRA1}}
\label{fig-LoRA}
\end{figure} 
The LoRA~\cite{Hu2021LoRA} is a low-cost fine-tuning method for large language models,which leverages the intrinsic low-rank characteristics(Fig.~\ref{fig-LoRA}) of large models. It adds bypass matrices to simulate full-parameter fine-tuning,thereby reducing the amount of parameters and time required for training. The LoRA~\cite{Hu2021LoRA} can achieve comparable or better results in content understanding and generation tasks compared to full-parameter fine-tuning,and it can be combined with other efficient fine-tuning methods.

The core formula of LoRA~\cite{Hu2021LoRA} is:
\begin{equation}
   W^{\prime}=W+\Delta W
\end{equation}

where \( W \) is the parameters of the pre-trained language model,\( W' \) is the parameters after fine-tuning,and \( \Delta W \) is the parameters that need to be updated. \( \Delta W \) consists of the product of two low-rank matrices,namely:
\[ \Delta W = AB^T \]
where \( A \in \mathbb{R}^{n \times d} \),\( B \in \mathbb{R}^{d \times m} \),and \( d \ll n \),with \( d \) being a very small integer,typically 1 or 2. Thus,the number of parameters in \( \Delta W \) is much less than in \( W \),thereby reducing the cost of fine-tuning.

the idea of LoRA~\cite{Hu2021LoRA} is to add a small residual to the parameters of a pre-trained language model to adapt it to downstream tasks. This residual is obtained by multiplying two low-dimensional matrices,which can be understood as a type of linear transformation used to adjust the model's output. This method leverages the intrinsic low-rank property of pre-trained language models,meaning that the model's parameter matrix can be approximately decomposed into the product of two low-rank matrices. As such,only a small number of parameters need to be trained to achieve good results.
\begin{table*}[htbp]
\centering
\begin{tabular}{c|c|ccccccccc}
\hline Mode & \begin{tabular}{l} 
\# Trainable \\
Parameters
\end{tabular} & MNLI & SST-2 & MRPC & CoLA & QNLI & QQP & RTE & STS-B & Avg. \\
\hline 
RoB $B_{\text {base }}$ & $5.0 \mathrm{M}$ & $\mathbf{87.6}$ & 94.8 & 90.2 & $\mathbf{63.6}$ & 92.8 & $\mathbf{91.9}$ & 78.7 & 91.2 & 86.4 \\
$\operatorname{RoB}_{\text {base }}($ BitFit) $*$ & $0.1 \mathrm{M}$ & 84.7 & 93.7 & $\mathbf{92.7}$ & 62.0 & 91.8 & 84.0 & 81.5 & 90.8 & 85.2 \\
$\operatorname{RoB}_{\text {base }}\left(\mathrm{Adpt}^{\mathrm{D}}\right)^*$ & $0.3 \mathrm{M}$ & $87.1_{ \pm .0}$ & $94.2_{ \pm .1}$ & $88.5_{ \pm 1.1}$ & $60.8_{ \pm .4}$ & $93.1_{ \pm .1}$ & $90.2_{ \pm .0}$ & $71.5_{ \pm 2.7}$ & $89.7_{ \pm .3}$ & 84.4 \\
$\operatorname{RoB}_{\text {base }}\left(\mathrm{Adpt}^{\mathrm{D}}\right)^*$ & $0.9 \mathrm{M}$ & $87.3_{ \pm .1}$ & $94.7_{ \pm .3}$ & $88.4_{ \pm .1}$ & $62.6_{ \pm .9}$ & $93.0_{ \pm .2}$ & $90.6_{ \pm .0}$ & $75.9_{ \pm 2.2}$ & $90.3_{ \pm .1}$ & 85.4 \\
$\operatorname{RoB}_{\text {base }}($ LoRA) & $0.3 \mathrm{M}$ & $87.5_{ \pm .3}$ & $\mathbf{9 5 . 1} 1_{ \pm .2}$ & $89.7_{ \pm .7}$ & $63.4_{ \pm 1.2}$ & $\mathbf{9 3 . 3} \mathbf{3}_{ \pm .3}$ & $90.8_{ \pm .1}$ & $\mathbf{86.6_{ \pm .7}}$ & $\mathbf{91.5_{ \pm .2}}$ & $\mathbf{87.2}$ \\
\hline 
RoB $_{\text {large }}(\mathrm{F}$ & $5.0 \mathrm{M}$ & 90.2 & 96.4 & 90.9 & 68.0 & 94.7 & 92.2 & 86.6 & 92.4 & 88.9 \\
$\operatorname{RoB}_{\text {large }}($ LoRA) & $0.8 \mathrm{M}$ & $\mathbf{9 0 . 6}_{ \pm .2}$ & $96.2_{ \pm .5}$ & $\mathbf{9 0 . 9}_{ \pm 1.2}$ & $\mathbf{6 8 .} \mathbf{2}_{ \pm 1.9}$ & $\mathbf{9 4 . 9}_{ \pm .3}$ & $91.6_{ \pm .1}$ & $\mathbf{8 7 . 4}_{ \pm 2.5}$ & $\mathbf{9 2 . 6}_{ \pm .2}$ & 89.0 \\
\hline 
$\operatorname{RoB}_{\text {large }}\left(\mathrm{Adpt}^{\mathrm{P}}\right) \dagger$ & $3.0 \mathrm{M}$ & $90.2_{ \pm .3}$ & $96.1_{ \pm .3}$ & $90.2_{ \pm .7}$ & $\mathbf{6 8 . 3 _ { \pm 1 . 0 }}$ & $\mathbf{9 4 . 8 _ { \pm . 2 }}$ & $\mathbf{9 1 . 9}_{ \pm .1}$ & $83.8_{ \pm 2.9}$ & $92.1_{ \pm .7}$ & 88.4 \\
$\operatorname{RoB}_{\text {large }}\left(\mathrm{Adpt}^{\mathrm{P}}\right) \dagger$ & $0.8 \mathrm{M}$ & $\mathbf{9 0 . 5} \pm .3$ & $\mathbf{9 6 . 6} \pm .2$ & $89.7_{ \pm 1.2}$ & $67.8_{ \pm 2.5}$ & $\mathbf{9 4 . 8} \mathbf{8}_{ \pm .3}$ & $91.7_{ \pm .2}$ & $80.1_{ \pm 2.9}$ & $91.9_{ \pm .4}$ & 87.9 \\
$\operatorname{RoB}_{\text {large }}\left(\mathrm{Adpt}^{\mathrm{H}}\right) \dagger$ & $6.0 \mathrm{M}$ & $89.9_{ \pm .5}$ & $96.2 \pm .3$ & $88.7_{ \pm 2.9}$ & $66.5_{ \pm 4.4}$ & $94.7_{ \pm .2}$ & $92.1_{ \pm .1}$ & $83.4_{ \pm 1.1}$ & $91.0_{ \pm 1.7}$ & 87.8 \\
$\operatorname{RoB} B_{\text {large }}\left(\mathrm{Adpt}^{\mathrm{H}}\right) \dagger$ & $0.8 \mathrm{M}$ & $90.3_{ \pm .3}$ & $96.3_{ \pm .5}$ & $87.7_{ \pm 1.7}$ & $66.3_{ \pm 2.0}$ & $94.7_{ \pm .2}$& $91.5_{ \pm .1}$ & $72.9_{ \pm 2.9}$ & $91.5_{ \pm .5}$ & 86.4 \\
$\operatorname{RoB}_{\text {large }}(\operatorname{LoRA}) \dagger$ & $0.8 \mathrm{M}$ & $\mathbf{9 0 . 6}_{ \pm .2}$ & $96.2_{ \pm .5}$ & $\mathbf{9 0 . 2} 2_{ \pm 1.0}$ & $68.2_{ \pm 1.9}$ & $\mathbf{9 4 . 8} \mathbf{8}_{ \pm .3}$ & $91.6_{ \pm .2}$ & $\mathbf{8 5 . 2}_{ \pm 1.1}$ & $\mathbf{9 2 . 3 _ { \pm . 5 }}$ & 88.6 \\
\hline
$\operatorname{DeB}_{\mathrm{XXL}}(\mathrm{FT})^*$ & $00.0 \mathrm{M}$ & 91.8 & 97.2 & 92.0 & 72.0 & 96.0 & 92.7 & 93.9 & 92.9 & 91.1 \\
$\operatorname{DeB}_{X X L}$ (LoRA) & $4.7 \mathrm{M}$ & $\mathbf{9 1 . 9}_{ \pm .2}$ & $96.9_{ \pm .2}$ & $\mathbf{9 2 . 6}+.6$ & $\mathbf{7 2 . 4}_{ \pm 1.1}$ & $\mathbf{9 6 . 0} \mathbf{0}_{ \pm .1}$ & $\mathbf{9 2 . 9}_{ \pm .1}$ & $\mathbf{9 4 . 9}_{ \pm .4}$ & $\mathbf{9 3 . 0}_{ \pm .2}$ & 91.3 \\
\hline
\end{tabular}
\captionsetup{justification=justified,singlelinecheck=false}
\caption{RoBERTa$_{\text{base}}$,RoBERTa$_{\text{large}}$,and DeBERTa$_{\text{XXL}}$ with different adaptation methods on the GLUE benchmark. We report the overall (matched and mismatched) accuracy for MNLI,Matthew’s correlation for CoLA,Pearson correlation for STS-B,and accuracy for other tasks. Higher is better for all metrics. * indicates numbers published in prior works. \dag indicates runs configured in a setup similar to~\cite{houlsby2019parameterefficient} for a fair comparison~\cite{Hu2021LoRA}.}\label{table:Lora-Result}
\end{table*}
The advantages of LoRA~\cite{Hu2021LoRA}  include that the ability to share pre-trained models and use them to construct many small LoRA~\cite{Hu2021LoRA}  modules to adapt to different tasks; making training more efficient and lowering hardware requirements; and the capability to be combined with other methods,such as prefix tuning. LoRA~\cite{Hu2021LoRA}  has been experimented with on multiple datasets,proving its effectiveness and scalability,The result is presented in Table~\ref{table:Lora-Result}.

QLoRA~\cite{Dettmers2023-QLoRa} is a low-cost fine-tuning method for large language models,based on the intrinsic low-rank characteristics of these models. It adds bypass matrices to simulate full-parameter fine-tuning,thereby reducing the amount of parameters and time required for training. The QLoRA~\cite{Dettmers2023-QLoRa} model can achieve results comparable to or better than full-parameter fine-tuning in content understanding and generation tasks,and can be combined with other efficient fine-tuning methods.  Application scenarios for the QLoRA~\cite{Dettmers2023-QLoRa} model include code generation,natural language inference,text summarization,and more.

\subsection{$(IA)^3$}

\begin{figure}[htbp]
\centering{\includegraphics[width=0.9\linewidth]{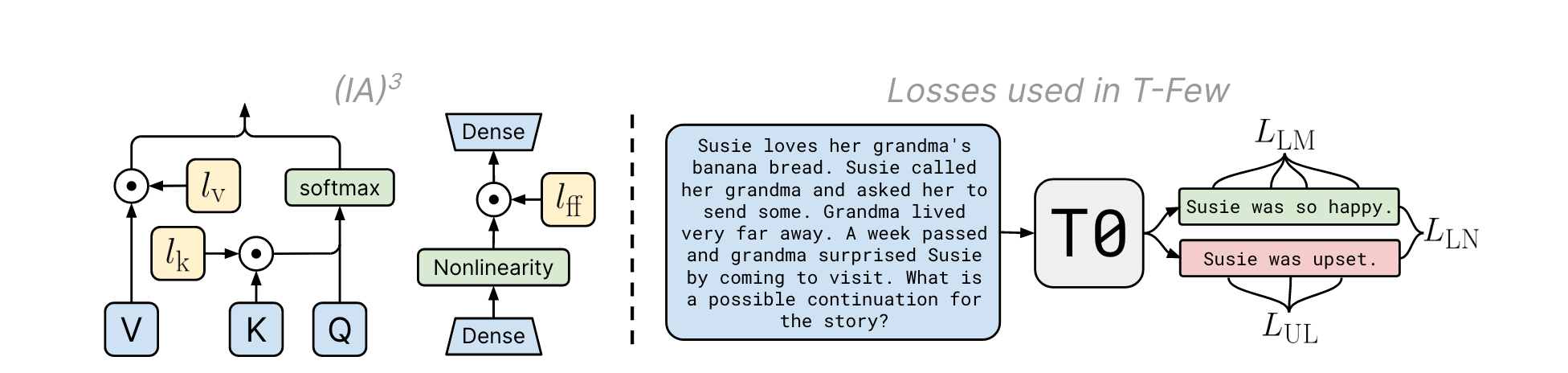}}
\caption{Diagram of $(IA)^3$.~\cite{NEURIPS2022_IA3}}
\label{fig-IA3}
\end{figure} 
$(IA)^3$~\cite{NEURIPS2022_IA3}(Fig.~\ref{fig-IA3}) is a technique used for fine-tuning large language models with limited labeled data. In contrast to the existing PEFT method,$(IA)^3$~\cite{NEURIPS2022_IA3} employs the product of intermediate activations rather than directly fine-tuning all parameters. This approach allows for updating fewer parameters,reducing computational costs,and enabling parameter sharing across multiple tasks. Additionally,$(IA)^3$~\cite{NEURIPS2022_IA3} utilizes a length-normalized loss function and an``unlikelihood'' loss function,contributing to improved performance in classification and multiple-choice tasks.

In the paper,the authors also compare the performance of $(IA)^3$~\cite{NEURIPS2022_IA3} with the existing PEFT methods such as LoRA and few-shot in-context learning (ICL). Experimental results demonstrate that IA3 outperforms LoRA~\cite{Hu2021LoRA} and ICL on various tasks while updating significantly fewer parameters. Furthermore,the authors propose a simple recipe called``T-Few'' that can be applied to new tasks without task-specific adjustments or modifications.

In summary,this paper introduces a novel fine-tuning technique,$(IA)^3$~\cite{NEURIPS2022_IA3},which enhances the performance of large language models with limited labeled data and allows for parameter sharing across multiple tasks. Additionally,the authors present a straightforward recipe,T-Few,for application to new tasks without the need for task-specific adjustments or modifications.

\subsection{BitFit}

BitFit~\cite{ben-zaken-etal-2022-bitfit} is a low-cost fine-tuning method designed for large language models. It leverages the inherent low-rank characteristics of large models by updating only the model's bias terms or a subset of them,thereby reducing the number of parameters and training time. BitFit~\cite{ben-zaken-etal-2022-bitfit} can achieve comparable or even superior performance to full-parameter fine-tuning on content understanding and generation tasks when dealing with small or medium-sized training data. It can also be used in conjunction with other efficient fine-tuning methods. BitFit~\cite{ben-zaken-etal-2022-bitfit} finds applications in various scenarios,including code generation,natural language inference,text summarization,and more.

\subsection{UniPELT,MAM Adapter}

Both UniPELT~\cite{mao-etal-2022-unipelt} and MAM Adapter~\cite{He2021-MAM-Adapter} are low-cost fine-tuning methods based on pretrained language models. They leverage the inherent low-rank characteristics of large models by adding auxiliary matrices to simulate full-parameter fine-tuning,thereby reducing the number of parameters and training time. These methods can achieve comparable or even superior performance to full-parameter fine-tuning on content understanding and generation tasks when dealing with small or medium-sized training data. Additionally,they can be used in conjunction with other efficient fine-tuning methods.

UniPELT~\cite{mao-etal-2022-unipelt} is a combination of LoRA~\cite{Hu2021LoRA},Prefix Tuning~\cite{li2021prefixtuning},and Adapter methods with gating mechanisms. It utilizes a small multi-layer perceptron (MLP) as a prefix,and the output of this MLP is concatenated with the input embeddings of the pretrained model,creating a new input representation. UniPELT~\cite{mao-etal-2022-unipelt} also incorporates LoRA's~\cite{Hu2021LoRA} reparameterization technique to modify the attention matrices of the pretrained model. Furthermore,it adds Adapters after the feed-forward sub-layers in each transformer layer to adjust the model's outputs. For each of these modules,it employs a gating mechanism to control their activation,allowing the model to automatically choose the most suitable approach for the current data or task.

MAM Adapter~\cite{He2021-MAM-Adapter} combines parallel Adapters with soft prompts (soft hints). It uses special tokens like [P1],[P2],etc.,as soft prompts and appends them before and after the input text to create a template. The embeddings of these soft prompts are trainable and can be optimized through backpropagation,enabling the model to learn semantic information relevant to the target task. MAM Adapter also adds Adapters after the feed-forward sub-layers in each transformer layer. It uses parallel placement of Adapters rather than sequential placement,as experiments have shown that this approach can improve both model performance and efficiency.

\section{Instruction Tuning}
Instruction Tuning is a method for training large language models that improves their performance on multi-task learning and unseen tasks by providing explicit instructions to the model. The core of this approach is to teach language models to follow natural language instructions,including prompts,positive or negative examples,and constraints,thereby performing better on training tasks and having better generalization capabilities on unseen tasks.

The motivation behind Instruction Tuning is to enhance the language model’s responsiveness to natural language processing instructions. The idea is that by using supervision to teach the language model to execute tasks described by instructions,the model will learn to follow instructions,even for tasks it has not seen before. This method often combines reinforcement learning techniques,rewarding the model for correctly executing instructed tasks,thus optimizing the generated results.

Unlike prompts,instructions are usually more detailed texts used to guide the model in performing specific operations or completing tasks. Instructions can be computer programs or scripts,or they can be instructive texts written by humans. The purpose of an instruction is to tell the model how to process data or perform a certain operation,rather than simply providing context or task-related information. Therefore,both prompts and instructions are texts used to guide the model in generating output,but their purposes and ways of use are different. Prompts are more often used to help the model understand tasks and context,while instructions are more often used to guide the model in performing specific operations or completing tasks.

\begin{figure}[htbp]
\centering{\includegraphics[width=0.9\linewidth]{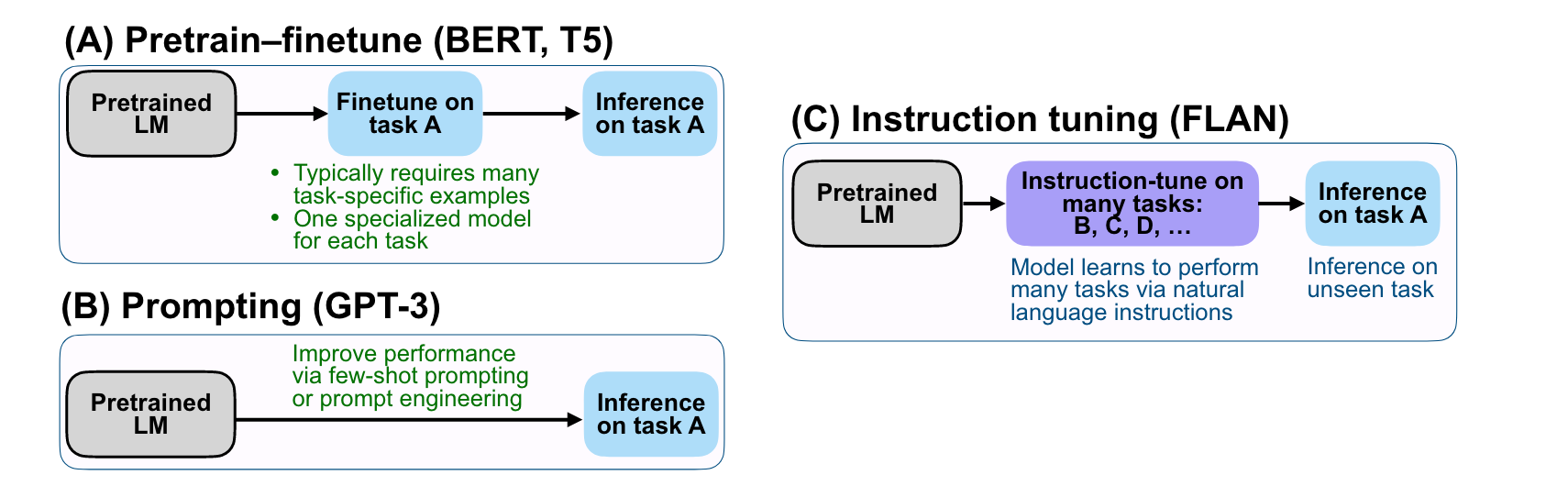}}
\caption{Comparing instruction tuning with pretrain–finetune and prompting.~\cite{Wei2021FinetunedLMFLAN}}
\label{fig-instruct-tuning}
\end{figure} 

The method of Instruction Tuning is used almost simultaneously by Google and OpenAI. Google Research proposed instruction-tuning in their 2021 paper ``Finetuned Language Models Are Zero-Shot Learners''\cite{Wei2021FinetunedLMFLAN}(Fig.~\ref{fig-instruct-tuning}). Google considers instruction-tuning a simple way to improve the zero-shot learning capabilities of language models. Similarly,OpenAI had the same idea with InstructGPT~\cite{Ouyang2022InstructGPT},which is the precursor to ChatGPT. Instructions provide more direct task guidance to help the model execute tasks.

\subsection{FLAN}
FLAN (Finetuning Language Models)~\cite{Wei2021FinetunedLMFLAN} is a method of fine-tuning large language models,which improves the model’s generalization ability to unseen instructions by fine-tuning on multiple natural language processing (NLP) tasks. The core idea of FLAN~\cite{Wei2021FinetunedLMFLAN}(Fig.~\ref{fig-instruct-tuning}) is to train language models using a mix of zero-shot prompts and few-shot prompts,thereby enhancing their performance across different tasks.
A significant feature of FLAN~\cite{Wei2021FinetunedLMFLAN} is that it can converge faster and higher with less fine-tuning on individual downstream tasks compared to the T5 model,meaning FLAN~\cite{Wei2021FinetunedLMFLAN} can serve as a more computation-efficient starting checkpoint for new tasks. Additionally,FLAN~\cite{Wei2021FinetunedLMFLAN} considers task balance and enrichment techniques in its design,especially using a mix of prompt settings (zero-shot,few-shot,and chain-of-thought) to improve performance across all settings. Research has found that such instruction fine-tuning significantly enhances the performance of various model categories (such as PaLM,T5,U-PaLM) across different prompt settings (zero-shot,few-shot,chain-of-thought) and evaluation benchmarks (such as MMLU,BBH,TyDiQA,MGSM,open-ended generation).

FLANv2~\cite{Longpre2023TheFCFLANv2} explores the design decisions of publicly available instruction tuning methods and provides a detailed analysis of the development of Flan 2022. Through careful ablation studies of the Flan task and method collection,the authors reveal the design decisions that enabled Flan-T5 to outperform previous work by 3-17\%+. The research found that task balance and enrichment techniques are key to effective instruction tuning,particularly using mixed prompt settings (zero-shot,few-shot,and chain-of-thought) can actually improve performance across all settings (2\%+).

Google researchers also found that FLAN significantly outperforms GPT-3 in many challenging benchmark tests,despite having fewer parameters than GPT-3. This indicates that even models with fewer parameters can achieve or surpass the performance of larger models through effective fine-tuning methods.

The research and application of FLAN demonstrate how fine-tuning and optimization can enhance the generalization ability and efficiency of large language models in the field,which is of great significance for the future development of NLP technology.

\subsection{InstructUIE}
\begin{figure}[htbp]
\centering{\includegraphics[width=0.9\linewidth]{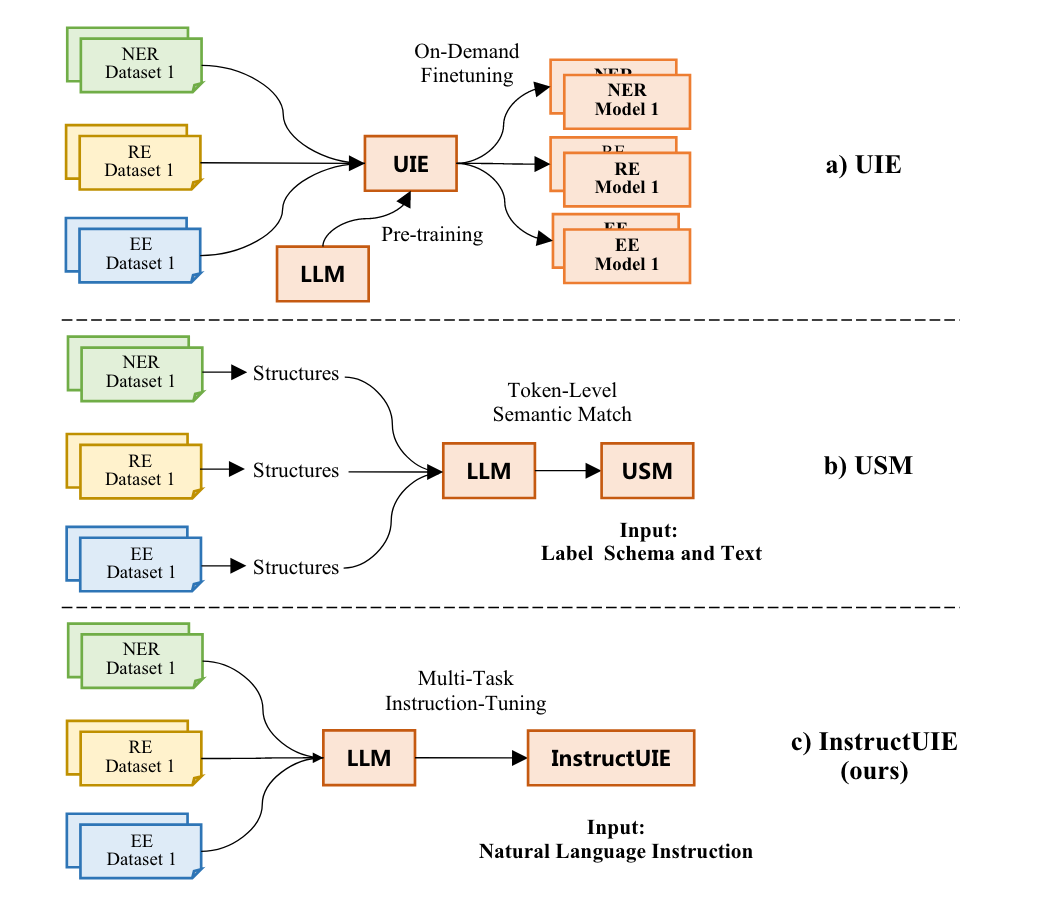}}
\caption{Illustration of different paradigms for solving unified information extraction task.~\cite{Wang2023InstructUIEMI}}.
\label{fig:uiewithinstructuie}
\end{figure} 
\begin{figure}[htbp]
\centering{\includegraphics[width=0.9\linewidth]{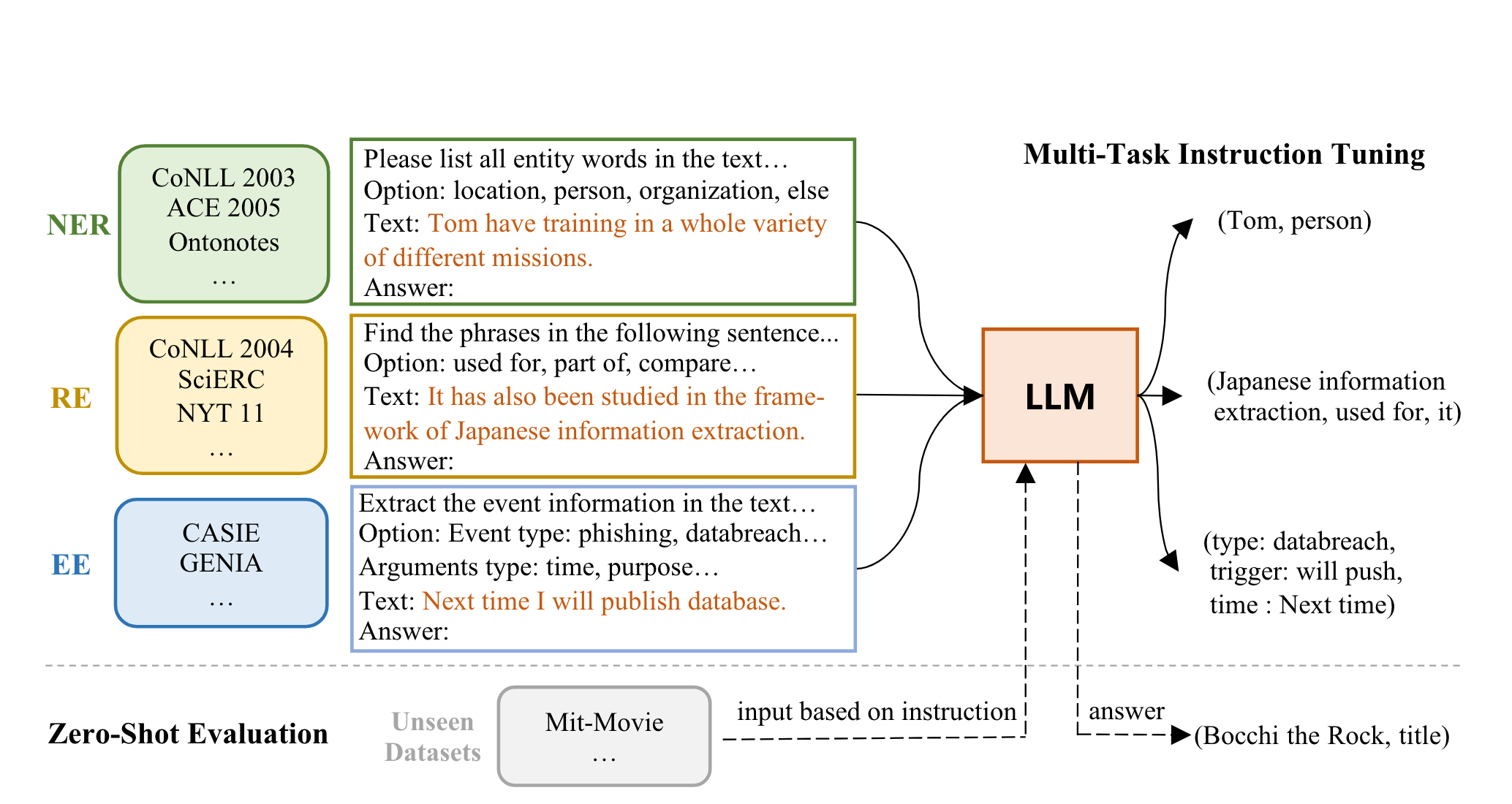}}
\caption{The overview framework of InstructUIE. The input consists of task instructions,options,and text. The output is a more understandable sentence converted from the original label structures\cite{Wang2023InstructUIEMI}}.
\label{fig:InstructUIE}
\end{figure} 
UIE (Unified Information Extraction)~\cite{Lu2022UnifiedSG}  is a unified framework for information extraction that encodes different extraction structures through structured extraction language and adaptively generates target extractions through a pattern-based prompting mechanism. The goal of UIE~\cite{Lu2022UnifiedSG} is to capture the common information extraction capabilities through a large-scale pre-trained text-to-structure model. InstructUIE~\cite{Wang2023InstructUIEMI} extends the functionality of UIE~\cite{Lu2022UnifiedSG},enabling it to unlock the potential of large language models for multi-task information extraction through instructive prompts.

InstructUIE~\cite{Wang2023InstructUIEMI}(Fig.~\ref{fig:uiewithinstructuie},~\ref{fig:InstructUIE}) is a unified information extraction framework based on instruction tuning,capable of unifying the modeling of various information extraction tasks and capturing dependencies between tasks. InstructUIE~\cite{Wang2023InstructUIEMI}  was proposed to address the difficulties that existing large models face in information extraction tasks. For example,gpt-3.5-turbo scored an F1 of 18.22 on the Ontonotes dataset,far below state-of-the-art performance. With instruction tuning,InstructUIE~\cite{Wang2023InstructUIEMI}  has been validated on 32 different information extraction datasets,and experimental results show that this method achieves performance comparable to Bert in supervised settings and significantly outperforms gpt3.5 and domain bests in zero-shot settings.

The main contributions of InstructUIE~\cite{Wang2023InstructUIEMI}  include:
\begin{enumerate}
\item Proposing an end-to-end framework for universal extraction,InstructUIE~\cite{Wang2023InstructUIEMI} ,which uses natural language instructions to guide large models in completing information extraction tasks.
\item Developing a benchmark called IE INSTRUCTIONS,consisting of 32 different information extraction datasets,allowing for consistent and standardized evaluation of various information extraction tasks.
\item Experimental results indicate that InstructUIE~\cite{Wang2023InstructUIEMI}  achieves performance comparable to Bert in supervised settings and significantly outperforms gpt3.5 and domain bests in zero-shot settings.
\end{enumerate}
InstructUIE~\cite{Wang2023InstructUIEMI}  represents information extraction tasks as natural language generation tasks,using descriptive instructions to help the model understand different tasks and constrain the output space. Additionally,to enable the model to capture common structural information and deepen its understanding of different semantics,InstructUIE~\cite{Wang2023InstructUIEMI}  introduces entity span extraction tasks and entity type tasks for NER,entity pair extraction tasks and entity pair relation identification tasks for relation extraction,and trigger extraction tasks and argument extraction tasks for event extraction.

\section{Alignment Tuning}
\subsection{RHLF}
\begin{figure}[htbp]
\centering{\includegraphics[width=0.9\linewidth]{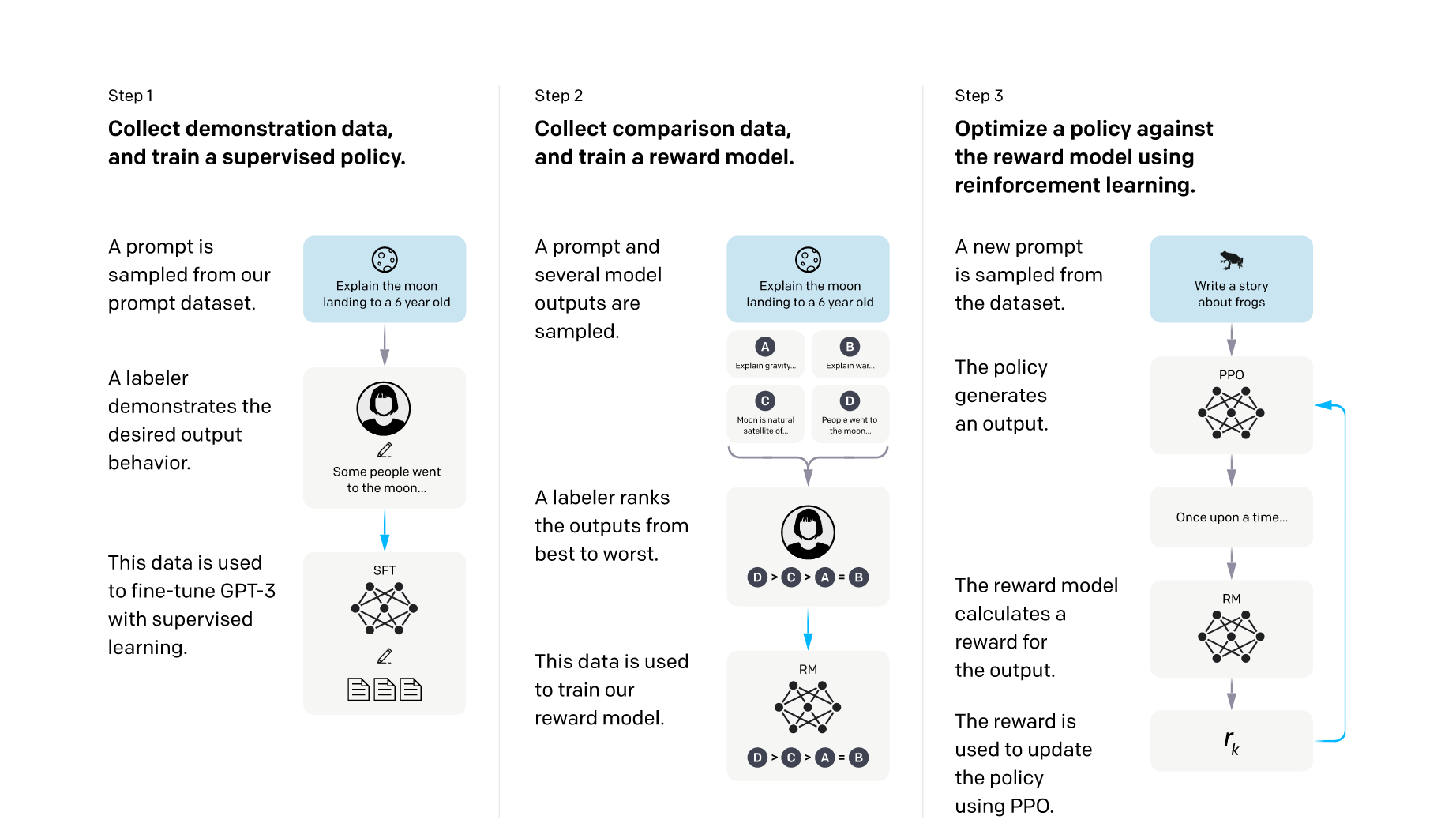}}
\caption{A diagram illustrating the three steps of our method: (1) supervised fine-tuning (SFT),(2) reward model (RM) training,and (3) reinforcement learning via proximal policy optimization (PPO) on this reward model. Blue arrows indicate that this data is used to train one of InstructGPT models. In Step 2,boxes A-D are samples from our models that get ranked by labelers.\cite{Ouyang2022TrainingLM_RLHF}}
\label{fig-RLHF}
\end{figure} 
RHLF~\cite{Ouyang2022TrainingLM_RLHF} Fine-tuning is a technique used to optimize large language models by leveraging human feedback. It enables language models to better understand and adapt to complex human preferences,thereby improving their performance and safety. The basic idea of RHLF~\cite{Ouyang2022TrainingLM_RLHF} Fine-tuning is as follows (Fig.~\ref{fig-RLHF}):

\begin{figure}[htbp]
\centering{\includegraphics[width=0.9\linewidth]{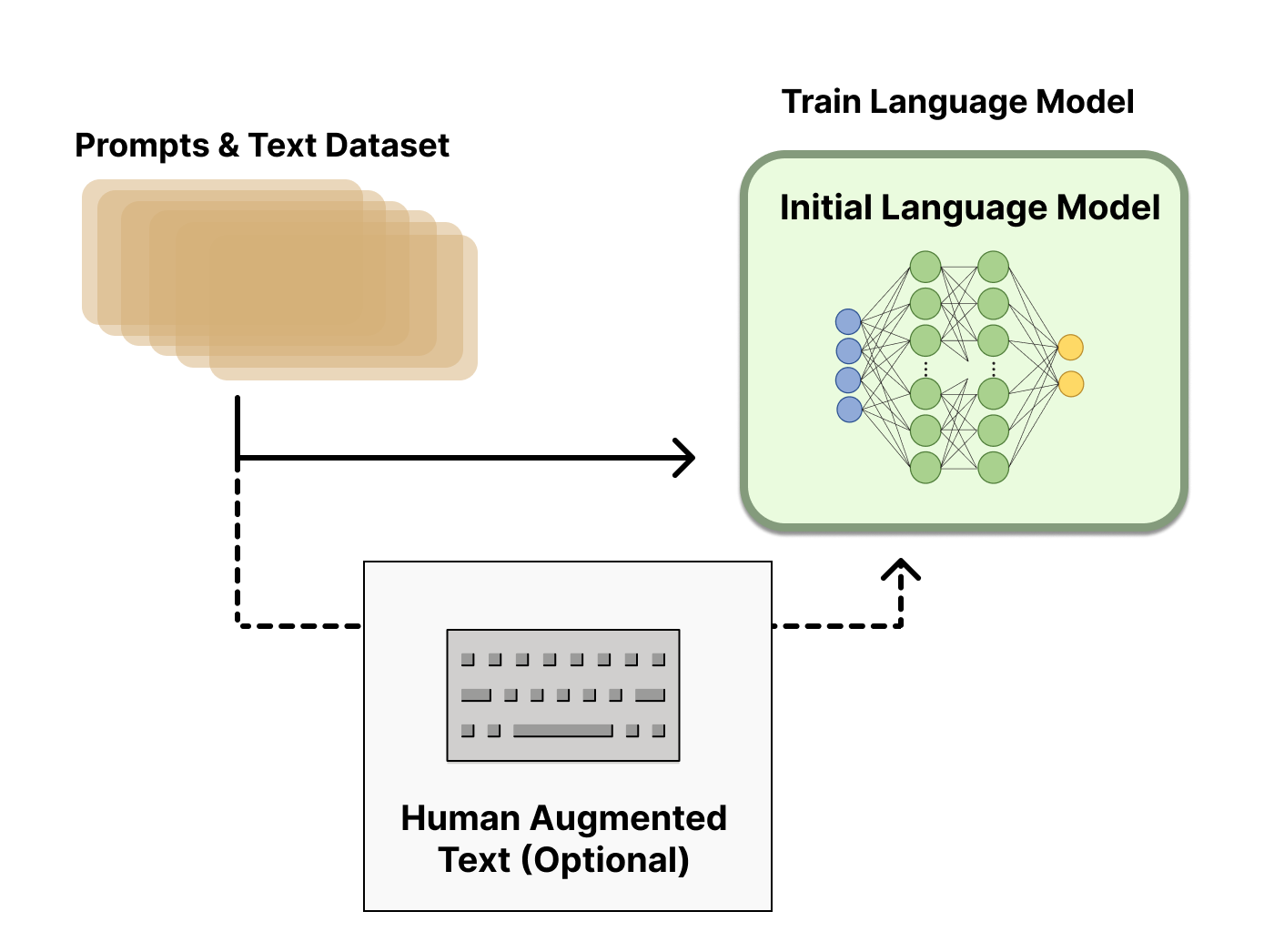}}
\caption{Pretraining a language model (LM).\cite{RLHFChatGPT}}
\label{fig-rlhf_pretraining}
\end{figure} 
\textbf{1.Supervised Fine-tuning(Fig.~\ref{fig-rlhf_pretraining}):} Collect human feedback on the model's outputs,including identifying which outputs are good,which are bad,or ranking different outputs.

\begin{figure}[htbp]
\centering{\includegraphics[width=0.9\linewidth]{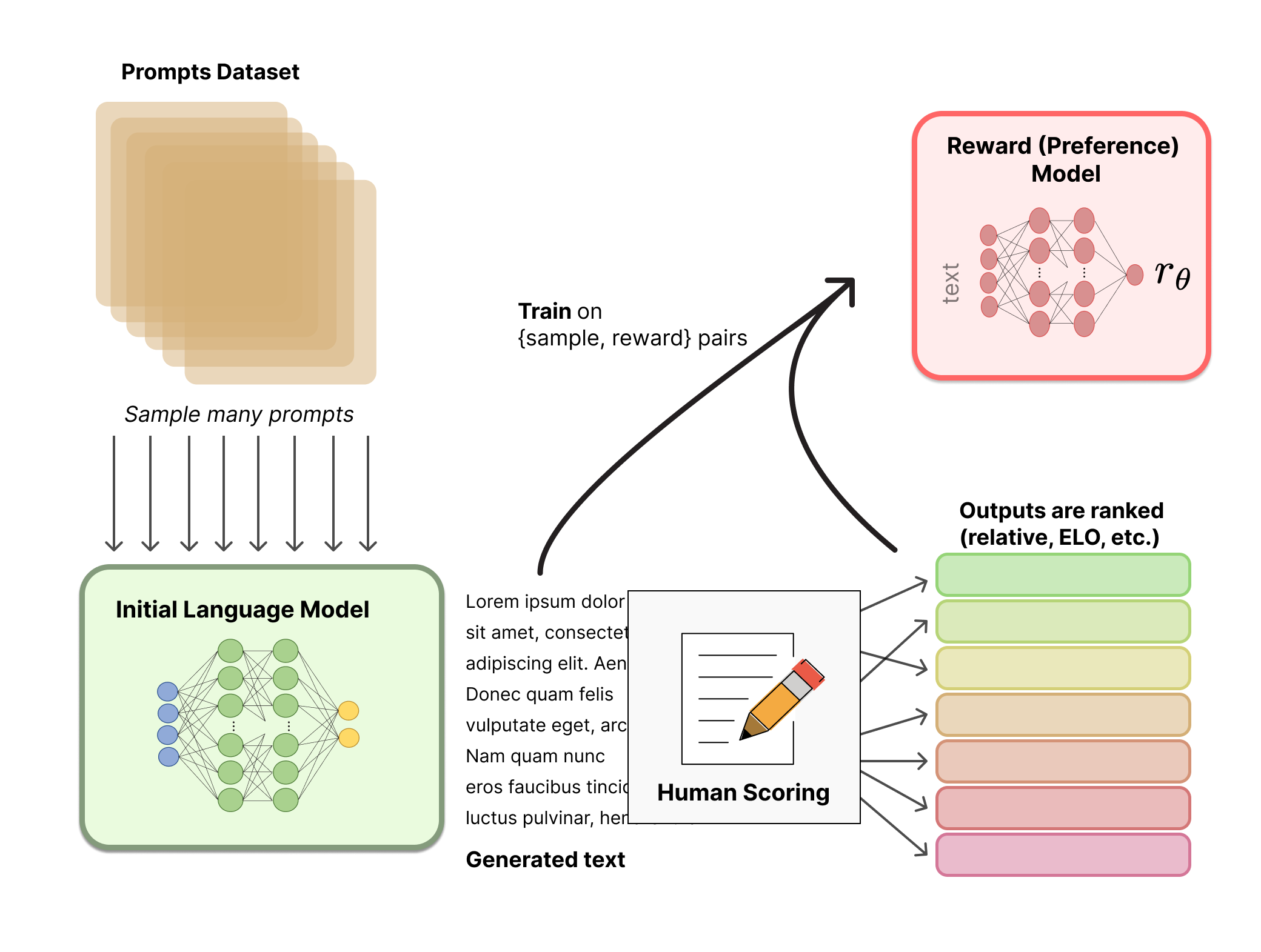}}
\caption{Gathering data and Training a Reward Mode.\cite{RLHFChatGPT}}
\label{fig-rlhf_reward-model}
\end{figure} 
\textbf{2. Reward Model Training(Fig.~\ref{fig-rlhf_reward-model}):} Train a reward model based on human feedback to predict the quality of the language model's outputs,assigning a score to each output.

\begin{figure}[htbp]
\centering{\includegraphics[width=0.9\linewidth]{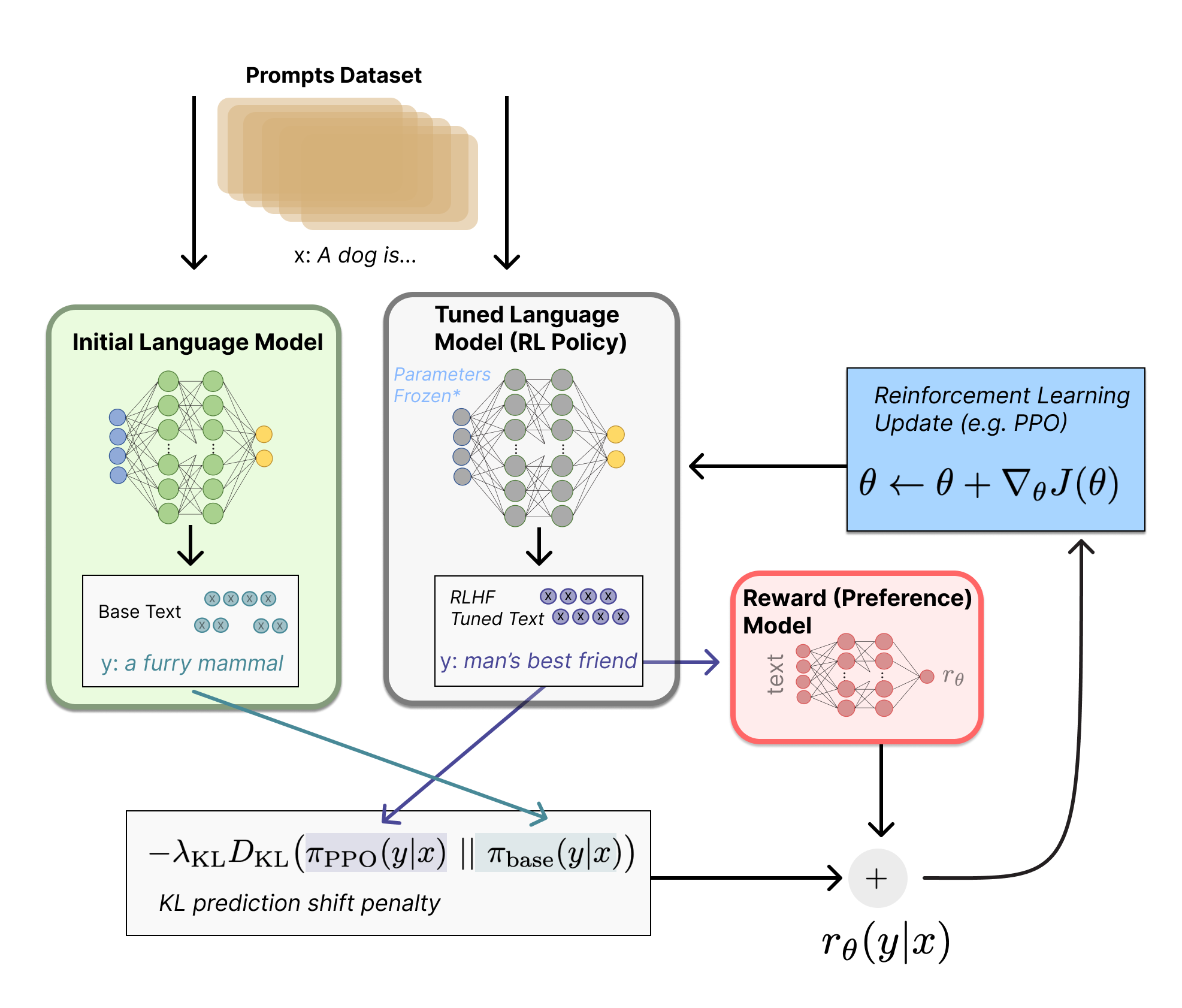}}
\caption{Fine-tuning the LM with Reinforcement Learning.\cite{RLHFChatGPT}}
\label{fig-rlhf_finetuningrlhf}
\end{figure} 
\textbf{3. Fine-tuning the LM with Reinforcement Learning~\cite{Schulman2017ProximalPO}(Fig.~\ref{fig-rlhf_finetuningrlhf}):} Use reinforcement learning methods to adjust the language model's parameters based on the scores from the reward model,making it more inclined to generate high-scoring outputs.Using PPO~\cite{Schulman2017ProximalPO} algorithm, fine-tune the policy model in multiple environments, and optimize an objective function. The objective function is a constrained optimization problem, which aims to maximize the output of the reward model and control the difference between the new and old policies

The key goal of RHLF Fine-tuning is to align the language model more closely with human preferences and expectations,enhancing its overall performance and safety.

The advantages of RLHF~\cite{Ouyang2022TrainingLM_RLHF} include:
\begin{itemize}
    \item It aligns the language model more closely with human expectations and values,rather than relying solely on datasets or fixed evaluation metrics.
    \item It enables the language model to exhibit creativity and diversity in its outputs,rather than generating repetitive or mundane responses.
    \item It enhances the robustness and exploratory nature of the language model,allowing it to handle semantic,logical,emotional,and other aspects of language understanding.
\end{itemize}

Applications of RLHF~\cite{RLHFChatGPT} include: 

\textbf{Dialogue Systems},RLHF can make dialogue systems provide more interesting,friendly,and reasonable responses,as seen in OpenAI's ChatGPT and InstructGPT~\cite{Ouyang2022InstructGPT}.

\textbf{Text Summarization},RLHF can lead to more concise,accurate,and useful summaries generated by text summarization systems,as demonstrated by DeepMind's Sparrow~\cite{Glaese2022Sparrow}.

\textbf{Natural Language Understanding},RLHF can help natural language understanding systems better handle semantic,logical,emotional,and other language-related challenges.

The TRL library,DeepSpeed library,and ColossalAI Chat are all toolkits used for RHLF (Reinforcement Learning from Human Feedback). They are designed to optimize large language models using human feedback to improve performance and safety. 

\textbf{TRL},developed by the Hugging Face team and based on the transformers library,supports various pre-trained language models and downstream tasks. The TRL library implements the RHLF method described in the InstructGPT paper~\cite{Ouyang2022InstructGPT},consisting of three steps: supervised fine-tuning,reward model training,and reinforcement learning optimization. The TRL library provides multiple text prompt templates and supports custom text prompts,making it applicable to various downstream tasks such as text classification,text generation,text summarization,and question-answering. Usage guidelines and example code for the TRL library can be found in the github~\footnote{Hugging face TRL Github:\label{section:trl}\url{https://github.com/huggingface/trl}}.

\textbf{DeepSpeed},developed by Microsoft and based on PyTorch,supports large-scale distributed training and inference. The DeepSpeed library offers the DeepSpeed-Chat module for training models like ChatGPT,also employing the RHLF approach with three steps: supervised fine-tuning,reward model training,and reinforcement learning optimization. The DeepSpeed-Chat module leverages DeepSpeed's mixed-precision engine,allowing seamless switching between inference and training modes,achieving efficient language generation and memory optimization. Usage instructions and example code for the DeepSpeed-Chat module can be found in the github \footnote{DeepSpeed Github:\label{section:deepspeed}\url{https://github.com/microsoft/DeepSpeed}}.

\textbf{ColossalAI Chat} is developed based on the ColossalAI framework and also supports various pre-trained language models and downstream tasks. ColossalAI Chat implements the RHLF method as part of the ColossalChat project,involving three steps: supervised data collection,supervised fine-tuning,reward model training,and reinforcement learning fine-tuning. ColossalAI Chat utilizes ColossalAI's distributed training and quantization inference capabilities,enabling training and deployment of large language models on memory-constrained devices. Usage guidelines and example code for ColossalAI Chat can be found in the github \footnote{ColossalAI Chat Github:\label{section:colossalai}\url{https://github.com/TongLi3701/ColossalChat}}.

In summary,the TRL library,DeepSpeed library,and ColossalAI Chat are excellent toolkits for RHLF,allowing you to adapt pre-trained language models to various downstream tasks efficiently,using minimal additional parameters or text prompts. You can use them to implement a wide range of language applications,such as text generation,text summarization,text classification,and more. If you're interested,you can refer to their official documentation and example code to learn about their usage methods and techniques.

\subsection{DPO}
Direct Preference Optimization (DPO)~\cite{Rafailov2023DirectPODPO} is a novel approach for fine-tuning large unsupervised language models (LMs) to align with human preferences. Unlike existing methods,DPO is a computationally efficient and stable algorithm that eliminates the need to fit a reward model,sample from the LM,or perform extensive hyperparameter tuning during fine-tuning123.
\begin{figure}[htbp]
\centering{\includegraphics[width=0.9\linewidth]{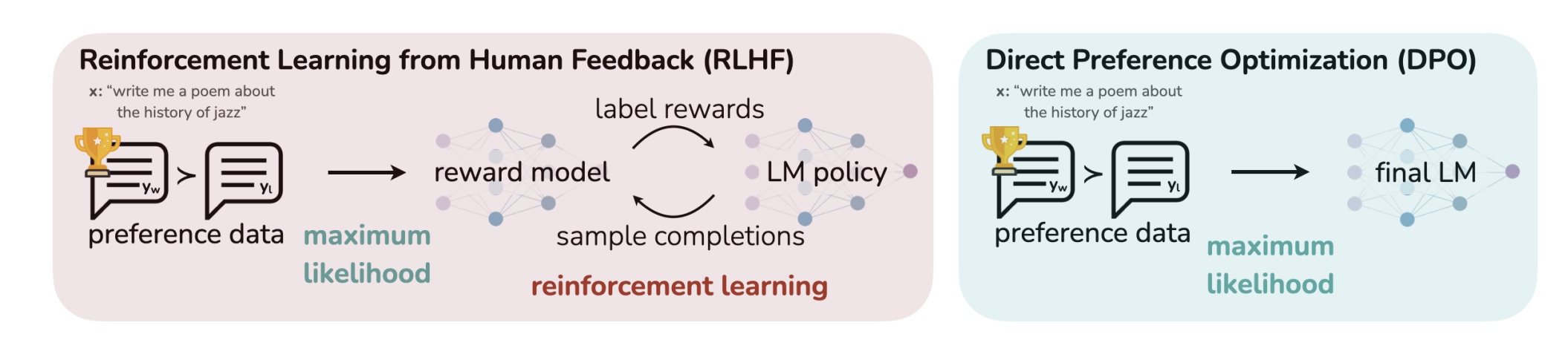}}
\caption{DPO optimizes for human preferences while avoiding reinforcement learning.~\cite{Rafailov2023DirectPODPO}}
\label{fig:DPO}
\end{figure} 
The core idea of DPO~\cite{Rafailov2023DirectPODPO}(Fig.~\ref{fig:DPO}) is to directly optimize the language model to match preference data,rather than through traditional reinforcement learning methods. This is achieved through a key insight: an analytic mapping from the reward function to the optimal RL policy,allowing the authors to transform the RL loss based on rewards and a reference model into a loss directly against the reference model.This mapping intuitively measures the consistency of a given reward function with given preference data. Thus,DPO~\cite{Rafailov2023DirectPODPO} starts with the optimal solution to the RLHF loss and derives a loss solely against the reference model through variable transformation.

The DPO~\cite{Rafailov2023DirectPODPO} method is now available in the TRL\textsuperscript{\ref{section:trl}} library and demonstrates how to fine-tune the latest Llama v2 7B parameter model on the stack-exchange preference dataset~\cite{DPOLLama2},which contains ranked answers to questions across various stack-exchange portals.

\subsection{NLHF}
Nash Learning from Human Feedback (NLHF)~\cite{Munos2023NashLF} is a novel approach for fine-tuning large language models (LLMs) through pairwise human feedback. This method begins by learning a preference model that is conditioned on two inputs given a prompt. It then pursues a strategy that consistently generates responses more preferred than those produced by any competing policy,thereby defining the Nash equilibrium of this preference model. The Nash-MD algorithm generates a sequence of policies,with the final iteration converging to the regularized Nash equilibrium. NLHF~\cite{Munos2023NashLF} offers an attractive pathway for preference learning and policy optimization,with the potential to advance the field of aligning LLMs with human preferences.

\section{AgentTuning}
Agent-Tuning is a fine-tuning technique that focuses on using agents to optimize pre-trained large language models (LLMs) so that they can better adapt to specific tasks or domains. In Agent-Tuning,agents refer to modules or components within the model that play specific roles and can be fine-tuned to improve the model’s performance on specific tasks.

The basic steps for fine-tuning with agents typically include the following aspects:

\textbf{Selecting the appropriate agents:} Choose one or more agents for fine-tuning based on your application needs. These agents can be part of the pre-trained model or new components designed for specific tasks.

\textbf{Preparing data:} Collect and prepare a dataset for fine-tuning. This data should be relevant to your task and representative enough for the agents to learn from.

\textbf{Fine-tuning agents:} Fine-tune the selected agents using your dataset. This may involve adjusting the agents parameters or interactive learning between agents.

\textbf{Evaluation and iteration:} After fine-tuning,evaluate the agents performance and iterate for improvement as needed.

\subsection{AgentTuning: Enabling Generalized Agent Abilities for LLMs}
\begin{figure}[htbp]
\centering{\includegraphics[width=0.9\linewidth]{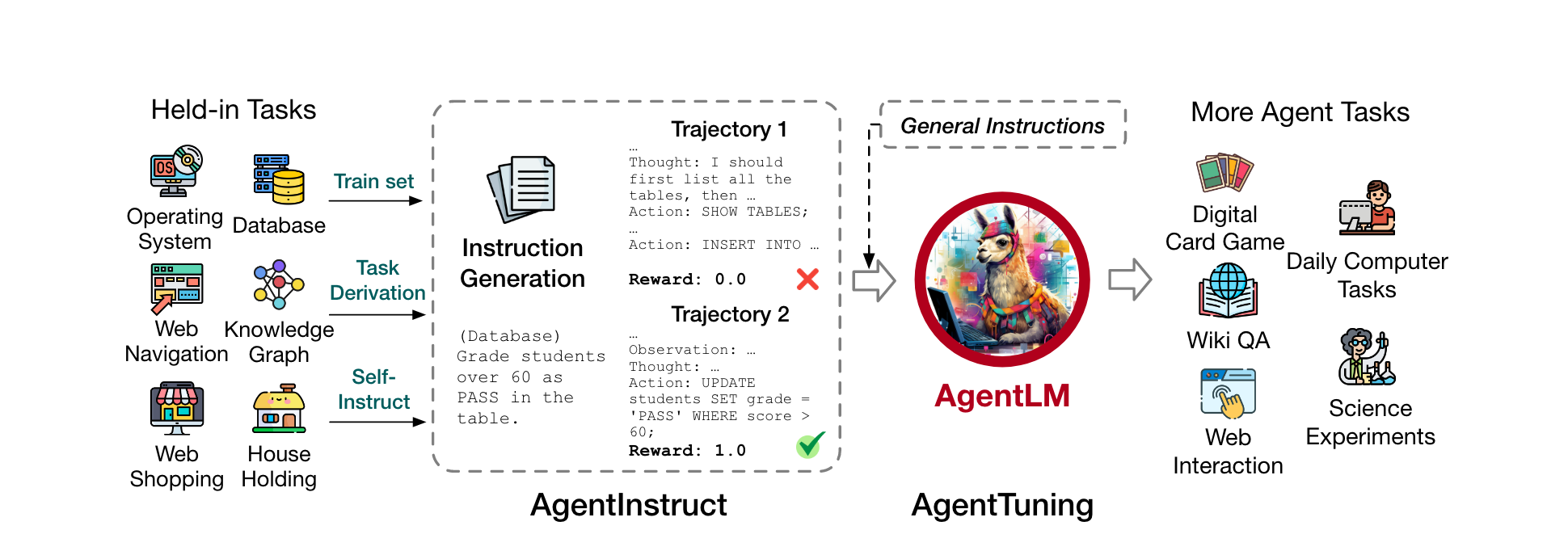}}
\caption{An overview of AgentInstruct and AgentTuning. The construction of AgentInstruct,consisting of instruction generation,trajectory interaction,and trajectory filter. AgentLM is finetuned using a mixture of AgentInstruct and general-domain instructions.~\cite{Zeng2023AgentTuningEG}}
\label{fig-Agent-Tuning}
\end{figure}

AgentTuning\cite{Zeng2023AgentTuningEG}(Fig.~\ref{fig-Agent-Tuning}) is designed to enhance the capabilities of large language models (LLMs) as agents in performing tasks,without compromising their general abilities. AgentTuning\cite{Zeng2023AgentTuningEG} combines a lightweight instruction tuning dataset,AgentInstruct,with open-source instructions from the general domain,adopting a mixed instruction tuning strategy to adjust the Llama 2 series,ultimately resulting in the AgentLM model.
The evaluation results of the paper show that AgentTuning\cite{Zeng2023AgentTuningEG} has improved the agent abilities of LLMs without harming their general capabilities. The AgentLM-70B\footnote{AgentLM-70B:\label{section:AgentLM-70B}\url{https://huggingface.co/THUDM/agentlm-70B}} model is comparable to GPT-3.5-turbo on unseen agent tasks,demonstrating a broad range of agent abilities. Moreover,the authors of the paper have open-sourced the AgentInstruct dataset and the AgentLM-7B\footnote{AgentLM-7B:\label{section:AgentLM-7B}\url{https://huggingface.co/THUDM/agentlm-7b}},AgentLM-13B\footnote{AgentLM-7B:\label{section:AgentLM-13B}\url{https://huggingface.co/THUDM/agentlm-13B}},AgentLM-70B models\textsuperscript{\ref{section:AgentLM-70B}},providing a powerful open-source LLM alternative for agent tasks\footnote{AgentTuning Github:\label{section:AgentTuning}\url{https://github.com/THUDM/AgentTuning}}.

AgentTuning\cite{Zeng2023AgentTuningEG} is significant for understanding how to enhance the performance of LLMs on specific tasks,especially in scenarios that require agent abilities. It offers new perspectives and methods for the further development and application of LLMs.

\section{RAG-Memory-Finetuning}
RAG-Memory-Finetuning is a technique to optimize the performance of large language models (LLMs) by combining Retrieval Augmented Generation (RAG) with fine-tuning. This approach aims to connect LLMs to external knowledge sources through a retrieval mechanism and combine it with generative capabilities to search and integrate relevant information from knowledge bases. The goal of RAG-Memory-Finetuning is to improve the consistency and reliability of the output and reduce hallucination issues.

Specifically,the RAG technique is implemented through the following steps:
\begin{enumerate}
\item Chunking an external domain-specific knowledge base into small documents,each about 150 words.
\item Creating embeddings using a pre-trained model and storing document vectors in a vector database.
\item When an input query is passed to the LLM,the most relevant information is retrieved from the external database using metrics such as cosine similarity and combined with the LLM as additional context.
\item This external context and the input prompt are passed together to the text generator to produce the output response.
\end{enumerate}
Responses generated by RAG are more factual,specific,and diverse. The parameter knowledge provided by traditional fine-tuning techniques is static,while RAG allows us to bypass retraining and obtain up-to-date information through retrieval-based generation to produce reliable outputs.
\begin{figure}[htbp]
\centering{\includegraphics[width=0.9\linewidth]{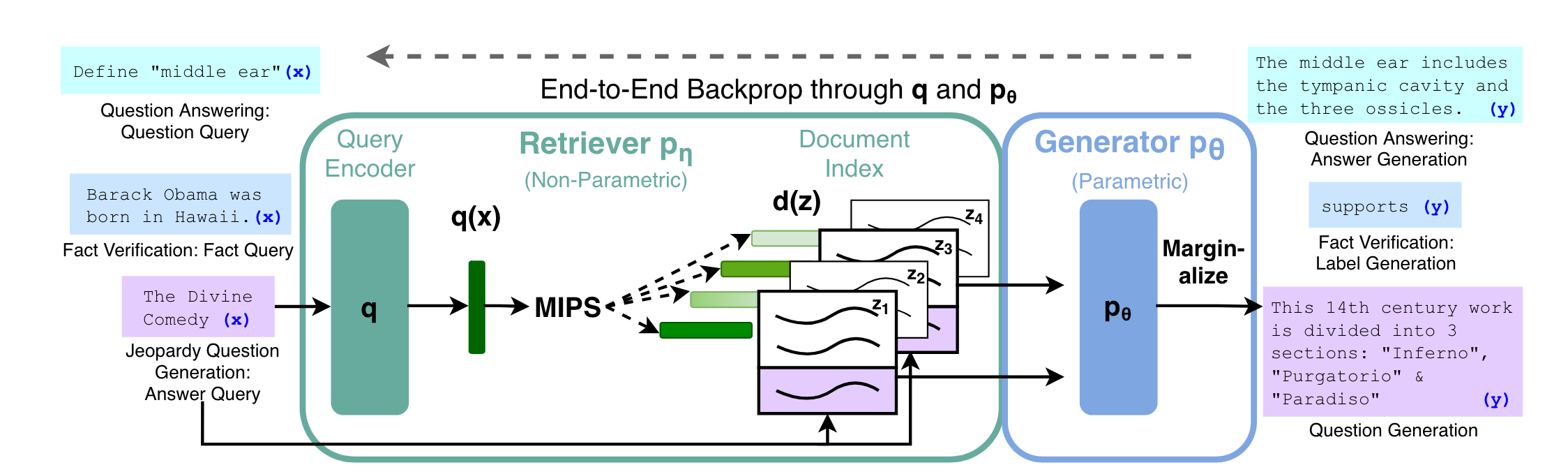}}
\caption{Overview of RAG.~\cite{Lewis2020RetrievalAugmentedGF}}
\label{fig-RAG}
\end{figure} 

\textbf{``Retrieval-Augmented Generation for Knowledge-Intensive NLP Tasks\cite{Lewis2020RetrievalAugmentedGF}(Fig.~\ref{fig-RAG}):''} This paper introduces the RAG model,which significantly enhances the performance of knowledge-intensive NLP tasks by combining a pre-trained generative model with non-parametric knowledge retrieval.
\begin{figure}[htbp]
\centering{\includegraphics[width=0.9\linewidth]{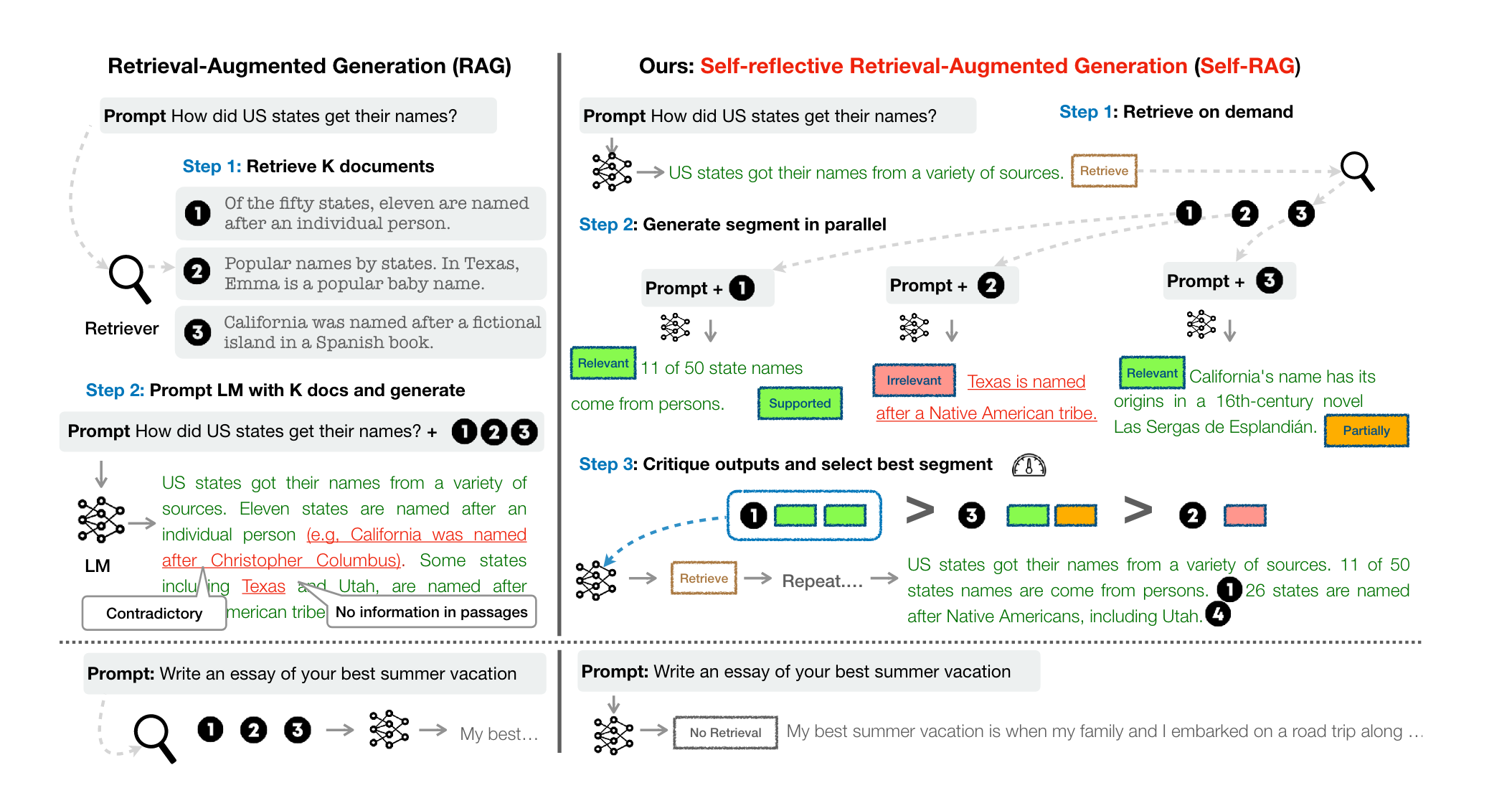}}
\caption{Overview of SELF-RAG.SELF-RAG learns to retrieve,critique,and generate text passages to enhance overall generation quality,factuality,and verifiability.~\cite{Asai2023SelfRAGLT}}
\label{fig-SELF-RAG}
\end{figure} 
\begin{algorithm}
\caption{SELF-RAG Inference}
\begin{algorithmic}[1]
\Require Generator LM \( \mathcal{M} \),Retriever \( \mathcal{R} \),Large-scale passage collections \( \{d_1,\ldots,d_N\} \)
\State \textbf{Input}: input prompt \( x \) and preceding generation \( y_{<t} \)
\State \textbf{Output}: next output segment \( y_t \)
\State \( \mathcal{M} \) predicts \fcolorbox{red}{white}{Retrieve} given \( (x,y_{<t}) \)
\If{\fcolorbox{red}{white}{Retrieve} \( == \) Yes}
    \State Retrieve relevant text passages \( D \) using \( \mathcal{R} \) given \( (x,y_{t-1}) \)
    \State \( \mathcal{M} \) predicts \fcolorbox{blue}{white}{IsRel} given \( x,d \) and \( y_t \) given \( x,d,y_{<t} \) for each \( d \in D \)
    \State \( \mathcal{M} \) predicts \fcolorbox{blue}{white}{IsSup} and \fcolorbox{blue}{white}{IsUse} given \( x,y_t,d \) for each \( d \in D \)
    \State Rank \( y_t \) based on \fcolorbox{blue}{white}{IsRel},\fcolorbox{blue}{white}{IsSup},\fcolorbox{blue}{white}{IsUse}
\ElsIf{\fcolorbox{red}{white}{Retrieve} \( == \) No}
    \State \( \mathcal{M}_{gen} \) predicts \( y_t \) given \( x \)
    \State \( \mathcal{M}_{gen} \) predicts \fcolorbox{blue}{white}{IsUse} given \( x,y_t \)
\EndIf
\end{algorithmic}
\label{alg:self-rag}
\end{algorithm}

\textbf{``Self-RAG: Learning to Retrieve,Generate,and Critique through Self-Reflection\cite{Asai2023SelfRAGLT}(Fig.~\ref{fig-SELF-RAG}):''} This paper introduces a new retrieval-augmented framework that enhances retrieval through self-reflection and also addresses the hallucination issues present in large models.

Algorithm~\ref{alg:self-rag} present an overview of SELF-RAG~\cite{Asai2023SelfRAGLT} at inference. For every $x$ and preceding generation $y<t$,the model decodes a retrieval token to evaluate the utility of retrieval. If retrieval is not required,the model predicts the next output segment,as it does in a standard LM. If retrieval is needed,the model generates: a critique token to evaluate the retrieved passage’s relevance,the next response segment,and a critique token to evaluate if the information in the response segment is supported by the passage. Finally,a new critique token evaluates the overall utility of the response.

\textbf{``REPLUG: Retrieval-Augmented Black-Box Language Models\cite{Shi2023REPLUGRB}:''} This research explores how to optimize retrieval results in black-box language models,such as those that only expose APIs without revealing embeddings,using a retrieval-augmented approach.

\textbf{``Atlas: Few-shot Learning with Retrieval Augmented Language Models\cite{Izacard2022FewshotLW}:''} The paper discusses methods for joint training of retrievers and language models,especially for knowledge-intensive tasks.

\textbf{``RA-DIT: RETRIEVAL-AUGMENTED DUAL INSTRUCTION TUNING\cite{Lin2023RADITRD}:''} A lightweight fine-tuning method that combines RAG and SFT is proposed to enhance the performance of retrieval-augmented language models.

\textbf{``Improving the Domain Adaptation of Retrieval Augmented Generation (RAG) Models for Open Domain Question Answering~\cite{Siriwardhana2022ImprovingTD}:''} This paper assesses the domain adaptability of RAG models in open-domain question answering (ODQA) tasks and proposes RAG-end2end,an extension of RAG that can adapt to specific domain knowledge bases by updating all components during training.

\textbf{``RAG Vs Fine-Tuning Vs Both: A Guide For Optimizing LLM Performance~\cite{RAGFineTuning}:''} This article provides a guide on the optimization strategies of RAG,fine-tuning,and their combination in large language models (LLMs).

\section{Experiments}

To validate the effectiveness of efficient parameter fine-tuning,different models are compared on five samll text classification datasets using the LoRA fine-tuning approach,and various model sizes are used to verify the impact of pre-trained parameter size on performance. The LoRA paper has already made a good comparison with other fine-tuning methods,confirming its superiority,and the results can be seen in Table~\ref{table:Lora-Result}.
\subsection{Datasets}\label{sec:datasets}
\textbf{AG News\footnote{AG News:\label{section:AG_News}\url{https://huggingface.co/datasets/ag_news}}} The AG News dataset is a text classification dataset that is a subset of the AG’s corpus of news articles. This dataset consists of the titles and description fields of news articles,covering the four largest categories: “World,” “Sports,” “Business,” and “Sci/Tech.” The AG News dataset contains 30,000 training samples and 1,900 test samples for each category. It is commonly used for text classification tasks in natural language processing,such as training and evaluating machine learning models.

\textbf{Auditor Sentiment\footnote{Auditor Sentiment:\label{section:auditor-sentiment}\url{https://huggingface.co/datasets/ihassan1/auditor-sentiment}}} Auditor Sentiment dataset is a text classification dataset available on Hugging Face Datasets. It is designed for the task of sentiment scoring and is monolingual,with annotations and language created by experts. The dataset is tagged with terms such as auditor,financial,and sentiment,indicating its focus on financial texts and their sentiment analysis1.

This dataset is likely to contain sentences from financial reports or news articles related to auditing,with labels indicating the sentiment of each sentence,such as positive,neutral,or negative. It can be used to train machine learning models for sentiment analysis in the financial domain,providing a valuable resource for researchers and practitioners interested in the intersection of natural language processing and finance.

\textbf{Shawhin/Imdb-Truncated\footnote{Shawhin/Imdb-Truncated:\label{section:imdb-truncated}\url{https://huggingface.co/datasets/shawhin/imdb-truncated}}} The shawhin/imdb-truncated dataset is a text classification dataset available on Hugging Face Datasets. This dataset is a truncated version of the IMDb movie review dataset,containing movie reviews with binary sentiment labels (positive or negative). It is commonly used for sentiment analysis tasks,such as training and evaluating machine learning models to identify the sentiment orientation of reviews.

As a truncated version of the IMDb dataset,the shawhin/imdb-truncated dataset likely contains fewer samples but still retains the basic structure and features of the original dataset. This makes it a suitable resource for quick experimentation and prototyping,especially when computational resources are limited.

\textbf{Financial PhraseBank\footnote{Financial-PhraseBank:\label{section:financial_phrasebank}\url{https://huggingface.co/datasets/financial_phrasebank/viewer/sentences_allagree}}} The Financial PhraseBank is a sentiment-annotated dataset containing sentences from English financial news. This dataset consists of 4,840 sentences,which are classified according to sentiment as positive,negative,or neutral. The purpose of the dataset is to establish a standard for manual annotation to benchmark different modeling techniques.

The annotations for the dataset were completed by 16 annotators with knowledge of financial markets,including researchers and master’s students from the Aalto University School of Business,with majors primarily in finance,accounting,and economics.

\textbf{Tweet Eval\footnote{Tweet Eval:\label{section:tweet_eval}\url{https://huggingface.co/datasets/tweet_eval/}}} The Tweet Eval dataset is a multi-task classification dataset specifically designed for Twitter text. It includes seven different classification tasks,covering a range of aspects such as irony,hate,offensiveness,stance,emojis,sentiment,and emotion.

These tasks are presented in the form of multi-category tweet classification,aiming to address various challenges in text classification. The size of the dataset varies,with the amount of data for different tasks ranging from 1K to over 1M,making it suitable for research of various scales.

All tasks are unified into the same benchmark,with each task’s dataset presented in the same format and fixed training,validation,and testing splits. This facilitates fair comparison and evaluation by researchers.

\subsection{Evaluation Metrics}
\textbf{Confusion Matrix} 

\begin{equation}
\begin{array}{c|cc}
\multicolumn{1}{c}{} & \multicolumn{2}{c}{\textbf{Predicted Class}} \\
\cline{2-3}
\textbf{Actual Class} & \text{Positive} & \text{Negative} \\
\hline
\text{Positive} & TP & FN \\
\text{Negative} & FP & TN \\
\end{array}
\end{equation}

Where TP stands for True Positives, FN stands for False Negatives, FP stands for False Positives, and TN stands for True Negatives.

The main evaluation metrics for classification models include the following:

\textbf{Accuracy}:
Accuracy is the proportion of true results (both true positives and true negatives) among the total number of cases examined. Its formula is:
$$\textbf{Accuracy} = \frac{\text{TP} + \text{TN}}{\text{TP} + \text{TN} + \text{FP} + \text{FN}}$$
where TP is true positive,TN is true negative,FP is false positive,and FN is false negative. Accuracy is suitable for balanced classes but can be misleading in imbalanced datasets.

\textbf{Precision}:
Precision is the proportion of true positive predictions in all positive predictions. Its formula is:
\begin{equation}
\textbf{Precision} = \frac{\text{TP}}{\text{TP} + \text{FP}}
\end{equation}
Precision is used when the cost of a false positive is high,such as in spam email detection.

\textbf{Recall}:
Recall is the proportion of true positive predictions in all actual positives. Its formula is:
\begin{equation}
\textbf{Recall} = \frac{\text{TP}}{\text{TP} + \text{FN}}
\end{equation}
Recall is used when the cost of a false negative is high,such as in disease screening.

\textbf{F1-score}:
The F1-score is the harmonic mean of precision and recall,balancing the two metrics. Its formula is:
\begin{equation}
\textbf{F1-score} = 2 \times \frac{\text{Precision} \times \text{Recall}}{\text{Precision} + \text{Recall}} \label{eq:F1-score}
\end{equation}
The F1-score is used when precision and recall are equally important.

These metrics are commonly used in machine learning for classification tasks to understand the performance of a model in different aspects. 

\subsection{Setting}
We have chosen to evaluate the $\mathbf{BERT}$ family and \textbf{Flan-T5} family models on the aforementioned datasets(\nameref{sec:datasets}). 

Since some datasets are imbalanced,the F1-score metric was used for evaluation. The models from the Bert family include:
\begin{itemize}
\item $\mathbf{BERT_{base}(FT)}$ full-parameter fine-tuned model.
\item $\mathbf{BERT_{base}(LoRA)}$ fine-tuned model.
\item $\mathbf{BERT_{large}(LoRA)}$ fine-tuned model.
\item $\mathbf{Flan-T5{base}(FT)}$ full-parameter fine-tuned model.
\item $\mathbf{Flan-T5{base}(LoRA)}$ fine-tuned model.
\item $\mathbf{Flan-T5{large}(LoRA)}$ fine-tuned model.
\item $\mathbf{Flan-T5{XL}(LoRA)}$ fine-tuned model.
\end{itemize}
During the fine-tuning process,for the $\mathbf{BERT}$ family models,
\begin{itemize}
\item $epoch=5$.
\item $learning\_rate=1e-3$.
\item $max\_seq\_len=2048$.
\item $batch\_size=64$.
\item $gradient\_accumulation\_steps=4$.
\end{itemize}
For the $\textbf{Flan-T5}$ family models, 
\begin{itemize}
\item $epoch=5$.
\item $learning\_rate=1e-3$.
\item $max\_seq\_len=2048$.
\item $batch\_size=8$.
\item $gradient\_accumulation\_steps=8$. 
\end{itemize}
The slight differences in the configuration of the $\mathbf{BERT}$family and $\textbf{Flan-T5}$ family models mainly take into account the significant differences in model size and the varying computational power requirements.

To save computational resources,the Bert family of models uses 32-bit precision,the Flan-T5 family uses 16-bit precision,and the hardware environment employs the Mac-M2 MPS acceleration card,without using 4-bit or 8-bit quantization.This statement refers to the use of different numerical precisions for loading models in order to optimize the computational efficiency,and the choice of hardware acceleration to support these operations.
\begin{table*}[htbp]
\centering
\begin{tabular}{c|c|ccccccccc}
\hline Mode & \begin{tabular}{l} 
\# Full \\Parameters\end{tabular} 
& \begin{tabular}{l} 
\# Trainable \\Parameters\end{tabular} & \begin{tabular} {l} AG\\News\end{tabular} & \begin{tabular}{l}Auditor\\Sentiment\end{tabular} & \begin{tabular}{l}Financial\\Phrasebank\end{tabular} &\begin{tabular}{l} Shawhin \\Imdb \\Truncated \end{tabular} & \begin{tabular}{l}Tweet\\Eval \\ Irony\end{tabular} & \begin{tabular}{l}Tweet\\Eval \\Stance\\abortion\end{tabular}  & Avg. \\
\hline 
$\textbf{BERT}_{\text {base }}(\textbf{FT}) $&$103.1 \mathrm{M}$& $103.1 \mathrm{M}_{\textcolor{mygreen}{100\%}}$ & 10.0 & 42.2 & 46.7 & 33.3 & 30.9 & 35.9 &  33.2  \\

$\textbf{BERT}_{\text{base}}(\textbf{LoRA})$ & $103.1 \mathrm{M}$ & $2.2 \mathrm{M}_{\textcolor{myred}{2.1\%}}$ & 91.7 & 86.6 & 99.6 & 84.6 & 66.7 & 64.2 & 82.2 \\

$\textbf{BERT}_{\text{large}}(\textbf{LoRA})$ & $324.1\mathrm{M}$ & $6.0\mathrm{M}_{\textcolor{myblue}{1.8\%}}$ & 93.6 & 87.3 & \textbf{98.2} & 82.5 & 70.0 & 64.0 & 82.6 \\

\hline 

$\textbf{FLAN-T5}_{\text{base}(\textbf{FT})}$ & $236.0\mathrm{M}$ & $236.0\mathrm{M}_{\textcolor{mygreen}{100\%}}$ & \textbf{98.8} & 88.2 & 96.7 & 88.6 & 66.8 & 65.3 & 84.1 \\

$\textbf{FLAN-T5}_{\text{base}(\textbf{LoRA})}$ & $236.0\mathrm{M}$ & $6.7\mathrm{M}_{\textcolor{mycyan}{2.78\%}}$ & 95.1 & 85.9 & 91.7 & 87.0 & 62.6 & 54.8 & 80.0 \\

$\textbf{FLAN-T5}_{\text{large}(\textbf{LoRA})}$ & $764.4\mathrm{M}$ & $18.0\mathrm{M}_{\textcolor{myred}{2.35\%}}$ & 95.1 & 87.4 & 97.4 & 90.0 & 65.0 & 59.3 & 82.4 \\

$\textbf{FLAN-T5}_{\text{XL}(\textbf{LoRA})}$ & $2.7\mathrm{G}$ & $35.9\mathrm{M}_{\textcolor{myblue}{1.31\%}}$ & 96.8 & \textbf{90.3} & \textbf{98.2} & \textbf{93.0} & \textbf{73.4} & \textbf{69.6} & \textbf{\textcolor{myred}{86.8}} \\

\hline
\end{tabular}
\captionsetup{justification=justified,singlelinecheck=false}
\caption{BERT$_{\text{base}}$,BERT$_{\text{large}}$,FLAN-T5$_{\text{base}}$,FLAN-T5$_{\text{large}}$,and FLAN-T5$_{\text{XL}}$ with different fine-tuning methods on six text classification tasks. We report the F1-score for each task.}

\label{table:LoRA-FT-Result}
\end{table*}

\subsection{Result}
\begin{figure}[htbp]
\centering{\includegraphics[width=0.9\linewidth]{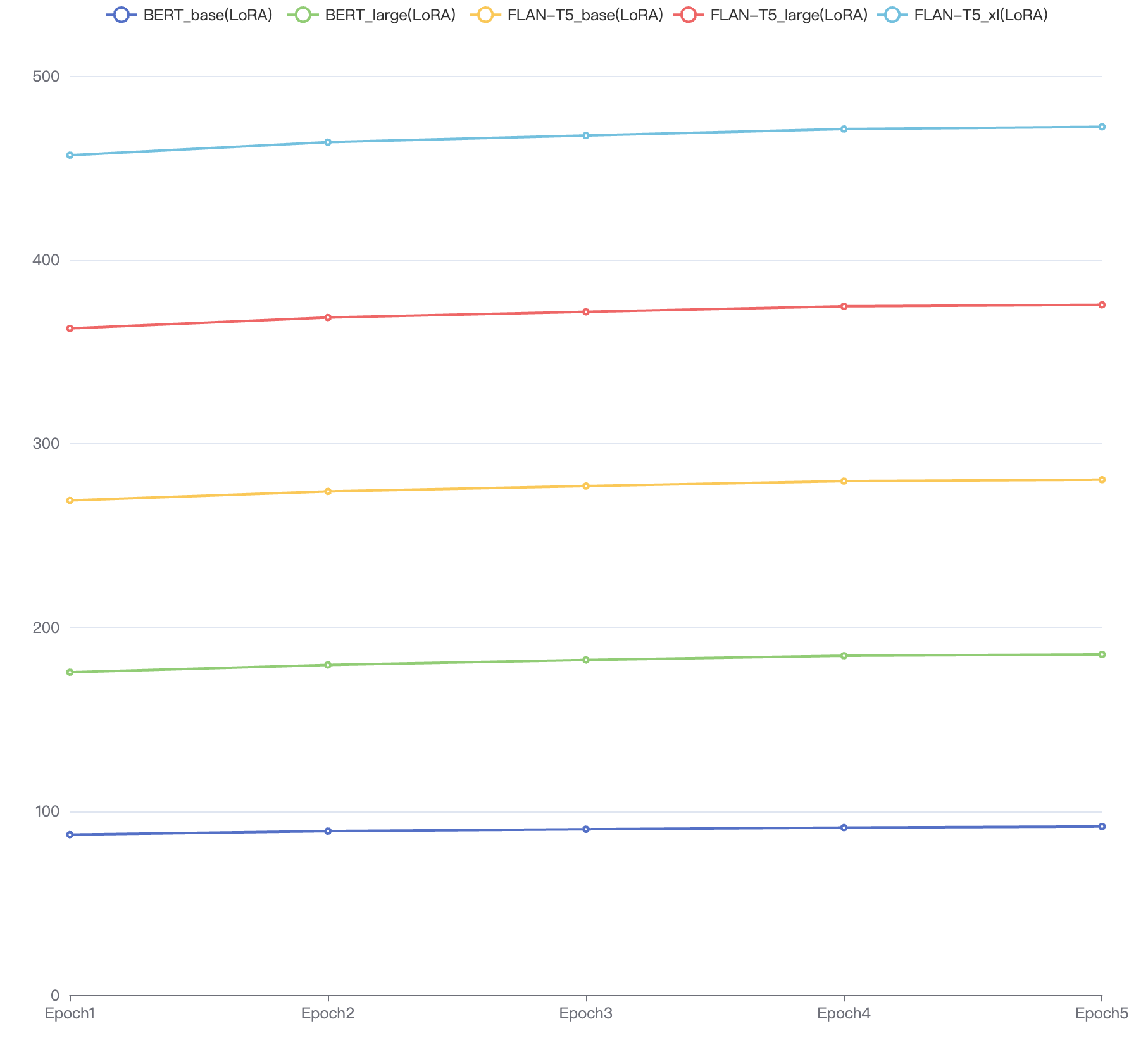}}
\caption{Comparison of F1-score(Equation:~\eqref{eq:F1-score}) performance across different epochs on \textbf{Age-news}~\ref{section:AG_News}.}
\label{fig:ag_news_epoch}
\end{figure} 

\begin{figure}[htbp]
\centering{\includegraphics[width=0.9\linewidth]{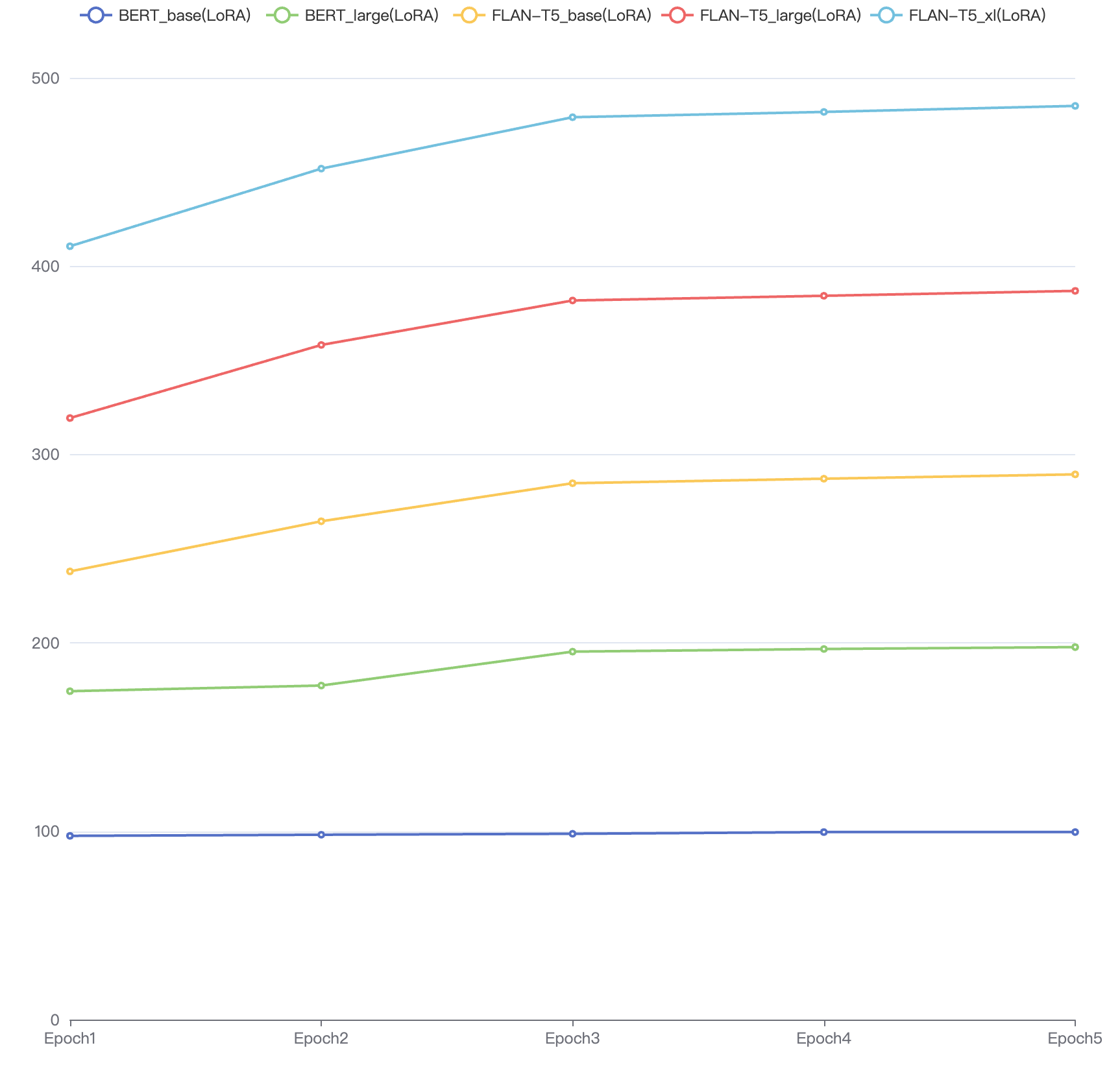}}
\caption{Comparison of F1-score(Equation:~\eqref{eq:F1-score}) performance across different epochs on \textbf{Financial-phrasebank}~\ref{section:financial_phrasebank}.}
\label{fig:financial_phrasebank_epoch_trends}
\end{figure} 
Based on Table~\ref{table:LoRA-FT-Result},the following conclusions can be drawn:
\begin{enumerate}
\item In most cases,performance based on LoRA fine-tuning is better than full-parameter fine-tuning.
\item In small-sample scenarios,the performance of the base model can even surpass that of the large model in some cases.
\item As the size of the model increases,the performance of the model improves.
\item Fine-tuning for a small number of epochs can achieve good performance(Fig.~\ref{fig:ag_news_epoch},~\ref{fig:financial_phrasebank_epoch_trends}),which is more evident as the model size increases(Fig.~\ref{fig:financial_phrasebank_epoch_trends}).
\item Flan-T5 is a Seq2SeqLM,and as shown in Table~\ref{table:LoRA-FT-Result},generative models also perform exceptionally well in classification tasks. 
\end{enumerate}

\section{Challenges and Future Directions}
To date,there is no observed upper limit to the parameters of large models,and it is evident that an increase in the number of model parameters generally yields additional benefits. Currently,the pre-training of large models,considering computational power,technology,and economic costs,is feasible for only a few large companies and institutions,highlighting the significance of fine-tuning large models. Further fine-tuning experiments are reserved for future research,such as:
\begin{enumerate}
\item Additional ablation studies,for instance,adjusting hyperparameters like max-sequence-length,learning rate,batch size,and gradient clipping.
\item Experimentation with larger models.
\item Trials with different tasks,such as text classification,named entity recognition,and conversational question answering.
\item Exploration of various fine-tuning methods,including Prefix-Tuning,Prompt-Tuning,RHLF,and Agent-Tuning.
\item Comparison of the T5~\cite{T5} model and the Flan-T5\footnote{Flan-T5:\label{flan-t5}\url{https://huggingface.co/google/flan-t5-base}} model post-instruction fine-tuning across different tasks.
\item Analysis of the rate of loss reduction under different parameter settings.
\item Contrast of the performance differences between 4-bit,8-bit quantization,and 16-bit,32-bit precision.
\end{enumerate}

\section{Conclusion}
This paper encapsulates the advancements in fine-tuning LLMs,underscoring the diverse methodologies like task-specific adaptation,few-shot learning,and innovative techniques like LoRA~\cite{Hu2021LoRA}. It emphasizes the role of these models in enhancing language understanding and generation across different domains. The paper concludes by noting the promising directions for future research,particularly in refining fine-tuning techniques for greater efficiency and effectiveness in handling complex NLP tasks.


\printbibliography

@misc{RNN,
title	= {Recurrent Neural Network Regularization},
author	= {Wojciech Zaremba and Ilya Sutskever and Oriol Vinyals},
year	= {2014},
URL	= {https://arxiv.org/abs/1409.2329}
}

@ARTICLE{CNN,
  author={LeCun, Y. and Boser, B. and Denker, J. S. and Henderson, D. and Howard, R. E. and Hubbard, W. and Jackel, L. D.},
  journal={Neural Computation}, 
  title={Backpropagation Applied to Handwritten Zip Code Recognition}, 
  year={1989},
  volume={1},
  number={4},
  pages={541-551},
  doi={10.1162/neco.1989.1.4.541}}

@article{Peng2022BEiTVM,
  title={BEiT v2: Masked Image Modeling with Vector-Quantized Visual Tokenizers},
  author={Zhiliang Peng and Li Dong and Hangbo Bao and Qixiang Ye and Furu Wei},
  journal={ArXiv},
  year={2022},
  volume={abs/2208.06366},
  url={https://api.semanticscholar.org/CorpusID:251554649}
}

@ARTICLE{MOE,
  author={Jacobs, Robert A. and Jordan, Michael I. and Nowlan, Steven J. and Hinton, Geoffrey E.},
  journal={Neural Computation}, 
  title={Adaptive Mixtures of Local Experts}, 
  year={1991},
  volume={3},
  number={1},
  pages={79-87},
  doi={10.1162/neco.1991.3.1.79}}

@article{Watson2022,
  title = {Broadly Applicable and Accurate Protein Design by Integrating Structure Prediction Networks and Diffusion Generative Models},
  author = {Watson, Joseph L. and Juergens, David and Bennett, Nathaniel R. and Trippe, Brian L. and Yim, Jason and Eisenach, Helen E. and Ahern, Woody and Borst, Andrew J. and Ragotte, Robert J. and Milles, Lukas F. and Wicky, Basile I. M. and Hanikel, Nikita and Pellock, Samuel J. and Courbet, Alexis and Sheffler, William and Wang, Jue and Venkatesh, Preetham and Sappington, Isaac and Torres, Susana Vázquez and Lauko, Anna and Bortoli, Valentin De and Mathieu, Emile and Barzilay, Regina and Jaakkola, Tommi S. and DiMaio, Frank and Baek, Minkyung and Baker, David},
  date = {2022},
  journaltitle = {bioRxiv : the preprint server for biology},
  shortjournal = {bioRxiv},
  eprint = {https://www.biorxiv.org/content/early/2022/12/10/2022.12.09.519842.full.pdf},
  publisher = {{Cold Spring Harbor Laboratory}},
  doi = {10.1101/2022.12.09.519842},
  url = {https://www.biorxiv.org/content/early/2022/12/10/2022.12.09.519842},
  elocation-id = {2022.12.09.519842}
}

@article{Marino2019OKVQAAV,
  title={OK-VQA: A Visual Question Answering Benchmark Requiring External Knowledge},
  author={Kenneth Marino and Mohammad Rastegari and Ali Farhadi and Roozbeh Mottaghi},
  journal={2019 IEEE/CVF Conference on Computer Vision and Pattern Recognition (CVPR)},
  year={2019},
  pages={3190-3199},
  url={https://api.semanticscholar.org/CorpusID:173991173}
}

@article{Ho2020DenoisingDP,
  title={Denoising Diffusion Probabilistic Models},
  author={Jonathan Ho and Ajay Jain and P. Abbeel},
  journal={ArXiv},
  year={2020},
  volume={abs/2006.11239},
  url={https://api.semanticscholar.org/CorpusID:219955663}
}

@article{Zhang2023AddingCC,
  title={Adding Conditional Control to Text-to-Image Diffusion Models},
  author={Lvmin Zhang and Anyi Rao and Maneesh Agrawala},
  journal={ArXiv},
  year={2023},
  volume={abs/2302.05543},
  url={https://api.semanticscholar.org/CorpusID:256827727}
}

@inproceedings{dai-etal-2019-transformer,
    title = "Transformer-{XL}: Attentive Language Models beyond a Fixed-Length Context",
    author = "Dai, Zihang  and
      Yang, Zhilin  and
      Yang, Yiming  and
      Carbonell, Jaime  and
      Le, Quoc  and
      Salakhutdinov, Ruslan",
    editor = "Korhonen, Anna  and
      Traum, David  and
      M{\`a}rquez, Llu{\'\i}s",
    booktitle = "Proceedings of the 57th Annual Meeting of the Association for Computational Linguistics",
    month = jul,
    year = "2019",
    address = "Florence, Italy",
    publisher = "Association for Computational Linguistics",
    url = "https://aclanthology.org/P19-1285",
    doi = "10.18653/v1/P19-1285",
    pages = "2978--2988",
}

@article{Miao2022GalvatronET,
  title={Galvatron: Efficient Transformer Training over Multiple GPUs Using Automatic Parallelism},
  author={Xupeng Miao and Yujie Wang and Youhe Jiang and Chunan Shi and Xiaonan Nie and Hailin Zhang and Bin Cui},
  journal={Proc. VLDB Endow.},
  year={2022},
  volume={16},
  pages={470-479},
  url={https://api.semanticscholar.org/CorpusID:254017888}
}

@article{Gou2020KnowledgeDA,
  title={Knowledge Distillation: A Survey},
  author={Jianping Gou and B. Yu and Stephen J. Maybank and Dacheng Tao},
  journal={International Journal of Computer Vision},
  year={2020},
  volume={129},
  pages={1789 - 1819},
  url={https://api.semanticscholar.org/CorpusID:219559263}
}

@inproceedings{Vaswani2017AttentionIA,
  title={Attention is All you Need},
  author={Ashish Vaswani and Noam M. Shazeer and Niki Parmar and Jakob Uszkoreit and Llion Jones and Aidan N. Gomez and Lukasz Kaiser and Illia Polosukhin},
  booktitle={Neural Information Processing Systems},
  year={2017},
  url={https://api.semanticscholar.org/CorpusID:13756489}
}

@inproceedings{tranformer_understand,
  title={The Illustrated Transformer},
  author={JayAlammar},
  year={2020},
  url={https://jalammar.github.io/illustrated-transformer/}
}

@inproceedings{LINGO-1,
  title={LINGO-1: Exploring Natural Language for Autonomous Driving},
  author={wayve},
  year={2023},
  url={https://wayve.ai/thinking/lingo-natural-language-autonomous-driving/}
}

@inproceedings{RLHFChatGPT,
  title={Illustrating Reinforcement Learning from Human Feedback (RLHF)},
  author={Nathan Lambert and Louis Castricato and Leandro von Werra and Alex Havrilla},
  year={2022},
  url={https://huggingface.co/blog/rlhf}
}

@inproceedings{FintuingLLM,
  title={Finetuning Falcon LLMs More Efficiently With LoRA and Adapters},
  author={Sebastian Raschka},
  year={2023},
  url={https://sebastianraschka.com/blog/2023/falcon-finetuning.html}
}

@inproceedings{FalcoEcosystem,
  title={The Falcon has landed in the Hugging Face ecosystem},
  author={Leandro von Werra and Younes Belkad and Sourab Mangrulkar and Lewis Tunstall and Olivier Dehaene and Pedro Cuenca and Philipp Schmid and Omar Sanseviero},
  year={2023},
  url={https://huggingface.co/blog/falcon}
}

@inproceedings{Radford2021LearningTVCLIP,
  title={Learning Transferable Visual Models From Natural Language Supervision},
  author={Alec Radford and Jong Wook Kim and Chris Hallacy and Aditya Ramesh and Gabriel Goh and Sandhini Agarwal and Girish Sastry and Amanda Askell and Pamela Mishkin and Jack Clark and Gretchen Krueger and Ilya Sutskever},
  booktitle={International Conference on Machine Learning},
  year={2021},
  url={https://api.semanticscholar.org/CorpusID:231591445}
}

@online{zhaoSurveyLargeLanguage2023,
  title = {A {{Survey}} of {{Large Language Models}}},
  author = {Zhao, Wayne Xin and Zhou, Kun and Li, Junyi and Tang, Tianyi and Wang, Xiaolei and Hou, Yupeng and Min, Yingqian and Zhang, Beichen and Zhang, Junjie and Dong, Zican and Du, Yifan and Yang, Chen and Chen, Yushuo and Chen, Zhipeng and Jiang, Jinhao and Ren, Ruiyang and Li, Yifan and Tang, Xinyu and Liu, Zikang and Liu, Peiyu and Nie, Jian-Yun and Wen, Ji-Rong},
  date = {2023-09-11},
  eprint = {2303.18223},
  eprinttype = {arxiv},
  eprintclass = {cs},
  url = {http://arxiv.org/abs/2303.18223},
  urldate = {2023-11-19},
  langid = {english},
  pubstate = {preprint},
}

@inproceedings{Devlin2019BERTPO,
  title={BERT: Pre-training of Deep Bidirectional Transformers for Language Understanding},
  author={Jacob Devlin and Ming-Wei Chang and Kenton Lee and Kristina Toutanova},
  booktitle={North American Chapter of the Association for Computational Linguistics},
  year={2019},
  url={https://api.semanticscholar.org/CorpusID:52967399}
}

@inproceedings{Radford2018ImprovingLU,
  title={Improving Language Understanding by Generative Pre-Training},
  author={Alec Radford and Karthik Narasimhan},
  year={2018},
  url={https://api.semanticscholar.org/CorpusID:49313245}
}

@inproceedings{Radford2019LanguageMAgpt2,
  title={Language Models are Unsupervised Multitask Learners},
  author={Alec Radford and Jeff Wu and Rewon Child and David Luan and Dario Amodei and Ilya Sutskever},
  year={2019},
  url={https://api.semanticscholar.org/CorpusID:160025533}
}

@inproceedings{gpt3,
author = {Brown, Tom B. and Mann, Benjamin and Ryder, Nick and Subbiah, Melanie and Kaplan, Jared and Dhariwal, Prafulla and Neelakantan, Arvind and Shyam, Pranav and Sastry, Girish and Askell, Amanda and Agarwal, Sandhini and Herbert-Voss, Ariel and Krueger, Gretchen and Henighan, Tom and Child, Rewon and Ramesh, Aditya and Ziegler, Daniel M. and Wu, Jeffrey and Winter, Clemens and Hesse, Christopher and Chen, Mark and Sigler, Eric and Litwin, Mateusz and Gray, Scott and Chess, Benjamin and Clark, Jack and Berner, Christopher and McCandlish, Sam and Radford, Alec and Sutskever, Ilya and Amodei, Dario},
title = {Language Models Are Few-Shot Learners},
year = {2020},
isbn = {9781713829546},
publisher = {Curran Associates Inc.},
address = {Red Hook, NY, USA},
booktitle = {Proceedings of the 34th International Conference on Neural Information Processing Systems},
articleno = {159},
numpages = {25},
location = {Vancouver, BC, Canada},
series = {NIPS'20}
}

@article{OpenAI2023GPT4TR,
  title={GPT-4 Technical Report},
  author={OpenAI},
  journal={ArXiv},
  year={2023},
  volume={abs/2303.08774},
  url={https://api.semanticscholar.org/CorpusID:257532815}
}

@article{Ouyang2022InstructGPT,
  title={Training language models to follow instructions with human feedback},
  author={Long Ouyang and Jeff Wu and Xu Jiang and Diogo Almeida and Carroll L. Wainwright and Pamela Mishkin and Chong Zhang and Sandhini Agarwal and Katarina Slama and Alex Ray and John Schulman and Jacob Hilton and Fraser Kelton and Luke E. Miller and Maddie Simens and Amanda Askell and Peter Welinder and Paul Francis Christiano and Jan Leike and Ryan J. Lowe},
  journal={ArXiv},
  year={2022},
  volume={abs/2203.02155},
  url={https://api.semanticscholar.org/CorpusID:246426909}
}

@inproceedings{Lewis2019BARTDS,
  title={BART: Denoising Sequence-to-Sequence Pre-training for Natural Language Generation, Translation, and Comprehension},
  author={Mike Lewis and Yinhan Liu and Naman Goyal and Marjan Ghazvininejad and Abdel-rahman Mohamed and Omer Levy and Veselin Stoyanov and Luke Zettlemoyer},
  booktitle={Annual Meeting of the Association for Computational Linguistics},
  year={2019},
  url={https://api.semanticscholar.org/CorpusID:204960716}
}

@article{Touvron2023LLaMA,
  title={LLaMA: Open and Efficient Foundation Language Models},
  author={Hugo Touvron and Thibaut Lavril and Gautier Izacard and Xavier Martinet and Marie-Anne Lachaux and Timoth{\'e}e Lacroix and Baptiste Rozi{\`e}re and Naman Goyal and Eric Hambro and Faisal Azhar and Aurelien Rodriguez and Armand Joulin and Edouard Grave and Guillaume Lample},
  journal={ArXiv},
  year={2023},
  volume={abs/2302.13971},
  url={https://api.semanticscholar.org/CorpusID:257219404}
}

@article{Touvron2023Llama2,
  title={Llama 2: Open Foundation and Fine-Tuned Chat Models},
  author={Hugo Touvron and Louis Martin and Kevin R. Stone and Peter Albert and Amjad Almahairi and Yasmine Babaei and Nikolay Bashlykov and Soumya Batra and Prajjwal Bhargava and Shruti Bhosale and Daniel M. Bikel and Lukas Blecher and Cristian Cant{\'o}n Ferrer and Moya Chen and Guillem Cucurull and David Esiobu and Jude Fernandes and Jeremy Fu and Wenyin Fu and Brian Fuller and Cynthia Gao and Vedanuj Goswami and Naman Goyal and Anthony S. Hartshorn and Saghar Hosseini and Rui Hou and Hakan Inan and Marcin Kardas and Viktor Kerkez and Madian Khabsa and Isabel M. Kloumann and A. V. Korenev and Punit Singh Koura and Marie-Anne Lachaux and Thibaut Lavril and Jenya Lee and Diana Liskovich and Yinghai Lu and Yuning Mao and Xavier Martinet and Todor Mihaylov and Pushkar Mishra and Igor Molybog and Yixin Nie and Andrew Poulton and Jeremy Reizenstein and Rashi Rungta and Kalyan Saladi and Alan Schelten and Ruan Silva and Eric Michael Smith and R. Subramanian and Xia Tan and Binh Tang and Ross Taylor and Adina Williams and Jian Xiang Kuan and Puxin Xu and Zhengxu Yan and Iliyan Zarov and Yuchen Zhang and Angela Fan and Melanie Kambadur and Sharan Narang and Aurelien Rodriguez and Robert Stojnic and Sergey Edunov and Thomas Scialom},
  journal={ArXiv},
  year={2023},
  volume={abs/2307.09288},
  url={https://api.semanticscholar.org/CorpusID:259950998}
}

@inproceedings{Dong2019UniLM,
  title={Unified Language Model Pre-training for Natural Language Understanding and Generation},
  author={Li Dong and Nan Yang and Wenhui Wang and Furu Wei and Xiaodong Liu and Yu Wang and Jianfeng Gao and M. Zhou and Hsiao-Wuen Hon},
  booktitle={Neural Information Processing Systems},
  year={2019},
  url={https://api.semanticscholar.org/CorpusID:147704286}
}

@inproceedings{Bao2020UniLMv2PL,
  title={UniLMv2: Pseudo-Masked Language Models for Unified Language Model Pre-Training},
  author={Hangbo Bao and Li Dong and Furu Wei and Wenhui Wang and Nan Yang and Xiaodong Liu and Yu Wang and Songhao Piao and Jianfeng Gao and Ming Zhou and Hsiao-Wuen Hon},
  booktitle={International Conference on Machine Learning},
  year={2020},
  url={https://api.semanticscholar.org/CorpusID:211572655}
}

@inproceedings{Ding2021CogViewMT,
  title={CogView: Mastering Text-to-Image Generation via Transformers},
  author={Ming Ding and Zhuoyi Yang and Wenyi Hong and Wendi Zheng and Chang Zhou and Da Yin and Junyang Lin and Xu Zou and Zhou Shao and Hongxia Yang and Jie Tang},
  booktitle={Neural Information Processing Systems},
  year={2021},
  url={https://api.semanticscholar.org/CorpusID:235212350}
}

@article{Ding2022CogView2FA,
  title={CogView2: Faster and Better Text-to-Image Generation via Hierarchical Transformers},
  author={Ming Ding and Wendi Zheng and Wenyi Hong and Jie Tang},
  journal={ArXiv},
  year={2022},
  volume={abs/2204.14217},
  url={https://api.semanticscholar.org/CorpusID:248476190}
}

@article{Bao2021BEiTBP,
  title={BEiT: BERT Pre-Training of Image Transformers},
  author={Hangbo Bao and Li Dong and Furu Wei},
  journal={ArXiv},
  year={2021},
  volume={abs/2106.08254},
  url={https://api.semanticscholar.org/CorpusID:235436185}
}

@article{Chen2021-CodeX,
  title={Evaluating Large Language Models Trained on Code},
  author={Mark Chen and Jerry Tworek and Heewoo Jun and Qiming Yuan and Henrique Ponde and Jared Kaplan and Harrison Edwards and Yura Burda and Nicholas Joseph and Greg Brockman and Alex Ray and Raul Puri and Gretchen Krueger and Michael Petrov and Heidy Khlaaf and Girish Sastry and Pamela Mishkin and Brooke Chan and Scott Gray and Nick Ryder and Mikhail Pavlov and Alethea Power and Lukasz Kaiser and Mohammad Bavarian and Clemens Winter and Philippe Tillet and Felipe Petroski Such and David W. Cummings and Matthias Plappert and Fotios Chantzis and Elizabeth Barnes and Ariel Herbert-Voss and William H. Guss and Alex Nichol and Igor Babuschkin and S. Arun Balaji and Shantanu Jain and Andrew Carr and Jan Leike and Joshua Achiam and Vedant Misra and Evan Morikawa and Alec Radford and Matthew M. Knight and Miles Brundage and Mira Murati and Katie Mayer and Peter Welinder and Bob McGrew and Dario Amodei and Sam McCandlish and Ilya Sutskever and Wojciech Zaremba},
  journal={ArXiv},
  year={2021},
  volume={abs/2107.03374},
  url={https://api.semanticscholar.org/CorpusID:235755472}
}

@article{Nakano2021WebGPTBQ,
  title={WebGPT: Browser-assisted question-answering with human feedback},
  author={Reiichiro Nakano and Jacob Hilton and S. Arun Balaji and Jeff Wu and Ouyang Long and Christina Kim and Christopher Hesse and Shantanu Jain and Vineet Kosaraju and William Saunders and Xu Jiang and Karl Cobbe and Tyna Eloundou and Gretchen Krueger and Kevin Button and Matthew Knight and Benjamin Chess and John Schulman},
  journal={ArXiv},
  year={2021},
  volume={abs/2112.09332},
  url={https://api.semanticscholar.org/CorpusID:245329531}
}

@article{Du2021GLaMES,
  title={GLaM: Efficient Scaling of Language Models with Mixture-of-Experts},
  author={Nan Du and Yanping Huang and Andrew M. Dai and Simon Tong and Dmitry Lepikhin and Yuanzhong Xu and Maxim Krikun and Yanqi Zhou and Adams Wei Yu and Orhan Firat and Barret Zoph and Liam Fedus and Maarten Bosma and Zongwei Zhou and Tao Wang and Yu Emma Wang and Kellie Webster and Marie Pellat and Kevin Robinson and Kathleen S. Meier-Hellstern and Toju Duke and Lucas Dixon and Kun Zhang and Quoc V. Le and Yonghui Wu and Z. Chen and Claire Cui},
  journal={ArXiv},
  year={2021},
  volume={abs/2112.06905},
  url={https://api.semanticscholar.org/CorpusID:245124124}
}

@article{Scao2022BLOOMA1,
  title={BLOOM: A 176B-Parameter Open-Access Multilingual Language Model},
  author={Teven Le Scao and Angela Fan and Christopher Akiki and Ellie Pavlick and Suzana Ili'c and Daniel Hesslow and Roman Castagn'e and Alexandra Sasha Luccioni and Franccois Yvon and Matthias Gall{\'e} and Jonathan Tow and Alexander M. Rush and Stella Rose Biderman and Albert Webson and Pawan Sasanka Ammanamanchi and Thomas Wang and Beno{\^i}t Sagot and Niklas Muennighoff and Albert Villanova del Moral and Olatunji Ruwase and Rachel Bawden and Stas Bekman and Angelina McMillan-Major and Iz Beltagy and Huu Nguyen and Lucile Saulnier and Samson Tan and Pedro Ortiz Suarez and Victor Sanh and Hugo Laurenccon and Yacine Jernite and Julien Launay and Margaret Mitchell and Colin Raffel and Aaron Gokaslan and Adi Simhi and Aitor Soroa Etxabe and Alham Fikri Aji and Amit Alfassy and Anna Rogers and Ariel Kreisberg Nitzav and Canwen Xu and Chenghao Mou and Chris C. Emezue and Christopher Klamm and Colin Leong and Daniel Alexander van Strien and David Ifeoluwa Adelani and Dragomir R. Radev and Eduardo Gonz'alez Ponferrada and Efrat Levkovizh and Ethan Kim and Eyal Natan and Francesco De Toni and G{\'e}rard Dupont and Germ{\'a}n Kruszewski and Giada Pistilli and Hady ElSahar and Hamza Benyamina and Hieu Trung Tran and Ian Yu and Idris Abdulmumin and Isaac Johnson and Itziar Gonzalez-Dios and Javier de la Rosa and Jenny Chim and Jesse Dodge and Jian Zhu and Jonathan Chang and Jorg Frohberg and Josephine L. Tobing and Joydeep Bhattacharjee and Khalid Almubarak and Kimbo Chen and Kyle Lo and Leandro von Werra and Leon Weber and Long Phan and Loubna Ben Allal and Ludovic Tanguy and Manan Dey and Manuel Romero Mu{\~n}oz and Maraim Masoud and Mar{\'i}a Grandury and Mario vSavsko and Max Huang and Maximin Coavoux and Mayank Singh and Mike Tian-Jian Jiang and Minh Chien Vu and Mohammad Ali Jauhar and Mustafa Ghaleb and Nishant Subramani and Nora Kassner and Nurulaqilla Khamis and Olivier Nguyen and Omar Espejel and Ona de Gibert and Paulo Villegas and Peter Henderson and Pierre Colombo and Priscilla Amuok and Quentin Lhoest and Rheza Harliman and Rishi Bommasani and Roberto L'opez and Rui Ribeiro and Salomey Osei and Sampo Pyysalo and Sebastian Nagel and Shamik Bose and Shamsuddeen Hassan Muhammad and Shanya Sharma Sharma and S. Longpre and So-maieh Nikpoor and S. Silberberg and Suhas Pai and Sydney Zink and Tiago Timponi Torrent and Timo Schick and Tristan Thrush and Valentin Danchev and Vassilina Nikoulina and Veronika Laippala and Violette Lepercq and Vrinda Prabhu and Zaid Alyafeai and Zeerak Talat and Arun Raja and Benjamin Heinzerling and Chenglei Si and Elizabeth Salesky and Sabrina J. Mielke and Wilson Y. Lee and Abheesht Sharma and Andrea Santilli and Antoine Chaffin and Arnaud Stiegler and Debajyoti Datta and Eliza Szczechla and Gunjan Chhablani and Han Wang and Harshit Pandey and Hendrik Strobelt and Jason Alan Fries and Jos Rozen and Leo Gao and Lintang Sutawika and M Saiful Bari and Maged S. Al-Shaibani and Matteo Manica and Nihal V. Nayak and Ryan Teehan and Samuel Albanie and Sheng Shen and Srulik Ben-David and Stephen H. Bach and Taewoon Kim and Tali Bers and Thibault F{\'e}vry and Trishala Neeraj and Urmish Thakker and Vikas Raunak and Xiang Tang and Zheng-Xin Yong and Zhiqing Sun and Shaked Brody and Y Uri and Hadar Tojarieh and Adam Roberts and Hyung Won Chung and Jaesung Tae and Jason Phang and Ofir Press and Conglong Li and Deepak Narayanan and Hatim Bourfoune and Jared Casper and Jeff Rasley and Max Ryabinin and Mayank Mishra and Minjia Zhang and Mohammad Shoeybi and Myriam Peyrounette and Nicolas Patry and Nouamane Tazi and Omar Sanseviero and Patrick von Platen and Pierre Cornette and Pierre Franccois Lavall'ee and R{\'e}mi Lacroix and Samyam Rajbhandari and Sanchit Gandhi and Shaden Smith and St{\'e}phane Requena and Suraj Patil and Tim Dettmers and Ahmed Baruwa and Amanpreet Singh and Anastasia Cheveleva and Anne-Laure Ligozat and Arjun Subramonian and Aur'elie N'ev'eol and Charles Lovering and Daniel H Garrette and Deepak R. Tunuguntla and Ehud Reiter and Ekaterina Taktasheva and Ekaterina Voloshina and Eli Bogdanov and Genta Indra Winata and Hailey Schoelkopf and Jan-Christoph Kalo and Jekaterina Novikova and Jessica Zosa Forde and Xiangru Tang and Jungo Kasai and Ken Kawamura and Liam Hazan and Marine Carpuat and Miruna Clinciu and Najoung Kim and Newton Cheng and Oleg Serikov and Omer Antverg and Oskar van der Wal and Rui Zhang and Ruochen Zhang and Sebastian Gehrmann and Shachar Mirkin and S. Osher Pais and Tatiana Shavrina and Thomas Scialom and Tian Yun and Tomasz Limisiewicz and Verena Rieser and Vitaly Protasov and Vladislav Mikhailov and Yada Pruksachatkun and Yonatan Belinkov and Zachary Bamberger and Zdenvek Kasner and Zdeněk Kasner and Amanda Pestana and Amir Feizpour and Ammar Khan and Amy Faranak and Ananda Santa Rosa Santos and Anthony Hevia and Antigona Unldreaj and Arash Aghagol and Arezoo Abdollahi and Aycha Tammour and Azadeh HajiHosseini and Bahareh Behroozi and Benjamin Ayoade Ajibade and Bharat Kumar Saxena and Carlos Mu{\~n}oz Ferrandis and Danish Contractor and David M. Lansky and Davis David and Douwe Kiela and Duong Anh Nguyen and Edward Tan and Emi Baylor and Ezinwanne Ozoani and Fatim Tahirah Mirza and Frankline Ononiwu and Habib Rezanejad and H.A. Jones and Indrani Bhattacharya and Irene Solaiman and Irina Sedenko and Isar Nejadgholi and Jan Passmore and Joshua Seltzer and Julio Bonis Sanz and Karen Fort and L{\'i}via Dutra and Mairon Samagaio and Maraim Elbadri and Margot Mieskes and Marissa Gerchick and Martha Akinlolu and Michael McKenna and Mike Qiu and Muhammed Ghauri and Mykola Burynok and Nafis Abrar and Nazneen Rajani and Nour Elkott and Nourhan Fahmy and Olanrewaju Samuel and Ran An and R. P. Kromann and Ryan Hao and Samira Alizadeh and Sarmad Shubber and Silas L. Wang and Sourav Roy and Sylvain Viguier and Thanh-Cong Le and Tobi Oyebade and Trieu Nguyen Hai Le and Yoyo Yang and Zachary Kyle Nguyen and Abhinav Ramesh Kashyap and Alfredo Palasciano and Alison Callahan and Anima Shukla and Antonio Miranda-Escalada and Ayush Kumar Singh and Benjamin Beilharz and Bo Wang and Caio Matheus Fonseca de Brito and Chenxi Zhou and Chirag Jain and Chuxin Xu and Cl{\'e}mentine Fourrier and Daniel Le'on Perin'an and Daniel Molano and Dian Yu and Enrique Manjavacas and Fabio Barth and Florian Fuhrimann and Gabriel Altay and Giyaseddin Bayrak and Gully Burns and Helena U. Vrabec and Iman I.B. Bello and Isha Dash and Ji Soo Kang and John Giorgi and Jonas Golde and Jose David Posada and Karthi Sivaraman and Lokesh Bulchandani and Lu Liu and Luisa Shinzato and Madeleine Hahn de Bykhovetz and Maiko Takeuchi and Marc P{\`a}mies and Mar{\'i}a Andrea Castillo and Marianna Nezhurina and Mario Sanger and Matthias Samwald and Michael Cullan and Michael Weinberg and M Wolf and Mina Mihaljcic and Minna Liu and Moritz Freidank and Myungsun Kang and Natasha Seelam and Nathan Dahlberg and Nicholas Michio Broad and Nikolaus Muellner and Pascale Fung and Patricia Haller and R. Chandrasekhar and Renata Eisenberg and Robert Martin and Rodrigo L. Canalli and Rosaline Su and Ruisi Su and Samuel Cahyawijaya and Samuele Garda and Shlok S Deshmukh and Shubhanshu Mishra and Sid Kiblawi and Simon Ott and Sinee Sang-aroonsiri and Srishti Kumar and Stefan Schweter and Sushil Pratap Bharati and Tanmay Laud and Th'eo Gigant and Tomoya Kainuma and Wojciech Kusa and Yanis Labrak and Yashasvi Bajaj and Y. Venkatraman and Yifan Xu and Ying Xu and Yu Xu and Zhee Xao Tan and Zhongli Xie and Zifan Ye and Mathilde Bras and Younes Belkada and Thomas Wolf},
  journal={ArXiv},
  year={2022},
  volume={abs/2211.05100},
  url={https://api.semanticscholar.org/CorpusID:253420279}
}

@inproceedings{Du2021GLMGL,
  title={GLM: General Language Model Pretraining with Autoregressive Blank Infilling},
  author={Zhengxiao Du and Yujie Qian and Xiao Liu and Ming Ding and Jiezhong Qiu and Zhilin Yang and Jie Tang},
  booktitle={Annual Meeting of the Association for Computational Linguistics},
  year={2021},
  url={https://api.semanticscholar.org/CorpusID:247519241}
}

@article{Black2022GPTNeoX20BAO,
  title={GPT-NeoX-20B: An Open-Source Autoregressive Language Model},
  author={Sid Black and Stella Rose Biderman and Eric Hallahan and Quentin G. Anthony and Leo Gao and Laurence Golding and Horace He and Connor Leahy and Kyle McDonell and Jason Phang and Michael Martin Pieler and USVSN Sai Prashanth and Shivanshu Purohit and Laria Reynolds and Jonathan Tow and Benqi Wang and Samuel Weinbach},
  journal={ArXiv},
  year={2022},
  volume={abs/2204.06745},
  url={https://api.semanticscholar.org/CorpusID:248177957}
}

@article{Zhang2022OPTOP,
  title={OPT: Open Pre-trained Transformer Language Models},
  author={Susan Zhang and Stephen Roller and Naman Goyal and Mikel Artetxe and Moya Chen and Shuohui Chen and Christopher Dewan and Mona T. Diab and Xian Li and Xi Victoria Lin and Todor Mihaylov and Myle Ott and Sam Shleifer and Kurt Shuster and Daniel Simig and Punit Singh Koura and Anjali Sridhar and Tianlu Wang and Luke Zettlemoyer},
  journal={ArXiv},
  year={2022},
  volume={abs/2205.01068},
  url={https://api.semanticscholar.org/CorpusID:248496292}
}

@inproceedings{Xue2020mT5AM,
  title={mT5: A Massively Multilingual Pre-trained Text-to-Text Transformer},
  author={Linting Xue and Noah Constant and Adam Roberts and Mihir Kale and Rami Al-Rfou and Aditya Siddhant and Aditya Barua and Colin Raffel},
  booktitle={North American Chapter of the Association for Computational Linguistics},
  year={2020},
  url={https://api.semanticscholar.org/CorpusID:225040574}
}

@article{Lepikhin2020GShardSG,
  title={GShard: Scaling Giant Models with Conditional Computation and Automatic Sharding},
  author={Dmitry Lepikhin and HyoukJoong Lee and Yuanzhong Xu and Dehao Chen and Orhan Firat and Yanping Huang and Maxim Krikun and Noam M. Shazeer and Z. Chen},
  journal={ArXiv},
  year={2020},
  volume={abs/2006.16668},
  url={https://api.semanticscholar.org/CorpusID:220265858}
}

@article{Thoppilan2022LaMDALM,
  title={LaMDA: Language Models for Dialog Applications},
  author={Romal Thoppilan and Daniel De Freitas and Jamie Hall and Noam M. Shazeer and Apoorv Kulshreshtha and Heng-Tze Cheng and Alicia Jin and Taylor Bos and Leslie Baker and Yu Du and Yaguang Li and Hongrae Lee and Huaixiu Steven Zheng and Amin Ghafouri and Marcelo Menegali and Yanping Huang and Maxim Krikun and Dmitry Lepikhin and James Qin and Dehao Chen and Yuanzhong Xu and Zhifeng Chen and Adam Roberts and Maarten Bosma and Yanqi Zhou and Chung-Ching Chang and I. A. Krivokon and Willard James Rusch and Marc Pickett and Kathleen S. Meier-Hellstern and Meredith Ringel Morris and Tulsee Doshi and Renelito Delos Santos and Toju Duke and Johnny Hartz S{\o}raker and Ben Zevenbergen and Vinodkumar Prabhakaran and Mark D{\'i}az and Ben Hutchinson and Kristen Olson and Alejandra Molina and Erin Hoffman-John and Josh Lee and Lora Aroyo and Ravi Rajakumar and Alena Butryna and Matthew Lamm and V. O. Kuzmina and Joseph Fenton and Aaron Cohen and Rachel Bernstein and Ray Kurzweil and Blaise Aguera-Arcas and Claire Cui and Marian Rogers Croak and Ed Huai-hsin Chi and Quoc Le},
  journal={ArXiv},
  year={2022},
  volume={abs/2201.08239},
  url={https://api.semanticscholar.org/CorpusID:246063428}
}

@misc{wu2021yuan,
      title={Yuan 1.0: Large-Scale Pre-trained Language Model in Zero-Shot and Few-Shot Learning}, 
      author={Shaohua Wu and Xudong Zhao and Tong Yu and Rongguo Zhang and Chong Shen and Hongli Liu and Feng Li and Hong Zhu and Jiangang Luo and Liang Xu and Xuanwei Zhang},
      year={2021},
      eprint={2110.04725},
      archivePrefix={arXiv},
      primaryClass={cs.CL}
}

@article{Glaese2022Sparrow,
  title={Improving alignment of dialogue agents via targeted human judgements},
  author={Amelia Glaese and Nathan McAleese and Maja Trkebacz and John Aslanides and Vlad Firoiu and Timo Ewalds and Maribeth Rauh and Laura Weidinger and Martin Chadwick and Phoebe Thacker and Lucy Campbell-Gillingham and Jonathan Uesato and Po-Sen Huang and Ramona Comanescu and Fan Yang and A. See and Sumanth Dathathri and Rory Greig and Charlie Chen and Doug Fritz and Jaume Sanchez Elias and Richard Green and Sovna Mokr'a and Nicholas Fernando and Boxi Wu and Rachel Foley and Susannah Young and Iason Gabriel and William S. Isaac and John F. J. Mellor and Demis Hassabis and Koray Kavukcuoglu and Lisa Anne Hendricks and Geoffrey Irving},
  journal={ArXiv},
  year={2022},
  volume={abs/2209.14375},
  url={https://api.semanticscholar.org/CorpusID:252596089}
}

@article{Koishekenov2022MemoryefficientNLLB,
  title={Memory-efficient NLLB-200: Language-specific Expert Pruning of a Massively Multilingual Machine Translation Model},
  author={Yeskendir Koishekenov and Vassilina Nikoulina and Alexandre B{\'e}rard},
  journal={ArXiv},
  year={2022},
  volume={abs/2212.09811},
  url={https://api.semanticscholar.org/CorpusID:254877227}
}

@article{Zhang2021CPM2LC,
  title={CPM-2: Large-scale Cost-effective Pre-trained Language Models},
  author={Zhengyan Zhang and Yuxian Gu and Xu Han and Shengqi Chen and Chaojun Xiao and Zhenbo Sun and Yuan Yao and Fanchao Qi and Jian Guan and Pei Ke and Yanzheng Cai and Guoyang Zeng and Zhixing Tan and Zhiyuan Liu and Minlie Huang and Wentao Han and Yang Liu and Xiaoyan Zhu and Maosong Sun},
  journal={AI Open},
  year={2021},
  volume={2},
  pages={216-224},
  url={https://api.semanticscholar.org/CorpusID:235490263}
}

@article{Sun2021ERNIE3L,
  title={ERNIE 3.0: Large-scale Knowledge Enhanced Pre-training for Language Understanding and Generation},
  author={Yu Sun and Shuohuan Wang and Shikun Feng and Siyu Ding and Chao Pang and Junyuan Shang and Jiaxiang Liu and Xuyi Chen and Yanbin Zhao and Yuxiang Lu and Weixin Liu and Zhihua Wu and Weibao Gong and Jianzhong Liang and Zhizhou Shang and Peng Sun and Wei Liu and Ouyang Xuan and Dianhai Yu and Hao Tian and Hua Wu and Haifeng Wang},
  journal={ArXiv},
  year={2021},
  volume={abs/2107.02137},
  url={https://api.semanticscholar.org/CorpusID:235731579}
}

@article{Chung2022ScalingILFLAN,
  title={Scaling Instruction-Finetuned Language Models},
  author={Hyung Won Chung and Le Hou and S. Longpre and Barret Zoph and Yi Tay and William Fedus and Eric Li and Xuezhi Wang and Mostafa Dehghani and Siddhartha Brahma and Albert Webson and Shixiang Shane Gu and Zhuyun Dai and Mirac Suzgun and Xinyun Chen and Aakanksha Chowdhery and Dasha Valter and Sharan Narang and Gaurav Mishra and Adams Wei Yu and Vincent Zhao and Yanping Huang and Andrew M. Dai and Hongkun Yu and Slav Petrov and Ed Huai-hsin Chi and Jeff Dean and Jacob Devlin and Adam Roberts and Denny Zhou and Quoc V. Le and Jason Wei},
  journal={ArXiv},
  year={2022},
  volume={abs/2210.11416},
  url={https://api.semanticscholar.org/CorpusID:253018554}
}

@article{Dosovitskiy2020VIT,
  title={An Image is Worth 16x16 Words: Transformers for Image Recognition at Scale},
  author={Alexey Dosovitskiy and Lucas Beyer and Alexander Kolesnikov and Dirk Weissenborn and Xiaohua Zhai and Thomas Unterthiner and Mostafa Dehghani and Matthias Minderer and Georg Heigold and Sylvain Gelly and Jakob Uszkoreit and Neil Houlsby},
  journal={ArXiv},
  year={2020},
  volume={abs/2010.11929},
  url={https://api.semanticscholar.org/CorpusID:225039882}
}

@article{Nijkamp2023CodeGen2LF,
  title={CodeGen2: Lessons for Training LLMs on Programming and Natural Languages},
  author={Erik Nijkamp and Hiroaki Hayashi and Caiming Xiong and Silvio Savarese and Yingbo Zhou},
  journal={ArXiv},
  year={2023},
  volume={abs/2305.02309},
  url={https://api.semanticscholar.org/CorpusID:258461229}
}

@article{Penedo2023TheRDFalcon,
  title={The RefinedWeb Dataset for Falcon LLM: Outperforming Curated Corpora with Web Data, and Web Data Only},
  author={Guilherme Penedo and Quentin Malartic and Daniel Hesslow and Ruxandra-Aim{\'e}e Cojocaru and Alessandro Cappelli and Hamza Alobeidli and Baptiste Pannier and Ebtesam Almazrouei and Julien Launay},
  journal={ArXiv},
  year={2023},
  volume={abs/2306.01116},
  url={https://api.semanticscholar.org/CorpusID:259063761}
}

@article{Shazeer2019FastMultiQueryAttention,
  title={Fast Transformer Decoding: One Write-Head is All You Need},
  author={Noam M. Shazeer},
  journal={ArXiv},
  year={2019},
  volume={abs/1911.02150},
  url={https://api.semanticscholar.org/CorpusID:207880429}
}

@article{Ramesh2022HierarchicalTIDALLE2,
  title={Hierarchical Text-Conditional Image Generation with CLIP Latents},
  author={Aditya Ramesh and Prafulla Dhariwal and Alex Nichol and Casey Chu and Mark Chen},
  journal={ArXiv},
  year={2022},
  volume={abs/2204.06125},
  url={https://api.semanticscholar.org/CorpusID:248097655}
}

@article{Rombach2021HighResolutionISStableDiffusion,
  title={High-Resolution Image Synthesis with Latent Diffusion Models},
  author={Robin Rombach and A. Blattmann and Dominik Lorenz and Patrick Esser and Bj{\"o}rn Ommer},
  journal={2022 IEEE/CVF Conference on Computer Vision and Pattern Recognition (CVPR)},
  year={2021},
  pages={10674-10685},
  url={https://api.semanticscholar.org/CorpusID:245335280}
}

@article{Chowdhery2022PaLMSL,
  title={PaLM: Scaling Language Modeling with Pathways},
  author={Aakanksha Chowdhery and Sharan Narang and Jacob Devlin and Maarten Bosma and Gaurav Mishra and Adam Roberts and Paul Barham and Hyung Won Chung and Charles Sutton and Sebastian Gehrmann and Parker Schuh and Kensen Shi and Sasha Tsvyashchenko and Joshua Maynez and Abhishek Rao and Parker Barnes and Yi Tay and Noam M. Shazeer and Vinodkumar Prabhakaran and Emily Reif and Nan Du and Benton C. Hutchinson and Reiner Pope and James Bradbury and Jacob Austin and Michael Isard and Guy Gur-Ari and Pengcheng Yin and Toju Duke and Anselm Levskaya and Sanjay Ghemawat and Sunipa Dev and Henryk Michalewski and Xavier Garc{\'i}a and Vedant Misra and Kevin Robinson and Liam Fedus and Denny Zhou and Daphne Ippolito and David Luan and Hyeontaek Lim and Barret Zoph and Alexander Spiridonov and Ryan Sepassi and David Dohan and Shivani Agrawal and Mark Omernick and Andrew M. Dai and Thanumalayan Sankaranarayana Pillai and Marie Pellat and Aitor Lewkowycz and Erica Moreira and Rewon Child and Oleksandr Polozov and Katherine Lee and Zongwei Zhou and Xuezhi Wang and Brennan Saeta and Mark D{\'i}az and Orhan Firat and Michele Catasta and Jason Wei and Kathleen S. Meier-Hellstern and Douglas Eck and Jeff Dean and Slav Petrov and Noah Fiedel},
  journal={J. Mach. Learn. Res.},
  year={2022},
  volume={24},
  pages={240:1-240:113},
  url={https://api.semanticscholar.org/CorpusID:247951931}
}

@article{Anil2023PaLM2T,
  title={PaLM 2 Technical Report},
  author={Rohan Anil and Andrew M. Dai and Orhan Firat and Melvin Johnson and Dmitry Lepikhin and Alexandre Tachard Passos and Siamak Shakeri and Emanuel Taropa and Paige Bailey and Z. Chen and Eric Chu and J. Clark and Laurent El Shafey and Yanping Huang and Kathleen S. Meier-Hellstern and Gaurav Mishra and Erica Moreira and Mark Omernick and Kevin Robinson and Sebastian Ruder and Yi Tay and Kefan Xiao and Yuanzhong Xu and Yujing Zhang and Gustavo Hernandez Abrego and Junwhan Ahn and Jacob Austin and Paul Barham and Jan A. Botha and James Bradbury and Siddhartha Brahma and Kevin Michael Brooks and Michele Catasta and Yongzhou Cheng and Colin Cherry and Christopher A. Choquette-Choo and Aakanksha Chowdhery and C Cr{\'e}py and Shachi Dave and Mostafa Dehghani and Sunipa Dev and Jacob Devlin and M. C. D'iaz and Nan Du and Ethan Dyer and Vladimir Feinberg and Fan Feng and Vlad Fienber and Markus Freitag and Xavier Garc{\'i}a and Sebastian Gehrmann and Lucas Gonz{\'a}lez and Guy Gur-Ari and Steven Hand and Hadi Hashemi and Le Hou and Joshua Howland and An Ren Hu and Jeffrey Hui and Jeremy Hurwitz and Michael Isard and Abe Ittycheriah and Matthew Jagielski and Wen Hao Jia and Kathleen Kenealy and Maxim Krikun and Sneha Kudugunta and Chang Lan and Katherine Lee and Benjamin Lee and Eric Li and Mu-Li Li and Wei Li and Yaguang Li and Jun Yu Li and Hyeontaek Lim and Han Lin and Zhong-Zhong Liu and Frederick Liu and Marcello Maggioni and Aroma Mahendru and Joshua Maynez and Vedant Misra and Maysam Moussalem and Zachary Nado and John Nham and Eric Ni and Andrew Nystrom and Alicia Parrish and Marie Pellat and Martin Polacek and Alex Polozov and Reiner Pope and Siyuan Qiao and Emily Reif and Bryan Richter and Parker Riley and Alexandra Ros and Aurko Roy and Brennan Saeta and Rajkumar Samuel and Renee Marie Shelby and Ambrose Slone and Daniel Smilkov and David R. So and Daniela Sohn and Simon Tokumine and Dasha Valter and Vijay Vasudevan and Kiran Vodrahalli and Xuezhi Wang and Pidong Wang and Zirui Wang and Tao Wang and John Wieting and Yuhuai Wu and Ke Xu and Yunhan Xu and Lin Wu Xue and Pengcheng Yin and Jiahui Yu and Qiaoling Zhang and Steven Zheng and Ce Zheng and Wei Zhou and Denny Zhou and Slav Petrov and Yonghui Wu},
  journal={ArXiv},
  year={2023},
  volume={abs/2305.10403},
  url={https://api.semanticscholar.org/CorpusID:258740735}
}

@article{T5,
author = {Raffel, Colin and Shazeer, Noam and Roberts, Adam and Lee, Katherine and Narang, Sharan and Matena, Michael and Zhou, Yanqi and Li, Wei and Liu, Peter J.},
title = {Exploring the Limits of Transfer Learning with a Unified Text-to-Text Transformer},
year = {2020},
issue_date = {January 2020},
publisher = {JMLR.org},
volume = {21},
number = {1},
issn = {1532-4435},
journal = {J. Mach. Learn. Res.},
articleno = {140},
numpages = {67}
}

@inproceedings{lewis-etal-2020-bart,
    title = "{BART}: Denoising Sequence-to-Sequence Pre-training for Natural Language Generation, Translation, and Comprehension",
    author = "Lewis, Mike  and
      Liu, Yinhan  and
      Goyal, Naman  and
      Ghazvininejad, Marjan  and
      Mohamed, Abdelrahman  and
      Levy, Omer  and
      Stoyanov, Veselin  and
      Zettlemoyer, Luke",
    editor = "Jurafsky, Dan  and
      Chai, Joyce  and
      Schluter, Natalie  and
      Tetreault, Joel",
    booktitle = "Proceedings of the 58th Annual Meeting of the Association for Computational Linguistics",
    month = jul,
    year = "2020",
    address = "Online",
    publisher = "Association for Computational Linguistics",
    url = "https://aclanthology.org/2020.acl-main.703",
    doi = "10.18653/v1/2020.acl-main.703",
    pages = "7871--7880",
}

@inproceedings{zhuang-etal-2021-robustly-roberta,
    title = "A Robustly Optimized {BERT} Pre-training Approach with Post-training",
    author = "Zhuang, Liu  and
      Wayne, Lin  and
      Ya, Shi  and
      Jun, Zhao",
    editor = "Li, Sheng  and
      Sun, Maosong  and
      Liu, Yang  and
      Wu, Hua  and
      Liu, Kang  and
      Che, Wanxiang  and
      He, Shizhu  and
      Rao, Gaoqi",
    booktitle = "Proceedings of the 20th Chinese National Conference on Computational Linguistics",
    month = aug,
    year = "2021",
    address = "Huhhot, China",
    publisher = "Chinese Information Processing Society of China",
    url = "https://aclanthology.org/2021.ccl-1.108",
    pages = "1218--1227",
    language = "English",
}

@article{Lan2019ALBERTAL,
  title={ALBERT: A Lite BERT for Self-supervised Learning of Language Representations},
  author={Zhenzhong Lan and Mingda Chen and Sebastian Goodman and Kevin Gimpel and Piyush Sharma and Radu Soricut},
  journal={ArXiv},
  year={2019},
  volume={abs/1909.11942},
  url={https://api.semanticscholar.org/CorpusID:202888986}
}

@inproceedings{pan-etal-2023-context,
    title = "What In-Context Learning {``}Learns{''} In-Context: Disentangling Task Recognition and Task Learning",
    author = "Pan, Jane  and
      Gao, Tianyu  and
      Chen, Howard  and
      Chen, Danqi",
    editor = "Rogers, Anna  and
      Boyd-Graber, Jordan  and
      Okazaki, Naoaki",
    booktitle = "Findings of the Association for Computational Linguistics: ACL 2023",
    month = jul,
    year = "2023",
    address = "Toronto, Canada",
    publisher = "Association for Computational Linguistics",
    url = "https://aclanthology.org/2023.findings-acl.527",
    doi = "10.18653/v1/2023.findings-acl.527",
    pages = "8298--8319",
}

@inproceedings{ASurveyIncontextLearning,
  title={A Survey on In-context Learning},
  author={Qingxiu Dong and Lei Li and Damai Dai and Ce Zheng and Zhiyong Wu and Baobao Chang and Xu Sun and Jingjing Xu and Zhifang Sui},
  year={2022},
  url={https://api.semanticscholar.org/CorpusID:255372865}
}

@article{Liu2021OPTOP,
  title={OPT: Omni-Perception Pre-Trainer for Cross-Modal Understanding and Generation},
  author={Jing Liu and Xinxin Zhu and Fei Liu and Longteng Guo and Zijia Zhao and Ming-Ting Sun and Weining Wang and Hanqing Lu and Shiyu Zhou and Jiajun Zhang and Jinqiao Wang},
  journal={ArXiv},
  year={2021},
  volume={abs/2107.00249},
  url={https://api.semanticscholar.org/CorpusID:235694336}
}

@article{White2023APP,
  title={A Prompt Pattern Catalog to Enhance Prompt Engineering with ChatGPT},
  author={Jules White and Quchen Fu and Sam Hays and Michael Sandborn and Carlos Olea and Henry Gilbert and Ashraf Elnashar and Jesse Spencer-Smith and Douglas C. Schmidt},
  journal={ArXiv},
  year={2023},
  volume={abs/2302.11382},
  url={https://api.semanticscholar.org/CorpusID:257079092}
}

@article{Argyle2022AnIA,
  title={An Information-theoretic Approach to Prompt Engineering Without Ground Truth Labels},
  author={Lisa P. Argyle and E. Busby and Nancy Fulda and Joshua R Gubler and Christopher Rytting and Taylor Sorensen and David Wingate},
  journal={Political Analysis},
  year={2022},
  volume={31},
  pages={337 - 351},
  url={https://api.semanticscholar.org/CorpusID:252280474}
}

@misc{ye2023prompt,
      title={Prompt Engineering a Prompt Engineer}, 
      author={Qinyuan Ye and Maxamed Axmed and Reid Pryzant and Fereshte Khani},
      year={2023},
      eprint={2311.05661},
      archivePrefix={arXiv},
      primaryClass={cs.CL}
}

@misc{howard2018universal,
      title={Universal Language Model Fine-tuning for Text Classification}, 
      author={Jeremy Howard and Sebastian Ruder},
      year={2018},
      eprint={1801.06146},
      archivePrefix={arXiv},
      primaryClass={cs.CL}
}

@inproceedings{howard-ruder-2018-universal,
    title = "Universal Language Model Fine-tuning for Text Classification",
    author = "Howard, Jeremy  and
      Ruder, Sebastian",
    editor = "Gurevych, Iryna  and
      Miyao, Yusuke",
    booktitle = "Proceedings of the 56th Annual Meeting of the Association for Computational Linguistics (Volume 1: Long Papers)",
    month = jul,
    year = "2018",
    address = "Melbourne, Australia",
    publisher = "Association for Computational Linguistics",
    url = "https://aclanthology.org/P18-1031",
    doi = "10.18653/v1/P18-1031",
    pages = "328--339",
}

@article{BioBERT2019,
    author = {Lee, Jinhyuk and Yoon, Wonjin and Kim, Sungdong and Kim, Donghyeon and Kim, Sunkyu and So, Chan Ho and Kang, Jaewoo},
    title = "{BioBERT: a pre-trained biomedical language representation model for biomedical text mining}",
    journal = {Bioinformatics},
    volume = {36},
    number = {4},
    pages = {1234-1240},
    year = {2019},
    month = {09},
    issn = {1367-4803},
    url = {https://doi.org/10.1093/bioinformatics/btz682}  
}

@article{ClinicalBERT2019,
  title={ClinicalBERT: Modeling Clinical Notes and Predicting Hospital Readmission},
  author={Kexin Huang and Jaan Altosaar and Rajesh Ranganath},
  journal={ArXiv},
  year={2019},
  volume={abs/1904.05342},
  url={https://api.semanticscholar.org/CorpusID:119308351}
}

@inproceedings{Finn2017ModelAgnosticMF,
  title={Model-Agnostic Meta-Learning for Fast Adaptation of Deep Networks},
  author={Chelsea Finn and P. Abbeel and Sergey Levine},
  booktitle={International Conference on Machine Learning},
  year={2017},
  url={https://api.semanticscholar.org/CorpusID:6719686}
}

@article{Geng2019FewShotTC,
  title={Few-Shot Text Classification with Induction Network},
  author={Ruiying Geng and Binhua Li and Yongbin Li and Yuxiao Ye and Ping Jian and Jian Sun},
  journal={ArXiv},
  year={2019},
  volume={abs/1902.10482},
  url={https://api.semanticscholar.org/CorpusID:67856277}
}

@article{Li2023TableGPTTG,
  title={Table-GPT: Table-tuned GPT for Diverse Table Tasks},
  author={Peng Li and Yeye He and Dror Yashar and Weiwei Cui and Song Ge and Haidong Zhang and Danielle Rifinski Fainman and Dongmei Zhang and Surajit Chaudhuri},
  journal={ArXiv},
  year={2023},
  volume={abs/2310.09263},
  url={https://api.semanticscholar.org/CorpusID:264127877}
}

@article{Hinton2015DistillingTK,
  title={Distilling the Knowledge in a Neural Network},
  author={Geoffrey E. Hinton and Oriol Vinyals and Jeffrey Dean},
  journal={ArXiv},
  year={2015},
  volume={abs/1503.02531},
  url={https://api.semanticscholar.org/CorpusID:7200347}
}

@inproceedings{Jiao2019TinyBERTDB,
  title={TinyBERT: Distilling BERT for Natural Language Understanding},
  author={Xiaoqi Jiao and Yichun Yin and Lifeng Shang and Xin Jiang and Xiao Chen and Linlin Li and Fang Wang and Qun Liu},
  booktitle={Findings},
  year={2019},
  url={https://api.semanticscholar.org/CorpusID:202719327}
}

@article{Ruder2017MultiTaskLearning,
  title={An Overview of Multi-Task Learning in Deep Neural Networks},
  author={Sebastian Ruder},
  journal={ArXiv},
  year={2017},
  volume={abs/1706.05098},
  url={https://api.semanticscholar.org/CorpusID:10175374}
}

@inproceedings{MultiTaskDeepNeuralNetworks,
    title = "Multi-Task Deep Neural Networks for Natural Language Understanding",
    author = "Liu, Xiaodong  and
      He, Pengcheng  and
      Chen, Weizhu  and
      Gao, Jianfeng",
    editor = "Korhonen, Anna  and
      Traum, David  and
      M{\`a}rquez, Llu{\'\i}s",
    booktitle = "Proceedings of the 57th Annual Meeting of the Association for Computational Linguistics",
    month = jul,
    year = "2019",
    address = "Florence, Italy",
    publisher = "Association for Computational Linguistics",
    url = "https://aclanthology.org/P19-1441",
    doi = "10.18653/v1/P19-1441",
    pages = "4487--4496",
}

@misc{houlsby2019parameterefficient,
      title={Parameter-Efficient Transfer Learning for NLP}, 
      author={Neil Houlsby and Andrei Giurgiu and Stanislaw Jastrzebski and Bruna Morrone and Quentin de Laroussilhe and Andrea Gesmundo and Mona Attariyan and Sylvain Gelly},
      year={2019},
      eprint={1902.00751},
      archivePrefix={arXiv},
      primaryClass={cs.LG}
}

@misc{li2021prefixtuning,
      title={Prefix-Tuning: Optimizing Continuous Prompts for Generation}, 
      author={Xiang Lisa Li and Percy Liang},
      year={2021},
      eprint={2101.00190},
      archivePrefix={arXiv},
      primaryClass={cs.CL}
}

@article{Ziegler2019FineTuningLM,
  title={Fine-Tuning Language Models from Human Preferences},
  author={Daniel M. Ziegler and Nisan Stiennon and Jeff Wu and Tom B. Brown and Alec Radford and Dario Amodei and Paul Christiano and Geoffrey Irving},
  journal={ArXiv},
  year={2019},
  volume={abs/1909.08593},
  url={https://api.semanticscholar.org/CorpusID:202660943}
}

@inproceedings{Mi2020ContinualLF,
  title={Continual Learning for Natural Language Generation in Task-oriented Dialog Systems},
  author={Fei Mi and Liangwei Chen and Mengjie Zhao and Minlie Huang and Boi Faltings},
  booktitle={Findings},
  year={2020},
  url={https://api.semanticscholar.org/CorpusID:222124847}
}

@misc{su2023roformer,
      title={RoFormer: Enhanced Transformer with Rotary Position Embedding}, 
      author={Jianlin Su and Yu Lu and Shengfeng Pan and Ahmed Murtadha and Bo Wen and Yunfeng Liu},
      year={2023},
      eprint={2104.09864},
      archivePrefix={arXiv},
      primaryClass={cs.CL}
}

@article{kl,
  title = {Kullback–{{Leibler}} Divergence: {{A}} Quantile Approach},
  author = {Sankaran, P.G. and Sunoj, S.M. and Nair, N. Unnikrishnan},
  date = {2016},
  journaltitle = {Statistics \& Probability Letters},
  volume = {111},
  pages = {72--79},
  issn = {0167-7152},
  doi = {10.1016/j.spl.2016.01.007},
  url = {https://www.sciencedirect.com/science/article/pii/S0167715216000067},
}

@article{Houlsby2019ParameterEfficientTL,
  title={Parameter-Efficient Transfer Learning for NLP},
  author={Neil Houlsby and Andrei Giurgiu and Stanislaw Jastrzebski and Bruna Morrone and Quentin de Laroussilhe and Andrea Gesmundo and Mona Attariyan and Sylvain Gelly},
  journal={ArXiv},
  year={2019},
  volume={abs/1902.00751},
  url={https://api.semanticscholar.org/CorpusID:59599816}
}

@inproceedings{ben-zaken-etal-2022-bitfit,
    title = "{B}it{F}it: Simple Parameter-efficient Fine-tuning for Transformer-based Masked Language-models",
    author = "Ben Zaken, Elad  and
      Goldberg, Yoav  and
      Ravfogel, Shauli",
    editor = "Muresan, Smaranda  and
      Nakov, Preslav  and
      Villavicencio, Aline",
    booktitle = "Proceedings of the 60th Annual Meeting of the Association for Computational Linguistics (Volume 2: Short Papers)",
    month = may,
    year = "2022",
    address = "Dublin, Ireland",
    publisher = "Association for Computational Linguistics",
    url = "https://aclanthology.org/2022.acl-short.1",
    doi = "10.18653/v1/2022.acl-short.1",
    pages = "1--9",
}

@article{Liu2021P-Tuning,
  title={GPT Understands, Too},
  author={Xiao Liu and Yanan Zheng and Zhengxiao Du and Ming Ding and Yujie Qian and Zhilin Yang and Jie Tang},
  journal={ArXiv},
  year={2021},
  volume={abs/2103.10385},
  url={https://api.semanticscholar.org/CorpusID:232269696}
}

@inproceedings{liu-etal-2022-p-tuning-v2,
    title = "{P}-Tuning: Prompt Tuning Can Be Comparable to Fine-tuning Across Scales and Tasks",
    author = "Liu, Xiao  and
      Ji, Kaixuan  and
      Fu, Yicheng  and
      Tam, Weng  and
      Du, Zhengxiao  and
      Yang, Zhilin  and
      Tang, Jie",
    editor = "Muresan, Smaranda  and
      Nakov, Preslav  and
      Villavicencio, Aline",
    booktitle = "Proceedings of the 60th Annual Meeting of the Association for Computational Linguistics (Volume 2: Short Papers)",
    month = may,
    year = "2022",
    address = "Dublin, Ireland",
    publisher = "Association for Computational Linguistics",
    url = "https://aclanthology.org/2022.acl-short.8",
    doi = "10.18653/v1/2022.acl-short.8",
    pages = "61--68",
}

@article{Lialin2023ScalingDT,
  title={Scaling Down to Scale Up: A Guide to Parameter-Efficient Fine-Tuning},
  author={Vladislav Lialin and Vijeta Deshpande and Anna Rumshisky},
  journal={ArXiv},
  year={2023},
  volume={abs/2303.15647},
  url={https://api.semanticscholar.org/CorpusID:257771591}
}

@article{Radford2022RobustSRWhisper,
  title={Robust Speech Recognition via Large-Scale Weak Supervision},
  author={Alec Radford and Jong Wook Kim and Tao Xu and Greg Brockman and Christine McLeavey and Ilya Sutskever},
  journal={ArXiv},
  year={2022},
  volume={abs/2212.04356},
  url={https://api.semanticscholar.org/CorpusID:252923993}
}

@inproceedings{Hu2021LoRA,
  title={LoRA: Low-Rank Adaptation of Large Language Models},
  author={J. Edward Hu and Yelong Shen and Phillip Wallis and Zeyuan Allen-Zhu and Yuanzhi Li and Shean Wang and Weizhu Chen},
  journal={ArXiv},
  year={2021},
  volume={abs/2106.09685},
  url={https://api.semanticscholar.org/CorpusID:235458009}
}

@article{LoRA1,
  title={Practical Tips for Finetuning LLMs Using LoRA (Low-Rank Adaptation)},
  author={SEBASTIAN RASCHKA},
  year={2023},
  url={https://magazine.sebastianraschka.com/p/practical-tips-for-finetuning-llms}
}

@article{Dettmers2023-QLoRa,
  title={QLoRA: Efficient Finetuning of Quantized LLMs},
  author={Tim Dettmers and Artidoro Pagnoni and Ari Holtzman and Luke Zettlemoyer},
  journal={ArXiv},
  year={2023},
  volume={abs/2305.14314},
  url={https://api.semanticscholar.org/CorpusID:258841328}
}

@inproceedings{lester-etal-2021-Prompt-Tuning,
    title = "The Power of Scale for Parameter-Efficient Prompt Tuning",
    author = "Lester, Brian  and
      Al-Rfou, Rami  and
      Constant, Noah",
    editor = "Moens, Marie-Francine  and
      Huang, Xuanjing  and
      Specia, Lucia  and
      Yih, Scott Wen-tau",
    booktitle = "Proceedings of the 2021 Conference on Empirical Methods in Natural Language Processing",
    month = nov,
    year = "2021",
    address = "Online and Punta Cana, Dominican Republic",
    publisher = "Association for Computational Linguistics",
    url = "https://aclanthology.org/2021.emnlp-main.243",
    doi = "10.18653/v1/2021.emnlp-main.243",
    pages = "3045--3059",
}

@inproceedings{NEURIPS2022_IA3,
  title = {Few-Shot Parameter-Efficient Fine-Tuning Is Better and Cheaper than in-Context Learning},
  booktitle = {Advances in Neural Information Processing Systems},
  author = {Liu, Haokun and Tam, Derek and Muqeeth, Mohammed and Mohta, Jay and Huang, Tenghao and Bansal, Mohit and Raffel, Colin A},
  editor = {Koyejo, S. and Mohamed, S. and Agarwal, A. and Belgrave, D. and Cho, K. and Oh, A.},
  date = {2022},
  volume = {35},
  pages = {1950--1965},
  publisher = {{Curran Associates, Inc.}},
  url = {https://proceedings.neurips.cc/paper_files/paper/2022/file/0cde695b83bd186c1fd456302888454c-Paper-Conference.pdf}
}

@inproceedings{mao-etal-2022-unipelt,
    title = "{U}ni{PELT}: A Unified Framework for Parameter-Efficient Language Model Tuning",
    author = "Mao, Yuning  and
      Mathias, Lambert  and
      Hou, Rui  and
      Almahairi, Amjad  and
      Ma, Hao  and
      Han, Jiawei  and
      Yih, Scott  and
      Khabsa, Madian",
    editor = "Muresan, Smaranda  and
      Nakov, Preslav  and
      Villavicencio, Aline",
    booktitle = "Proceedings of the 60th Annual Meeting of the Association for Computational Linguistics (Volume 1: Long Papers)",
    month = may,
    year = "2022",
    address = "Dublin, Ireland",
    publisher = "Association for Computational Linguistics",
    url = "https://aclanthology.org/2022.acl-long.433",
    doi = "10.18653/v1/2022.acl-long.433",
    pages = "6253--6264",
}

@article{He2021-MAM-Adapter,
  title={Towards a Unified View of Parameter-Efficient Transfer Learning},
  author={Junxian He and Chunting Zhou and Xuezhe Ma and Taylor Berg-Kirkpatrick and Graham Neubig},
  journal={ArXiv},
  year={2021},
  volume={abs/2110.04366},
  url={https://api.semanticscholar.org/CorpusID:238583580}
}

@article{Ouyang2022TrainingLM_RLHF,
  title={Training language models to follow instructions with human feedback},
  author={Long Ouyang and Jeff Wu and Xu Jiang and Diogo Almeida and Carroll L. Wainwright and Pamela Mishkin and Chong Zhang and Sandhini Agarwal and Katarina Slama and Alex Ray and John Schulman and Jacob Hilton and Fraser Kelton and Luke E. Miller and Maddie Simens and Amanda Askell and Peter Welinder and Paul Francis Christiano and Jan Leike and Ryan J. Lowe},
  journal={ArXiv},
  year={2022},
  volume={abs/2203.02155},
  url={https://api.semanticscholar.org/CorpusID:246426909}
}

@article{Razzhigaev2023KandinskyAI,
  title={Kandinsky: an Improved Text-to-Image Synthesis with Image Prior and Latent Diffusion},
  author={Anton Razzhigaev and Arseniy Shakhmatov and Anastasia Maltseva and V.Ya. Arkhipkin and Igor Pavlov and Ilya Ryabov and Angelina Kuts and Alexander Panchenko and Andrey Kuznetsov and Denis Dimitrov},
  journal={ArXiv},
  year={2023},
  volume={abs/2310.03502},
  url={https://api.semanticscholar.org/CorpusID:263671912}
}

@article{Yang2023AutoGPTFO,
  title={Auto-GPT for Online Decision Making: Benchmarks and Additional Opinions},
  author={Hui Yang and Sifu Yue and Yunzhong He},
  journal={ArXiv},
  year={2023},
  volume={abs/2306.02224},
  url={https://api.semanticscholar.org/CorpusID:259075577}
}

@article{Zeng2023AgentTuningEG,
  title={AgentTuning: Enabling Generalized Agent Abilities for LLMs},
  author={Aohan Zeng and Mingdao Liu and Rui Lu and Bowen Wang and Xiao Liu and Yuxiao Dong and Jie Tang},
  journal={ArXiv},
  year={2023},
  volume={abs/2310.12823},
  url={https://api.semanticscholar.org/CorpusID:264306101}
}

@article{Luo2023LCMLoRAAU,
  title={LCM-LoRA: A Universal Stable-Diffusion Acceleration Module},
  author={Simian Luo and Yiqin Tan and Suraj Patil and Daniel Gu and Patrick von Platen and Apolin'ario Passos and Longbo Huang and Jian Li and Hang Zhao},
  journal={ArXiv},
  year={2023},
  volume={abs/2311.05556},
  url={https://api.semanticscholar.org/CorpusID:265067414}
}

@article{Saharia2022PhotorealisticTDImagen,
  title={Photorealistic Text-to-Image Diffusion Models with Deep Language Understanding},
  author={Chitwan Saharia and William Chan and Saurabh Saxena and Lala Li and Jay Whang and Emily L. Denton and Seyed Kamyar Seyed Ghasemipour and Burcu Karagol Ayan and Seyedeh Sara Mahdavi and Raphael Gontijo Lopes and Tim Salimans and Jonathan Ho and David J. Fleet and Mohammad Norouzi},
  journal={ArXiv},
  year={2022},
  volume={abs/2205.11487},
  url={https://api.semanticscholar.org/CorpusID:248986576}
}

@article{Blattmann2023AlignYLVideoLDM,
  title={Align Your Latents: High-Resolution Video Synthesis with Latent Diffusion Models},
  author={A. Blattmann and Robin Rombach and Huan Ling and Tim Dockhorn and Seung Wook Kim and Sanja Fidler and Karsten Kreis},
  journal={2023 IEEE/CVF Conference on Computer Vision and Pattern Recognition (CVPR)},
  year={2023},
  pages={22563-22575},
  url={https://api.semanticscholar.org/CorpusID:258187553}
}

@article{Ronneberger2015UNetCN,
  title={U-Net: Convolutional Networks for Biomedical Image Segmentation},
  author={Olaf Ronneberger and Philipp Fischer and Thomas Brox},
  journal={ArXiv},
  year={2015},
  volume={abs/1505.04597},
  url={https://api.semanticscholar.org/CorpusID:3719281}
}

@article{Liu2023ReasonFFXAgent,
  title={Reason for Future, Act for Now: A Principled Framework for Autonomous LLM Agents with Provable Sample Efficiency},
  author={Zhihan Liu and Hao Hu and Shenao Zhang and Hongyi Guo and Shuqi Ke and Boyi Liu and Zhaoran Wang},
  journal={ArXiv},
  year={2023},
  volume={abs/2309.17382},
  url={https://api.semanticscholar.org/CorpusID:263310943}
}

@article{Kirillov2023SegmentA,
  title={Segment Anything},
  author={Alexander Kirillov and Eric Mintun and Nikhila Ravi and Hanzi Mao and Chloe Rolland and Laura Gustafson and Tete Xiao and Spencer Whitehead and Alexander C. Berg and Wan-Yen Lo and Piotr Doll{\'a}r and Ross B. Girshick},
  journal={ArXiv},
  year={2023},
  volume={abs/2304.02643},
  url={https://api.semanticscholar.org/CorpusID:257952310}
}

@article{Oquab2023DINOv2LR,
  title={DINOv2: Learning Robust Visual Features without Supervision},
  author={Maxime Oquab and Timoth'ee Darcet and Th{\'e}o Moutakanni and Huy Q. Vo and Marc Szafraniec and Vasil Khalidov and Pierre Fernandez and Daniel Haziza and Francisco Massa and Alaaeldin El-Nouby and Mahmoud Assran and Nicolas Ballas and Wojciech Galuba and Russ Howes and Po-Yao (Bernie) Huang and Shang-Wen Li and Ishan Misra and Michael G. Rabbat and Vasu Sharma and Gabriel Synnaeve and Huijiao Xu and Herv{\'e} J{\'e}gou and Julien Mairal and Patrick Labatut and Armand Joulin and Piotr Bojanowski},
  journal={ArXiv},
  year={2023},
  volume={abs/2304.07193},
  url={https://api.semanticscholar.org/CorpusID:258170077}
}

@inproceedings{Driess2023PaLMEAE,
  title={PaLM-E: An Embodied Multimodal Language Model},
  author={Danny Driess and F. Xia and Mehdi S. M. Sajjadi and Corey Lynch and Aakanksha Chowdhery and Brian Ichter and Ayzaan Wahid and Jonathan Tompson and Quan Ho Vuong and Tianhe Yu and Wenlong Huang and Yevgen Chebotar and Pierre Sermanet and Daniel Duckworth and Sergey Levine and Vincent Vanhoucke and Karol Hausman and Marc Toussaint and Klaus Greff and Andy Zeng and Igor Mordatch and Peter R. Florence},
  booktitle={International Conference on Machine Learning},
  year={2023},
  url={https://api.semanticscholar.org/CorpusID:257364842}
}

@inproceedings{rt22023arxiv,
    title={RT-2: Vision-Language-Action Models Transfer Web Knowledge to Robotic Control},
    author={Anthony Brohan and Noah Brown and Justice Carbajal and Yevgen Chebotar and Xi Chen and Krzysztof Choromanski and Tianli Ding and Danny Driess and Avinava Dubey and Chelsea Finn and Pete Florence and Chuyuan Fu and Montse Gonzalez Arenas and Keerthana Gopalakrishnan and Kehang Han and Karol Hausman and Alex Herzog and Jasmine Hsu and Brian Ichter and Alex Irpan and Nikhil Joshi and Ryan Julian and Dmitry Kalashnikov and Yuheng Kuang and Isabel Leal  and Lisa Lee and Tsang-Wei Edward Lee and Sergey Levine and Yao Lu and Henryk Michalewski and Igor Mordatch and Karl Pertsch and Kanishka Rao and Krista Reymann and Michael Ryoo and Grecia Salazar and Pannag Sanketi and Pierre Sermanet and Jaspiar Singh and Anikait Singh and Radu Soricut and Huong Tran and Vincent Vanhoucke and Quan Vuong and Ayzaan Wahid and Stefan Welker and Paul Wohlhart and  Jialin Wu and Fei Xia and Ted Xiao and Peng Xu and Sichun Xu and Tianhe Yu and Brianna Zitkovich},
    booktitle={arXiv preprint arXiv:2307.15818},
    year={2023}
}

@inproceedings{ghosal-etal-2022-cicero,
    title = "{CICERO}: A Dataset for Contextualized Commonsense Inference in Dialogues",
    author = "Ghosal, Deepanway  and
      Shen, Siqi  and
      Majumder, Navonil  and
      Mihalcea, Rada  and
      Poria, Soujanya",
    editor = "Muresan, Smaranda  and
      Nakov, Preslav  and
      Villavicencio, Aline",
    booktitle = "Proceedings of the 60th Annual Meeting of the Association for Computational Linguistics (Volume 1: Long Papers)",
    month = may,
    year = "2022",
    address = "Dublin, Ireland",
    publisher = "Association for Computational Linguistics",
    url = "https://aclanthology.org/2022.acl-long.344",
    doi = "10.18653/v1/2022.acl-long.344", 
}

@article{bousmalis2023robocat,
  title={RoboCat: A Self-Improving Foundation Agent for Robotic Manipulation},
  author={Bousmalis, Konstantinos and Vezzani, Giulia and Rao, Dushyant and Devin, Coline and Lee, Alex X and Bauza, Maria and Davchev, Todor and Zhou, Yuxiang and Gupta, Agrim and Raju, Akhil and others},
  journal={arXiv preprint arXiv:2306.11706},
  year={2023}
}

@article{Singhal2023MedPaLM2,
  title={Towards Expert-Level Medical Question Answering with Large Language Models},
  author={K. Singhal and Tao Tu and Juraj Gottweis and Rory Sayres and Ellery Wulczyn and Le Hou and Kevin Clark and Stephen R. Pfohl and Heather J. Cole-Lewis and Darlene Neal and Mike Schaekermann and Amy Wang and Mohamed Amin and S. Lachgar and P. A. Mansfield and Sushant Prakash and Bradley Green and Ewa Dominowska and Blaise Ag{\"u}era y Arcas and Nenad Toma{\vs}ev and Yun Liu and Renee C Wong and Christopher Semturs and Seyedeh Sara Mahdavi and Jo{\"e}lle K. Barral and Dale R. Webster and Greg S Corrado and Yossi Matias and Shekoofeh Azizi and Alan Karthikesalingam and Vivek Natarajan},
  journal={ArXiv},
  year={2023},
  volume={abs/2305.09617},
  url={https://api.semanticscholar.org/CorpusID:258715226}
}

@inproceedings{Garza2023TimeGPT1,
  title={TimeGPT-1},
  author={Azul Garza and Max Mergenthaler-Canseco},
  year={2023},
  url={https://api.semanticscholar.org/CorpusID:263672111}
}

@inproceedings{Gu2023MambaLS,
  title={Mamba: Linear-Time Sequence Modeling with Selective State Spaces},
  author={Albert Gu and Tri Dao},
  year={2023},
  url={https://api.semanticscholar.org/CorpusID:265551773}
}

@article{Gu2021EfficientlyMLS4,
  title={Efficiently Modeling Long Sequences with Structured State Spaces},
  author={Albert Gu and Karan Goel and Christopher R'e},
  journal={ArXiv},
  year={2021},
  volume={abs/2111.00396},
  url={https://api.semanticscholar.org/CorpusID:240354066}
}

@article{Li2022CompetitionlevelCGAlphaCode,
  title={Competition-level code generation with AlphaCode},
  author={Yujia Li and David H. Choi and Junyoung Chung and Nate Kushman and Julian Schrittwieser and R{\'e}mi Leblond and Tom and Eccles and James Keeling and Felix Gimeno and Agustin Dal Lago and Thomas Hubert and Peter Choy and Cyprien de and Masson d’Autume and Igor Babuschkin and Xinyun Chen and Po-Sen Huang and Johannes Welbl and Sven Gowal and Alexey and Cherepanov and James Molloy and Daniel Jaymin Mankowitz and Esme Sutherland Robson and Pushmeet Kohli and Nando de and Freitas and Koray Kavukcuoglu and Oriol Vinyals},
  journal={Science},
  year={2022},
  volume={378},
  pages={1092 - 1097},
  url={https://api.semanticscholar.org/CorpusID:246527904}
}

@inproceedings{AlphaCode2_Tech_Report,
  title={AlphaCode 2 Technical Report},
  author={AlphaCode Team and Google DeepMind},
  year={2023},
  url={https://storage.googleapis.com/deepmind-media/AlphaCode2/AlphaCode2_Tech_Report.pdf}
}

@inproceedings{gemini_1_report,
  title={Gemini: A Family of Highly Capable Multimodal Models},
  author={Gemini Team},
  year={2023},
  url={https://storage.googleapis.com/deepmind-media/gemini/gemini_1_report.pdf}
}

@inproceedings{GeminiIntro,
  title={Introducing Gemini: our largest and most capable AI model},
  author={Sundar Pichai and Demis Hassabis},
  year={2023},
  url={https://blog.google/technology/ai/google-gemini-ai}
}

@article{AlphaGo,
author = {Silver, David and Huang, Aja and Maddison, Christopher and Guez, Arthur and Sifre, Laurent and Driessche, George and Schrittwieser, Julian and Antonoglou, Ioannis and Panneershelvam, Veda and Lanctot, Marc and Dieleman, Sander and Grewe, Dominik and Nham, John and Kalchbrenner, Nal and Sutskever, Ilya and Lillicrap, Timothy and Leach, Madeleine and Kavukcuoglu, Koray and Graepel, Thore and Hassabis, Demis},
year = {2016},
month = {01},
pages = {484-489},
title = {Mastering the game of Go with deep neural networks and tree search},
volume = {529},
journal = {Nature},
doi = {10.1038/nature16961}
}

@article{Min2022RethinkingTRICL,
  title={Rethinking the Role of Demonstrations: What Makes In-Context Learning Work?},
  author={Sewon Min and Xinxi Lyu and Ari Holtzman and Mikel Artetxe and Mike Lewis and Hannaneh Hajishirzi and Luke Zettlemoyer},
  journal={ArXiv},
  year={2022},
  volume={abs/2202.12837},
  url={https://api.semanticscholar.org/CorpusID:247155069}
}

@article{Xie2021AnEOICL,
  title={An Explanation of In-context Learning as Implicit Bayesian Inference},
  author={Sang Michael Xie and Aditi Raghunathan and Percy Liang and Tengyu Ma},
  journal={ArXiv},
  year={2021},
  volume={abs/2111.02080},
  url={https://api.semanticscholar.org/CorpusID:241035330}
}

@article{Wei2022ChainOTCoT,
  title={Chain of Thought Prompting Elicits Reasoning in Large Language Models},
  author={Jason Wei and Xuezhi Wang and Dale Schuurmans and Maarten Bosma and Ed Huai-hsin Chi and F. Xia and Quoc Le and Denny Zhou},
  journal={ArXiv},
  year={2022},
  volume={abs/2201.11903},
  url={https://api.semanticscholar.org/CorpusID:246411621}
}

@article{Kojima2022LargeLM,
  title={Large Language Models are Zero-Shot Reasoners},
  author={Takeshi Kojima and Shixiang Shane Gu and Machel Reid and Yutaka Matsuo and Yusuke Iwasawa},
  journal={ArXiv},
  year={2022},
  volume={abs/2205.11916},
  url={https://api.semanticscholar.org/CorpusID:249017743}
}

@article{Qiao2023TaskWeaverAC,
  title={TaskWeaver: A Code-First Agent Framework},
  author={Bo Qiao and Liqun Li and Xu Zhang and Shilin He and Yu Kang and Chaoyun Zhang and Fangkai Yang and Hang Dong and Jue Zhang and Lu Wang and Ming-Jie Ma and Pu Zhao and Si Qin and Xiaoting Qin and Chao Du and Yong Xu and Qingwei Lin and S. Rajmohan and Dongmei Zhang},
  journal={ArXiv},
  year={2023},
  volume={abs/2311.17541},
  url={https://api.semanticscholar.org/CorpusID:265498341}
}

@inproceedings{Wu2023AutoGenEN,
  title={AutoGen: Enabling Next-Gen LLM Applications via Multi-Agent Conversation},
  author={Qingyun Wu and Gagan Bansal and Jieyu Zhang and Yiran Wu and Beibin Li and Erkang Zhu and Li Jiang and Xiaoyun Zhang and Shaokun Zhang and Jiale Liu and Ahmed Hassan Awadallah and Ryen W White and Doug Burger and Chi Wang},
  year={2023},
  url={https://api.semanticscholar.org/CorpusID:263611068}
}

@article{Lewis2020RetrievalAugmentedGF,
  title={Retrieval-Augmented Generation for Knowledge-Intensive NLP Tasks},
  author={Patrick Lewis and Ethan Perez and Aleksandara Piktus and Fabio Petroni and Vladimir Karpukhin and Naman Goyal and Heinrich Kuttler and Mike Lewis and Wen-tau Yih and Tim Rockt{\"a}schel and Sebastian Riedel and Douwe Kiela},
  journal={ArXiv},
  year={2020},
  volume={abs/2005.11401},
  url={https://api.semanticscholar.org/CorpusID:218869575}
}

@article{Asai2023SelfRAGLT,
  title={Self-RAG: Learning to Retrieve, Generate, and Critique through Self-Reflection},
  author={Akari Asai and Zeqiu Wu and Yizhong Wang and Avirup Sil and Hannaneh Hajishirzi},
  journal={ArXiv},
  year={2023},
  volume={abs/2310.11511},
  url={https://api.semanticscholar.org/CorpusID:264288947}
}

@article{Shi2023REPLUGRB,
  title={REPLUG: Retrieval-Augmented Black-Box Language Models},
  author={Weijia Shi and Sewon Min and Michihiro Yasunaga and Minjoon Seo and Rich James and Mike Lewis and Luke Zettlemoyer and Wen-tau Yih},
  journal={ArXiv},
  year={2023},
  volume={abs/2301.12652},
  url={https://api.semanticscholar.org/CorpusID:256389797}
}

@article{Izacard2022FewshotLW,
  title={Few-shot Learning with Retrieval Augmented Language Models},
  author={Gautier Izacard and Patrick Lewis and Maria Lomeli and Lucas Hosseini and Fabio Petroni and Timo Schick and Jane A. Yu and Armand Joulin and Sebastian Riedel and Edouard Grave},
  journal={ArXiv},
  year={2022},
  volume={abs/2208.03299},
  url={https://api.semanticscholar.org/CorpusID:251371732}
}

@article{Lin2023RADITRD,
  title={RA-DIT: Retrieval-Augmented Dual Instruction Tuning},
  author={Xi Victoria Lin and Xilun Chen and Mingda Chen and Weijia Shi and Maria Lomeli and Rich James and Pedro Rodriguez and Jacob Kahn and Gergely Szilvasy and Mike Lewis and Luke Zettlemoyer and Scott Yih},
  journal={ArXiv},
  year={2023},
  volume={abs/2310.01352},
  url={https://api.semanticscholar.org/CorpusID:263605962}
}

@article{Sanh2019DistilBERTAD,
  title={DistilBERT, a distilled version of BERT: smaller, faster, cheaper and lighter},
  author={Victor Sanh and Lysandre Debut and Julien Chaumond and Thomas Wolf},
  journal={ArXiv},
  year={2019},
  volume={abs/1910.01108},
  url={https://api.semanticscholar.org/CorpusID:203626972}
}

@article{Yao2022ReActSR,
  title={ReAct: Synergizing Reasoning and Acting in Language Models},
  author={Shunyu Yao and Jeffrey Zhao and Dian Yu and Nan Du and Izhak Shafran and Karthik Narasimhan and Yuan Cao},
  journal={ArXiv},
  year={2022},
  volume={abs/2210.03629},
  url={https://api.semanticscholar.org/CorpusID:252762395}
}

@article{Pandya2023AutomatingCS,
  title={Automating Customer Service using LangChain: Building custom open-source GPT Chatbot for organizations},
  author={Keivalya Pandya and Mehfuza S. Holia},
  journal={ArXiv},
  year={2023},
  volume={abs/2310.05421},
  url={https://api.semanticscholar.org/CorpusID:263830717}
}

@article{Wu2022PromptChainerCL,
  title={PromptChainer: Chaining Large Language Model Prompts through Visual Programming},
  author={Tongshuang Sherry Wu and Ellen Jiang and Aaron Donsbach and Jeff Gray and Alejandra Molina and Michael Terry and Carrie J. Cai},
  journal={CHI Conference on Human Factors in Computing Systems Extended Abstracts},
  year={2022},
  url={https://api.semanticscholar.org/CorpusID:247447133}
}

@article{Zhang2023IgnitingLI,
  title={Igniting Language Intelligence: The Hitchhiker's Guide From Chain-of-Thought Reasoning to Language Agents},
  author={Zhuosheng Zhang and Yao Yao and Aston Zhang and Xiangru Tang and Xinbei Ma and Zhiwei He and Yiming Wang and Mark B. Gerstein and Rui Wang and Gongshen Liu and Hai Zhao},
  journal={ArXiv},
  year={2023},
  volume={abs/2311.11797},
  url={https://api.semanticscholar.org/CorpusID:265295180}
}

@article{Siriwardhana2022ImprovingTD,
  title={Improving the Domain Adaptation of Retrieval Augmented Generation (RAG) Models for Open Domain Question Answering},
  author={Shamane Siriwardhana and Rivindu Weerasekera and Elliott Wen and Tharindu Kaluarachchi and Rajib Kumar Rana and Suranga Nanayakkara},
  journal={Transactions of the Association for Computational Linguistics},
  year={2022},
  volume={11},
  pages={1-17},
  url={https://api.semanticscholar.org/CorpusID:252735056}
}

@inproceedings{RAGFineTuning,
  title={RAG Vs Fine-Tuning Vs Both: A Guide For Optimizing LLM Performance},
  author={Pratik Bhavsar},
  year={2023},
  url={https://www.rungalileo.io/blog/optimizing-llm-performance-rag-vs-finetune-vs-both}
}

@inproceedings{Longpre2023TheFCFLANv2,
  title={The Flan Collection: Designing Data and Methods for Effective Instruction Tuning},
  author={S. Longpre and Le Hou and Tu Vu and Albert Webson and Hyung Won Chung and Yi Tay and Denny Zhou and Quoc V. Le and Barret Zoph and Jason Wei and Adam Roberts},
  booktitle={International Conference on Machine Learning},
  year={2023},
  url={https://api.semanticscholar.org/CorpusID:256415991}
}

@article{Wei2021FinetunedLMFLAN,
  title={Finetuned Language Models Are Zero-Shot Learners},
  author={Jason Wei and Maarten Bosma and Vincent Zhao and Kelvin Guu and Adams Wei Yu and Brian Lester and Nan Du and Andrew M. Dai and Quoc V. Le},
  journal={ArXiv},
  year={2021},
  volume={abs/2109.01652},
  url={https://api.semanticscholar.org/CorpusID:237416585}
}

@article{Wang2023InstructUIEMI,
  title={InstructUIE: Multi-task Instruction Tuning for Unified Information Extraction},
  author={Xiao Wang and Wei Zhou and Can Zu and Han Xia and Tianze Chen and Yuan Zhang and Rui Zheng and Junjie Ye and Qi Zhang and Tao Gui and Jihua Kang and J. Yang and Siyuan Li and Chunsai Du},
  journal={ArXiv},
  year={2023},
  volume={abs/2304.08085},
  url={https://api.semanticscholar.org/CorpusID:258179792}
}

@inproceedings{Munos2023NashLF,
  title={Nash Learning from Human Feedback},
  author={R{\'e}mi Munos and Michal Valko and Daniele Calandriello and Mohammad Gheshlaghi Azar and Mark Rowland and Zhaohan Daniel Guo and Yunhao Tang and Matthieu Geist and Thomas M'esnard and Andrea Michi and Marco Selvi and Sertan Girgin and Nikola Momchev and Olivier Bachem and Daniel Jaymin Mankowitz and Doina Precup and Bilal Piot},
  year={2023},
  url={https://api.semanticscholar.org/CorpusID:265609682}
}

@inproceedings{Lu2022UnifiedSG,
  title={Unified Structure Generation for Universal Information Extraction},
  author={Yaojie Lu and Qing Liu and Dai Dai and Xinyan Xiao and Hongyu Lin and Xianpei Han and Le Sun and Hua Wu},
  booktitle={Annual Meeting of the Association for Computational Linguistics},
  year={2022},
  url={https://api.semanticscholar.org/CorpusID:247619149}
}

@article{Jiang2023Mistral7,
  title={Mistral 7B},
  author={Albert Qiaochu Jiang and Alexandre Sablayrolles and Arthur Mensch and Chris Bamford and Devendra Singh Chaplot and Diego de Las Casas and Florian Bressand and Gianna Lengyel and Guillaume Lample and Lucile Saulnier and L'elio Renard Lavaud and Marie-Anne Lachaux and Pierre Stock and Teven Le Scao and Thibaut Lavril and Thomas Wang and Timoth{\'e}e Lacroix and William El Sayed},
  journal={ArXiv},
  year={2023},
  volume={abs/2310.06825},
  url={https://api.semanticscholar.org/CorpusID:263830494}
}

@article{Ainslie2023GQATG,
  title={GQA: Training Generalized Multi-Query Transformer Models from Multi-Head Checkpoints},
  author={Joshua Ainslie and James Lee-Thorp and Michiel de Jong and Yury Zemlyanskiy and Federico Lebr'on and Sumit K. Sanghai},
  journal={ArXiv},
  year={2023},
  volume={abs/2305.13245},
  url={https://api.semanticscholar.org/CorpusID:258833177}
}

@article{Wang2023VoyagerAO,
  title={Voyager: An Open-Ended Embodied Agent with Large Language Models},
  author={Guanzhi Wang and Yuqi Xie and Yunfan Jiang and Ajay Mandlekar and Chaowei Xiao and Yuke Zhu and Linxi (Jim) Fan and Anima Anandkumar},
  journal={ArXiv},
  year={2023},
  volume={abs/2305.16291},
  url={https://api.semanticscholar.org/CorpusID:258887849}
}

@article{Park2023GenerativeAI,
  title={Generative Agents: Interactive Simulacra of Human Behavior},
  author={Joon Sung Park and Joseph C. O'Brien and Carrie J. Cai and Meredith Ringel Morris and Percy Liang and Michael S. Bernstein},
  journal={Proceedings of the 36th Annual ACM Symposium on User Interface Software and Technology},
  year={2023},
  url={https://api.semanticscholar.org/CorpusID:258040990}
}

@article{Zeng2022GLM130BAO,
  title={GLM-130B: An Open Bilingual Pre-trained Model},
  author={Aohan Zeng and Xiao Liu and Zhengxiao Du and Zihan Wang and Hanyu Lai and Ming Ding and Zhuoyi Yang and Yifan Xu and Wendi Zheng and Xiao Xia and Weng Lam Tam and Zixuan Ma and Yufei Xue and Jidong Zhai and Wenguang Chen and P. Zhang and Yuxiao Dong and Jie Tang},
  journal={ArXiv},
  year={2022},
  volume={abs/2210.02414},
  url={https://api.semanticscholar.org/CorpusID:252715691}
}

@article{Rafailov2023DirectPODPO,
  title={Direct Preference Optimization: Your Language Model is Secretly a Reward Model},
  author={Rafael Rafailov and Archit Sharma and Eric Mitchell and Stefano Ermon and Christopher D. Manning and Chelsea Finn},
  journal={ArXiv},
  year={2023},
  volume={abs/2305.18290},
  url={https://api.semanticscholar.org/CorpusID:258959321}
}

@inproceedings{DPOLLama2,
  title={Fine-tune Llama 2 with DPO},
  author={Kashif Rasul and Younes Belkada and Leandro von Werra},
  booktitle={Annual Meeting of the Association for Computational Linguistics},
  year={2022},
  url={https://huggingface.co/blog/dpo-trl}
}

@article{Schulman2017ProximalPO,
  title={Proximal Policy Optimization Algorithms},
  author={John Schulman and Filip Wolski and Prafulla Dhariwal and Alec Radford and Oleg Klimov},
  journal={ArXiv},
  year={2017},
  volume={abs/1707.06347},
  url={https://api.semanticscholar.org/CorpusID:28695052}
}

\end{document}